\documentclass{article}

\usepackage[main, final]{neurips_2026}

\usepackage{graphicx}
\usepackage{amsmath}
\usepackage{amssymb}
\usepackage{booktabs}
\usepackage{algorithm}
\usepackage{algpseudocode}
\usepackage{xspace}

\usepackage{amsmath,amsfonts,bm}

\def\eqref#1{equation~\ref{#1}}

\def\1{\bm{1}}

\def\vn{{\bm{n}}}

\def\vr{{\bm{r}}}
\def\vs{{\bm{s}}}
\def\vt{{\bm{t}}}

\def\vv{{\bm{v}}}

\def\vx{{\bm{x}}}

\def\vz{{\bm{z}}}

\def\mI{{\bm{I}}}

\DeclareMathAlphabet{\mathsfit}{\encodingdefault}{\sfdefault}{m}{sl}
\SetMathAlphabet{\mathsfit}{bold}{\encodingdefault}{\sfdefault}{bx}{n}

\newcommand{\E}{\mathbb{E}}

\newcommand{\ourmethod}{FlowR2A}

\usepackage{amsthm}
\usepackage{tablefootnote}
\usepackage{multirow}
\usepackage{color, colortbl}
\usepackage{pifont}

\usepackage[utf8]{inputenc} 
\usepackage[T1]{fontenc}    
\usepackage{hyperref}       
\usepackage{url}            
\usepackage{booktabs}       
\usepackage{amsfonts}       
\usepackage{nicefrac}       
\usepackage{microtype}      
\usepackage{xcolor}         
\usepackage{paralist}       
\usepackage{cleveref}
\usepackage{subcaption}

\usepackage[font=small,skip=4pt]{caption}
\renewcommand{\arraystretch}{0.95}

\title{FlowR2A: Learning Reward-to-Action Distribution for Multimodal Driving Planning}

\newcommand{\authorskip}{\hspace{2.5mm}}
\newcommand{\institutionskip}{\hspace{5.0mm}}

\author{Xirui Li\textsuperscript{\mdseries1} \authorskip
Zhe Liu\textsuperscript{\mdseries1$\dag$} \authorskip
Xiaoqing Ye\textsuperscript{\mdseries2,3*} \authorskip
Wenhua Han\textsuperscript{\mdseries2} \\
\textbf{Yifeng Pan}\textsuperscript{\mdseries2} \authorskip
\textbf{Junyu Han}\textsuperscript{\mdseries2} \authorskip
\textbf{Hengshuang Zhao}\textsuperscript{\mdseries1*} \\
\textsuperscript{1}The University of Hong Kong \institutionskip
\textsuperscript{2}Changan Automobile \institutionskip
\textsuperscript{3}Wuhan University\\
\vspace{1mm}
\small{\textsuperscript{$\dag$}project lead\hspace{0.5cm}\textsuperscript{*}corresponding author}\\
{\tt\small \url{https://lixirui142.github.io/flowr2a-ad}}
}

\begin{document}

\maketitle

\begin{abstract}
Multimodal driving planning faces a long-standing tension between two paradigms: scoring-based methods benefit from dense reward supervision but are confined to a fixed action vocabulary, while anchor-based methods generate proposals dynamically yet suffer from sparse supervision constrained to a single ground-truth trajectory.
In this work, we propose \ourmethod{}, which resolves this tension by reframing simulation-based rewards from discriminative targets into generative conditions. By learning the reward-conditioned action distribution from dense trajectory-reward pairs with a flow-matching decoder, \ourmethod{} unifies the dense supervision of scoring-based methods with the proposal generation of anchor-based methods in a single generative model, forcing the model to internalize the correlation between an action and its outcomes in safety, progress, comfort, and rule compliance.
To balance hard safety constraints against soft progress objectives, we introduce fine-grained per-timestep reward conditioning and reward noise augmentation.
The generative formulation naturally supports controllable test-time sampling via reward guidance and anchored sampling, producing high-quality proposals.
\ourmethod{} achieves state-of-the-art results on the NAVSIM v1 and v2 benchmarks, with multimodal proposals of substantially higher quality than prior methods.
\end{abstract}

\section{Introduction}

End-to-end autonomous driving (E2E-AD) has emerged as a promising paradigm that maps raw sensor input directly to planning output through a differentiable model~\cite{hu2023planning,jiang2023vad,chitta2022transfuser}. A typical E2E-AD model consists of a perception encoder that extracts scene observations from sensor input and a plan decoder that produces planning actions on top of these observations. Early efforts predict a single action that imitates the human trajectory~\cite{hu2023planning,jiang2023vad,chitta2022transfuser,chen2024ppad,weng2024paradrive,li2024enhancing,li2024ego}. Given the inherent uncertainty and multimodality of driving behavior, recent research has shifted toward multimodal planning~\cite{chen2024vadv2,li2024hydra,liao2024diffusiondrive,guo2025ipad}, where the final action is selected from multiple proposals.

\begin{figure}[t]
\begin{center}
\includegraphics[width=\linewidth]{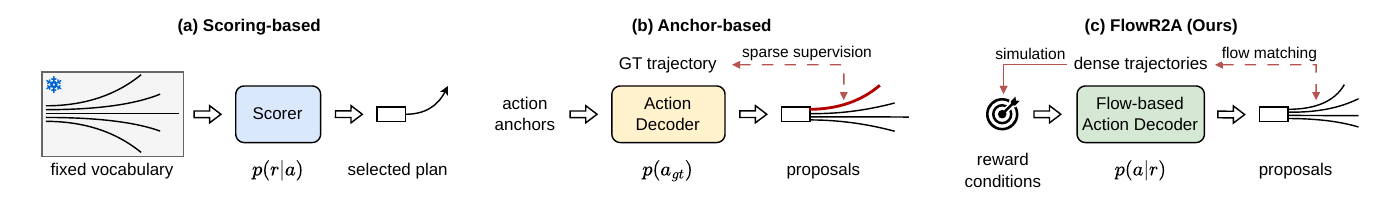}
\end{center}
\caption{Comparison of multimodal planning paradigms. (a) Scoring-based methods select from a large fixed trajectory vocabulary. (b) Anchor-based methods decode multiple proposals from action anchors and sparsely supervise them with GT. (c) Our method \ourmethod{} learns the reward-conditioned action distribution $p(a|r)$ from dense trajectories, where $a$ is the action and $r$ is the reward. Red lines indicate training supervision.}
\label{fig:teaser}
\end{figure}

Existing multimodal planners fall into two paradigms: scoring-based and anchor-based methods, illustrated in Fig.~\ref{fig:teaser}.
Scoring-based methods~\cite{chen2024vadv2,li2024hydra,li2025hydra,yao2026drivesuprim,li2025generalized} use a large, fixed action vocabulary as candidates and train a scorer to evaluate each candidate with simulation-based reward labels, selecting the best one as output. The key insight of this paradigm is that dense reward supervision over the entire action vocabulary provides a rich, comprehensive signal about the relationship between actions and their outcomes. However, scoring-based methods are fundamentally discriminative: they learn $p(r|a)$ to rank actions, but cannot transfer this knowledge to generate new proposals. Their output is therefore constrained to the fixed vocabulary, limiting adaptability to dynamic real-world scenarios.

Anchor-based methods~\cite{guo2025ipad,liao2024diffusiondrive,zou2025diffusiondrivev2,kirby2026driving} address this rigidity by decoding proposals dynamically from a set of action anchors and applying a winner-takes-all loss that supervises only the proposal closest to the single ground-truth (GT) trajectory. While this enables scene-adaptive proposals, the sparse GT supervision introduces two fundamental limitations. First, many anchors receive no training signal per scene, resulting in low-quality or degenerate proposals as shown in our experiments (Tab.~\ref{tab:proposal_quant}, Fig.~\ref{fig:qual}). Second, targeting the single GT trajectory inherits well-known imitation learning pathologies~\cite{dauner2023parting,dauner2024navsim}, including shortcut learning from ego status and unawareness of action consequences.

The two paradigms reveal a common tension: dense supervision and generative proposal modeling have so far been mutually exclusive. Scoring-based methods enjoy dense action-reward supervision but are bounded by their discriminative nature. Anchor-based methods can generate proposals, yet are supervised by a single GT trajectory per scene. This raises a natural question: can we combine dense reward supervision with generative proposal modeling in a single framework?

In this work, we propose \ourmethod{}, a multimodal planning framework that learns the reward-conditioned action distribution $p(a|r)$ from dense trajectory-reward pairs, resolving the above tension with a simple reframing. A dense action vocabulary can be paired with simulation-based rewards characterizing safety, progress, comfort, and rule compliance, yielding dense action-reward pairs that span the action space. While prior work~\cite{li2024hydra,yao2026drivesuprim} treats these rewards as discriminative targets to predict, we instead treat them as conditions to learn the generative reward-to-action distribution.
Under this formulation, each action-reward pair becomes a valid training sample, and the model is forced to internalize the correlation between an action and its outcomes. We construct fine-grained reward signals from rule-based simulation~\cite{dauner2024navsim} and train a flow-matching-based~\cite{liu2022flow,lipman2022flow,esser2024scaling} action decoder to recover clean trajectories from noise under reward conditioning. At inference, this generative formulation enables classifier-free guidance conditioned on high rewards and anchored sampling from any reference trajectory, yielding diverse, high-quality proposals.

A practical challenge arises in conditioning on high rewards. High-reward actions often sit close to the boundary of the feasible region, requiring the generative decoder to approach it without crossing into the infeasible side. We address this through two designs. First, we replace key safety and compliance rewards with per-timestep counterparts, which provide a more general and finer-grained signal that sharpens the hard constraints.
Second, we corrupt the continuous reward values with minor Gaussian noise during training, which acts as label smoothing and prevents the decoder from over-relying on them. Together, these designs effectively balance the conflicting objectives.

We evaluate \ourmethod{} on both NAVSIM v1 and v2 benchmarks~\cite{dauner2024navsim,cao2025pseudo}, achieving state-of-the-art performance. More importantly, our model produces multimodal proposals of substantially higher quality than prior methods, validating the benefit of modeling the entire conditional action distribution. Our contributions are:

\begin{compactitem}
    \item A new paradigm for multimodal driving planning that learns the reward-to-action distribution $p(a|r)$, unifying the dense supervision of scoring-based methods with the generation ability of anchor-based methods, and naturally supporting controllable test-time sampling.
    \item A reward construction recipe that effectively balances conflicting objectives with fine-grained per-timestep reward conditioning and reward noise augmentation.
    \item State-of-the-art results on the NAVSIM benchmark with higher proposal quality than previous methods, validating the advantage of learning the full conditional action distribution.
\end{compactitem}

\section{Preliminaries}\label{sec:pre}

Flow matching~\cite{lipman2022flow, albergo2022building, liu2022flow} provides a general perspective on generative modeling by defining a probability path between the noise distribution and the data distribution. It defines a forward process that linearly combines the data sample $\vx\sim p_{\mathrm{data}}(\vx)$ and the noise $\bm{\epsilon}\sim \mathcal{N}(0, \mI)$ into a noisy sample $\vz_t=a_t\vx+b_t\bm{\epsilon}$, where $a_t, b_t$ are noise schedules at time $t\in[0,1]$. Following rectified flow~\cite{liu2022flow}, we use a straight path with a linear schedule,
\begin{equation}\label{eq:forward}
    \vz_t = t\vx + (1-t)\bm{\epsilon},
\end{equation}
so $\vz_0=\bm{\epsilon}$ is pure noise and $\vz_1=\vx$ is the clean sample. The flow velocity is the time derivative of $\vz_t$,
\begin{equation}\label{eq:velocity}
    \vv = \frac{d\vz_t}{dt} = \vx-\bm{\epsilon}.
\end{equation}
Flow-based methods~\cite{esser2024scaling, lipman2022flow} train a model $\vv_\theta(\vz_t,t)$ to match $\vv$ via the velocity-matching loss
\begin{equation}
    \mathcal{L} = \E_{t,\vx,\bm{\epsilon}}\,\|\vv_\theta(\vz_t,t) - \vv\|^2.
\end{equation}
At inference, we draw clean samples by solving the ODE $d\vz_t = \vv_\theta(\vz_t,t)\,dt$ from $\vz_0\sim\mathcal{N}(0,\mI)$ at $t=0$ to $\vz_1$ at $t=1$. We use a 20-step Euler solver~\cite{esser2024scaling, euler1792institutiones} in this work.

\section{Method}
\begin{figure*}[t]
    \centering
    \includegraphics[width=0.9\linewidth]{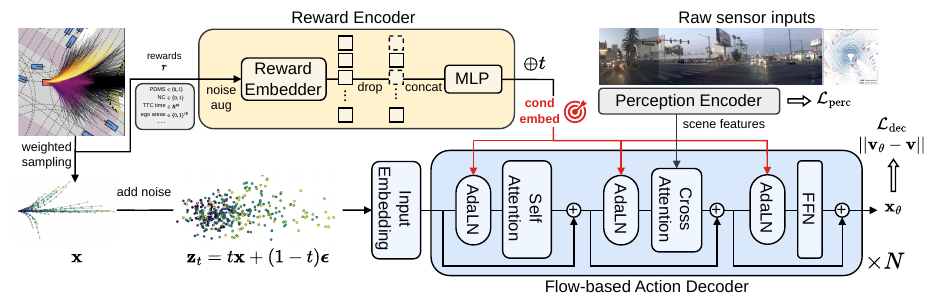}
    \caption{\ourmethod{} structure and training pipeline. We randomly sample action-reward pairs to produce noisy samples $\vz_t$. The reward encoder embeds rewards $r$ into a condition embedding. The flow-based action decoder predicts the clean sample based on scene features from the perception encoder and the condition injected via AdaLN, which is supervised by the velocity-matching loss. }
\label{fig:framework}
\end{figure*}

\ourmethod{} is a multimodal planning method that learns the reward-conditioned action distribution $p(a|r)$ (with scene context $s$ omitted from the condition for brevity). We construct fine-grained reward signals that capture safety, progress, comfort, and rule compliance, and pair them with a dense action vocabulary to form training samples $(a,r)$ that span the action space. The end-to-end model couples a perception encoder that produces scene features, a reward encoder that maps reward signals into a condition embedding, a flow-based action decoder that generates trajectory proposals, and a mode selector that picks the final output. The model is trained with a flow-matching objective on dense $(a,r)$ pairs and supports controllable test-time sampling via reward guidance and anchored sampling. Fig.~\ref{fig:framework} shows the model structure and training pipeline; Fig.~\ref{fig:inference} shows the inference pipeline.

\subsection{Reward Condition} \label{sec:reward}

\noindent \textbf{Dense Action-Reward Pairs.} Following scoring-based methods~\cite{chen2024vadv2, li2024hydra, li2025generalized}, we discretize the continuous action space into a dense action vocabulary $\mathcal{V}_a$ containing 8192 four-second trajectories clustered from 700K nuPlan trajectories~\cite{Caesar2021nuplan}. For every training scene, we simulate each vocabulary trajectory using the NAVSIM simulator~\cite{dauner2024navsim} and record the resulting reward labels, yielding dense action-reward pairs $(a,r)$ that span the action space and serve as training samples for our generative model of $p(a|r)$.

\noindent \textbf{Fine-grained Reward Signals.} NAVSIM evaluates a plan trajectory by closed-loop simulation and reports a PDM score that summarizes the overall quality~\cite{dauner2024navsim}. A single scalar score is too coarse for conditioning. For example, given a zero-score condition, the model cannot tell whether the action collides with another agent or simply stays stationary. We therefore expose the underlying submetrics, namely no at-fault collisions (NC), drivable area compliance (DAC), driving direction compliance (DDC), traffic light compliance (TLC), ego progress (EP), time to collision (TTC), lane keeping (LK), and history comfort (HC). These metrics cover safety (NC, TTC), rule compliance (DAC, DDC, TLC), progress (EP), and comfort (LK, HC).
Together with the PDM score, they form a fine-grained signal that enables the model to resolve the conditional action distribution with high fidelity.

\noindent \textbf{Balancing Hard and Soft Objectives.} Hard constraints and soft objectives conflict under naive high-reward conditioning. High-reward actions tend to sit on the decision boundary of binary hard constraints, which generative models smooth across in continuous action space. Naive conditioning on a high reward pushes the decoder toward aggressive proposals that sacrifice hard constraints to maximize progress. We confirm this empirically. Without our refinements, increasing the target reward at inference degrades overall performance rather than improving it (Fig.~\ref{fig:reward_ablation}).

We address this with two designs that strengthen hard constraints and soften continuous signals. First, we replace the binary TTC and DAC labels with per-timestep arrays. The TTC-time array stores the projected collision time at each future timestep within a max detection horizon, and the ego-area array~\cite{guo2025ipad} marks whether the ego is on-road and on-route at each timestep. Both arrays preserve the hard constraint at higher temporal resolution, providing a stronger conditioning signal than a single binary label. Second, we corrupt the continuous rewards (EP and PDM score) with minor Gaussian noise during training, treating them as random variables $\tilde{r}_k \sim \mathcal{N}(r_k, \sigma^2)$ with noise scale $\sigma$, which acts as label smoothing and prevents the decoder from over-relying on these signals. The final reward set is $\mathcal{R}=\{r_k\}$, comprising the two per-timestep arrays and the seven other scalar metrics.

\subsection{Model Architecture} \label{sec:model}

\noindent \textbf{Perception Encoder.} We adopt Transfuser~\cite{chitta2022transfuser} as the perception backbone to encode multi-view images and a bird's-eye-view (BEV) LiDAR feature map. We combine the backbone output with the encoded ego status and driving command to obtain a set of scene features $\vs$ comprising context tokens and agent tokens. $\vs$ is supervised with auxiliary losses on agent detection and BEV semantic segmentation. We also train an imitation learning head with GT trajectories, which provides an optional anchor at inference (Sec.~\ref{sec:inference}). Together these supervisions form the perception loss $\mathcal{L}_{\mathrm{perc}}$.

\noindent \textbf{Reward Encoder.} The reward encoder maps the heterogeneous reward signals in $\mathcal{R}$ into a single condition embedding $\vr_c$ that is fed to the action decoder. Each reward $r_k$ is independently mapped into a feature embedding $\vr_k^{\mathrm{emb}}$ by a per-reward embedder, and the embeddings are concatenated and passed through an MLP to produce $\vr_c$. 
To enable test-time classifier-free guidance~\cite{ho2022classifier} and conditioning on arbitrary subsets of rewards, we randomly replace each $\vr_k^{\mathrm{emb}}$ with a per-reward null token $\vn_k$ during training. Overall, the reward encoder is formulated as,
\begin{equation}\label{eq:reward}
    \begin{split}
        \vr_{k}^{\mathrm{emb}} &=   \begin{cases}
                    \vn_k, & \text{if drop } r_k \\
                    \text{Embed}(r_k),   & \text{otherwise} \\
                \end{cases} \\
        \vr_{c} &= \text{MLP}(\text{concat}[\vr_{k}^{\mathrm{emb}}]).
    \end{split}
\end{equation}
The continuous-reward noise augmentation introduced above is applied to $r_k$ before embedding.

\noindent \textbf{Flow-based Action Decoder.} The reward-conditioned formulation is agnostic to the choice of generative model. We adopt flow matching because it models continuous action distributions well and natively supports inference-time guidance, which enables reward-guided proposal sampling. We instantiate the action decoder as a flow-based generative model over the continuous action space. As described in Sec.~\ref{sec:pre}, it learns to recover clean trajectories $\vx$ from noisy inputs $\vz_t$ during training and generates action proposals through iterative denoising at inference. Following JiT~\cite{li2025back}, we adopt $\vx$-prediction, where the decoder predicts a clean sample $\vx_\theta(\vz_t, t, c)$ from the noisy input $\vz_t$, time $t$, and the condition $c=(\vs, \vr_c)$ comprising scene features and the reward embedding. Training uses the velocity-matching loss,
\begin{equation}
        \mathcal{L}_{\mathrm{dec}} = \E_{t,\vx,\bm{\epsilon}} \|\vv_{\theta}(\vz_t,t,c) - \vv\|^2,
\end{equation}
where the predicted $\vx_\theta$ is converted to velocity by $\vv_\theta = (\vx_\theta - \vz_t)/(1-t)$ based on Eqs.~\ref{eq:forward} and~\ref{eq:velocity}.

The decoder embeds the noisy trajectory $\vz_t$ into a sequence of tokens with sinusoidal positional encoding, then applies four stacked transformer blocks. Each block consists of self-attention across trajectory tokens, cross-attentions to the scene features $\vs$ (context and agent tokens), and a feed-forward layer, with residual connections around each module. The reward condition embedding is injected via adaptive layer normalization (AdaLN)~\cite{karras2019style, perez2018film, peebles2023scalable}, using the concatenation of $\vr_c$ and the time embedding $\vt$ as the modulation signal. The final block outputs the predicted clean sample $\vx_\theta(\vz_t, t, c)$.

\noindent \textbf{Mode Selector.} The mode selector ranks the proposals from the action decoder and returns the highest-scoring trajectory as the final output. Following scoring-based methods~\cite{li2024hydra,yao2026drivesuprim}, it is a lightweight two-layer transformer that attends to the scene features and predicts a set of NAVSIM subscores through shallow heads, which are aggregated into a single ranking score. We supervise the heads with a multi-head prediction loss $\mathcal{L}_{\mathrm{sel}}$ that matches the predicted subscores to their ground-truth values. See App.~\ref{app:selector} for full details.

\subsection{Training} \label{sec:training}

We train \ourmethod{} in two stages. The first stage trains the whole model end-to-end. The second stage finetunes the mode selector on proposals from the frozen decoder.

\noindent \textbf{Training the Action Decoder (Stage 1).} For each step, we sample action-reward pairs $(a,r)$ from $\mathcal{V}_a$ with weight inversely proportional to the score density, since vocabulary scores are heavily skewed toward zero and rare high-quality samples would otherwise be drowned out. The trajectory is corrupted by Gaussian noise at $t\sim\text{Uniform}(0,1)$, fed to the decoder with the reward condition, and supervised by $\mathcal{L}_{\mathrm{dec}}$. We train the full model jointly with the perception and selector losses as auxiliary objectives that refine the scene features,
\begin{equation}
        \mathcal{L}_{\mathrm{train}} = \mathcal{L}_{\mathrm{dec}} + w_{\mathrm{perc}}\mathcal{L}_{\mathrm{perc}} + w_{\mathrm{sel}}\mathcal{L}_{\mathrm{sel}}.
\end{equation}

\noindent \textbf{Training the Mode Selector (Stage 2).} The stage-1 selector sees only vocabulary trajectories, which differ from decoder proposals at inference. In stage-2 training, we close this gap by freezing all other components and optimizing the selector with $\mathcal{L}_{\mathrm{sel}}$ on online proposals labeled via simulation. Since the reward-guided decoder is biased toward high-quality samples, we mix in random vocabulary trajectories to keep the selector calibrated across the full quality range.

\begin{figure*}[t]
    \centering
    \includegraphics[width=\linewidth]{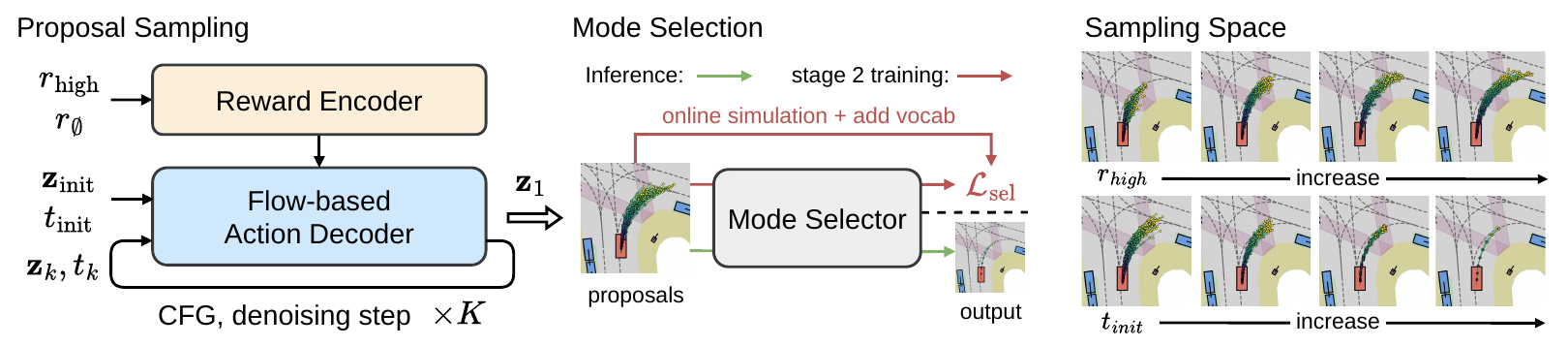}
    \caption{\ourmethod{} inference pipeline. (Left) The action decoder samples each proposal by denoising from a noisy anchor at $t_{\mathrm{init}}$ under CFG with the high-reward condition $r_{\mathrm{high}}$, together producing multiple proposal candidates. (Middle) The mode selector ranks proposals at inference and is supervised by simulated GT labels during stage-2 training. (Right) Sampling space spanned by $r_{\mathrm{high}}$ and $t_{\mathrm{init}}$. See Fig.~\ref{fig:sampling_space} for the full version.}
\label{fig:inference}
\end{figure*}

\subsection{Inference} \label{sec:inference}

At inference, \ourmethod{} samples action proposals by solving the ODE $d\vz_t = \vv_\theta(\vz_t,t,c)\,dt$ with the trained flow-based decoder. Built on the generative formulation, the decoder exposes a controllable test-time sampling interface over reward target and anchor noise level, instantiated via classifier-free guidance~\cite{ho2022classifier} and zero-shot editing~\cite{meng2021sdedit}. Fig.~\ref{fig:inference} overviews the inference pipeline.

\noindent \textbf{Reward Guidance.} We use classifier-free guidance (CFG)~\cite{ho2022classifier} to steer the decoder toward the high-reward region of $p(a|r)$,
\begin{equation}
    \vv_g = \vv_\theta(\vz_t,t,r_\emptyset) + w_g\big(\vv_\theta(\vz_t,t,r_{\mathrm{high}}) - \vv_\theta(\vz_t,t,r_\emptyset)\big),
\end{equation}
where $w_g$ is the CFG scale and $r_\emptyset$ is the empty condition obtained by setting every $\vr_k^{\mathrm{emb}}=\vn_k$ in Eq.~\ref{eq:reward}. The guided velocity $\vv_g$ replaces $\vv_\theta$ in the denoising step, so $w_g>1$ amplifies the reward direction and pulls the sample toward high-quality actions. We instantiate $r_{\mathrm{high}}$ on a subset of $\mathcal{R}$ (App.~\ref{app:cfg_subset}), and fix each reward entry to its maximal value except the target PDM score, which is left as a sampling control to be set by the strategy below. In what follows, sampling $r_{\mathrm{high}}$ refers to sampling this target score.

\noindent \textbf{Anchored Sampling.} The decoder also supports zero-shot anchored sampling~\cite{meng2021sdedit}. Given any trajectory $\vx_{\mathrm{anchor}}$ as the anchor and an initial denoising time $t_{\mathrm{init}}\in[0,1]$, we form a noisy sample by reusing the forward process of Eq.~\ref{eq:forward},
\begin{equation}
    \vz_{\mathrm{init}} = t_{\mathrm{init}}\vx_{\mathrm{anchor}} + (1-t_{\mathrm{init}})\bm{\epsilon},
\end{equation}
and start the denoising ODE from $\vz_{\mathrm{init}}$ at $t=t_{\mathrm{init}}$ instead of from pure noise. The remaining denoising steps run under the reward condition, biasing the output toward the anchor's coarse structure. Unlike anchor-based methods~\cite{liao2024diffusiondrive} that train on a fixed anchor set, our decoder accepts any trajectory as anchor at inference without additional training. Anchored sampling is optional, and the decoder generates high-quality proposals from pure noise as well. By default, we use the IL head output as the anchor.

\noindent \textbf{Sampling Strategy.} Reward guidance and anchored sampling together define a sampling space spanned by the target score and the initial noise level. The PDM score in $r_{\mathrm{high}}$ relates to the target progress level. The initial time $t_{\mathrm{init}}$ controls anchor adherence, ranging from $t_{\mathrm{init}}=0$ (pure-noise sampling) to $t_{\mathrm{init}}=1$ (output equal to the anchor). To produce diverse proposals, we sample both controls uniformly per proposal, drawing the target score from $\text{Uniform}(s_{\mathrm{min}},s_{\mathrm{max}})$ and $t_{\mathrm{init}}$ from $\text{Uniform}(t_{\mathrm{min}},t_{\mathrm{max}})$. The resulting candidates are scored by the mode selector, and the highest-scoring trajectory is returned as the final output.

\begin{table*}[t]
\caption{Results with closed-loop metrics on NAVSIM v1 \texttt{navtest} benchmark. The performance of our method is averaged over three inferences. Following recent works~\cite{zou2025diffusiondrivev2}, we compare against methods using lightweight image backbones ResNet-34~\cite{he2016deep} and V2-99~\cite{lee2019energy}. Methods are grouped by single-proposal and multi-proposal. For our single-proposal entry, $r_{\mathrm{high}}$ and $t_{\mathrm{init}}$ are fixed to representative values rather than randomly sampled (App.~\ref{app:inference_per_experiment}).}
\centering
\resizebox{1.0\textwidth}{!}
{
 \begin{tabular}{lccccccc>{\columncolor[gray]{0.9}}c}
    \toprule
    Method & \# Proposals & Img. Backbone & NC $\uparrow$ &DAC $\uparrow$ & TTC $\uparrow$& Comf. $\uparrow$ & EP $\uparrow$ &  \textbf{PDMS} $\uparrow$  \\
    \midrule
    UniAD~\cite{hu2023planning} & 1 & ResNet-34  & 97.8 & 91.9 & 92.9 & 100 & 78.8 & 83.4 \\
    Transfuser~\cite{chitta2022transfuser} & 1 & ResNet-34  & 97.7 & 92.8 & 92.8 & 100 & 79.2 & 84.0 \\
    PARA-Drive~\cite{weng2024paradrive} & 1 & ResNet-34  & 97.9 & 92.4 & 93.0 & 99.8 & 79.3 & 84.0 \\
    DRAMA~\cite{yuan2024drama} & 1 & ResNet-34 & 98.0 & 93.1 & 94.8 & 100 & 80.1 & 85.5 \\
    ARTEMIS~\cite{feng2025artemis} & 1 & ResNet-34 & 98.3 & 95.1 & 94.3 & 100 & 81.4 & 87.0 \\
    \midrule
    \ourmethod{} (Ours) & 1 & ResNet-34  & \textbf{98.6} & \textbf{97.3} & \textbf{95.3} &  100 &  \textbf{84.9} & \textbf{90.0} \\
    \midrule
    VADv2~\cite{chen2024vadv2} & 8192 & ResNet-34 & 97.2 & 89.1 & 91.6 & 100 & 76.0 & 80.9 \\
    Hydra-MDP~\cite{li2024hydra} & 8192 & ResNet-34 & 98.3 & 96.0 & 94.6 & 100 & 78.7 & 86.5 \\
    Hydra-MDP++~\cite{li2025hydra} & 8192 & ResNet-34 & 97.6 & 96.0 & 93.1 & 100 & 80.4 & 86.6 \\
    Hydra-MDP~\cite{li2024hydra} & 8192 & V2-99 & 98.4 & 97.8 & 93.9 & 100 & 86.5 & 90.3 \\
    Hydra-MDP++~\cite{li2025hydra} & 8192 & V2-99 & 98.6 & \textbf{98.6} & 95.1 & 100 & 85.7 & 91.0 \\
    DriveSuprim~\cite{yao2026drivesuprim} & 8192 & ResNet-34 & 97.8 & 97.3 & 93.6 & 100 & 86.7 & 89.9 \\
    DiffusionDrive~\cite{liao2024diffusiondrive} & 20 & ResNet-34 & 98.2  & 96.2  & 94.7  & 100  & 82.2  & 88.1 \\
    WoTE~\cite{li2025end} & 256 & ResNet-34 & 98.5 & 96.8 & 94.9 & 99.9 & 81.9 & 88.3 \\
    GoalFlow~\cite{xing2025goalflow} & 256 & V2-99 & 98.4 & 98.3 & 94.6 & 100 & 85.0 & 90.3 \\
    DiffusionDriveV2~\cite{zou2025diffusiondrivev2} & 800 & ResNet-34 & 98.3  & 97.9  & 94.8  & 99.9 & 87.5  & 91.2 \\
    iPad~\cite{guo2025ipad} & 64 & ResNet-34  & 98.6 & 98.3 & 94.9 &  100 &  88.0 & 91.7 \\
    \midrule
    \ourmethod{} (Ours) & 60 & ResNet-34  & \textbf{98.8} & 98.0 & \textbf{96.0} &  100 &  \textbf{90.1} & \textbf{92.8} \\
    \bottomrule
\end{tabular}}
\label{tab:navsim}
\end{table*}

\begin{table*}[tb]
\centering
\caption{Results with closed-loop metrics on NAVSIM v2 \texttt{navtest} benchmark. The performance of our method is averaged over three inferences. Following recent works~\cite{zou2025diffusiondrivev2}, we compare against methods using lightweight image backbones ResNet-34~\cite{he2016deep} and V2-99~\cite{lee2019energy}.}
\resizebox{\textwidth}{!}{
\begin{tabular}{lccccccccccc}
    \toprule
    Method 
    & Img. Backbone
    & NC $\uparrow$ 
    & DAC $\uparrow$ 
    & DDC $\uparrow$ 
    & TLC $\uparrow$ 
    & EP $\uparrow$ 
    & TTC $\uparrow$ 
    & LK $\uparrow$ 
    & HC $\uparrow$ 
    & EC $\uparrow$ 
    & \cellcolor{gray!20}\textbf{EPDMS} $\uparrow$ \\
    \midrule
    Ego Status MLP 
        & ResNet-34
        & 93.1 & 77.9 & 92.7 & 99.6 & 86.0 & 91.5 & 89.4 & 98.3 & 85.4 
        & \cellcolor{gray!20}64.0 \\
    Transfuser~\cite{chitta2022transfuser}
        & ResNet-34
        & 96.9 & 89.9 & 97.8 & 99.7 & 87.1 & 95.4 & 92.7 & 98.3 & \textbf{87.2}
        & \cellcolor{gray!20}76.7 \\ 
    Hydra-MDP++~\cite{li2025hydra}
        & ResNet-34
        & 97.2 & 97.5 & 99.4 & 99.6 & 83.1 & 96.5 & 94.4 & 98.2 & 70.9 
        & \cellcolor{gray!20}81.4 \\
    DriveSuprim~\cite{yao2026drivesuprim}
        & ResNet-34
        & 97.5 & 96.5 & 99.4 & 99.6 & 88.4 & 96.6 & 95.5 & 98.3 & 77.0 
        & \cellcolor{gray!20}83.1 \\
    ARTEMIS~\cite{feng2025artemis}
        & ResNet-34
        & 98.3 & 95.1 & 98.6 & 99.8 & 81.5 & 97.4 & 96.5 & 98.3 & -
        & \cellcolor{gray!20}83.1 \\
    \midrule
    Hydra-MDP++~\cite{li2025hydra} 
        & V2-99
        & 98.4 & 98.0 & 99.4 & 99.8 & 87.5 & 97.7 & 95.3 & 98.3 & 77.4 
        & \cellcolor{gray!20}85.1 \\
    DriveSuprim~\cite{yao2026drivesuprim}
        & V2-99
        & 97.8 & 97.9 & \textbf{99.5} & \textbf{99.9} & 90.6 & 97.1 & \textbf{96.6} & 98.3 & 77.9 
        & \cellcolor{gray!20}86.0 \\
    \midrule
    \ourmethod{} (Ours) 
        & ResNet-34
        & \textbf{98.9} & \textbf{98.1} & 99.1 & 99.7 & \textbf{91.5} 
        & \textbf{98.5} & 95.0 & 98.3 & 65.2
        & \cellcolor{gray!20}\textbf{88.9} \\
    \bottomrule
\end{tabular}
}
\label{tab:navsim_v2}
\end{table*}

\section{Experiments}

We evaluate \ourmethod{} on the NAVSIM v1~\cite{dauner2024navsim} and v2~\cite{cao2025pseudo} benchmarks against scoring-based and anchor-based multimodal planners to validate that learning the reward-conditioned action distribution $p(a|r)$ yields both higher overall performance and substantially better proposal quality.

\noindent \textbf{Dataset.}
NAVSIM~\cite{dauner2024navsim} is built upon OpenScene~\cite{openscene2023}, a compact redistribution of the nuPlan dataset~\cite{Caesar2021nuplan}, and curates real-world non-trivial driving scenes where the future plan cannot be directly inferred from history. NAVSIM is split into \texttt{navtrain}, containing 103k training frames, and \texttt{navtest}, containing 12k evaluation frames.

\noindent \textbf{Metrics.} NAVSIM~\cite{dauner2024navsim,cao2025pseudo} computes closed-loop metrics for each planned trajectory by log-replay simulation. For NAVSIM-v1~\cite{dauner2024navsim}, metrics include no at-fault collisions (NC), driving area compliance (DAC), time-to-collision (TTC), comfort, and ego progress (EP), aggregated into the Predictive Driver Model Score (PDMS). NAVSIM-v2~\cite{cao2025pseudo} further improves the evaluation using Extended PDMS (EPDMS), adding submetrics including driving direction compliance (DDC), traffic light compliance (TLC), lane keeping (LK), and extended comfort (EC).

\noindent \textbf{Implementation Details.}
The perception backbone takes as input a front-view image stitched from the front, left, and right cameras together with a rasterized 2D BEV LiDAR feature map aggregating 4 recent frames for temporal context. We train end-to-end on \texttt{navtrain} for 100 epochs using AdamW ($\text{lr}=3\times10^{-4}$, cosine annealing to $10^{-6}$), with a total batch size of 64 across 4 NVIDIA H20 GPUs. The mode selector is trained for an additional two epochs in the second stage. At inference, we perform 20 denoising steps with CFG scale $w_g=5$ and sample target score $r_{\mathrm{high}} \in [0.9, 1.0]$, initial denoising time $t_{\mathrm{init}} \in [0.5, 0.9]$, generating 60 proposals by default. See the appendix for full details.

\subsection{Main Results}
\noindent \textbf{Results on NAVSIM-v1.} Tab.~\ref{tab:navsim} compares \ourmethod{} against existing methods on the NAVSIM-v1~\cite{dauner2024navsim} \texttt{navtest} split. \ourmethod{} achieves state-of-the-art 92.8 PDMS, outperforming all prior methods by $\geq$1.1 PDMS, with margins of $\geq$0.9 on TTC and $\geq$2.1 on EP. We highlight that our method leads in both safety (NC, TTC) and progress (EP) metrics, which are typically in tension.
Even when sampling a single proposal, \ourmethod{} attains on par or better performance on safety metrics (NC, TTC) among multimodal methods, indicating consistently feasible proposals. We analyze this further in Sec.~\ref{sec:proposal_quality}.

\noindent \textbf{Results on NAVSIM-v2.} On the NAVSIM-v2~\cite{cao2025pseudo} \texttt{navtest} split (Tab.~\ref{tab:navsim_v2}), \ourmethod{} again leads with 88.9 EPDMS and the best scores on safety and progress submetrics (NC, TTC, EP). One submetric where \ourmethod{} underperforms is extended comfort (EC), which measures dynamic consistency across consecutive frames. Since our action decoder naturally produces multimodal proposals, enforcing such inter-frame consistency is the role of the mode selector, while our current selector scores each frame independently without modeling temporal correlation.

\begin{table}[t]
\centering
\caption{Quantitative comparison of proposal quality. We report PDMS, EP, and TTC on \texttt{navtest} when selecting among different numbers of proposal candidates. Mean and standard deviation of all proposals' scores are reported to reflect the average quality of proposals. Ours generates 64 proposals in total to match iPad~\cite{guo2025ipad}.}
\label{tab:proposal_quant}
\small
\resizebox{\textwidth}{!}{
\begin{tabular}{l ccc ccc ccc ccc}
\toprule
& \multicolumn{3}{c}{1 Proposal} & \multicolumn{3}{c}{4 Proposals} & \multicolumn{3}{c}{All Proposals} & \multicolumn{3}{c}{Mean over All Proposals} \\
\cmidrule(lr){2-4} \cmidrule(lr){5-7} \cmidrule(lr){8-10} \cmidrule(lr){11-13}
Method & PDMS & EP & TTC & PDMS & EP & TTC & PDMS & EP & TTC & PDMS & EP & TTC \\
\midrule
DiffusionDrive~\cite{liao2024diffusiondrive} & 69.7 & 65.9 & 78.6 & 85.3 & 79.0 & 93.6 & 88.1 & 82.2 & 94.7 & 60.3\tiny$\pm$33.1 & 57.5\tiny$\pm$31.6 & 73.1\tiny$\pm$28.4 \\
iPad~\cite{guo2025ipad} & 83.3 & 69.4 & \textbf{95.7} & 89.7 & 84.8 & 94.6 & 91.5 & 87.8 & 94.8 & 77.9\tiny$\pm$22.4 & 73.5\tiny$\pm$22.1 & 86.4\tiny$\pm$19.5 \\
\ourmethod{} (Ours) & \textbf{89.4} & \textbf{84.2} & 94.7 & \textbf{91.8} & \textbf{87.9} & \textbf{95.7} & \textbf{92.9} & \textbf{90.1} & \textbf{96.0} & \textbf{89.4}\tiny$\pm$7.4 & \textbf{84.1}\tiny$\pm$8.5 & \textbf{94.9}\tiny$\pm$6.1 \\
\bottomrule
\end{tabular}
}
\end{table}

\begin{figure*}[t]
\centering
\begin{minipage}[t]{0.48\textwidth}
    \centering
    \includegraphics[width=\textwidth]{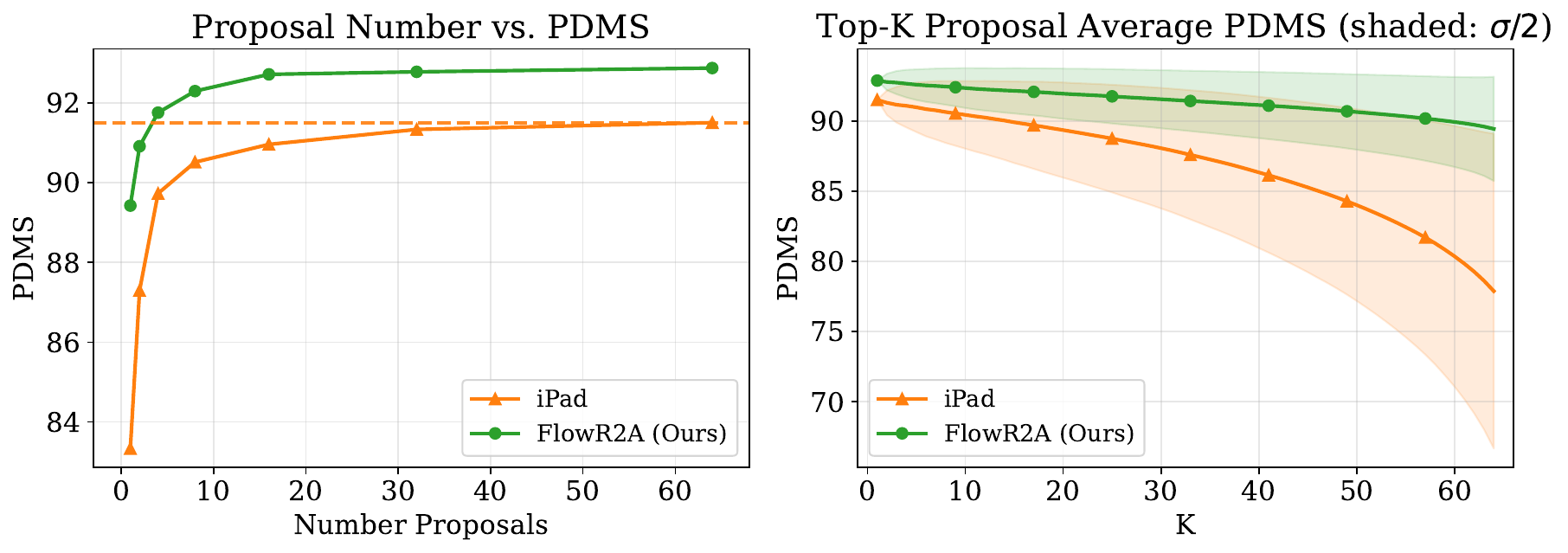}
    \caption{Comparing proposal quality with iPad~\cite{guo2025ipad}.
    (Left) PDMS under different proposal numbers. (Right) Average score of top-k proposals. Here $\sigma$ indicates standard deviation.
    \label{fig:topk_compare}}
\end{minipage}
\hfill
\begin{minipage}[t]{0.48\textwidth}
    \centering
    \includegraphics[width=\textwidth]{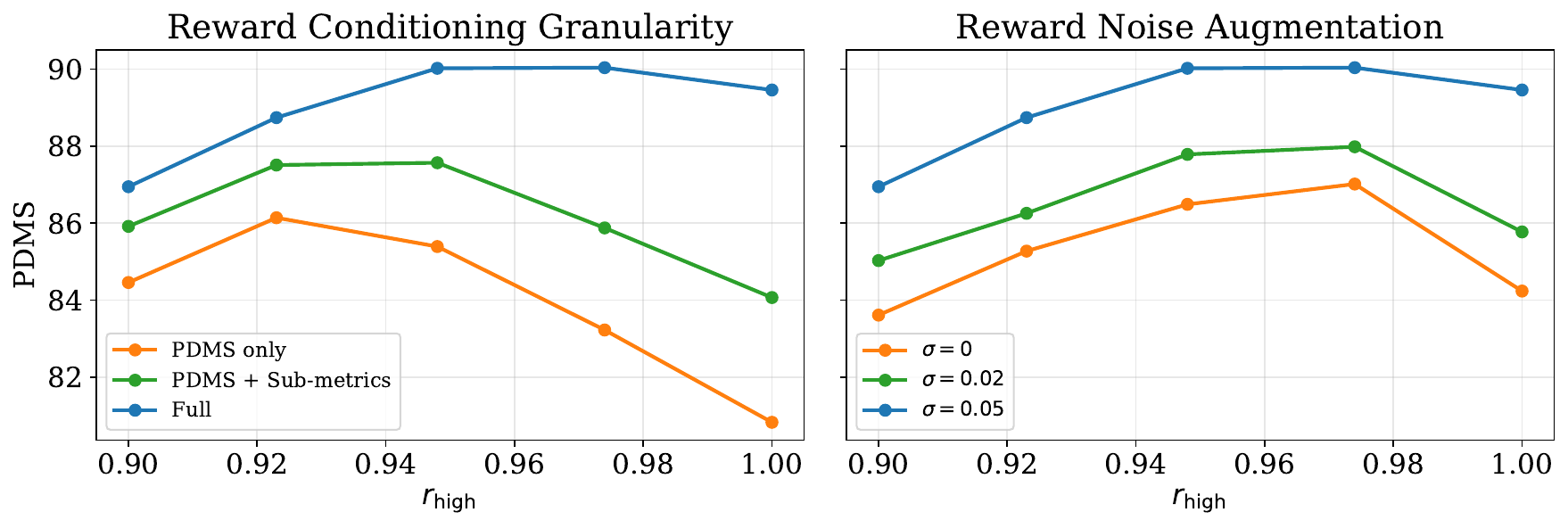}
    \caption{Ablation on reward conditioning. Evaluated on single proposal under different $r_{\mathrm{high}}$. (Left) Reward condition granularity effect. (Right) Reward noise augmentation effect. \label{fig:reward_ablation}}
\end{minipage}
\end{figure*}

\subsection{Proposal Quality}
\label{sec:proposal_quality}

A core promise of learning $p(a|r)$ from dense supervision is that each sampled proposal should fall within the feasible action distribution, not just the one selected by the scorer. We test this claim against the strong anchor-based baselines, DiffusionDrive~\cite{liao2024diffusiondrive} and iPad~\cite{guo2025ipad}.

\noindent \textbf{Quantitative Comparison.}
Tab.~\ref{tab:proposal_quant} reports the performance across different proposal counts, and the mean$\pm$std over all proposals.
We highlight two observations.
First, \ourmethod{} reaches strong performance with very few proposals, surpassing iPad's full PDMS (64 proposals) within 4 proposals.
Second, the gap is most pronounced when averaging performance over all proposals. \ourmethod{} obtains an average proposal PDMS exceeding iPad by +11.5 and DiffusionDrive by +29.1, with significantly lower standard deviation, indicating that our decoder produces consistent on-distribution proposals rather than scattered candidates.
Fig.~\ref{fig:topk_compare} visualizes this trend across $K$ proposals. \ourmethod{} dominates iPad in PDMS at every generated proposal count (left) and in top-$K$ average proposal score (right), with a markedly tighter spread.
In addition, the 1-proposal column already matches the full system on safety, and the scorer contributes primarily to progress, showing that hard constraints are absorbed into $p(a|r)$ at sampling time, while selection only resolves soft trade-offs.

\begin{table}[t]
\centering
\caption{Diversity and worst-case proposal quality on \texttt{navtest}. Diversity follows the metric of DiffusionDriveV2~\cite{zou2025diffusiondrivev2}, computed over all proposals and over proposals with PDMS$>$0. Worst-case metrics are evaluated on the lowest-PDMS proposal of each scene. $^\dagger$Number taken from DiffusionDriveV2~\cite{zou2025diffusiondrivev2}.}
\label{tab:diversity_worst}
\small
\resizebox{\textwidth}{!}{
\begin{tabular}{l cc cccccc}
\toprule
& \multicolumn{2}{c}{Diversity} & \multicolumn{6}{c}{Worst-Case Proposal} \\
\cmidrule(lr){2-3} \cmidrule(lr){4-9}
Method & All $\uparrow$ & PDMS$>$0 $\uparrow$ & NC $\uparrow$ & DAC $\uparrow$ & TTC $\uparrow$ & Comf. $\uparrow$ & EP $\uparrow$ & PDMS $\uparrow$ \\
\midrule
Diffusion Policy$^\dagger$ & 0.001 & 0.001 & -- & -- & -- & -- & -- & -- \\
DiffusionDrive~\cite{liao2024diffusiondrive} & \textbf{0.419} & \textbf{0.328} & 54.4 & 52.8 & 41.0 & 79.5 & 8.9 & 7.5 \\
DiffusionDriveV2~\cite{zou2025diffusiondrivev2} & 0.385 & 0.279 & 52.9 & 39.3 & 43.9 & 76.0 & 0.6 & 0.4 \\
iPad~\cite{guo2025ipad} & 0.144 & 0.135 & 65.0 & 63.7 & 49.7 & 81.1 & 22.5 & 20.2 \\
\ourmethod{} (Ours) & 0.239 & 0.239 & \textbf{93.6} & \textbf{89.6} & \textbf{79.8} & \textbf{99.5} & \textbf{63.9} & \textbf{69.3} \\
\bottomrule
\end{tabular}
}
\end{table}

\noindent \textbf{Diversity and Worst-Case Quality.}
Higher average quality could come from proposals collapsing around the GT. Tab.~\ref{tab:diversity_worst} rules this out. \ourmethod{} keeps reasonable diversity, far above a naive diffusion policy, and retains all of it after filtering out unsafe proposals (PDMS$>$0). In contrast, the diversity of DiffusionDrive and DiffusionDriveV2 drops substantially, since part of it comes from degenerate proposals.
The worst proposal per scene shows the gap more clearly. Baselines degrade to near-zero worst-case PDMS, while \ourmethod{} raises it from 20.2 (iPad) to 69.3, with consistent gains on every submetric. Its remaining low-quality proposals are mostly low-progress rather than unsafe.

\newcommand{\vizimg}[1]{\includegraphics[trim={0 600 0 0},clip,width=0.225\linewidth]{#1}}
\newcommand{\methodlabel}[1]{\rotatebox{90}{\parbox{1.9cm}{\centering\small\textbf{#1}}}}

\begin{figure*}[t]
\centering
\setlength{\tabcolsep}{1pt}
\renewcommand{\arraystretch}{0.4}
\begin{tabular}{@{}c cccc@{}}
\methodlabel{Diffusion-\\Drive~\cite{liao2024diffusiondrive}} &
\vizimg{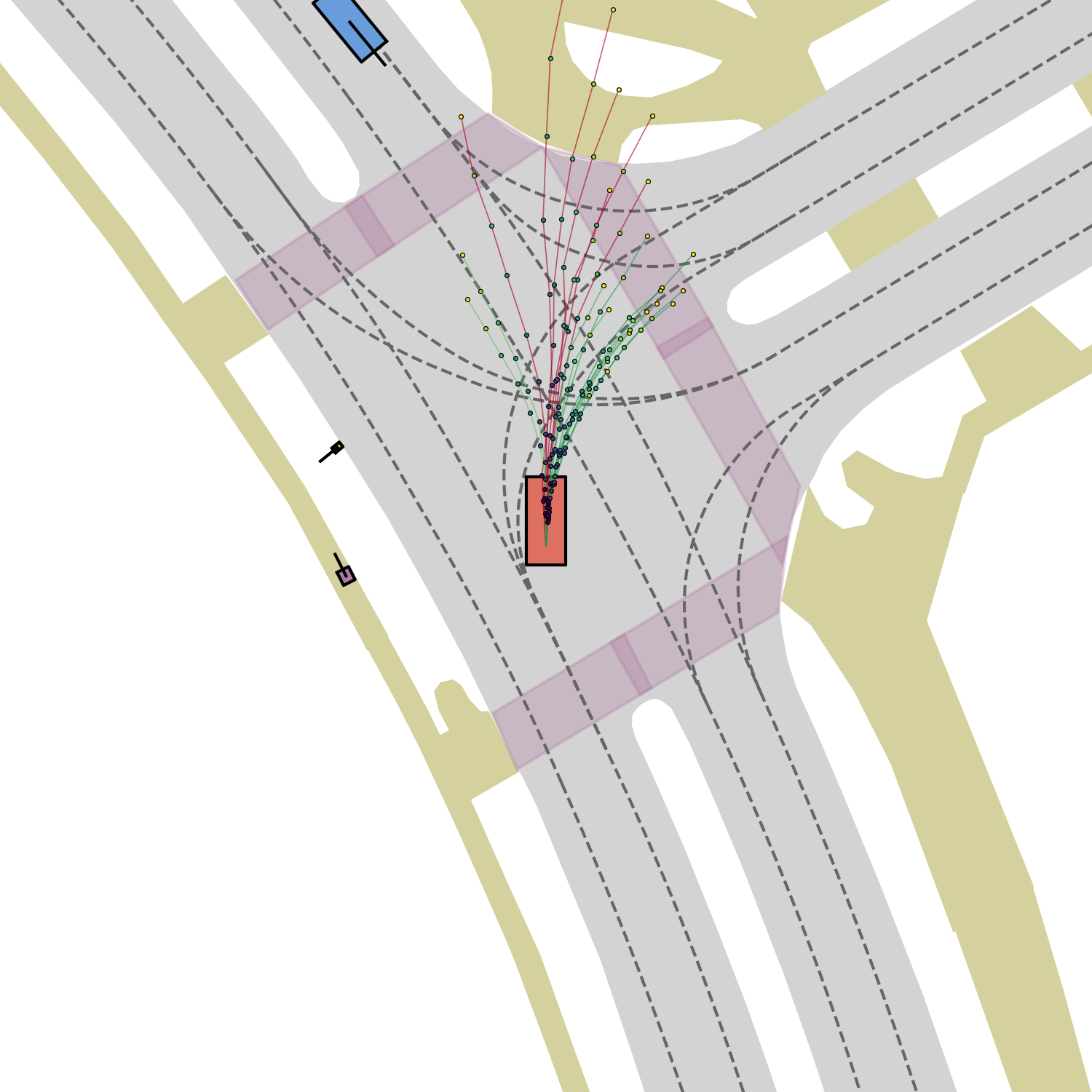} &
\vizimg{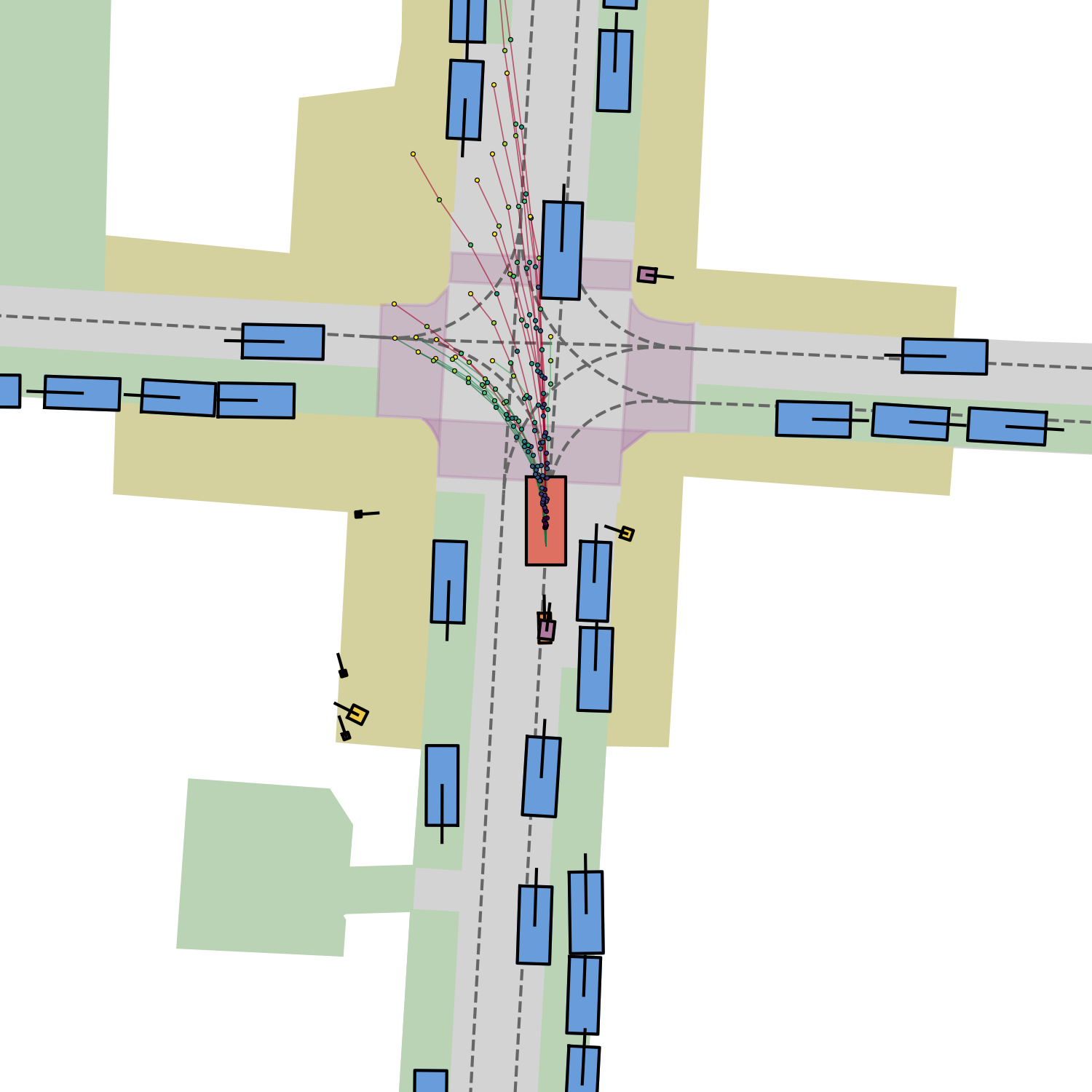} &
\vizimg{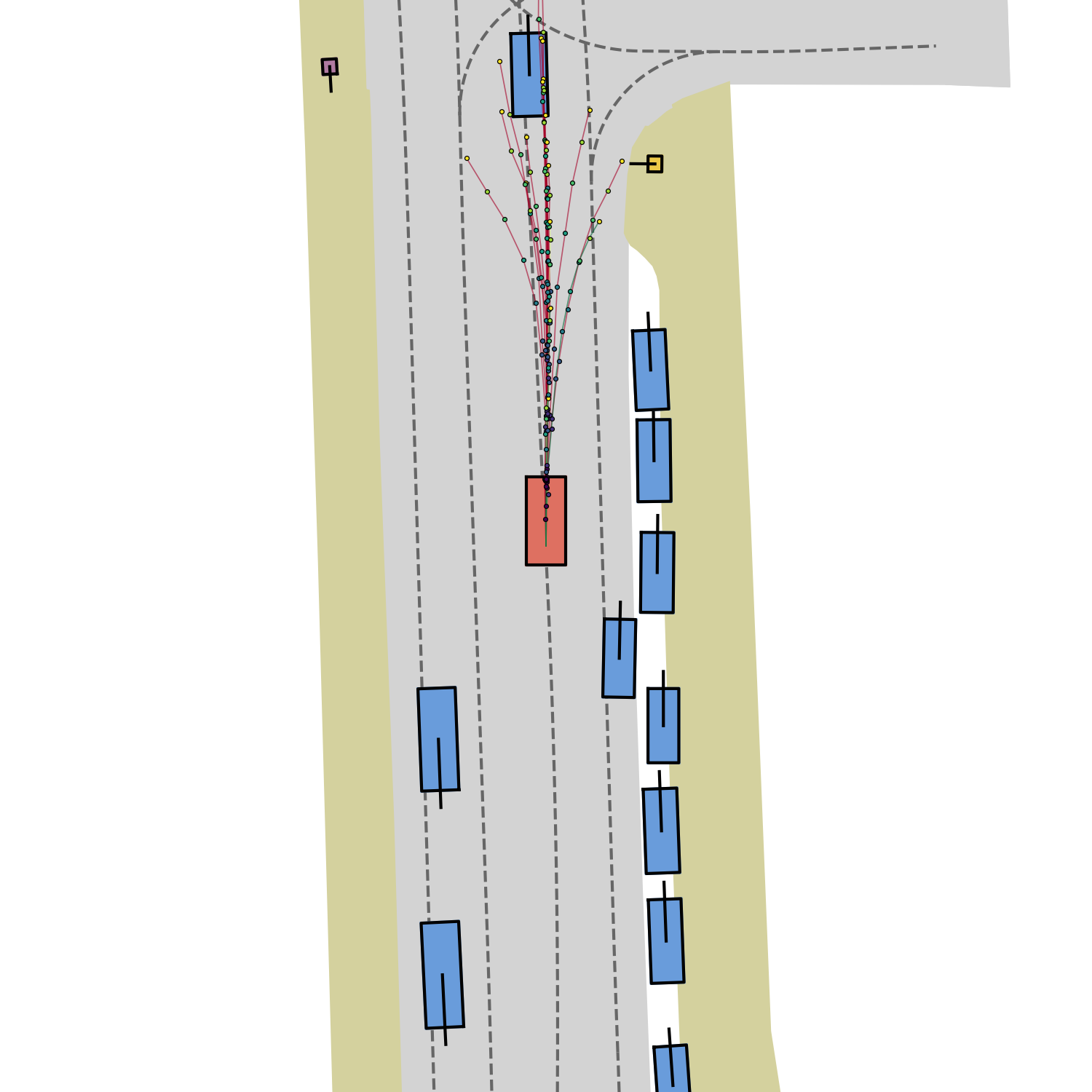} &
\vizimg{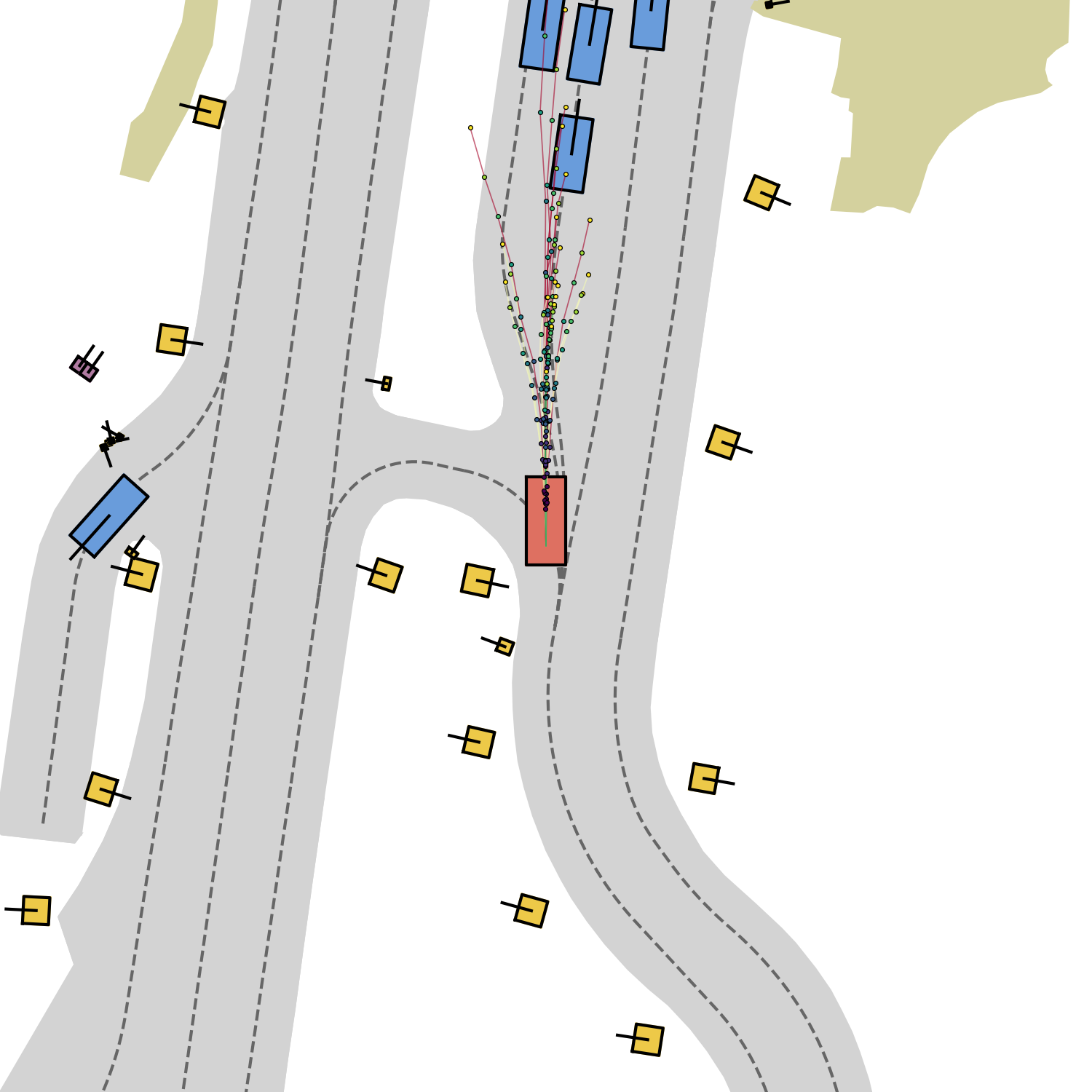}\\
\methodlabel{Diffusion-\\DriveV2~\cite{zou2025diffusiondrivev2}} &
\vizimg{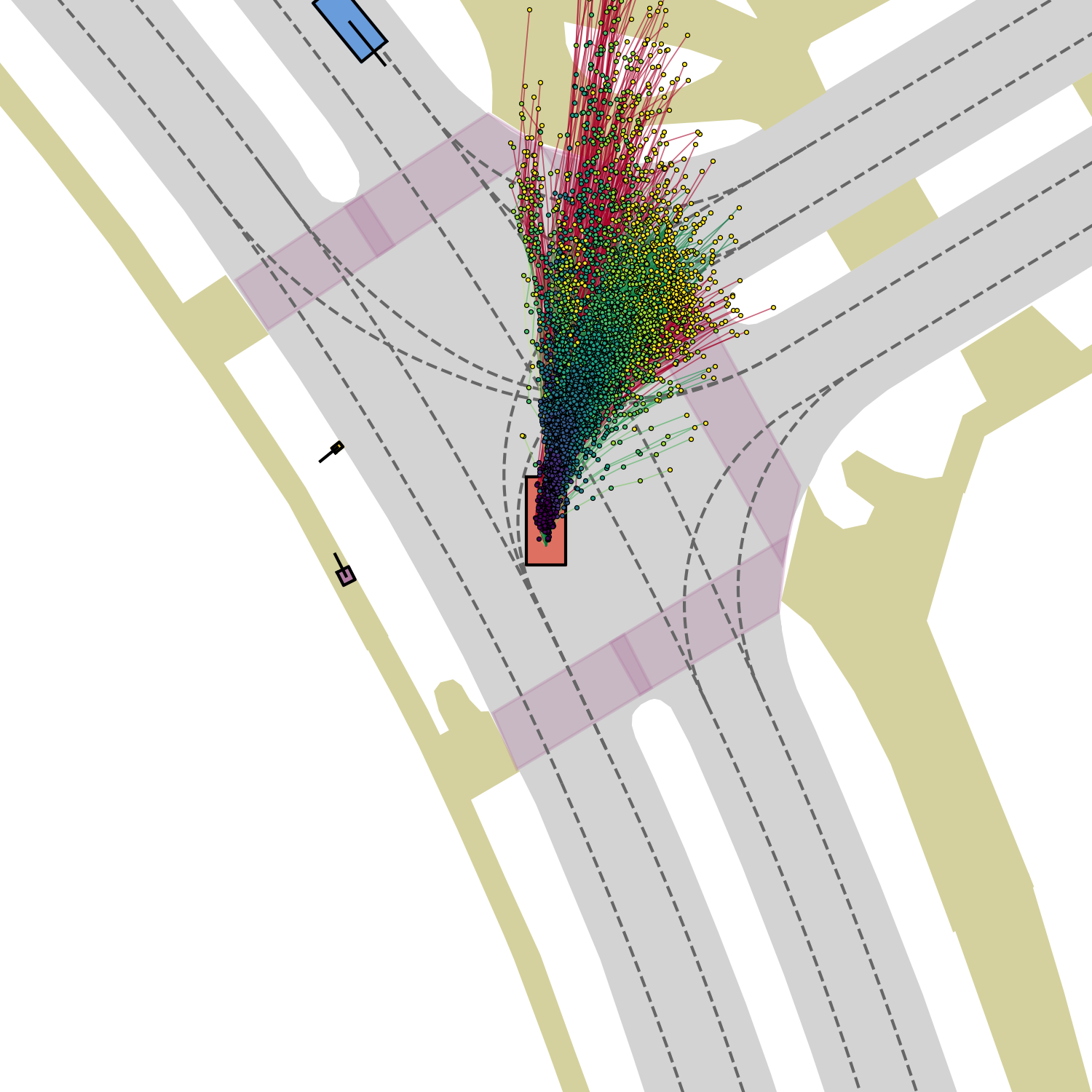} &
\vizimg{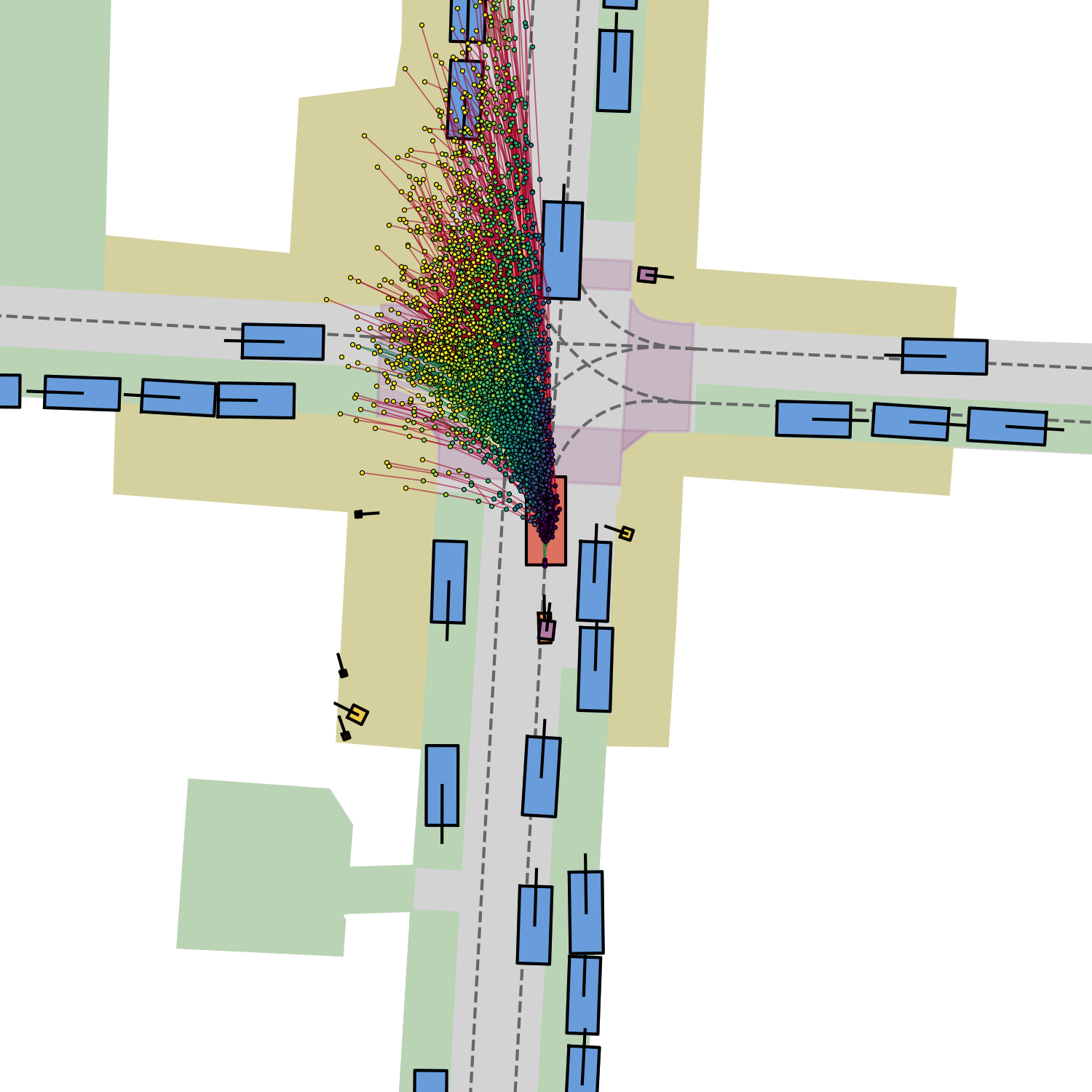} &
\vizimg{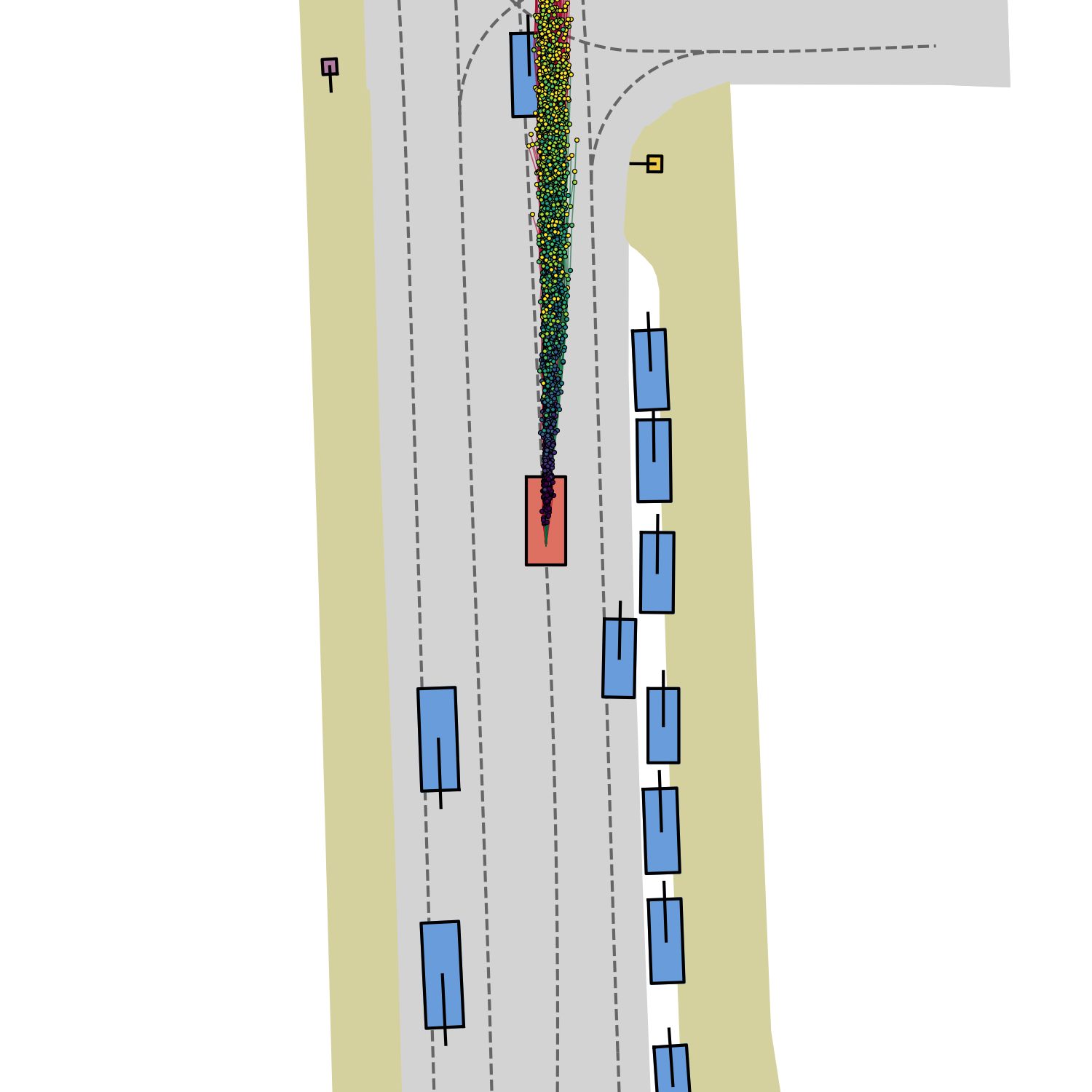} &
\vizimg{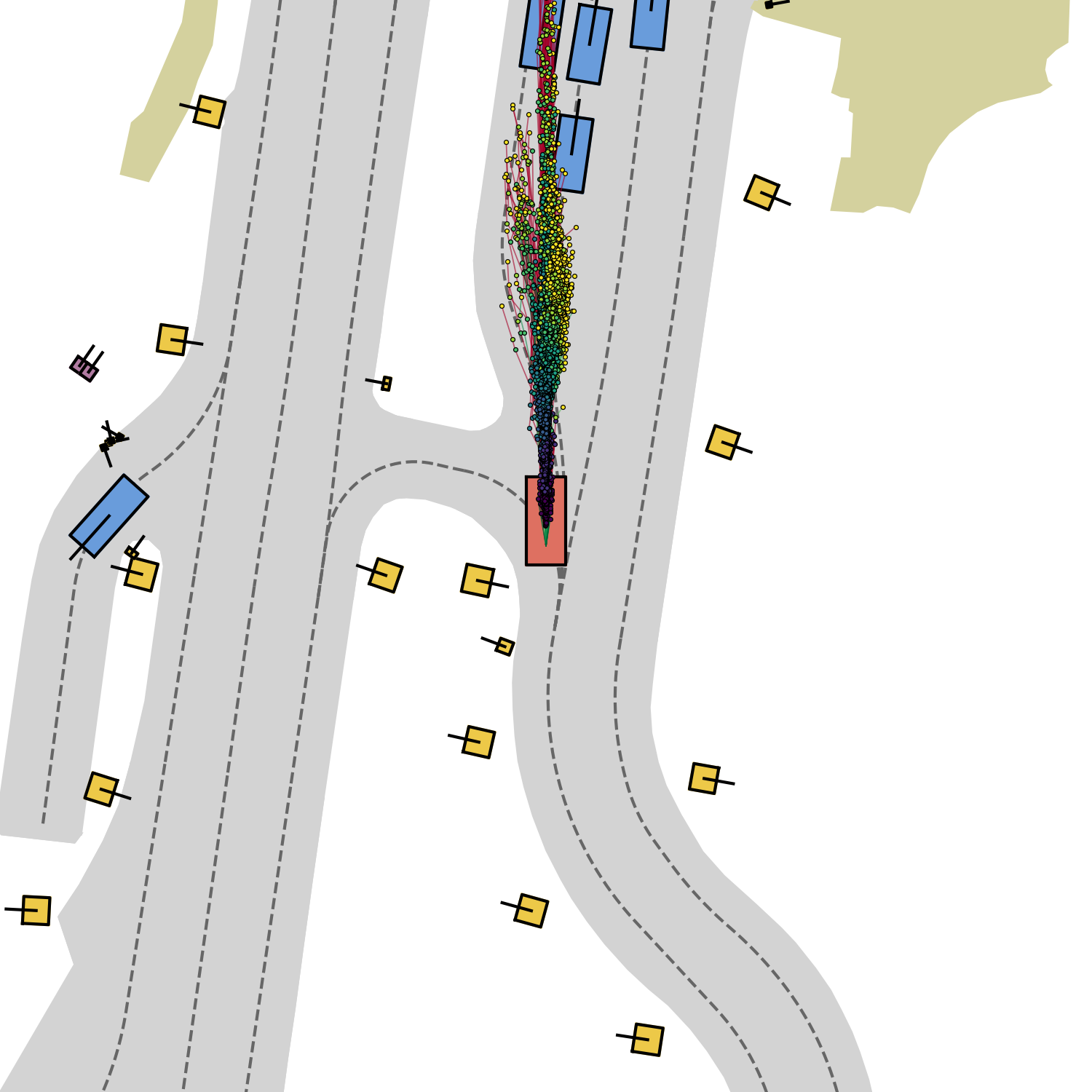}\\
\methodlabel{iPad~\cite{guo2025ipad}} &
\vizimg{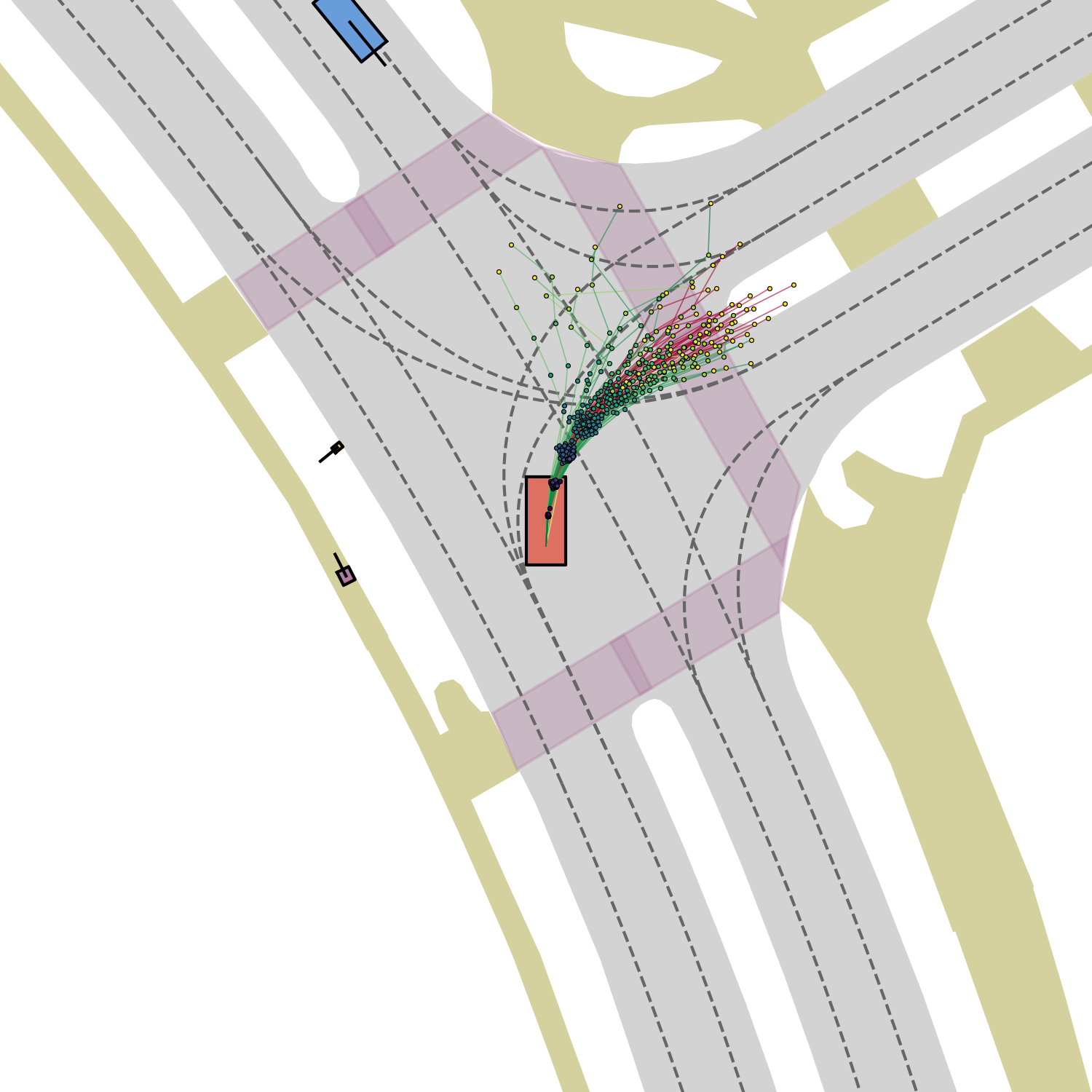} &
\vizimg{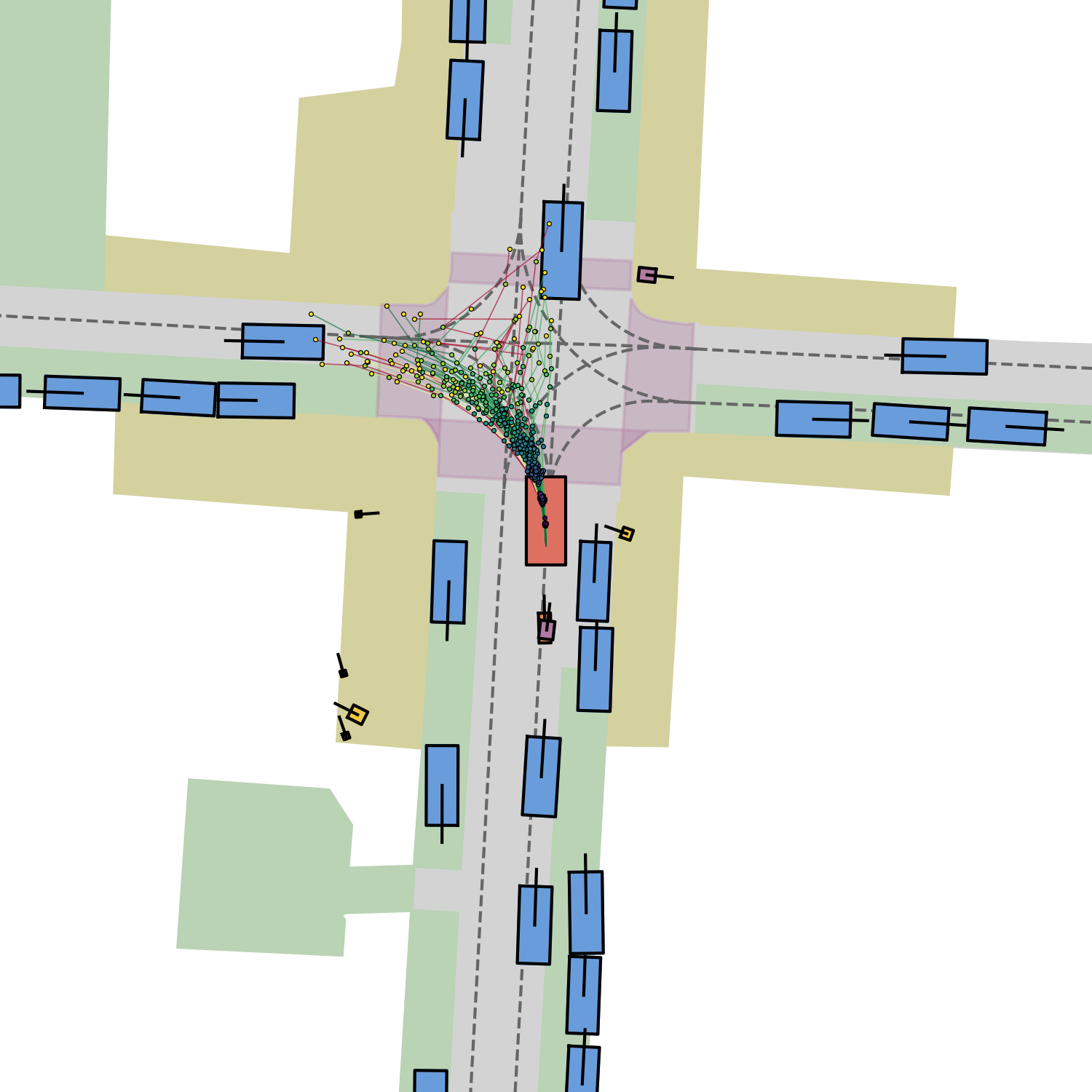} &
\vizimg{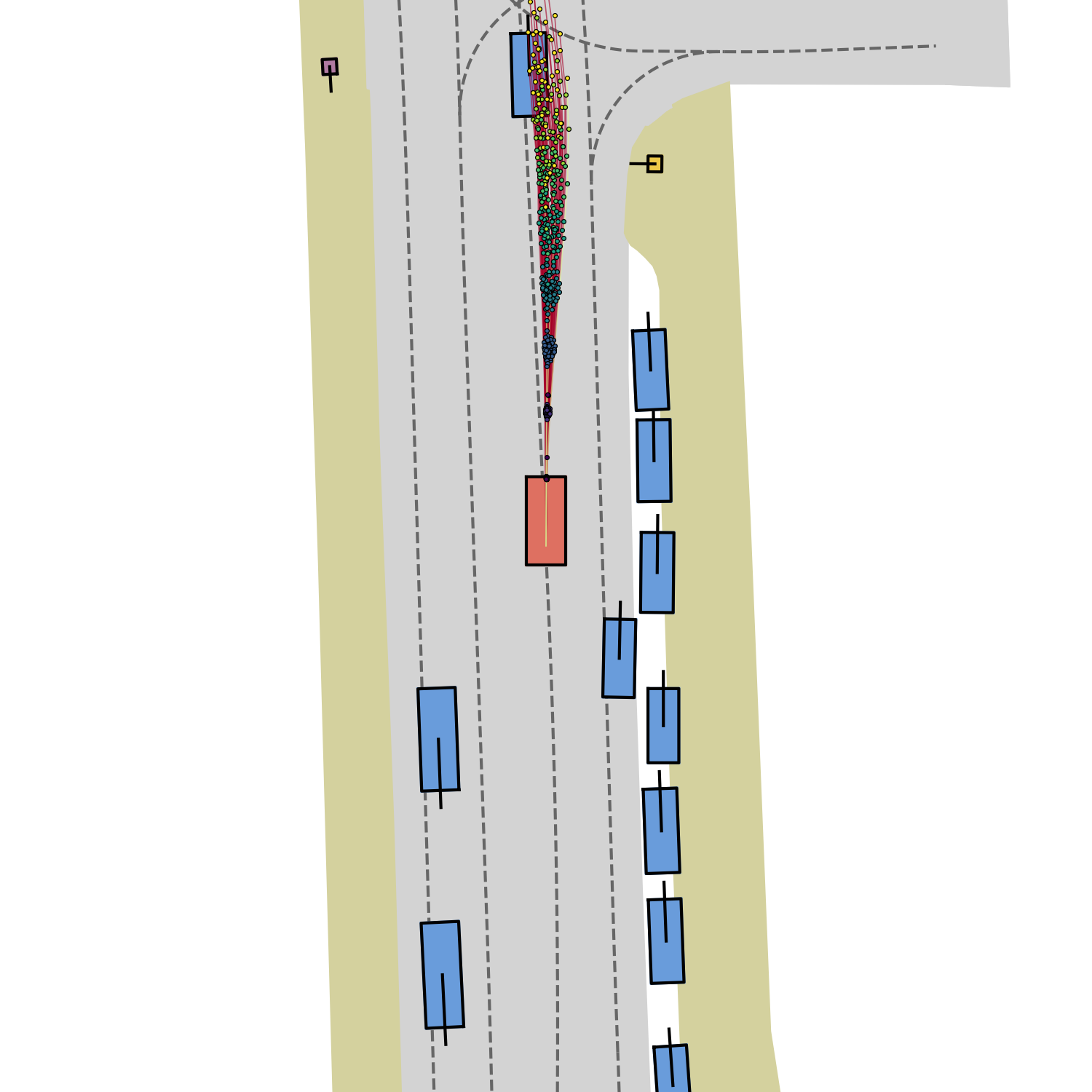} &
\vizimg{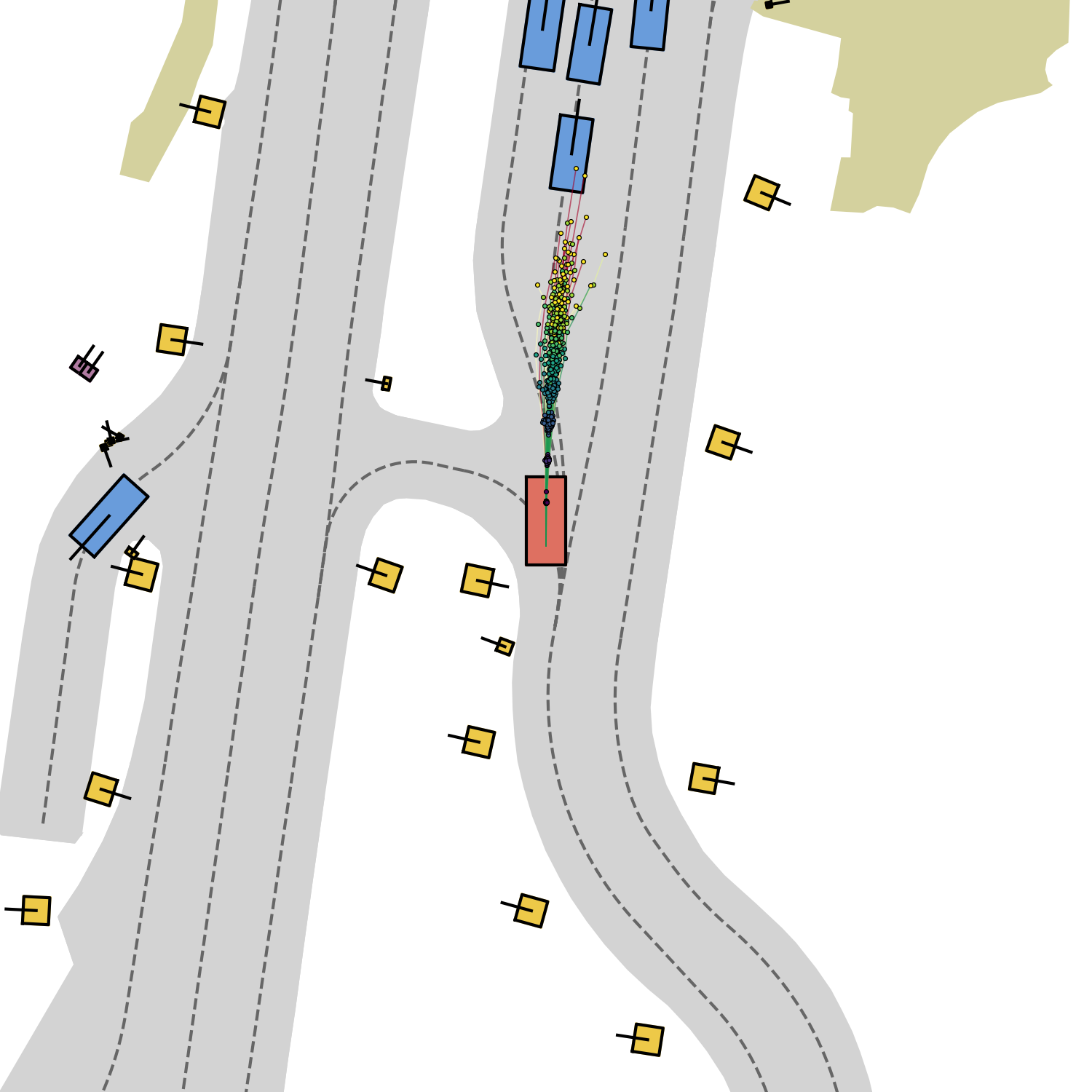}\\
\methodlabel{\ourmethod{} (Ours)} &
\vizimg{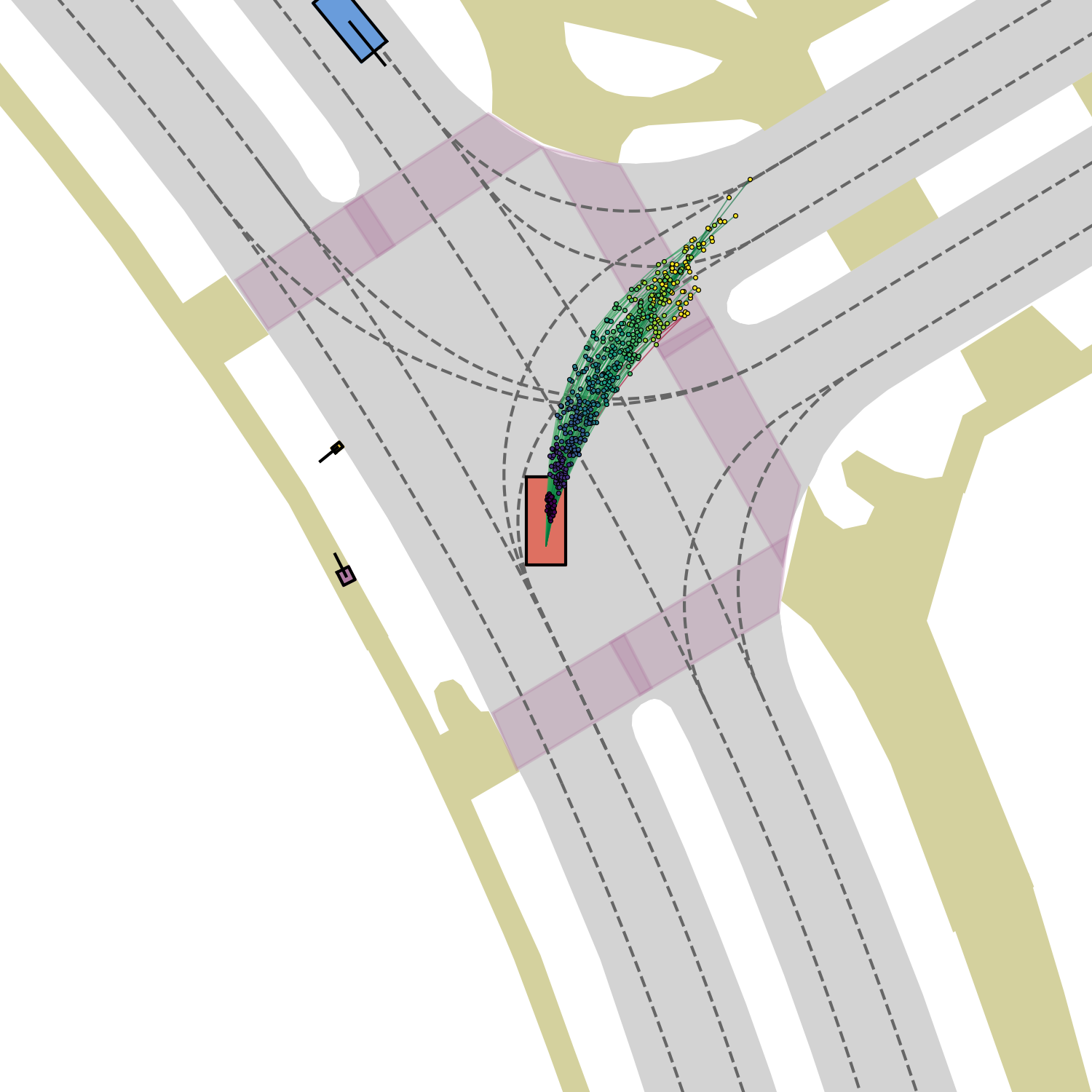} &
\vizimg{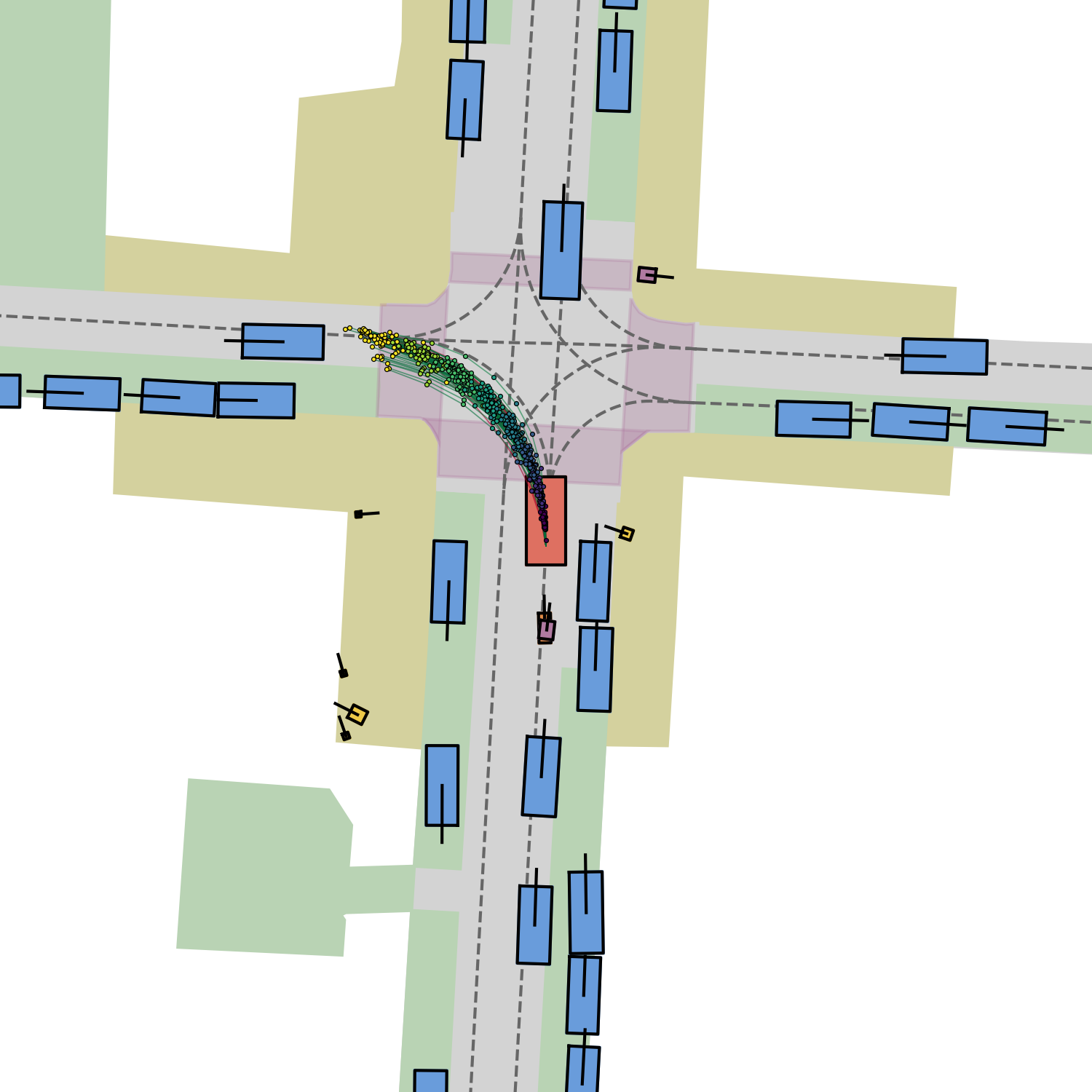} &
\vizimg{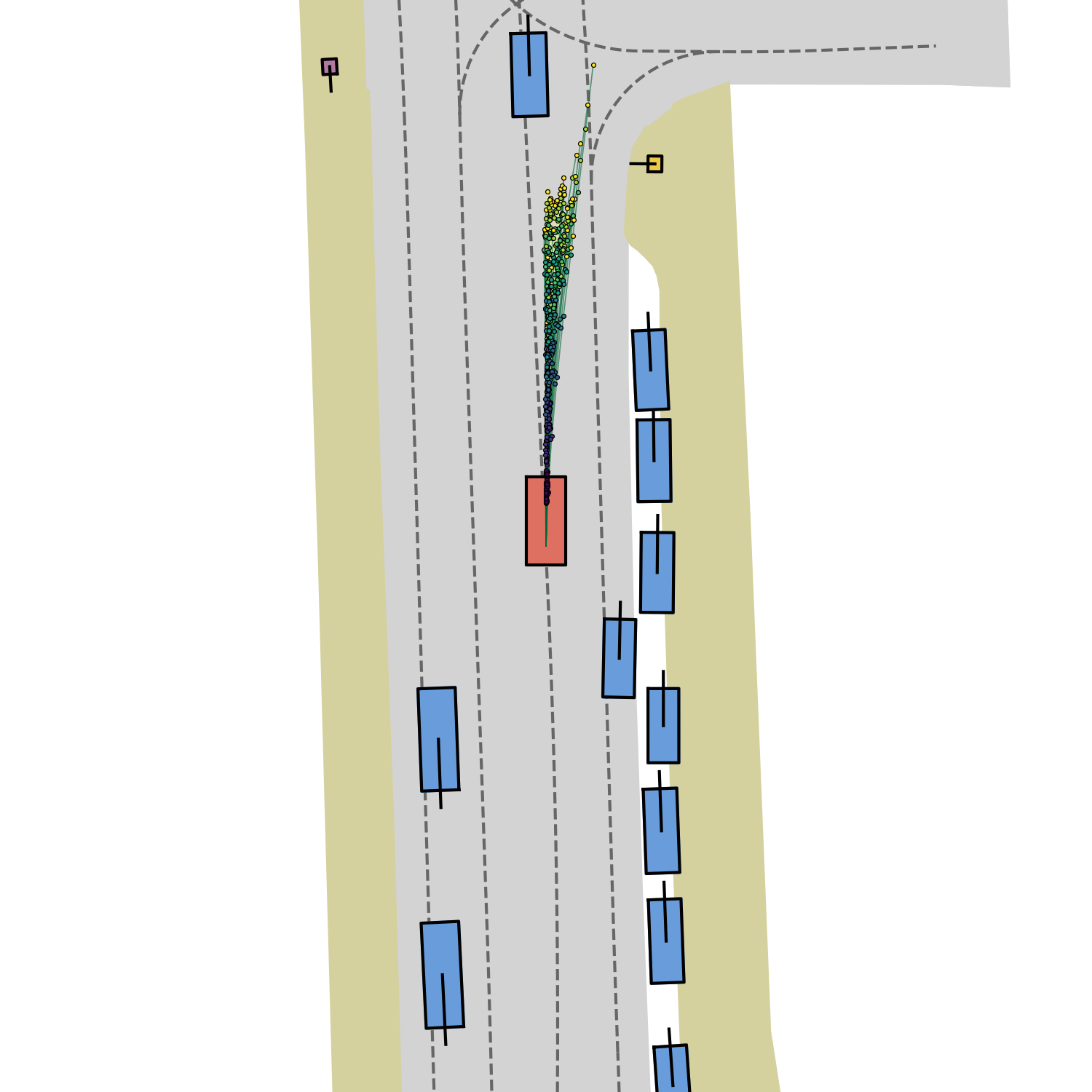} &
\vizimg{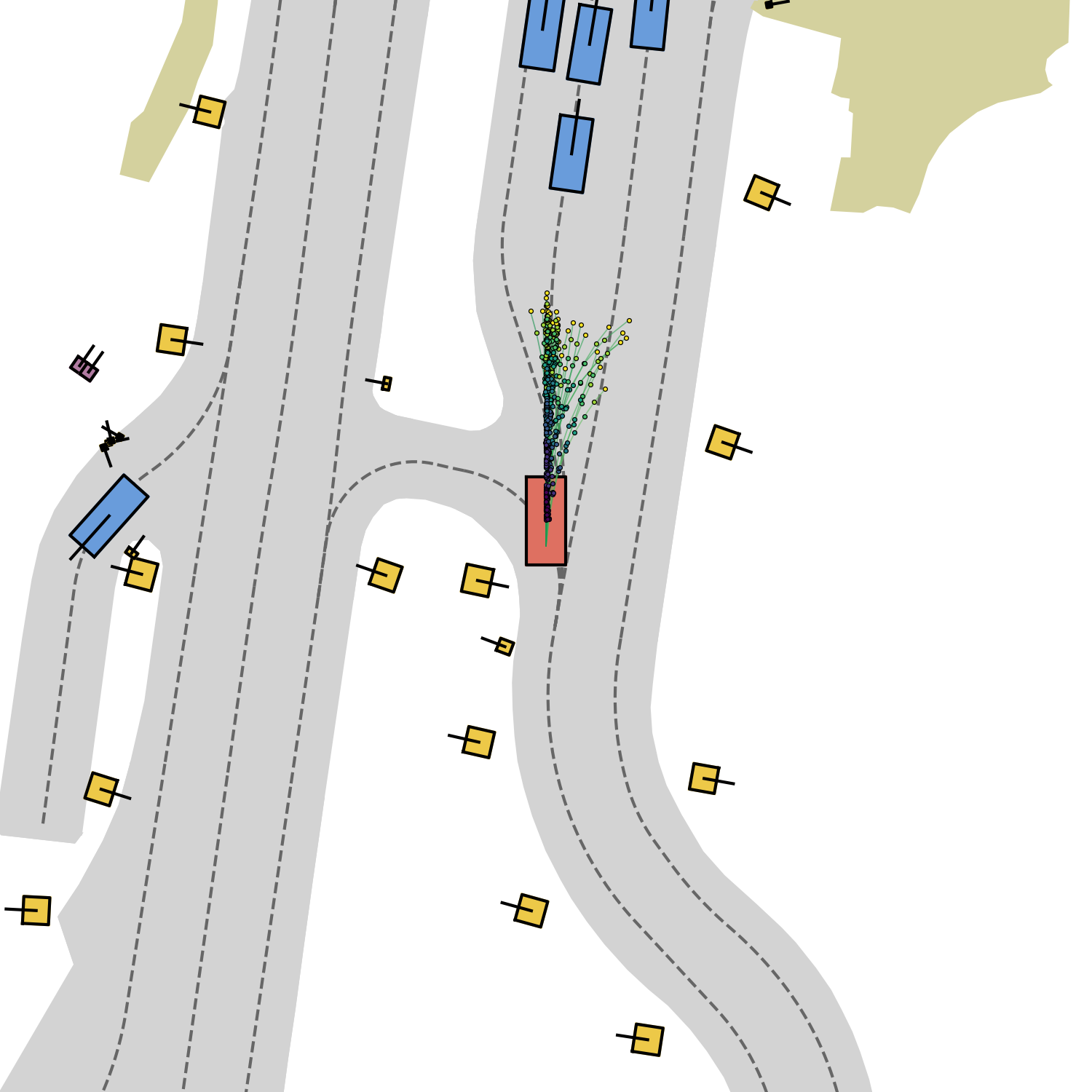}\\
& \small Right & \small Left & \small Forward & \small Forward \\
\end{tabular}
\caption{Qualitative comparison of proposal quality. Trajectories are colored by PDMS from 0 ({\color[RGB]{200,30,30}red}) to 1 ({\color[RGB]{30,150,30}green}). Driving command for each scene is labeled below. See App.~\ref{app:qual_extended} for additional scenes.}
\label{fig:qual}
\vspace{-6pt}
\end{figure*}

\noindent \textbf{Qualitative Comparison.}
Fig.~\ref{fig:qual} compares BEV proposals on representative scenes.
DiffusionDrive and DiffusionDriveV2 produce highly divergent proposals from fixed anchors, many of which receive negative scores.
iPad noticeably tightens the proposal set, but still emits irregular low-quality trajectories on complex scenes.
\ourmethod{} consistently concentrates proposals within a feasible action distribution across all scenes while preserving meaningful multimodality, demonstrating its ability to generate high-quality candidates at sampling time.

\subsection{Ablation Studies}
\begin{table}[t]
\centering
\caption{Ablation studies on reward conditioning. Evaluated on single proposal.}
\label{tab:ablation_combined}
\begin{minipage}{0.54\linewidth}
\centering
\subcaption{Reward condition granularity}
\label{tab:ablation_reward_condition}
\resizebox{\textwidth}{!}{
\begin{tabular}{lccc}
\toprule
 & Score-only & +Sub-rewards & +Per-timestep \\
\midrule
EP $\uparrow$  & 83.3 & \textbf{84.3} & 84.0 \\
TTC $\uparrow$ & 86.9 & 88.8 & \textbf{94.9} \\
\cellcolor{gray!20}PDMS $\uparrow$ & \cellcolor{gray!20}84.5 & \cellcolor{gray!20}86.7 & \cellcolor{gray!20}\textbf{89.4} \\
\bottomrule
\end{tabular}
}
\end{minipage}
\hfill
\begin{minipage}{0.43\linewidth}
\centering
\subcaption{Reward noise augmentation}
\label{tab:ablation_reward_augmentation}
\resizebox{\textwidth}{!}{
\begin{tabular}{lccc}
\toprule
 noise scale $\sigma$ & $\sigma\!=\!0$ & $\sigma\!=\!0.02$ & $\sigma\!=\!0.05$ \\
\midrule
EP $\uparrow$  & 79.1 & 82.8 & \textbf{84.0} \\
TTC $\uparrow$ & 91.2 & 92.2 & \textbf{94.9} \\
\cellcolor{gray!20}PDMS $\uparrow$ & \cellcolor{gray!20}84.8 & \cellcolor{gray!20}87.3 & \cellcolor{gray!20}\textbf{89.4} \\
\bottomrule
\end{tabular}
}
\end{minipage}
\end{table}

\noindent \textbf{Ablation on Reward Condition Granularity.} Tab.~\ref{tab:ablation_reward_condition} ablates the reward condition granularity discussed in Sec.~\ref{sec:reward}. Finer conditioning significantly improves proposal quality, most strikingly on TTC (88.8 to 94.9, +6.1). This confirms that exposing the model to where along the trajectory a constraint is violated is essential for the decoder to internalize hard safety constraints.

\begin{figure*}[t]
\centering
\begin{minipage}[t]{0.45\textwidth}
    \centering
    \includegraphics[width=\textwidth]{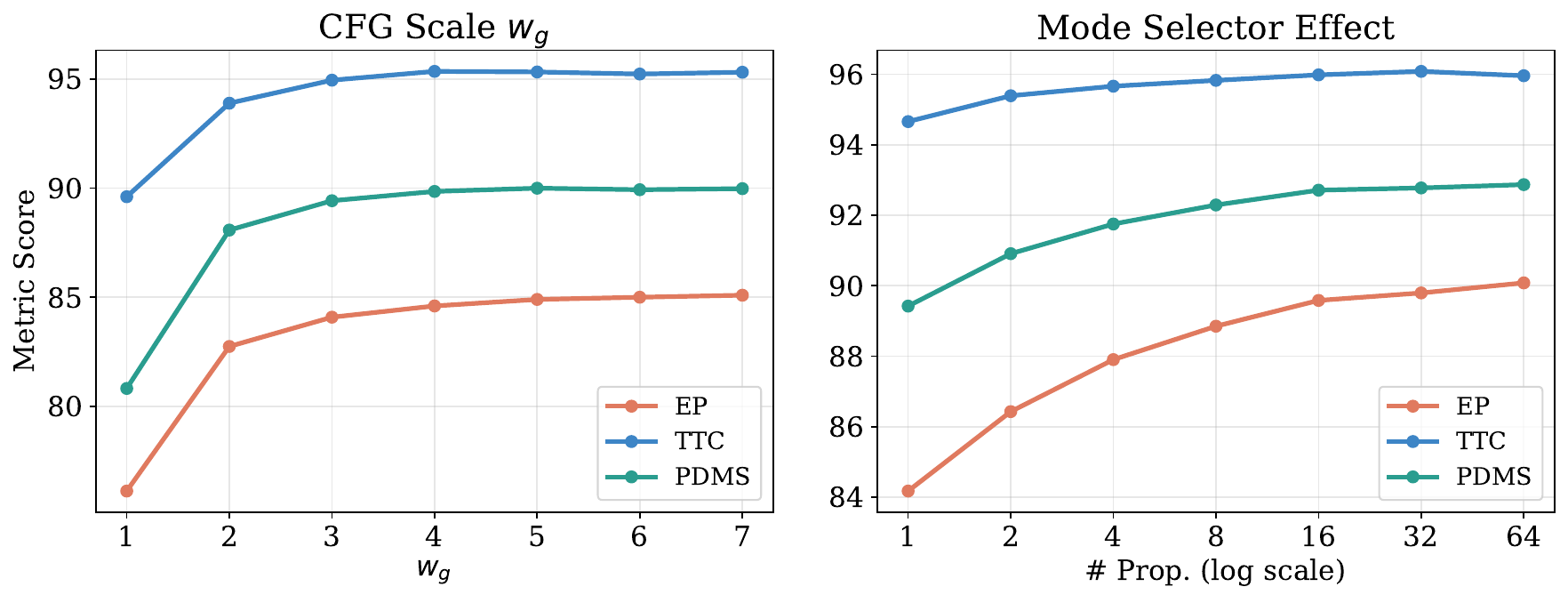}
    \caption{Ablation on CFG (left) and mode selector effect (right). \label{fig:ab_cfg_scorer}}
\end{minipage}
\hfill
\begin{minipage}[t]{0.52\textwidth}
    \centering
    \includegraphics[width=0.48\linewidth]{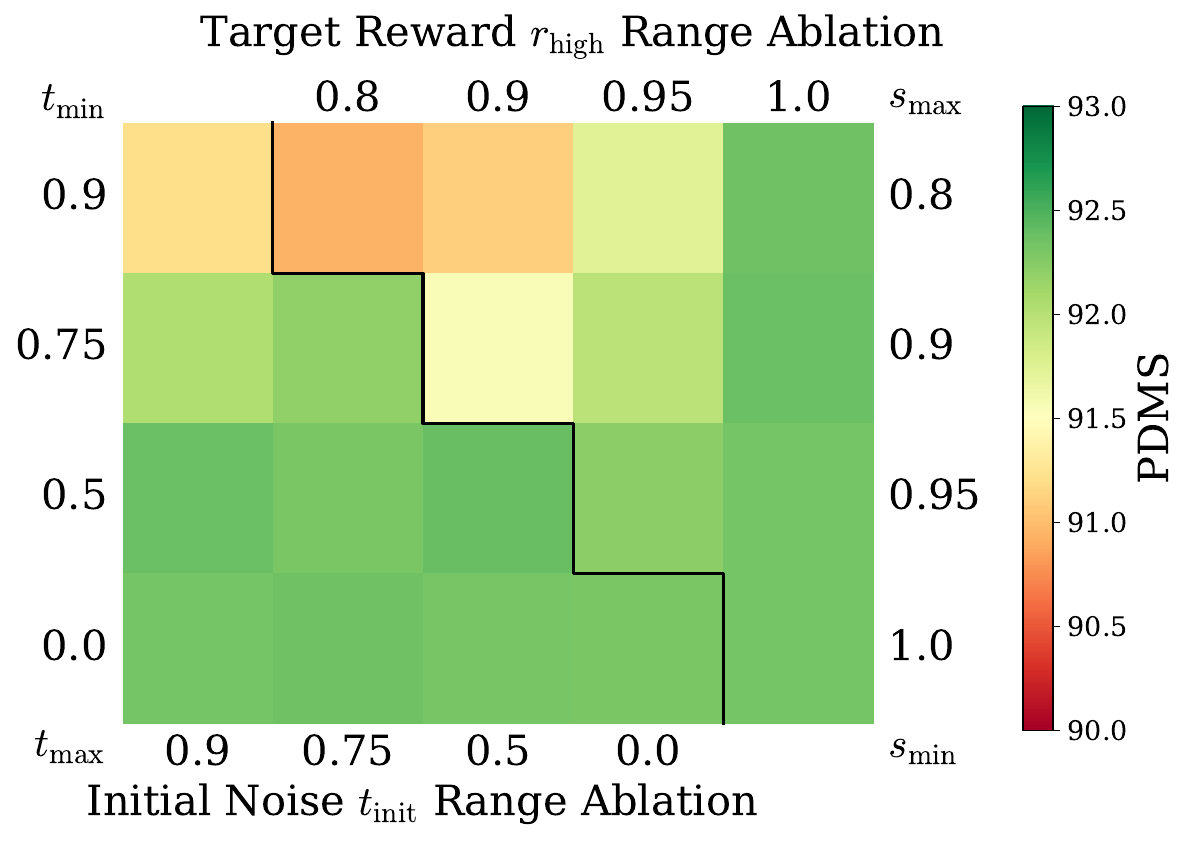}
    \includegraphics[width=0.48\linewidth]{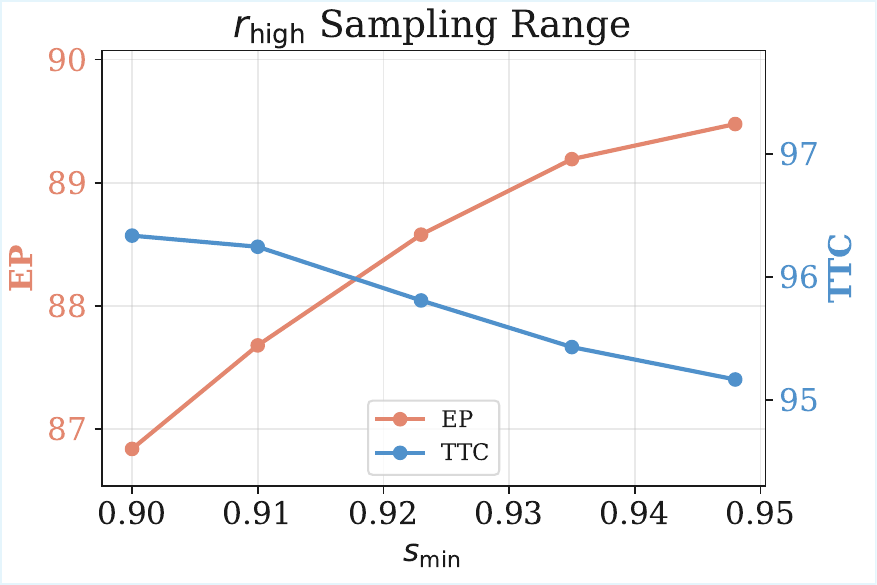}
    \caption{Ablation on sampling space (left) and inference-time objective balancing (right). \label{fig:abl_sampling}}
\end{minipage}
\end{figure*}

\noindent \textbf{Ablation on Reward Noise Augmentation.} Tab.~\ref{tab:ablation_reward_augmentation} studies the effect of noise augmentation on continuous rewards. Without noise ($\sigma{=}0$), the decoder treats rewards as trajectory identifiers rather than quality indicators, causing PDMS to collapse when conditioned on high target scores at inference (Fig.~\ref{fig:reward_ablation} right). Larger noise scales force the model to map a band of high rewards to feasible high-quality actions, rather than relying on trajectory cues leaked by the precise reward value.
Fig.~\ref{fig:reward_ablation} shows that the proposed per-timestep rewards and the noise augmentation together resolve the performance degradation with high reward conditioning.

\noindent \textbf{Effect of Mode Selector.} Fig.~\ref{fig:ab_cfg_scorer} right shows how the mode selector improves overall performance. As we increase the number of proposals it scores, the main gain is concentrated on progress (EP, 84 to 90), while safety metrics remain saturated. This confirms that the mode selector primarily refines soft objectives on top of an already-feasible proposal pool.

\noindent \textbf{Effect of CFG.}
Fig.~\ref{fig:ab_cfg_scorer} left shows the effect of classifier-free guidance~\cite{ho2022classifier} (CFG) at inference. CFG is essential for high performance, consistent with the standard finding in conditional generative models. We use CFG scale 5, at which performance saturates.

\noindent \textbf{Ablation on Sampling Strategy.} Fig.~\ref{fig:abl_sampling} ablates the sampling strategies of Sec.~\ref{sec:inference}. The left panel sweeps the sampling ranges of the target reward and initial noise. Given high rewards ($s_{\mathrm{max}}=1.0$) with enough denoising steps ($t_{\min} \leq 0.5$), \ourmethod{} achieves consistently high performance.
The right panel shows that adjusting the sampling range of $r_{\mathrm{high}}$ trades off between objectives at inference. With the same range length of $0.05$, a lower $s_{\min}$ yields more conservative proposals, and vice versa.

\section{Related Works}
\noindent \textbf{End-to-End Autonomous Driving.} End-to-end autonomous driving maps raw sensor input directly to planning output through a single differentiable model. UniAD~\cite{hu2023planning} pioneers this direction by integrating perception, prediction, and planning into one model. VAD~\cite{jiang2023vad} replaces dense BEV features with vectorized scene representations, and Transfuser~\cite{chitta2022transfuser} fuses multi-view images with LiDAR in the perception backbone. A line of follow-up works~\cite{chen2024ppad,weng2024paradrive,li2024enhancing,li2024ego,sun2024sparsedrive,liu2025unilion,zheng2024genad} continues to refine under the single-trajectory planning paradigm. Besides, some advanced methods~\cite{liu2026drivepi,li2026sgdrive} adopt a vision-language-action model to obtain a promising trajectory.

\noindent \textbf{Multimodal Driving Planning.} Multimodal planners fall into two categories, scoring-based~\cite{chen2024vadv2,li2024hydra,li2025generalized,yao2026drivesuprim} and anchor-based~\cite{liao2024diffusiondrive,guo2025ipad,kirby2026driving,xing2025goalflow}. Scoring-based methods select from a large fixed action vocabulary using a dedicated scorer. Starting from VADv2~\cite{chen2024vadv2}, Hydra-MDP~\cite{li2024hydra} and its extensions~\cite{li2025hydra,li2025generalized} train learning-based scorers with fine-grained simulation-based supervision to enhance selection performance. DriveSuprim~\cite{yao2026drivesuprim} refines scoring with a two-stage coarse-to-fine scheme.

Anchor-based methods decode proposals dynamically from a set of action anchors. DiffusionDrive~\cite{liao2024diffusiondrive} predicts proposals from fixed anchors via truncated diffusion. DiffusionDriveV2~\cite{zou2025diffusiondrivev2} adds RL post-training and dense trajectory scoring. GoalFlow~\cite{xing2025goalflow} pairs flow matching with goal-point anchors. iPad~\cite{guo2025ipad} uses a proposal-centric framework that iteratively refines proposals.
Our method unifies the dense supervision of scoring-based methods with the generation ability of anchor-based methods by learning a reward-conditioned action distribution from dense trajectory-reward pairs.

\noindent \textbf{Reward-Conditioned Policies and Offline RL.} Our reward-to-action distribution learning is conceptually related to a line of offline RL methods that learn conditional policies from logged data, including reward-conditioned behavioral cloning~\cite{kumar2019reward,schmidhuber2019reinforcement,emmons2022rvs}, return-conditioned sequence modeling~\cite{chen2021decision}, goal-conditioned supervised learning~\cite{gcsl_ghosh2021}, and diffusion-based planning~\cite{janner2022planning}. These methods condition on a single scalar reward, return, or goal, whereas we condition on a per-timestep multi-signal reward vector covering safety, progress, comfort, and rule compliance, and apply this formulation to driving planning with rule-based simulation rewards. Recent works also explore multi-component reward parameterizations~\cite{nauman2026reward} and treat classifier-free guidance as a policy improvement operator~\cite{frans2025diffusion}.

\section{Conclusion}
In this paper, we propose a novel multimodal driving planning framework, \ourmethod{}, to learn the reward-conditioned action distribution. Our core contribution unifies dense reward supervision with generative action modeling through fine-grained reward signals. Experiments show that the resulting action decoder generates high-quality multimodal proposals consistently within the feasible action distribution.
We hope this reward-conditioned generative modeling paradigm motivates future work that explores richer reward signals and extends to other policy learning scenarios beyond driving.

\begin{ack}
This work is supported by the Hong Kong Research Grant Council General Research Fund (No.~17213925), National Natural Science Foundation of China (No.~62422606), Shenzhen Loop Area Institute Fund (No.~FPF10120250006), and HKU Shanghai Intelligent Computing Research Center.
\end{ack}

\bibliographystyle{plain}
\bibliography{main}

\appendix

\clearpage

\section{Limitations and Future Directions}\label{app:limitations}

\noindent \textbf{Limitations.} The quality of the reward-conditioned action distribution $p(a|r)$ is bounded by the fidelity of the reward signals used as the condition, and in this work all signals come from the NAVSIM rule-based simulator. Inaccuracies in any subscore propagate into supervision. For instance, the NAVSIM ego-area check sometimes returns false detections due to gaps between adjacent area polygons, which enters training as label noise on the per-timestep ego-area array. A separate limitation lies in the mode selector, which is not a focus of this work. Our lightweight two-layer transformer occasionally promotes an unsafe proposal even when the action decoder produces a feasible majority (Fig.~\ref{fig:failure_cases}), and because it scores each frame independently it gives no explicit pressure for inter-frame consistency, which is reflected in the lower Two-Frame Extended Comfort score. Finally, evaluation is restricted to NAVSIM, since other planning datasets lack a comparable reward labeling pipeline.

\noindent \textbf{Future Directions.} A central question is how \ourmethod{} extends to settings where a NAVSIM-style rule-based simulator is unavailable. The generative formulation only requires a function from candidate action to reward vector, so the simulator can be replaced by a learning-based reward model, by hand-crafted proxy metrics, or by a pretrained trajectory scorer such as GTRS~\cite{li2025generalized}. All of these alternatives are substantially faster than rule-based simulation and can support online labeling with random or model-proposed trajectories during training, rather than the one-time offline pass we use here. Beyond reward sourcing, designing more fine-grained rewards that capture aspects beyond the NAVSIM rule set, building a stronger mode selector with cross-frame context, and applying \ourmethod{} to broader planning datasets are natural next steps.

\noindent \textbf{Broader Impacts.} \ourmethod{} targets autonomous driving planning, where deployment carries direct safety implications. Bias or blind spots in the reward signals can propagate into the learned action distribution, so practical use would require independent runtime monitoring and human oversight. Our evaluation is restricted to closed-loop simulation on NAVSIM and we make no claims about real-world deployment.

\section{Comparison with Prior Generative Planners}\label{app:gen_planners}

Several recent end-to-end planners adopt a flow-matching or diffusion-based trajectory decoder~\cite{liao2024diffusiondrive,xing2025goalflow,liu2025beyond}. \ourmethod{} shares this generative formulation but differs in what the decoder is trained to model. Prior methods supervise the decoder with a single ground-truth trajectory per scene and inject multimodality through external mechanisms such as fixed anchors, goal-point selection, or scorer outputs. \ourmethod{} instead supervises the decoder with dense action-reward pairs that span the action space, so the learned distribution $p(a|r)$ is multimodal by construction.

\noindent \textbf{Naive Diffusion or Flow Policies.} A baseline approach trains the generative decoder on the GT trajectory under a diffusion or flow-matching loss. Since the supervision target is deterministic given the scene, the decoder has no incentive to use the noise input as an information channel and learns to predict the GT directly from the scene features. The generative formulation collapses into single-trajectory regression, and proposals from different noise samples converge to one mode at convergence. DiffusionDrive~\cite{liao2024diffusiondrive} reports this collapse explicitly when applying a vanilla diffusion policy on top of Transfuser~\cite{chitta2022transfuser}.

\noindent \textbf{DiffusionDrive.} DiffusionDrive addresses the collapse with anchored truncated diffusion. It clusters training trajectories into 20 anchors and uses each anchor as the start of a truncated diffusion schedule. Per-anchor reconstruction is supervised by a winner-takes-all loss in which only the anchor closest to the GT receives the reconstruction signal, while a separate classification head ranks anchors. This forces proposals to span a multimodal anchor space, which is the source of the reported diversity. The supervision per winning anchor, however, remains a single-GT objective. Each anchor selected as the winner is regressed to that scene's GT, so the per-anchor distribution can still collapse to one trajectory. The procedure spans modes across a fixed anchor set rather than within a scene, and the anchor set is shared across all scenes.

\noindent \textbf{GoalFlow.} GoalFlow~\cite{xing2025goalflow} decouples multimodality from the flow decoder by introducing a separate goal-point construction module that selects a single high-scoring endpoint and conditions the flow on this goal. The flow decoder is then trained with the standard rectified-flow velocity-matching loss against the single GT. The flow itself is therefore subject to the same degeneration as the naive policy, and the diversity comes from varying the goal-point condition rather than from sampling the flow. This view is consistent with the reported observation that even a single denoising step is sufficient at inference, indicating that the noise input carries little information.

\noindent \textbf{CATG.} CATG~\cite{liu2025beyond} adds richer conditioning, including a trajectory anchor and a target endpoint selected from the top-100 candidates of a pretrained GTRS~\cite{li2025generalized} scorer, together with a per-trajectory ego-progress label. The flow decoder is again trained with a flow-matching loss against the single GT. The richer conditioning sharpens the conditional distribution, which makes per-condition mode collapse more likely rather than less. CATG mitigates the resulting drivable-area-compliance issues with three inference-time constraints over the velocity field and the flow start. The method effectively functions as a flow-matching post-processor over GTRS proposals, at the cost of 100 candidates and 100 sampling steps per scene.

\noindent \textbf{Difference of \ourmethod{}.} The common pattern across these methods is to train a generative trajectory decoder under single-GT supervision and to inject multimodality externally through anchors, goal points, or scorer outputs. Adding scene-conditional information does not resolve the underlying tension, since the supervision remains deterministic given the chosen condition. \ourmethod{} addresses this at the supervision level. The reward-conditioned distribution $p(a|r)$ is trained on dense action-reward pairs that span the action vocabulary in each scene. So the same condition maps to multiple possible actions instead of a single one, necessitating the generative formulation and forcing the decoder to rely on its noisy sample input. Multimodality is a property of the learned distribution rather than explicit anchors, and the trained decoder produces high-quality proposals through reward conditioning alone.

\section{Additional Experiments}\label{app:additional}

\subsection{Controllability}\label{app:controllability}

\begin{table}[t]
\centering
\caption{Controllability of the reward condition on \texttt{navtest}. Evaluated on single proposal.}
\label{tab:controllability}
\begin{minipage}{0.52\linewidth}
\centering
\subcaption{Target score}
\label{tab:control_score}
\resizebox{\textwidth}{!}{
\begin{tabular}{lcccc}
\toprule
Target score & NC $\uparrow$ & TTC $\uparrow$ & EP $\uparrow$ & \cellcolor{gray!20}PDMS $\uparrow$ \\
\midrule
0.90 & \textbf{98.9} & \textbf{96.4} & 75.9 & \cellcolor{gray!20}87.1 \\
0.95 & 98.6 & 95.6 & 85.1 & \cellcolor{gray!20}\textbf{90.3} \\
1.00 & 98.0 & 92.6 & \textbf{88.4} & \cellcolor{gray!20}89.6 \\
\bottomrule
\end{tabular}
}
\end{minipage}
\hfill
\begin{minipage}{0.45\linewidth}
\centering
\subcaption{Comfort condition}
\label{tab:control_comfort}
\resizebox{\textwidth}{!}{
\begin{tabular}{lccc}
\toprule
Comfort target & NC $\uparrow$ & DAC $\uparrow$ & \cellcolor{gray!20}HC $\uparrow$ \\
\midrule
1 (comfortable) & \textbf{98.1} & \textbf{96.6} & \cellcolor{gray!20}\textbf{98.3} \\
unset & 98.0 & \textbf{96.6} & \cellcolor{gray!20}96.2 \\
0 (uncomfortable) & 92.3 & 90.5 & \cellcolor{gray!20}18.0 \\
\bottomrule
\end{tabular}
}
\end{minipage}
\end{table}

The reward condition directly steers proposal generation. Tab.~\ref{tab:control_score} varies the target score: a lower target yields more conservative proposals with higher safety, and a higher target yields more aggressive proposals with higher progress (see also Fig.~\ref{fig:sampling_space}). Tab.~\ref{tab:control_comfort} toggles the comfort condition, and the generated proposals follow it accordingly. We report the NAVSIM v2 history comfort (HC) metric here, since the v1 comfort metric labels almost all trajectories as comfortable.

\subsection{Latency Analysis}\label{app:latency}

\begin{table}[t]
\centering
\caption{Inference latency breakdown on NAVSIM \texttt{navtest}. Per-sample latency in milliseconds, measured on a single NVIDIA H20 GPU with batch size 1 over 1000 samples. $N$ is the number of proposals, $K$ is the total Euler solver steps over $[0,1]$, and $t_{\min}$ is the lower bound of the initial noise sampling range. The first row is the default setting reported in the main paper.}
\label{tab:latency}
\setlength{\tabcolsep}{5pt}
\small
\begin{tabular}{ccccccccc}
\toprule
\multirow{2}{*}{$N$} & \multirow{2}{*}{$K$} & \multirow{2}{*}{$t_{\min}$} & \multicolumn{4}{c}{Component Latency (ms)} & \multirow{2}{*}{Total (ms)} & \multirow{2}{*}{PDMS $\uparrow$} \\
\cmidrule(lr){4-7}
& & & Perception & Reward Enc. & Denoising & Selector & & \\
\midrule
60 & 20 & 0.50 & 13.8 & 2.6 & 70.2 & 4.3 & 91.3 & \textbf{92.8} \\
 4 & 20 & 0.50 & 14.2 & 2.7 & 60.6 & 4.2 & 82.0 & 91.8   \\
60 & 20 & 0.75 & 14.1 & 2.7 & 36.9 & 4.5 & 58.7 & 92.3 \\
60 & 10 & 0.50 & 12.8 & 2.5 & 33.5 & 4.0 & 53.2 & 92.2 \\
60 & 10 & 0.70 & 14.4 & 2.7 & 22.2 & 4.4 & 44.1 & 92.0 \\
\bottomrule
\end{tabular}
\end{table}

Tab.~\ref{tab:latency} reports the per-component latency of \ourmethod{}. The denoising loop dominates total latency, exceeding 75\% of the per-frame cost in the default 20-step setting, while perception, reward encoding, and the mode selector together stay below 22\,ms across all configurations. The proposal count has little effect on latency, with only a 9\,ms reduction when going from 60 to 4 proposals, leaving the sequential denoising steps as the dominant factor. The generative formulation allows us to customize the denoising process for efficiency trade-off by changing the total Euler solver steps $K$ and the minimum initial denoising time $t_{\min}$. $K$ defines the discretization of the full $[0,1]$ time interval, and only the steps from $t_{\mathrm{init}}$ onward are executed, so the realized number of forward passes per proposal is $\lceil K\,(1-t_{\mathrm{init}})\rceil$. Raising $t_{\min}$ from $0.50$ to $0.75$ (rows 1, 3) drops latency from 91 to 59\,ms at a 0.5 PDMS regression, halving $K$ from 20 to 10 (rows 1, 4) gives a similar speed-up at 0.6 PDMS regression, and combining $t_{\min}{=}0.70$ with $K{=}10$ (row 5) reaches 44\,ms per frame, more than $2\times$ faster than the default while staying within 0.8 PDMS. Both controls are chosen at inference time, so the same trained model can run at any of these speed-quality settings.

\subsection{Number of Denoising Steps}\label{app:denoising_steps}

\begin{table}[t]
\centering
\caption{PDMS vs. number of denoising steps on NAVSIM v1 \texttt{navtest}. Number of proposals $N{=}60$. Per-sample latency measured on a single NVIDIA H20 GPU with batch size 1 over 1000 samples.}
\label{tab:denoising_steps}
\setlength{\tabcolsep}{8pt}
\small
\begin{tabular}{cccc}
\toprule
$K$ & Denoising (ms) & Total (ms) & PDMS $\uparrow$ \\
\midrule
10 & 33.5 & 53.2 & 92.2 \\
20 & 70.2 & 91.3 & \textbf{92.8} \\
30 & 104.3 & 125.1 & 92.7 \\
40 & 140.0 & 161.4 & 92.7 \\
\bottomrule
\end{tabular}
\end{table}

Tab.~\ref{tab:denoising_steps} sweeps the number of denoising steps $K$ under default sampling setting. PDMS saturates at $K{=}20$ and does not improve when $K$ is raised further, while latency grows roughly linearly with $K$. Halving $K$ to 10 gives a $1.7\times$ speed-up at a 0.6 PDMS regression. We therefore use $K{=}20$ as the default.

\subsection{Mode Selector Aggregation Weights}\label{app:scorer_weights}

\begin{table}[t]
\centering
\caption{Effect of mode selector aggregation weights on NAVSIM v1 \texttt{navtest}. The weights $(w_{\mathrm{EP}}, w_{\mathrm{TTC}}, w_{\mathrm{HC}})$ are applied to the weighted subscores in Eq.~\ref{eq:agg}. The first row is our default; the second row uses the official NAVSIM weights.}
\label{tab:scorer_weights}
\setlength{\tabcolsep}{6pt}
\small
\begin{tabular}{lcccccc}
\toprule
$(w_{\mathrm{EP}}, w_{\mathrm{TTC}}, w_{\mathrm{HC}})$ & NC $\uparrow$ & DAC $\uparrow$ & EP $\uparrow$ & TTC $\uparrow$ & Comf. $\uparrow$ & PDMS $\uparrow$ \\
\midrule
$(1,1,1)$ (default)        & 98.8 & 98.0 & \textbf{90.1} & 96.0 & 100 & 92.8 \\
$(5,5,2)$ & 98.8 & 97.9 & 90.0 & 95.9 & 100 & 92.8 \\
$(1,2,1)$                   & 99.0 & 98.0 & 89.0 & 96.8 & 100 & 92.8 \\
$(1,3,1)$                   & \textbf{99.1} & \textbf{98.1} & 88.0 & \textbf{97.1} & 100 & 92.6 \\
\bottomrule
\end{tabular}
\end{table}

Tab.~\ref{tab:scorer_weights} reports performance on \texttt{navtest} under different aggregation weights $w_k$ in Eq.~\ref{eq:agg}. The final PDMS is stable across settings, including the official NAVSIM weights $(w_{\mathrm{EP}}, w_{\mathrm{TTC}}, w_{\mathrm{HC}}){=}(5,5,2)$. Beyond robustness, the weights also serve as a free inference-time knob to balance EP and TTC. Increasing $w_{\mathrm{TTC}}$ relative to $w_{\mathrm{EP}}$ smoothly trades progress for safety without retraining, raising TTC from $96.0$ to $97.1$ at the cost of EP dropping from $90.1$ to $88.0$, while PDMS stays within $0.3$.

\section{Qualitative Results}\label{app:qualitative}

\subsection{Sampling Space Visualization}\label{app:sampling_space}

\newcommand{\sampleimg}[2]{\includegraphics[width=0.175\linewidth]{figs/sampling_space/rhigh#1_tinit#2.png}}
\newcommand{\sampletinitlabel}[1]{\rotatebox{90}{\parbox{1.7cm}{\centering\small$t_{\mathrm{init}}{=}#1$}}}

\begin{figure*}[t]
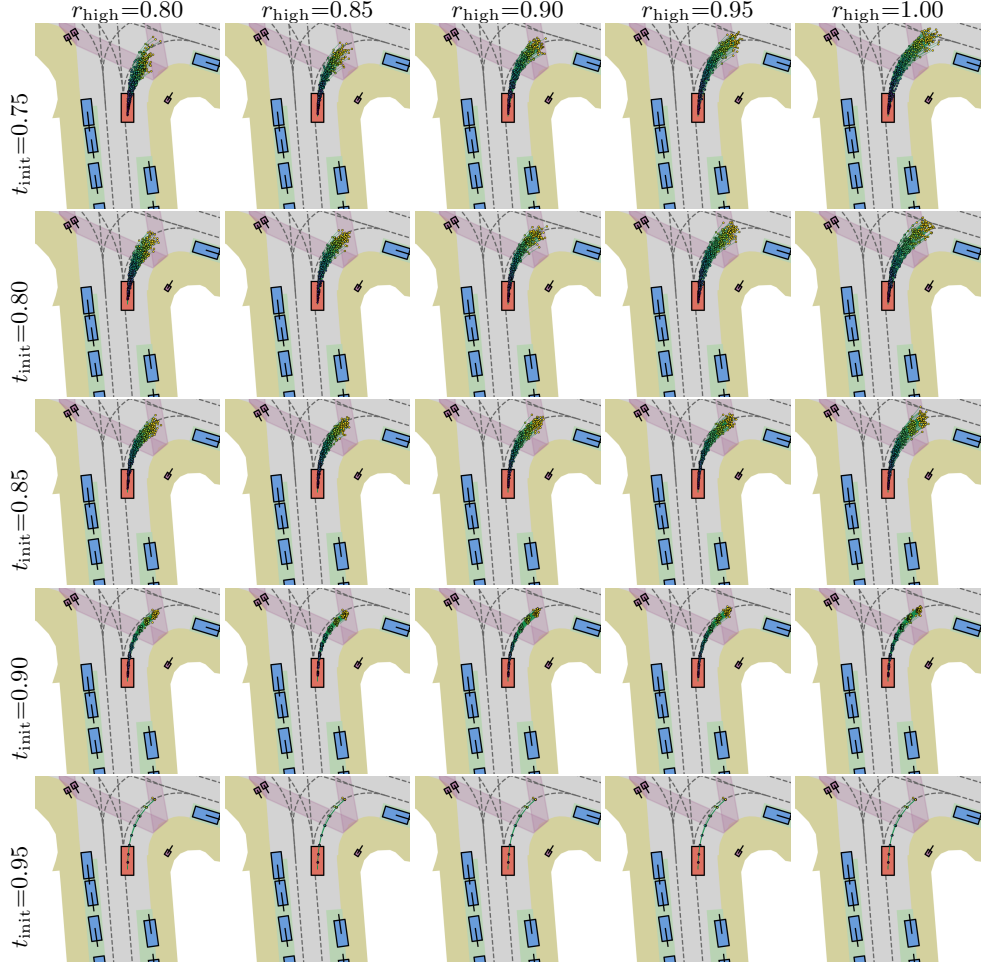

\centering
\setlength{\tabcolsep}{1pt}
\renewcommand{\arraystretch}{0.4}
\begin{tabular}{@{}c ccccc@{}}
 & \small$r_{\mathrm{high}}{=}0.80$ & \small$r_{\mathrm{high}}{=}0.85$ & \small$r_{\mathrm{high}}{=}0.90$ & \small$r_{\mathrm{high}}{=}0.95$ & \small$r_{\mathrm{high}}{=}1.00$ \\
\sampletinitlabel{0.75} &
\sampleimg{0.80}{0.75} & \sampleimg{0.85}{0.75} & \sampleimg{0.90}{0.75} & \sampleimg{0.95}{0.75} & \sampleimg{1.00}{0.75} \\
\sampletinitlabel{0.80} &
\sampleimg{0.80}{0.80} & \sampleimg{0.85}{0.80} & \sampleimg{0.90}{0.80} & \sampleimg{0.95}{0.80} & \sampleimg{1.00}{0.80} \\
\sampletinitlabel{0.85} &
\sampleimg{0.80}{0.85} & \sampleimg{0.85}{0.85} & \sampleimg{0.90}{0.85} & \sampleimg{0.95}{0.85} & \sampleimg{1.00}{0.85} \\
\sampletinitlabel{0.90} &
\sampleimg{0.80}{0.90} & \sampleimg{0.85}{0.90} & \sampleimg{0.90}{0.90} & \sampleimg{0.95}{0.90} & \sampleimg{1.00}{0.90} \\
\sampletinitlabel{0.95} &
\sampleimg{0.80}{0.95} & \sampleimg{0.85}{0.95} & \sampleimg{0.90}{0.95} & \sampleimg{0.95}{0.95} & \sampleimg{1.00}{0.95} \\
\end{tabular}
\caption{Sampling-space visualization of \ourmethod{} on a single \texttt{navtest} scene. The grid sweeps the high-reward target $r_{\mathrm{high}}$ across columns and the initial denoising time $t_{\mathrm{init}}$ across rows, with all other inference settings fixed. Each cell shows 60 proposals from a single sampling configuration.\label{fig:sampling_space}}
\end{figure*}

Fig.~\ref{fig:sampling_space} sweeps the two sampling controls on a single scene. The column sweep varies $r_{\mathrm{high}}$ within the high-reward region, from $0.80$ to $1.00$. Within this range the proposals are already high-quality, so the visible effect is a shift toward higher ego progress, with trajectories reaching further along the route as $r_{\mathrm{high}}$ grows. Moving down the rows ($t_{\mathrm{init}}$ from $0.75$ to $0.95$) anchors the denoising trajectory closer to the IL head output, which reduces the spatial spread of proposals and converges the set toward a single mode.

\subsection{Reward Score Distribution}\label{app:reward_viz}

\begin{figure*}[t]
\centering
\begin{subfigure}[t]{0.162\linewidth}\includegraphics[width=\linewidth]{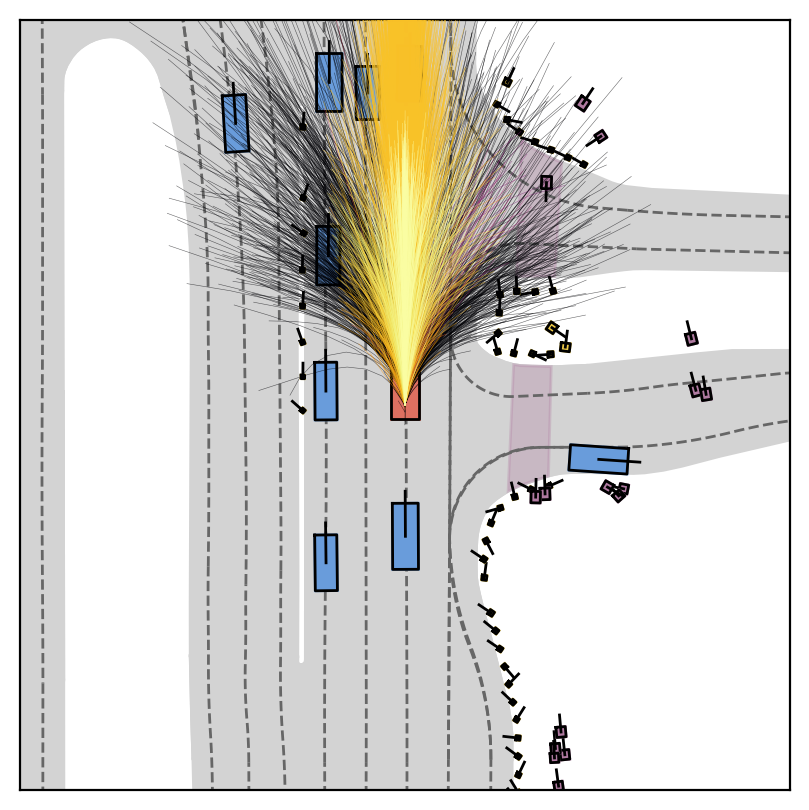}\caption{PDMS}\end{subfigure}\hfill
\begin{subfigure}[t]{0.162\linewidth}\includegraphics[width=\linewidth]{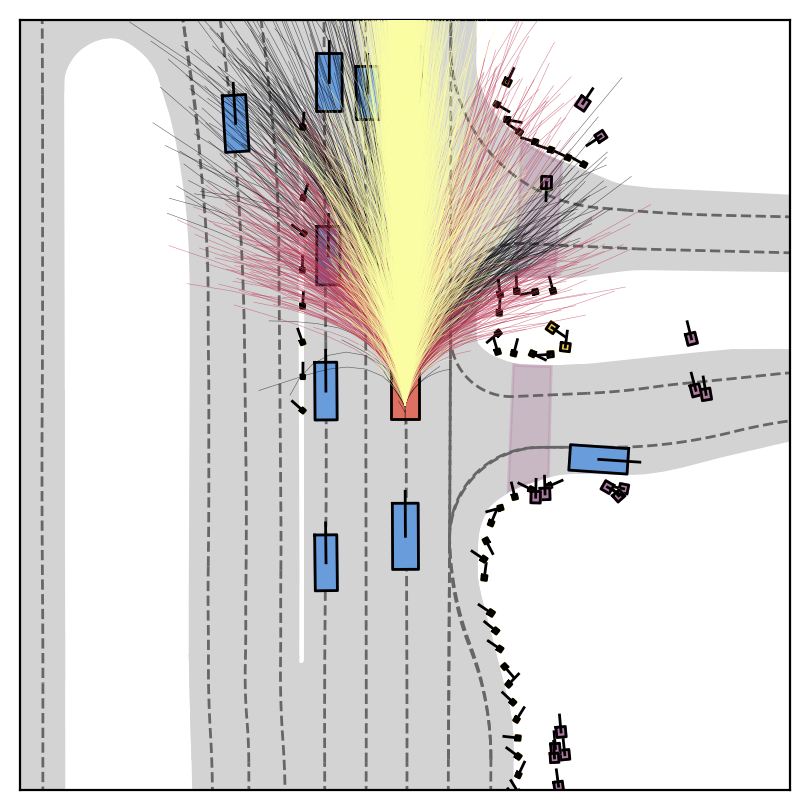}\caption{NC}\end{subfigure}\hfill
\begin{subfigure}[t]{0.162\linewidth}\includegraphics[width=\linewidth]{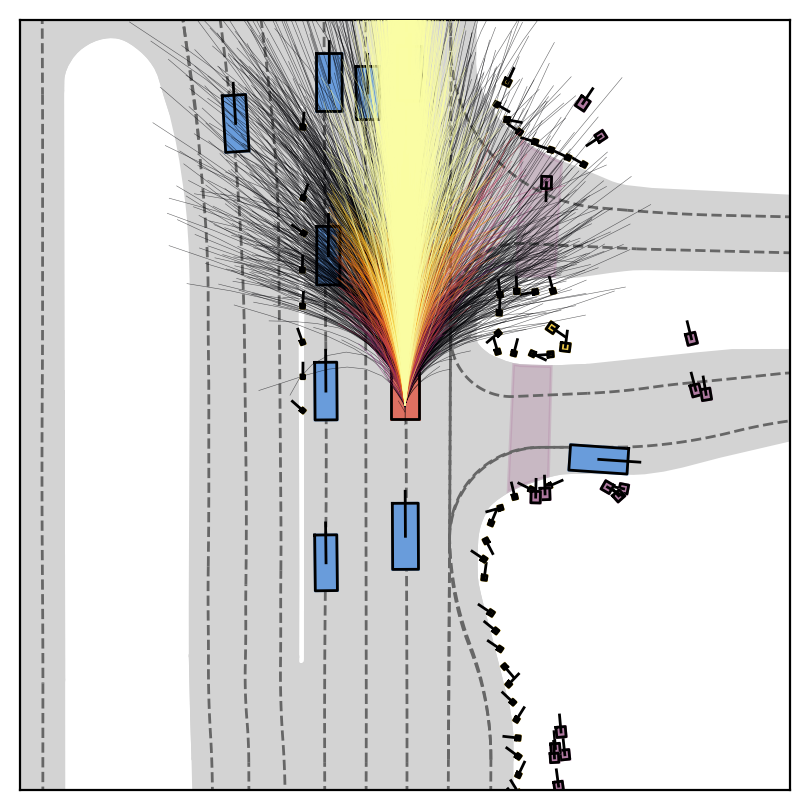}\caption{EP}\end{subfigure}\hfill
\begin{subfigure}[t]{0.162\linewidth}\includegraphics[width=\linewidth]{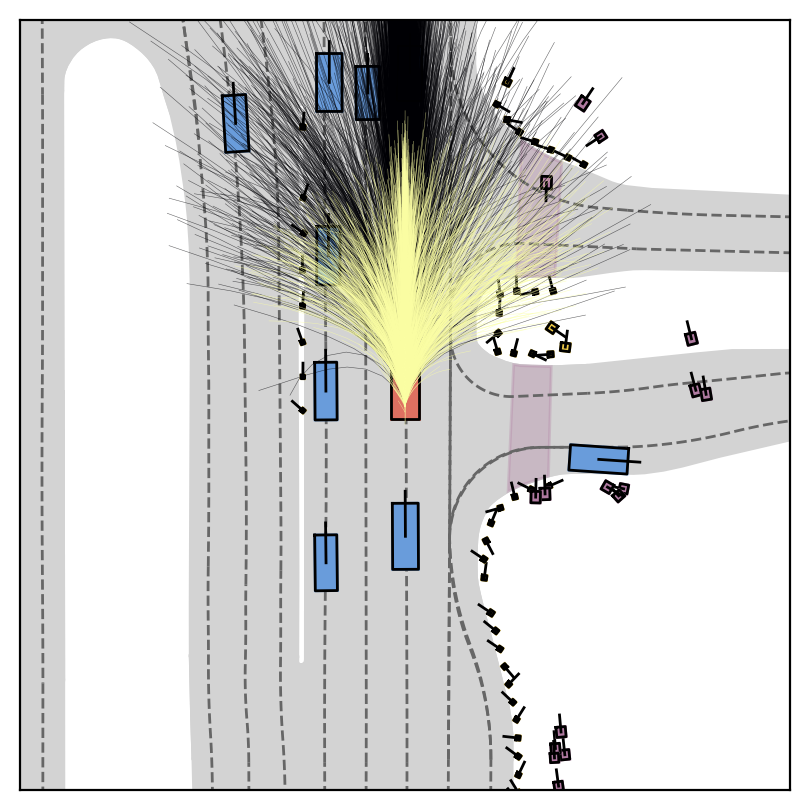}\caption{HC}\end{subfigure}\hfill
\begin{subfigure}[t]{0.162\linewidth}\includegraphics[width=\linewidth]{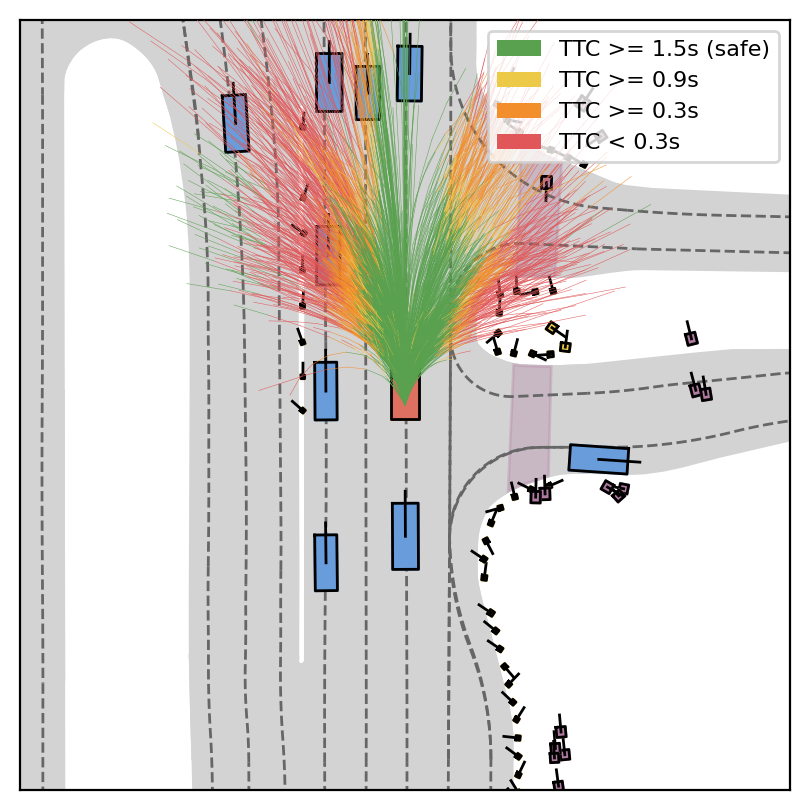}\caption{TTC-time}\end{subfigure}\hfill
\begin{subfigure}[t]{0.162\linewidth}\includegraphics[width=\linewidth]{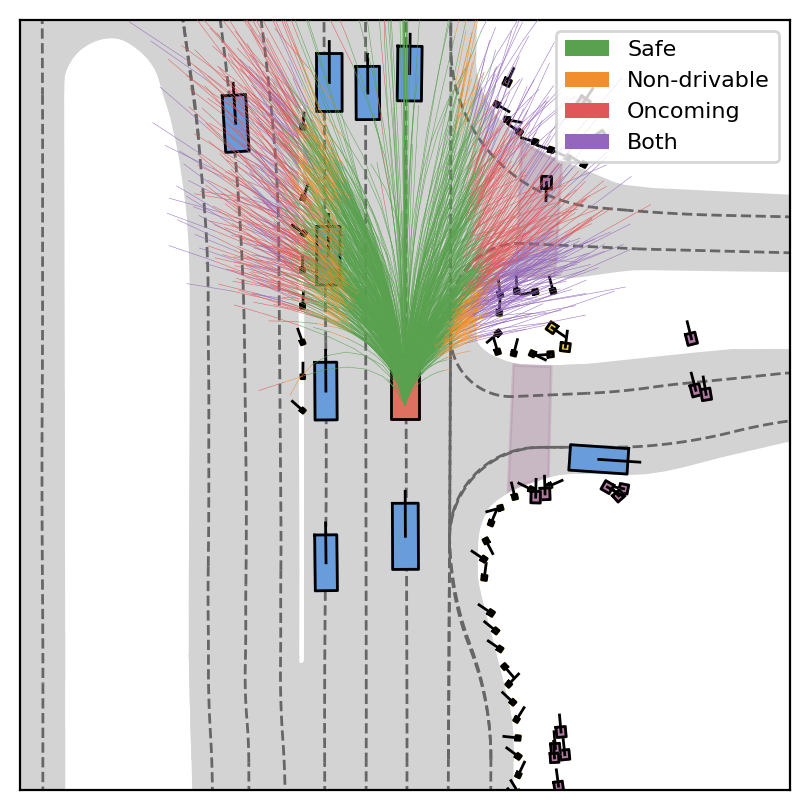}\caption{Ego-area}\end{subfigure}
\caption{Reward score distribution over the action vocabulary on a single \texttt{navtest} scene. Each panel colors the 8192 vocabulary trajectories by one reward signal. PDMS, NC, EP, and HC use a brighter color to denote higher score; TTC-time and ego-area use a categorical palette and selectively show trajectories with violations.\label{fig:reward_viz}}
\end{figure*}

Fig.~\ref{fig:reward_viz} visualizes the reward labels assigned to the full 8192-trajectory vocabulary for one scene. The distributions are highly non-uniform: most candidate trajectories receive low aggregate PDMS, while high-quality actions occupy a sparse subset of the vocabulary. Different reward signals also carve the action space in complementary ways. Scalar metrics such as EP and HC describe soft preferences over progress and smoothness, whereas the per-timestep TTC-time and ego-area labels localize hard safety and drivable-area violations along the trajectory. This heterogeneity motivates both our inverse-density sampling strategy during training and the fine-grained reward conditioning used by the decoder.

\subsection{Failure Cases}\label{app:failure_cases}

\begin{figure*}[t]
\centering
\includegraphics[width=\linewidth]{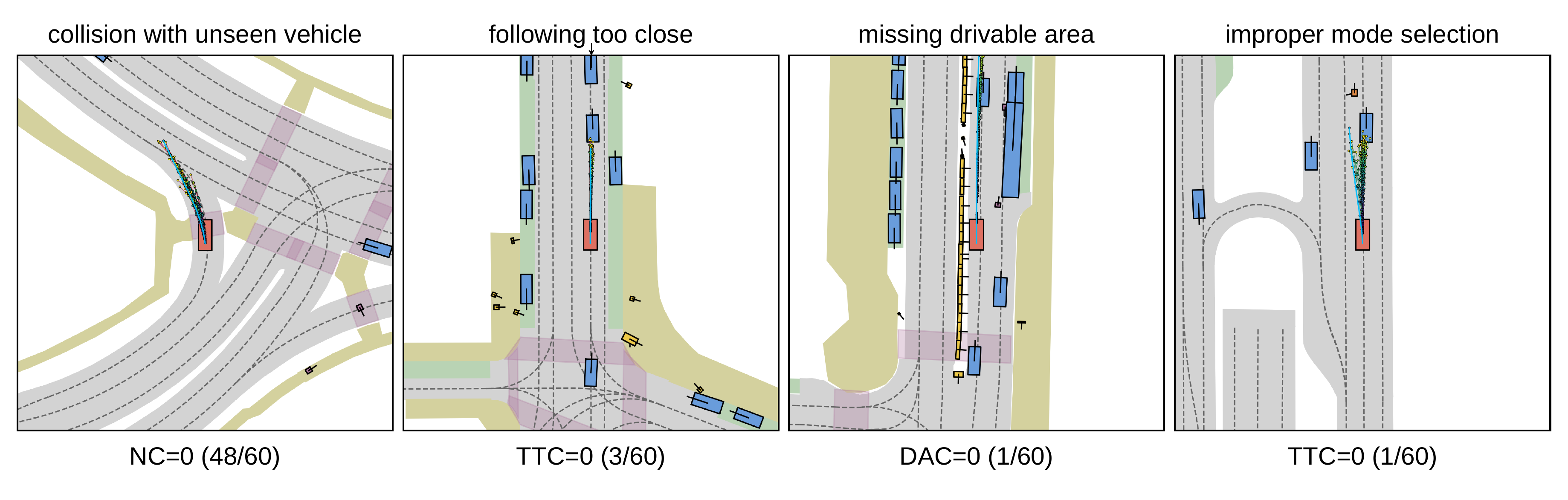}
\caption{Failure cases of \ourmethod{} on \texttt{navtest}. For each scene, the top label names the failure mode and the bottom label reports the failed metric together with the count of failing proposals out of the 60 sampled proposals. Selected proposal is colored in \textcolor{blue}{blue}.\label{fig:failure_cases}}
\end{figure*}

Fig.~\ref{fig:failure_cases} shows four representative failures of \ourmethod{} on \texttt{navtest}. From left to right, the cases are a collision with a vehicle that becomes visible only mid-trajectory and is missed by the perception encoder, following the lead vehicle too closely so fails TTC, drifting near the drivable area edge so fails DAC, and a mode-selection error that ranks an unsafe proposal above feasible alternatives.

These cases suggest that the current performance of \ourmethod{} is primarily bounded by perception coverage and mode selector quality. Of the two, the selector is the more actionable bottleneck. In scenes where the action decoder already produces a feasible majority, the selector still occasionally promotes one of the few failing proposals, leaving headroom that a stronger ranking model could recover. Improving the selector is a promising direction for future work.

\subsection{Extended Comparisons}\label{app:qual_extended}

Fig.~\ref{fig:qual_all_1}, Fig.~\ref{fig:qual_all_2} and Fig.~\ref{fig:qual_all_3} extend the qualitative comparison in the main paper to additional \texttt{navtest} scenes.

\section{NAVSIM Benchmark and PDM Score}\label{app:navsim}

We give a self-contained description of the NAVSIM~\cite{dauner2024navsim,cao2025pseudo} benchmark and the closed-loop PDM score used for evaluation and reward construction in the main paper. Sec.~\ref{app:navsim_sim} describes the dataset and the simulation pipeline. Sec.~\ref{app:metrics} lists every subscore. Sec.~\ref{app:pdm_agg} gives the aggregation formulas for PDMS (v1) and EPDMS (v2). Sec.~\ref{app:v1v2} summarizes the differences between the two versions.

\subsection{Dataset and Simulation Pipeline}\label{app:navsim_sim}

\noindent \textbf{Dataset.} NAVSIM is built on top of OpenScene~\cite{openscene2023}, a compact redistribution of nuPlan~\cite{Caesar2021nuplan}. It curates real-world non-trivial driving scenes where the future plan cannot be directly extrapolated from history, and is split into \texttt{navtrain} (103k frames) and \texttt{navtest} (12k frames). Each frame provides multi-view camera images, LiDAR point clouds, ego status, driving command, surrounding agent tracks, map elements, and a 4-second future ground-truth trajectory.

\noindent \textbf{Simulation.} A planned trajectory is evaluated by closed-loop log-replay simulation. The 4-second plan, given as eight ego-frame waypoints at 0.5\,s intervals, is converted into a global interpolated trajectory and fed to an LQR tracker that drives a kinematic bicycle model at 10\,Hz. The output is a sequence of 41 states at 0.1\,s intervals. All subscores are computed on this simulated state sequence rather than on the raw planned waypoints.

\noindent \textbf{PDM Reference Trajectory.} The official scorer also evaluates a rule-based reference trajectory produced by the PDM-Closed planner~\cite{dauner2023parting}.
The reference is used only to normalize ego progress (Sec.~\ref{app:metrics}).

\subsection{Subscores}\label{app:metrics}

NAVSIM subscores split into multiplicative metrics that gate the final score on hard violations, and weighted metrics that contribute additively to a quality term.

\noindent \textbf{No At-Fault Collisions (NC, multiplicative).} For each timestep, the scorer checks whether the ego polygon intersects any tracked object. A collision is at-fault if it is a front collision, a stopped-object collision, or a lateral collision while the ego sits in multiple lanes or non-drivable area. NC is set to 0 on an at-fault collision with another road user, 0.5 on collisions with static objects, and is 1 otherwise. The minimum value across all timesteps is taken.

\noindent \textbf{Drivable Area Compliance (DAC, multiplicative).} DAC is 0 if any corner of the ego bounding box lies outside all drivable polygons at any timestep, otherwise 1.

\noindent \textbf{Driving Direction Compliance (DDC).} Multiplicative in v2, disabled in v1. It accumulates oncoming-direction motion per timestep, with the ego displacement set to 0 when the ego center is on a route lane or the ego is in an intersection. The maximum oncoming progress within a sliding window is then thresholded into a score of $\{0, 0.5, 1\}$.

\noindent \textbf{Traffic Light Compliance (TLC, v2 only, multiplicative).} TLC is 0 if the ego polygon intersects any red-light token at any timestep, otherwise 1.

\noindent \textbf{Ego Progress (EP, weight 5).} The raw progress is the centerline-projected displacement along the trajectory, clipped to be non-negative. It is normalized by a per-scene constant equal to the maximum raw progress across the agent and the PDM reference trajectory after multiplying by the multiplicative score, so that a trajectory that fails a multiplicative gate does not raise the normalization constant. EP defaults to 1 when the reference progress is too little to be informative.

\noindent \textbf{Time-to-Collision (TTC, weight 5).} At each timestep, the ego is forward-projected at constant velocity and heading over a $1.0$\,s window, and the resulting expanded polygon is checked against the same at-fault rules as NC. TTC is 0 if any projected polygon collides at fault, otherwise 1. A stopped ego and previously collided objects are skipped.

\noindent \textbf{Comfort (C, v1 only, weight 2).} Six dynamic quantities (longitudinal acceleration, lateral acceleration, total jerk, longitudinal jerk, yaw rate, yaw acceleration) are computed on the simulated states with a Savitzky-Golay filter and thresholded against fixed bounds. C is 1 only if all six quantities stay within bounds at every timestep.

\noindent \textbf{History Comfort (HC, v2 only, weight 2).} HC replaces v1's comfort. The simulated states are prepended with the past human ego states before applying the same six-quantity comfort check. HC penalizes discontinuities at the boundary between history and the planned trajectory, where a velocity or acceleration mismatch produces a jerk spike under the Savitzky-Golay filter.

\noindent \textbf{Two-Frame Extended Comfort (EC, v2 only, weight 2).} EC is computed across two consecutive scene frames. The overlapping portion of their simulated state sequences is used to compute root-mean-square differences in acceleration magnitude, jerk magnitude, yaw rate, and yaw acceleration. EC is 1 if all four RMS differences stay below their thresholds, otherwise 0. EC is undefined for the first frame in a scene, in which case the EC term is dropped from the weighted average.

\noindent \textbf{Lane Keeping (LK, v2 only, weight 2).} The lateral distance from the ego center to the lane centerline is computed at each timestep with intersections excluded. LK is 0 if the deviation exceeds a small threshold for a sustained duration, otherwise 1.

\noindent \textbf{Human Penalty Filter (v2 only).} A post-processing step that re-runs the human reference trajectory through the same scoring pipeline and overrides the agent's score to 1 on any subscore where the human also fails. This forgives failures that are unavoidable given the scene.

\subsection{Aggregation}\label{app:pdm_agg}

\noindent \textbf{NAVSIM v1.} The PDM score combines the multiplicative metrics NC and DAC with the weighted metrics EP, TTC, and C,
\begin{equation}\label{eq:pdms}
    \mathrm{PDMS} = \mathrm{NC} \cdot \mathrm{DAC} \cdot \frac{5\,\mathrm{EP} + 5\,\mathrm{TTC} + 2\,\mathrm{C}}{12}.
\end{equation}
DDC is technically present in v1 but enters with weight 0, so it does not contribute.

\noindent \textbf{NAVSIM v2.} The Extended PDM score promotes DDC to a multiplicative gate, adds TLC as a second new gate, replaces C with HC, and introduces LK and EC as additional weighted terms,
\begin{equation}\label{eq:epdms}
    \mathrm{EPDMS} = \mathrm{NC} \cdot \mathrm{DAC} \cdot \mathrm{DDC} \cdot \mathrm{TLC} \cdot \frac{5\,\mathrm{EP} + 5\,\mathrm{TTC} + 2\,\mathrm{LK} + 2\,\mathrm{HC} + 2\,\mathrm{EC}}{16}.
\end{equation}
When EC is undefined for a frame (no previous adjacent scene), the EC term is dropped from the numerator and the denominator becomes 14.

\subsection{Differences Between v1 and v2}\label{app:v1v2}

The two versions share most subscore definitions but differ in five places.
\begin{compactitem}
    \item DDC is promoted from a weight-0 weighted term in v1 to a multiplicative gate in v2, and additionally excludes intersections from oncoming progress accumulation.
    \item TLC is added in v2 as a second new multiplicative gate.
    \item Comfort (C) in v1 is replaced by History Comfort (HC) in v2, which prepends 1.5\,s of past human ego states before the comfort check, and Two-Frame Extended Comfort (EC) is added as a separate weighted term that measures dynamic consistency across consecutive scene frames.
    \item Lane Keeping (LK) is added in v2 as a weighted term.
    \item Human penalty filter is added in v2.
\end{compactitem}
EP and TTC keep the same form in both versions up to minor implementation differences in the normalization and the iteration bound.

\section{Implementation Details}\label{app:impl}

This section provides the full implementation details of \ourmethod{}, organized in the same way as the method section. Sec.~\ref{app:reward_construction}: reward construction, Sec.~\ref{app:architecture}: model architecture, Sec.~\ref{app:training}: two-stage training recipe, Sec.~\ref{app:inference_details}: inference configuration.

\subsection{Reward Construction}\label{app:reward_construction}

\noindent \textbf{Dense Action-Reward Pairs.} We follow scoring-based methods~\cite{chen2024vadv2,li2024hydra,li2025generalized} and use a dense action vocabulary $\mathcal{V}_a$ of 8192 four-second trajectories obtained by clustering 700K trajectories from nuPlan~\cite{Caesar2021nuplan}, each containing 8 ego-frame waypoints at 0.5\,s spacing. For every \texttt{navtrain} scene, we simulate all vocabulary trajectories through the NAVSIM simulator and record the resulting reward labels, producing dense action-reward pairs $(a,r)$ that span the action space.

\noindent \textbf{Scalar Rewards.} The scalar rewards in $\mathcal{R}$ are the NAVSIM v2 subscores NC, DDC, TLC, EP, LK, HC and the overall PDM score (Sec.~\ref{app:metrics}). Two-Frame Extended Comfort is excluded as it is not defined within a single scene. EP is replaced by a safety-oriented variant. We first gate it by a safety mask that is $1$ when both NC$=1$ and TTC$=1$ and $0$ otherwise, then divide by the maximum masked EP across $\mathcal{V}_a$ in the scene.

\noindent \textbf{Per-Timestep Array Rewards.} We replace the binary DAC and TTC subscores with per-timestep arrays that preserve the hard constraint at higher temporal resolution. The arrays are computed with the adapted NAVSIM v1 PDM scorer.
\begin{compactitem}
    \item \textbf{TTC-time array.} At each timestep, the ego is forward-projected at constant velocity over a $1.5$\,s detection horizon and we record the minimum projected collision time, set to the horizon when no collision is detected. The array is sampled at 40 timesteps, yielding a 40-dim array per trajectory.
    \item \textbf{Ego-area array.} At each timestep we record two binary flags, on-road (the ego polygon lies inside the drivable area) and on-route (the ego center is on the route lane consistent with the driving direction), sampled at 8 timesteps to yield an $(8,2)$ array per trajectory.
\end{compactitem}

\noindent \textbf{Continuous-Reward Noise Augmentation.} During training, we corrupt EP and the PDM score by adding Gaussian noise from $\mathcal{N}(0,\sigma^2)$ with $\sigma=0.05$, acting as label smoothing on these dense scalars. No noise is added at inference.

\noindent \textbf{Simulation Cost.} Reward construction is a one-time offline cost. Following the GTRS~\cite{li2025generalized} protocol, scoring $\mathcal{V}_a$ on \texttt{navtrain} for NAVSIM v2 scalar rewards takes around one day on 32 machines. We compute the per-timestep rewards on 4 nodes of 8 machines, taking approximately 8 hours. Simulation runs on CPU only and does not require GPU resources.

\subsection{Model Architecture}\label{app:architecture}

\subsubsection{Perception Encoder}\label{app:perception}

We adopt the Transfuser~\cite{chitta2022transfuser} backbone without modification. The backbone outputs a set of BEV feature tokens, which we concatenate with an encoded status feature that embeds the ego status and driving command. Positional encodings are added to the concatenated tokens to form the context tokens. A set of agent tokens is then queried from the context tokens through cross-attention. Together, the context tokens and agent tokens form the scene features $\vs$ used by the downstream action decoder.

Two auxiliary losses follow the Transfuser recipe. A BEV semantic segmentation loss supervises the backbone BEV features, and an agent prediction loss is applied to the queried agent tokens. We keep the same imitation-learning (IL) head of Transfuser, an MLP supervised with the L1 loss against the GT trajectory, that produces the optional inference-time anchor (Sec.~\ref{sec:inference}). These three losses sum into $\mathcal{L}_{\mathrm{perc}}$.

\subsubsection{Reward Encoder}\label{app:reward_encoder}

The reward encoder maps each reward in $\mathcal{R}$ into a 256-dim embedding using a per-reward embedder selected by reward type. Discrete rewards use learnable embedding tables, scalar continuous rewards use sinusoidal positional encoding, and the per-timestep array rewards (TTC-time, ego-area) use an MLP. Each reward also has a learned null token $\vn_k$ that replaces its embedding when the reward is dropped. The resulting embeddings are concatenated and passed through an MLP to produce the 256-dim condition embedding $\vr_c$.

\noindent \textbf{Reward Dropout Policy.} We use a three-mode dropout schedule during training. With probability $0.5$ we keep all rewards, with probability $0.1$ we drop all rewards together to form the empty condition $r_\emptyset$ used at inference, and with probability $0.4$ we drop each reward independently with per-reward probability $0.5$. The all-drop event ensures that $r_\emptyset$ is observed often enough for the model to learn the unconditional distribution required by classifier-free guidance, while the independent dropout exposes the model to a wide range of partial conditions for arbitrary subset conditioning at inference.

\subsubsection{Flow-based Action Decoder}\label{app:decoder}

\noindent \textbf{Trajectory Representation.} The decoder operates on 4-second trajectories represented as 8 waypoints ($x,y,\theta$) where $\theta$ is the ego heading. The $(x,y)$ is normalized to $[-2, 2]$, and the heading is decoupled into $(\sin\theta, \cos\theta)$. So the shape of a trajectory is $8\times 4$.

\noindent \textbf{Architecture.} We stack four transformer blocks with hidden dim 256, FFN dim 1024, and 8 attention heads. The input trajectory is encoded by a sinusoidal positional embedding followed by an MLP projected to hidden dim, and a learned positional embedding is added along the waypoint dimension. Each block applies, in order, self-attention, cross-attention to the context tokens, cross-attention to the agent tokens, and an FFN, with residual connections around each module. AdaLN modulation is applied before every attention and FFN module, and dropout is added around the agent cross-attention to reduce overfitting. After the final block, a linear head projects each waypoint feature back to 4 to produce the predicted clean trajectory.

\noindent \textbf{Conditioning.} The time $t$ is embedded by sinusoidal positional encoding followed by a 2-layer MLP. The AdaLN modulation input is the concatenation of the time embedding and the reward embedding $\vr_c$, projected per block into the scale and shift parameters.

\subsubsection{Mode Selector}\label{app:selector}

\noindent \textbf{Architecture.} The mode selector is a lightweight two-layer transformer that scores trajectory proposals from the action decoder. It first encodes the input trajectory with sinusoidal positional embeddings, then applies the same transformer block as the action decoder with three modifications. We drop the trajectory self-attention module, replace AdaLN with ordinary layer normalization, and add a grid-sample-based cross-attention module that attends to the BEV features. The output feature is mapped to a set of subscore predictions through shallow MLP heads.

\noindent \textbf{Prediction Heads.} We attach one prediction head per NAVSIM subscore, covering NC, DAC, TTC, EP, and HC. Two auxiliary heads predict the per-timestep TTC-time and ego-area arrays used in the reward condition (Sec.~\ref{sec:reward}). The auxiliary heads increase the resolution of the selector's awareness of safety and rule compliance. The auxiliary heads are enabled only in stage 2 (App.~\ref{app:training_stage2}); stage 1 trains the five subscore heads only (App.~\ref{app:training_stage1}). Our selector score follows the NAVSIM PDMS aggregation formulation rather than replicating it exactly. In particular, we predict v2 HC instead of v1 Comf., as the v1 Comf. computation has a known issue that labels almost all candidate trajectories as $1$ and provides little ranking signal, whereas v2 HC corrects this by prepending past ego states before the comfort check.

\noindent \textbf{Continuous TTC Label.} The official TTC subscore in NAVSIM is binary and gives no gradient near the decision boundary. We replace it with a continuous label $\bar{s}_{\mathrm{TTC}} = \min(ttc_{\mathrm{min}}, t_{\mathrm{bound}})/t_{\mathrm{bound}}$, where $ttc_{\mathrm{min}}$ is the minimum projected collision time across the trajectory horizon and $t_{\mathrm{bound}}$ is the maximum detection bound. The label takes value $0$ at an actual collision and $1$ when no collision is detected within $t_{\mathrm{bound}}$ at any timestep. We set $t_{\mathrm{bound}}=2$ seconds in practice.

\noindent \textbf{Training Loss.} The five subscore heads (NC, DAC, TTC, EP, HC) and the per-timestep ego-area auxiliary head are supervised with binary cross-entropy (BCE) loss. The per-timestep TTC-time auxiliary head is supervised with MSE since its target is not bounded in $[0,1]$. The total selector loss is the sum of all head losses,
\begin{equation}
    \mathcal{L}_{\mathrm{sel}} = \sum_{k} \mathcal{L}_k,
\end{equation}
where $\mathcal{L}_k$ is the BCE or MSE term for head $k$.

\noindent \textbf{Final Score Aggregation.} Following the official NAVSIM aggregation, the final ranking score is
\begin{equation}\label{eq:agg}
    s_{\mathrm{final}} = \Big(\prod_{k\in\mathcal{M}} \hat{s}_k\Big) \cdot \Big(\sum_{k\in\mathcal{W}} w_k \hat{s}_k\Big),
\end{equation}
where $\mathcal{M}=\{NC,DAC\}$ are the multiplicative metrics and $\mathcal{W}=\{EP,TTC,HC\}$ are the weighted metrics. We set all weights $w_k$ to $1$ to equally treat all weighted scores. The final performance of \ourmethod{} is stable under different weight choices (App.~\ref{app:scorer_weights}).

\subsection{Training Details}\label{app:training}

\subsubsection{Input Construction}\label{app:training_inputs}

The image input is a $1024\times 256$ stitched view from the front, front-left, and front-right cameras. The LiDAR input aggregates four frames at 0.5\,Hz, covering a 2-second history from $-1.5$\,s to the current frame. History points are transformed into the current ego frame and rasterized into 2D BEV histograms based on local point statistics. We use both above-ground and below-ground points, producing an 8-channel BEV grid as the LiDAR input.

\subsubsection{Trajectory Sampling}\label{app:idsw}

The dense action vocabulary contains many trajectories that fail safety or compliance checks in any given scene, so the empirical PDM-score distribution over all $(a,r)$ pairs is heavily skewed toward zero. Uniform sampling from this set under-represents the rare high-score samples that the decoder needs to model the high-quality region of $p(a|r)$. To rebalance, we sample $(a,r)$ pairs with weight inversely proportional to the score density,
\begin{equation}\label{eq:idsw}
    w(s) \propto \frac{1}{p(s)^\alpha},
\end{equation}
where $p(s)$ is the empirical density of the PDM score $s$, estimated per scene with a 1D Gaussian KDE over the vocabulary scores, and $\alpha\in[0,1]$ controls the strength of the rebalancing. A small constant is added to $p(s)$ for numerical stability before inverting. At $\alpha=0$, $w$ reduces to a constant and sampling is uniform across all trajectories. At $\alpha=1$, the score density is fully canceled and sampling is uniform across score bins. We use $\alpha=0.6$ in practice, which lifts the rare high-score samples while still retaining enough common low-score actions to model the full distribution.

At each training step, we separately sample 20 trajectories for the action decoder and the mode selector using this weighting. 

\subsubsection{Stage 1: End-to-End Training}\label{app:training_stage1}

We train all components jointly on \texttt{navtrain} for 100 epochs with AdamW (default $\beta$, weight decay $10^{-4}$). The learning rate follows a cosine schedule from $3\times 10^{-4}$ to $10^{-6}$ with a 3-epoch warmup. We use a total batch size of 64 across 4 NVIDIA H20 GPUs.

The full stage-1 objective is a weighted sum of per-component losses with weights chosen to balance their gradient scales. The velocity-matching loss $\mathcal{L}_{\mathrm{dec}}$ has weight 40. Inside $\mathcal{L}_{\mathrm{perc}}$, the agent class and box prediction losses have weights 10 and 1, the BEV semantic loss has weight 14, and the IL head L1 loss has weight 10. The selector loss $\mathcal{L}_{\mathrm{sel}}$ has weight 10.

In stage 1, we mainly treat the mode selector supervision as an auxiliary signal to refine scene features, and use a reduced version of selector loss. The per-timestep TTC-time and ego-area heads are not included, and the mode selector is trained to predict v2 subscores NC, DAC, EP, TTC, and HC.

\noindent \textbf{Numerical Stability of $\mathcal{L}_{\mathrm{dec}}$.} The velocity conversion $\vv_\theta = (\vx_\theta - \vz_t)/(1-t)$ in Sec.~\ref{sec:model} diverges as $t\to 1$. To stabilize training, we clip the denominator to a minimum of $0.05$, computing $\vv_\theta = (\vx_\theta - \vz_t)/\max(1-t, 0.05)$ when sampling $t$ near 1. The clip is loss-only and does not affect the inference ODE solver.

\subsubsection{Stage 2: Mode Selector Finetune}\label{app:training_stage2}

In stage 2, we freeze every component except the mode selector and finetune the selector to close the gap between vocabulary trajectories seen at stage-1 and decoder proposals seen at inference. We train with a batch size of 256 across 8 H20 GPUs for 2 epochs. Other settings match stage 1.

For each training scene, we generate decoder proposals using the same sampling configuration as inference, and combine them with 32 random vocabulary trajectories drawn under inverse-density weighting. Generated proposals are scored online by the NAVSIM simulator to obtain ground-truth subscores. The selector is then trained to predict these subscores together with the per-timestep TTC-time and ego-area arrays under $\mathcal{L}_{\mathrm{sel}}$ defined in App.~\ref{app:selector}.

\subsection{Inference}\label{app:inference_details}

\subsubsection{Default Sampling Configuration}\label{app:inference_hparams}

Unless otherwise stated, all experiments share the following sampling configuration. We use the Euler scheduler with 20 denoising steps, discretizing time $t\in[0,1]$. The sampling uses the CFG scale $w_g{=}5$ and takes the IL head output as the anchor for zero-shot anchored sampling. The two sampling controls $r_{\mathrm{high}}$ and $t_{\mathrm{init}}$ are drawn independently and uniformly per proposal, with the target score in $[s_{\min},s_{\max}]{=}[0.9,1.0]$ and the initial noise level in $[t_{\min},t_{\max}]{=}[0.5,0.9]$. We generate 60 proposals and rank them with the mode selector under unit aggregation weights $w_k{=}1$ (App.~\ref{app:selector}). The same model trained once on \texttt{navtrain} is used for both NAVSIM v1 and v2 evaluation.

\subsubsection{Per-Experiment Sampling Settings}\label{app:inference_per_experiment}

A few experiments deviate from the default. The differences are limited to the number of proposals, the $t_{\min}$ lower bound, and how the single-proposal setting in Tab.~\ref{tab:navsim} is constructed. We summarize the settings in Tab.~\ref{tab:inference_settings}.

\begin{table}[ht]
\centering
\small
\caption{Inference settings per experiment. Default refers to the configuration in App.~\ref{app:inference_hparams}.\label{tab:inference_settings}}
\begin{tabular}{lccc}
\toprule
Experiment & \# proposals & $r_{\mathrm{high}}$ & $t_{\mathrm{init}}$ \\
\midrule
Tab.~\ref{tab:navsim} (multi-proposal) & 60 & default & default \\
Tab.~\ref{tab:navsim} (single-proposal) & 1 & fixed $0.95$ & fixed $0.5$ \\
Tab.~\ref{tab:navsim_v2} & 60 & default & $[0.75, 0.9]$ \\
Tab.~\ref{tab:proposal_quant} & 64 & default & default \\
Tab.~\ref{tab:ablation_reward_condition}, Tab.~\ref{tab:ablation_reward_augmentation} & 1 & default & default \\
Fig.~\ref{fig:topk_compare} & 64 & default & default \\
Fig.~\ref{fig:reward_ablation} & 1 & default & default \\
Fig.~\ref{fig:qual} & 64 & default & default \\
Fig.~\ref{fig:ab_cfg_scorer} & 1 & default & default \\
Fig.~\ref{fig:abl_sampling} (left) & 10 & varies (sweep) & varies (sweep) \\
Fig.~\ref{fig:abl_sampling} (right) & 16 & varies (sweep) & default \\
\bottomrule
\end{tabular}
\end{table}

The single-proposal column of Tab.~\ref{tab:navsim} is the only configuration that fixes $r_{\mathrm{high}}$ and $t_{\mathrm{init}}$ to representative values, since random sampling would inject noise into a single-shot evaluation. Every other single-proposal experiment draws the two controls from the default ranges, matching the behavior at multi-proposal inference. Tab.~\ref{tab:proposal_quant} uses 64 proposals to be comparable to iPad~\cite{guo2025ipad}.

\subsubsection{Reward Subset for Classifier-Free Guidance}\label{app:cfg_subset}

The high-reward condition $r_{\mathrm{high}}$ used for classifier-free guidance covers a subset of $\mathcal{R}$ rather than all reward signals. The subset comprises three scalar rewards (NC, HC, and the target PDM score) and the two per-timestep rewards (TTC-time and ego-area). All entries except the target score are set to their maximal values, signaling no collision, full history comfort, on-road and on-route at every timestep, and no near collision at every timestep. The target score is randomized over $[s_{\min},s_{\max}]$ and acts as the diversification axis described in Sec.~\ref{sec:inference}. The remaining rewards in $\mathcal{R}$ are dropped from $r_{\mathrm{high}}$ by replacing their embedding with the per-reward null token $\vn_k$ in Eq.~\ref{eq:reward}.

The choice of subset reflects what we can confidently prescribe at inference time. The optimal value of most rewards is scene-dependent and cannot be set in advance, so we leave them empty and let the model balance them through the learned $p(a|r)$. The PDM target score serves as an overall indicator for high-quality actions. We additionally pin the rewards that encode hard constraints, namely NC and the two per-timestep arrays for safety and drivable-area compliance, to enforce them across all proposals. HC is included as a stable, easy-to-satisfy regularizer that smooths the dynamics of the resulting trajectories.

\newcommand{\vizimgall}[1]{\includegraphics[width=0.19\linewidth]{#1}}
\newcommand{\romarkall}[1]{\rotatebox{90}{\parbox{2.5cm}{\centering\small\textbf{#1}}}}

\begin{figure*}[t]
\centering
\setlength{\tabcolsep}{1pt}
\renewcommand{\arraystretch}{0.3}
\begin{tabular}{@{}c cccc@{}}
& \textbf{DiffusionDrive} & \textbf{DiffusionDriveV2} & \textbf{iPad} & \textbf{\ourmethod{} (Ours)} \\[2pt]
\romarkall{Forward} &
\vizimgall{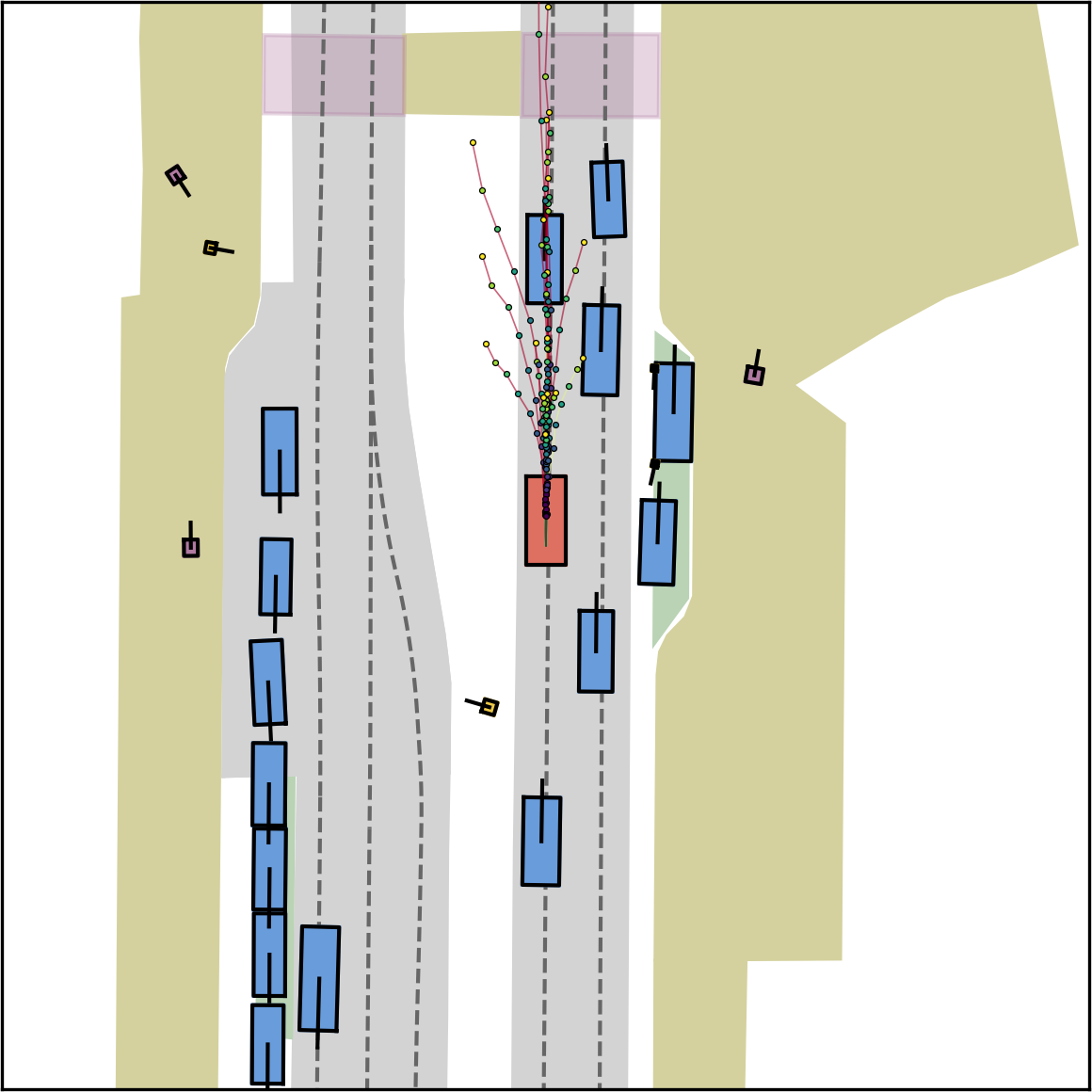} & \vizimgall{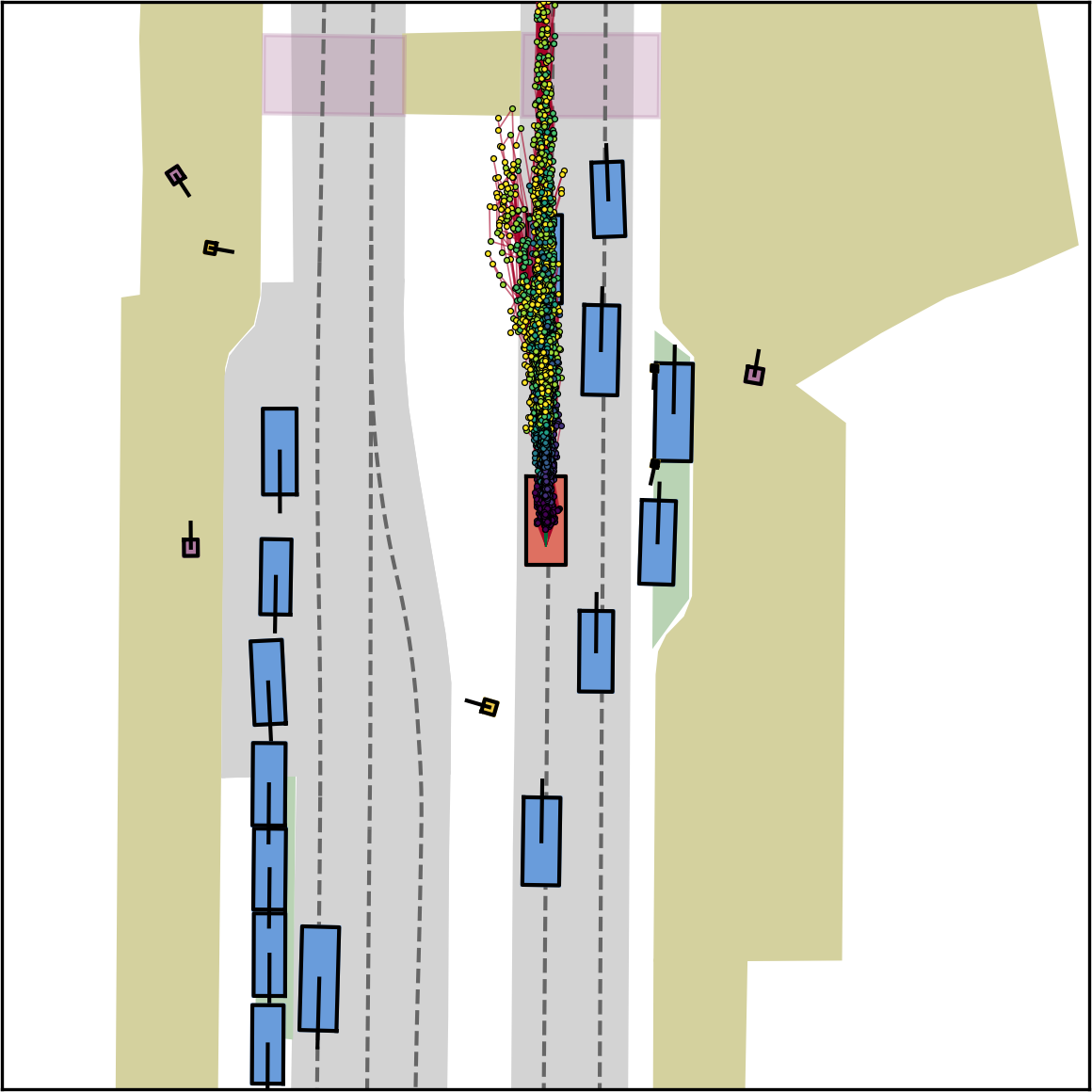} & \vizimgall{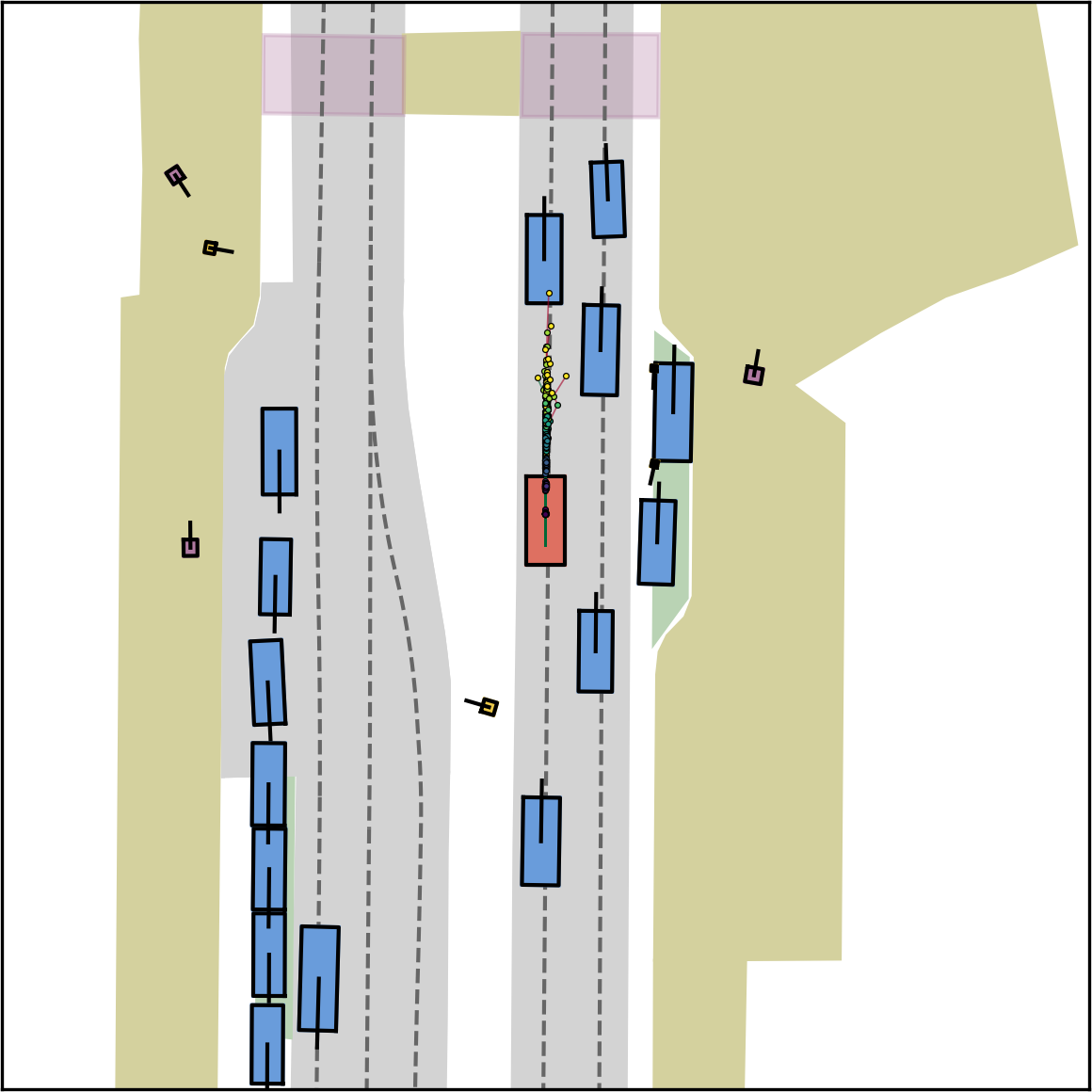} & \vizimgall{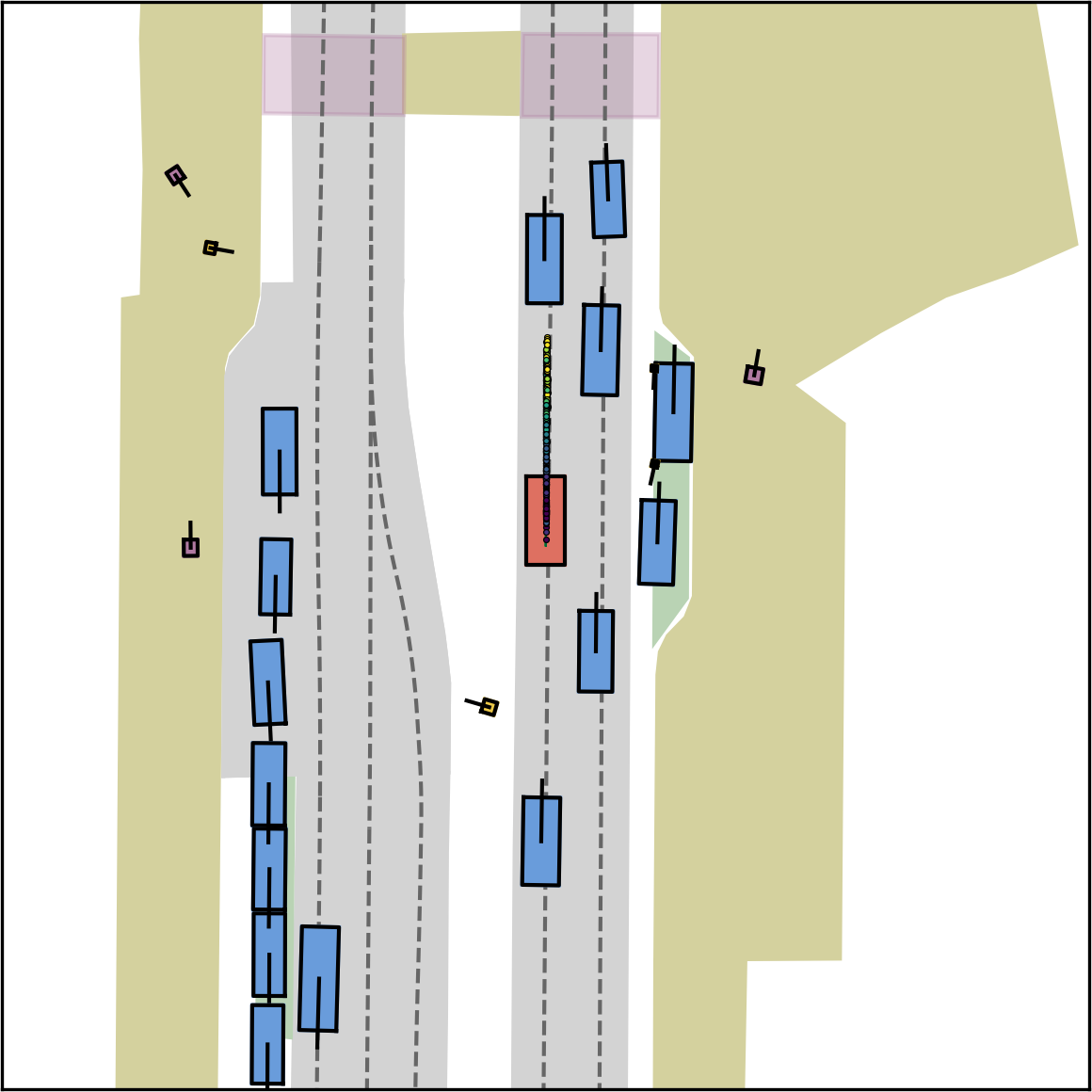} \\
 &
\vizimgall{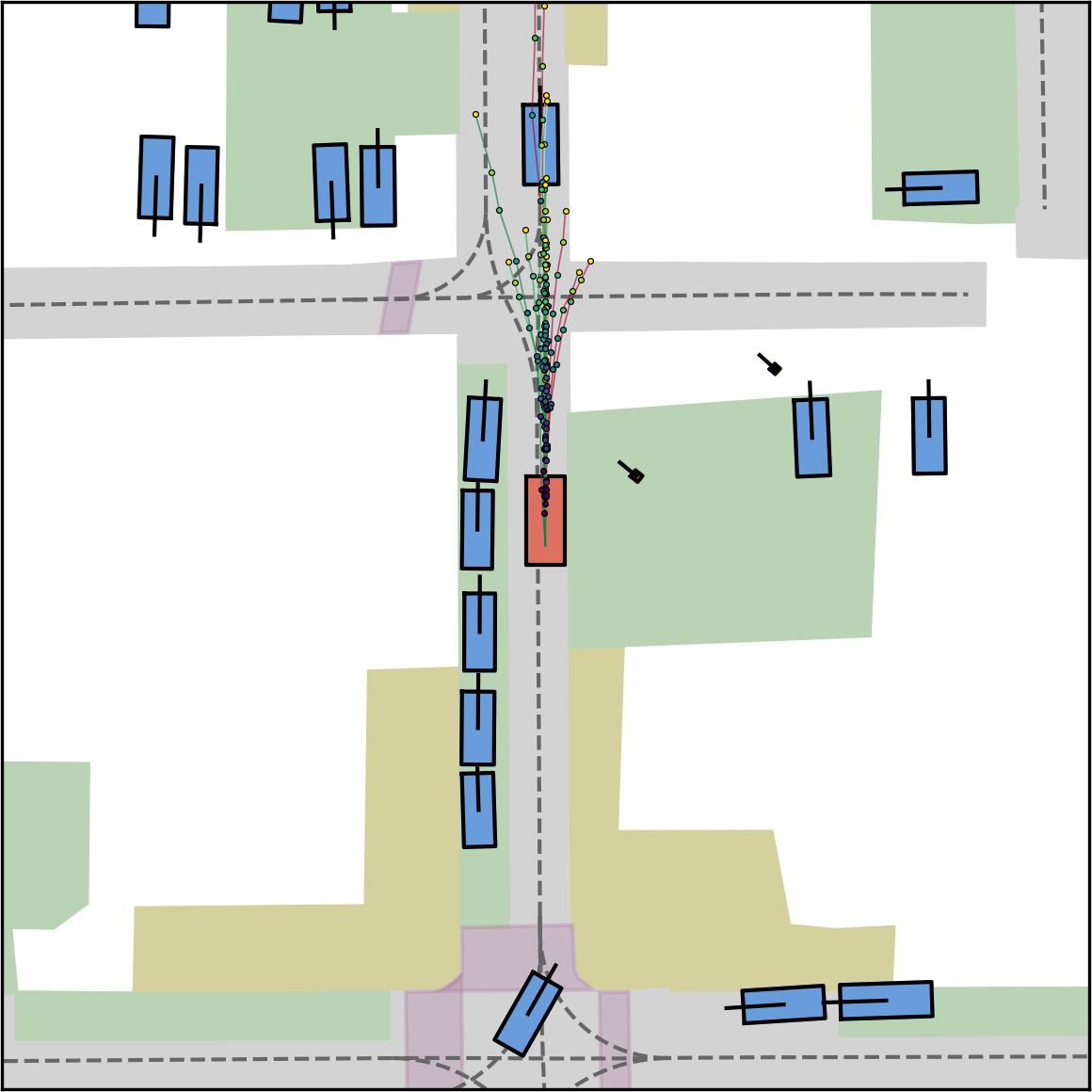} & \vizimgall{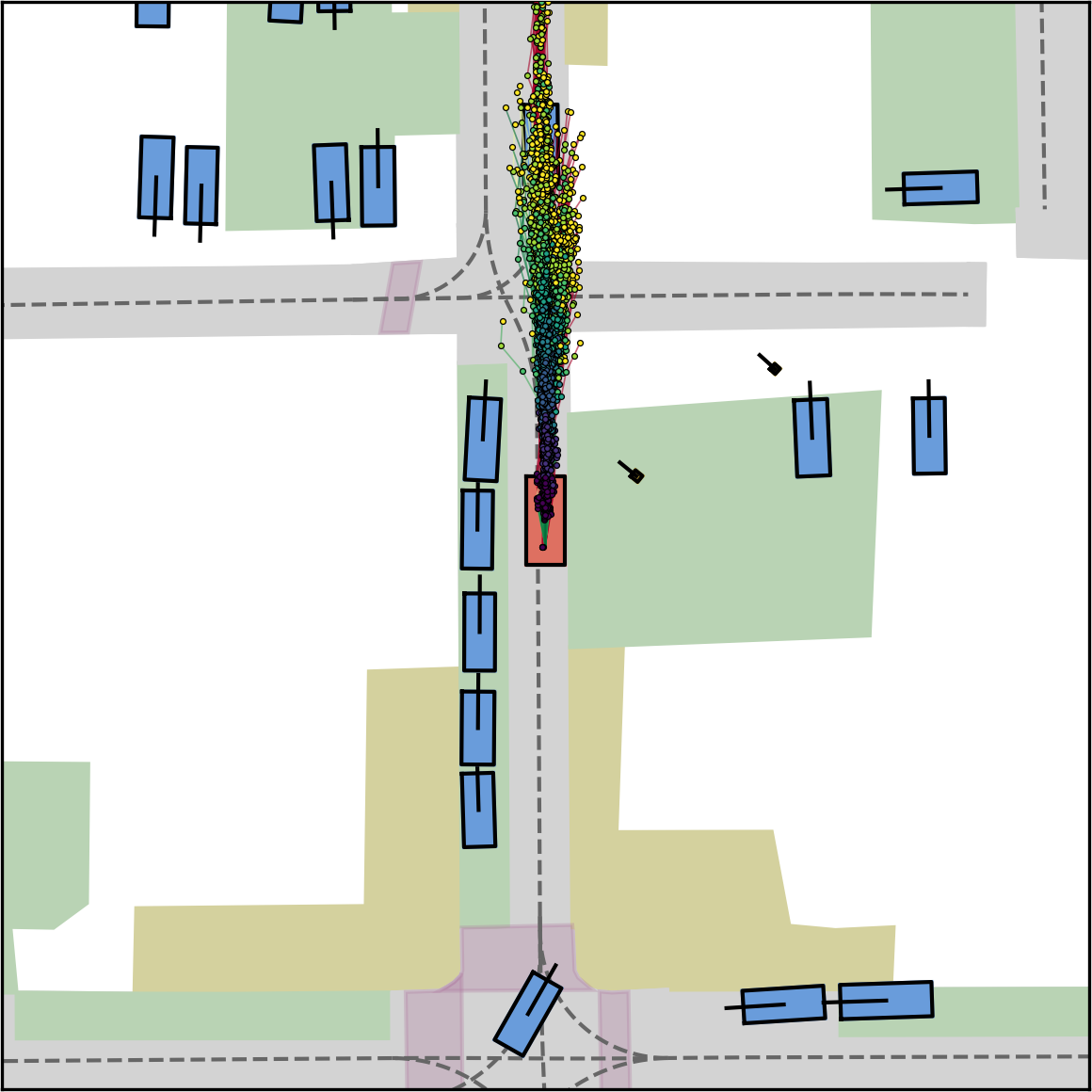} & \vizimgall{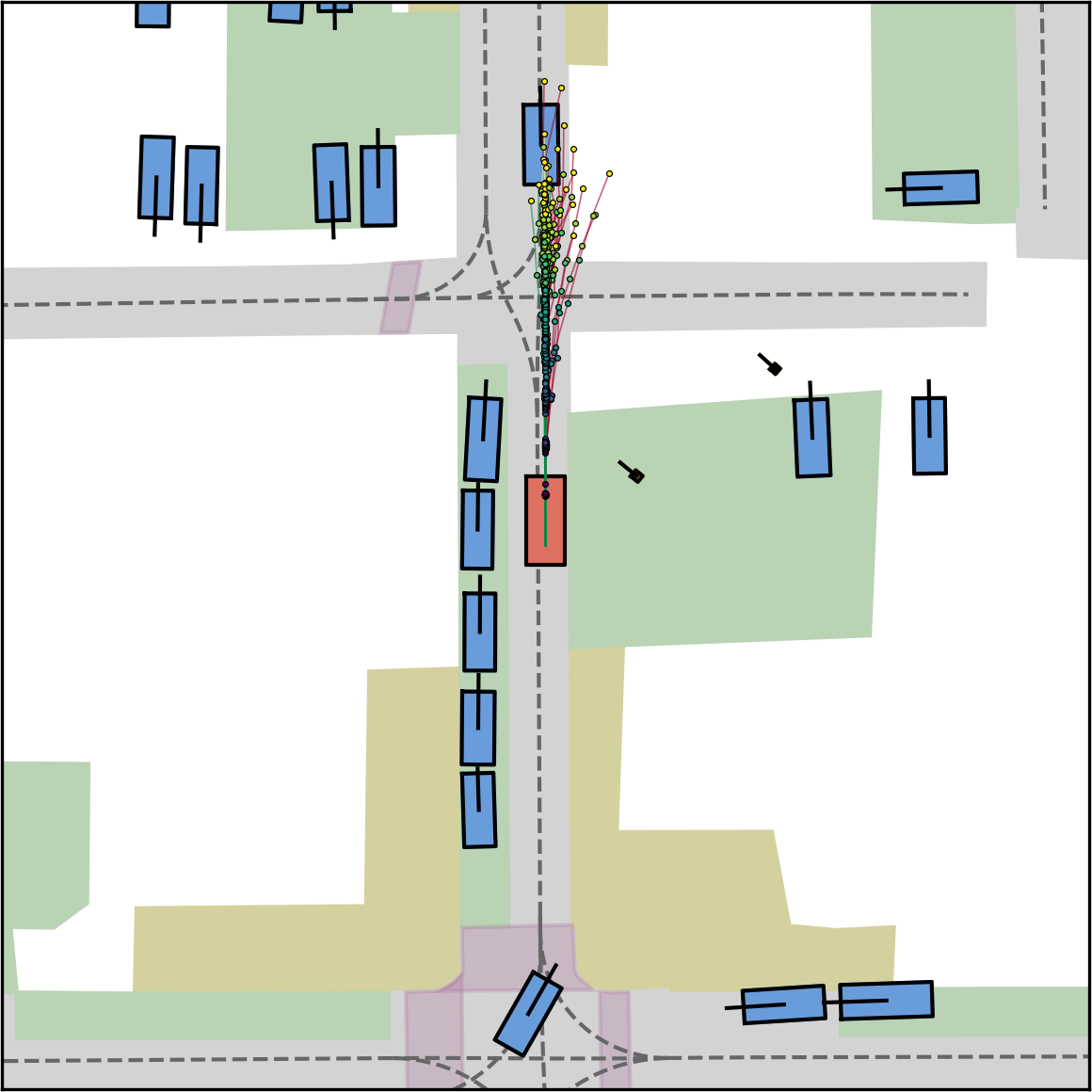} & \vizimgall{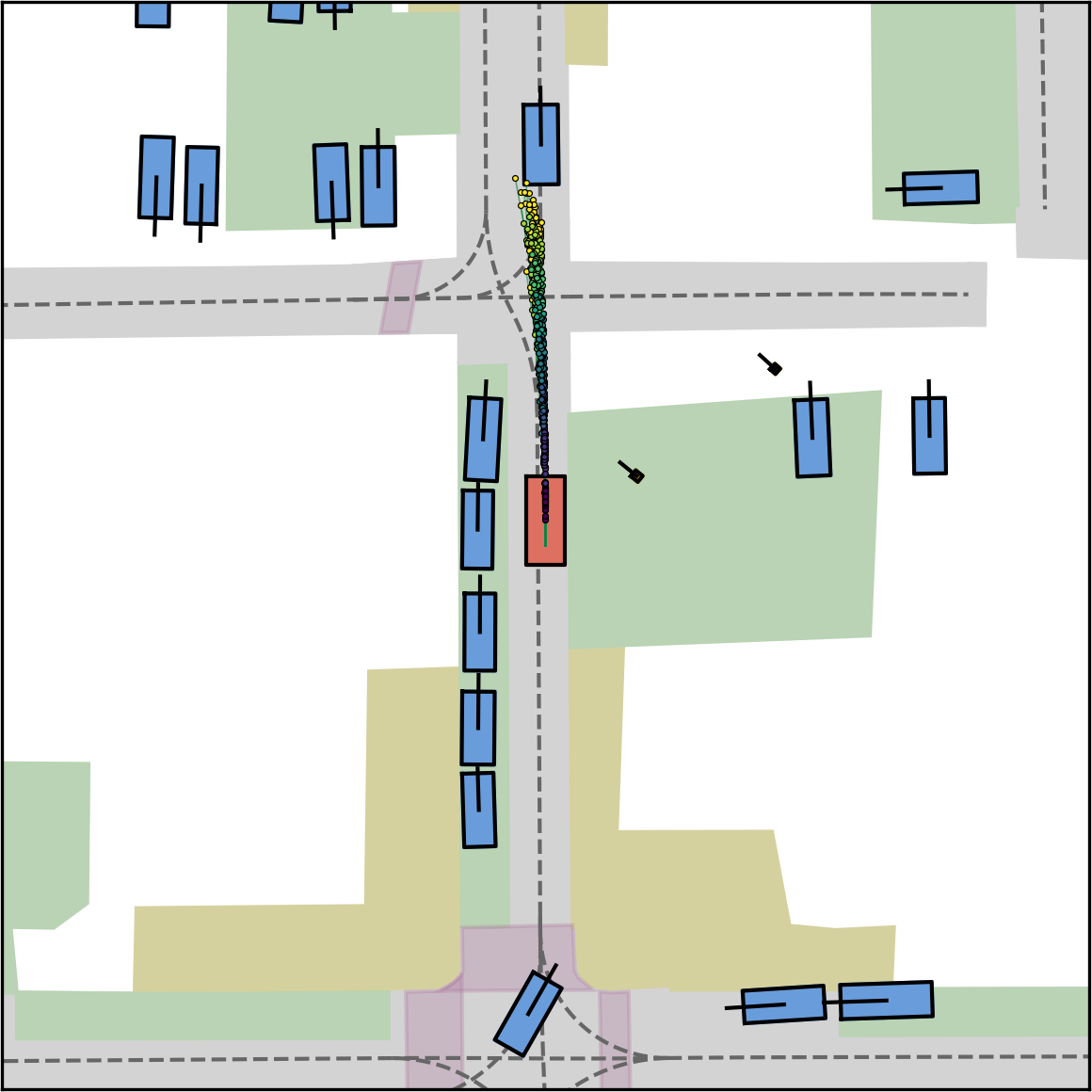} \\
 &
\vizimgall{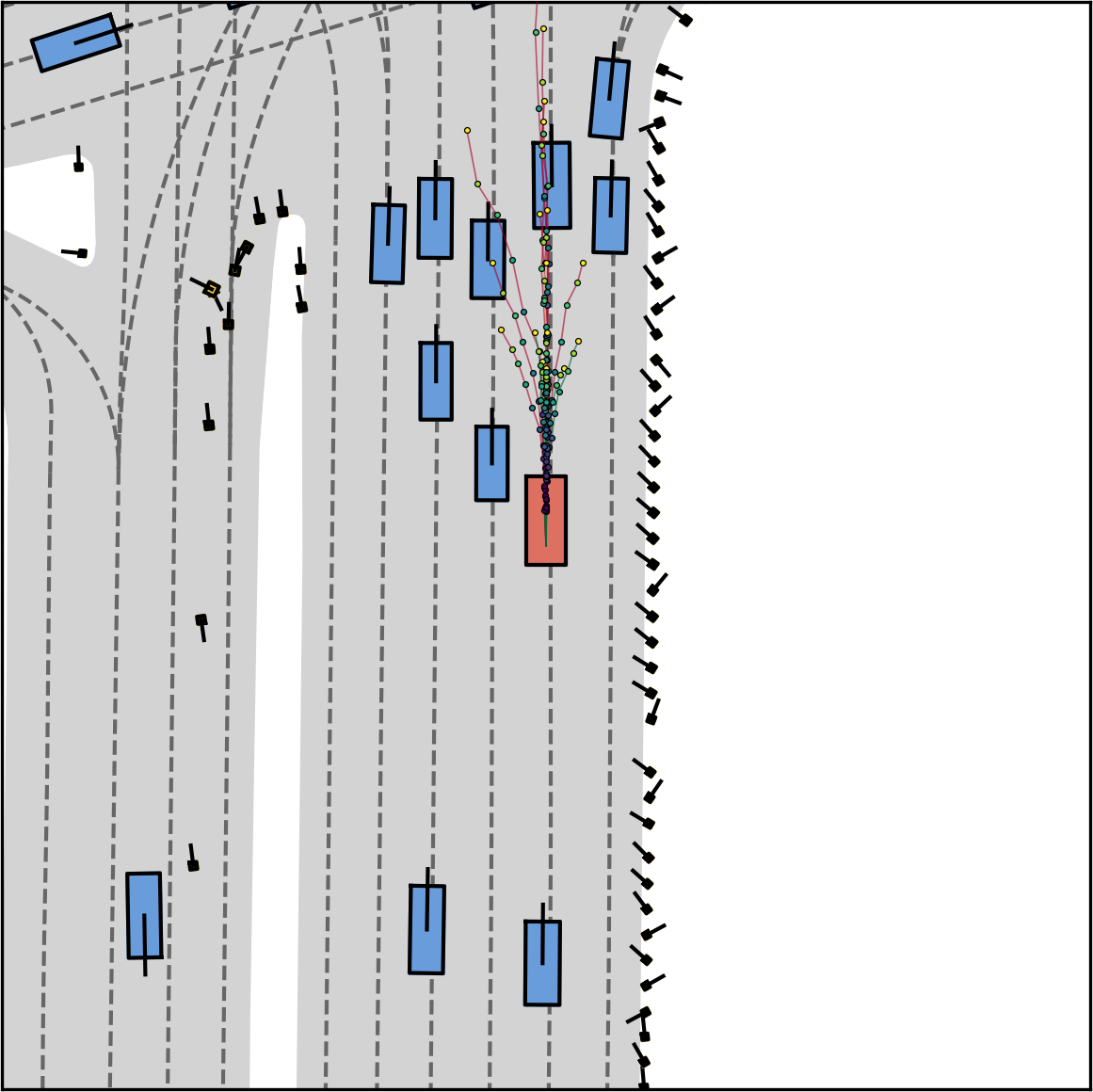} & \vizimgall{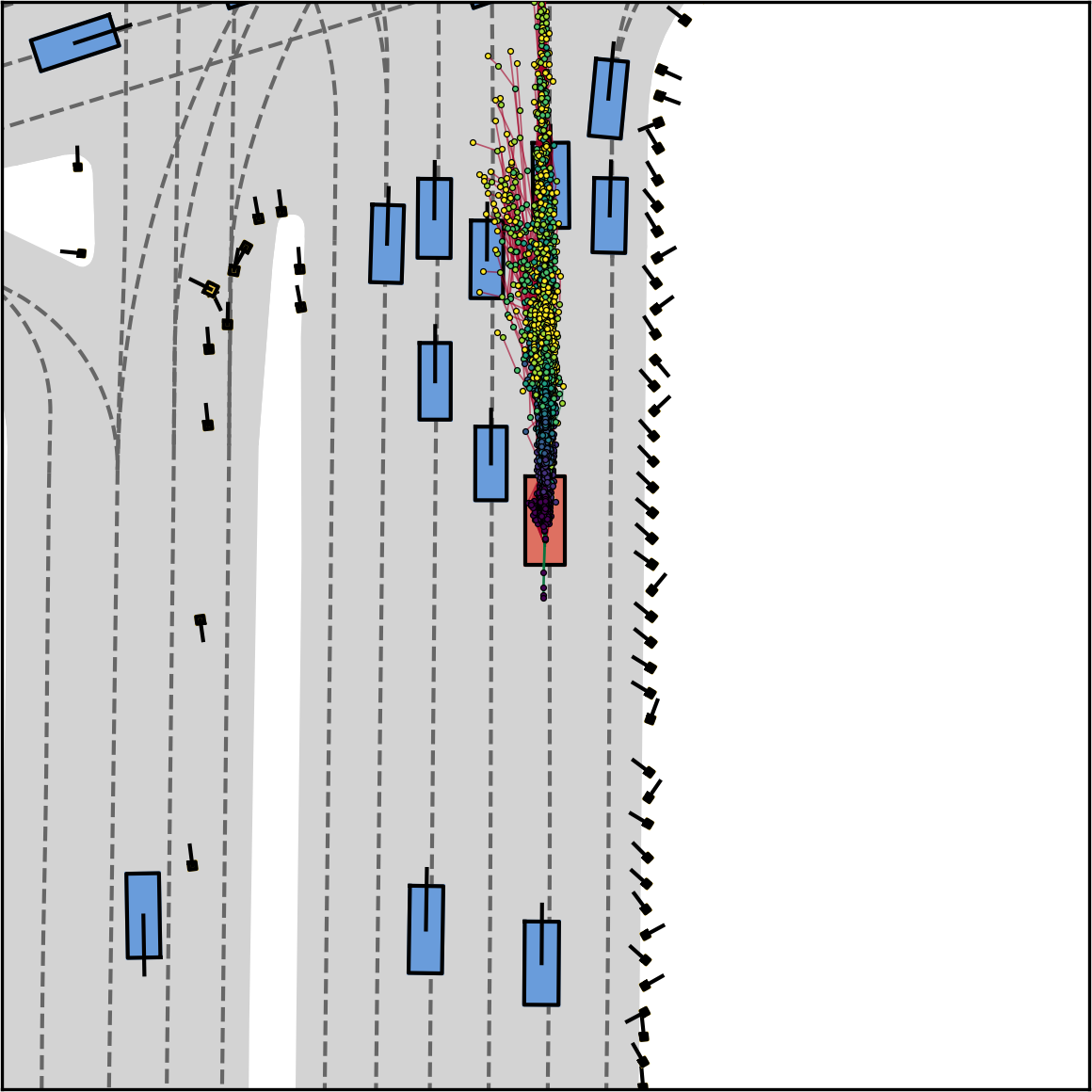} & \vizimgall{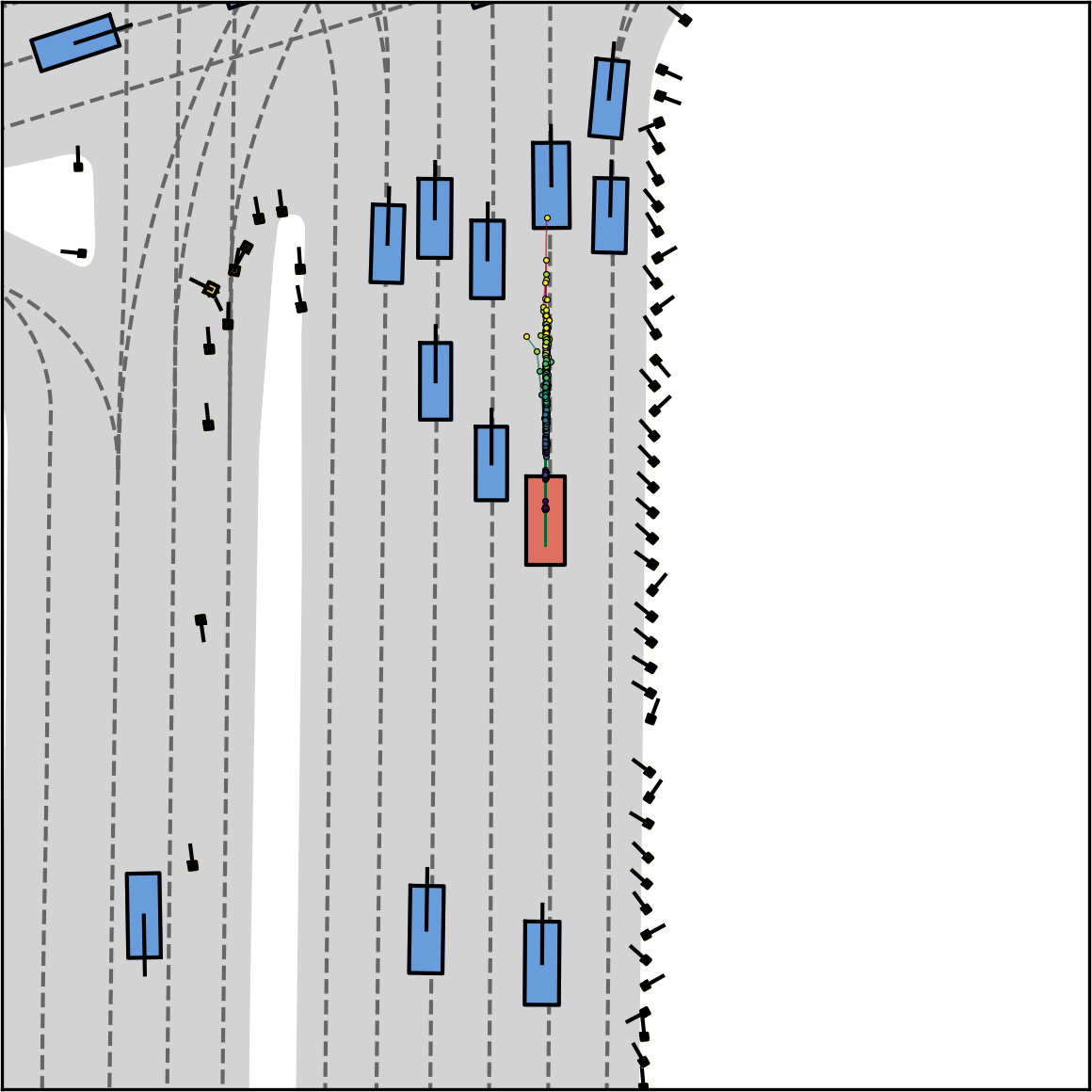} & \vizimgall{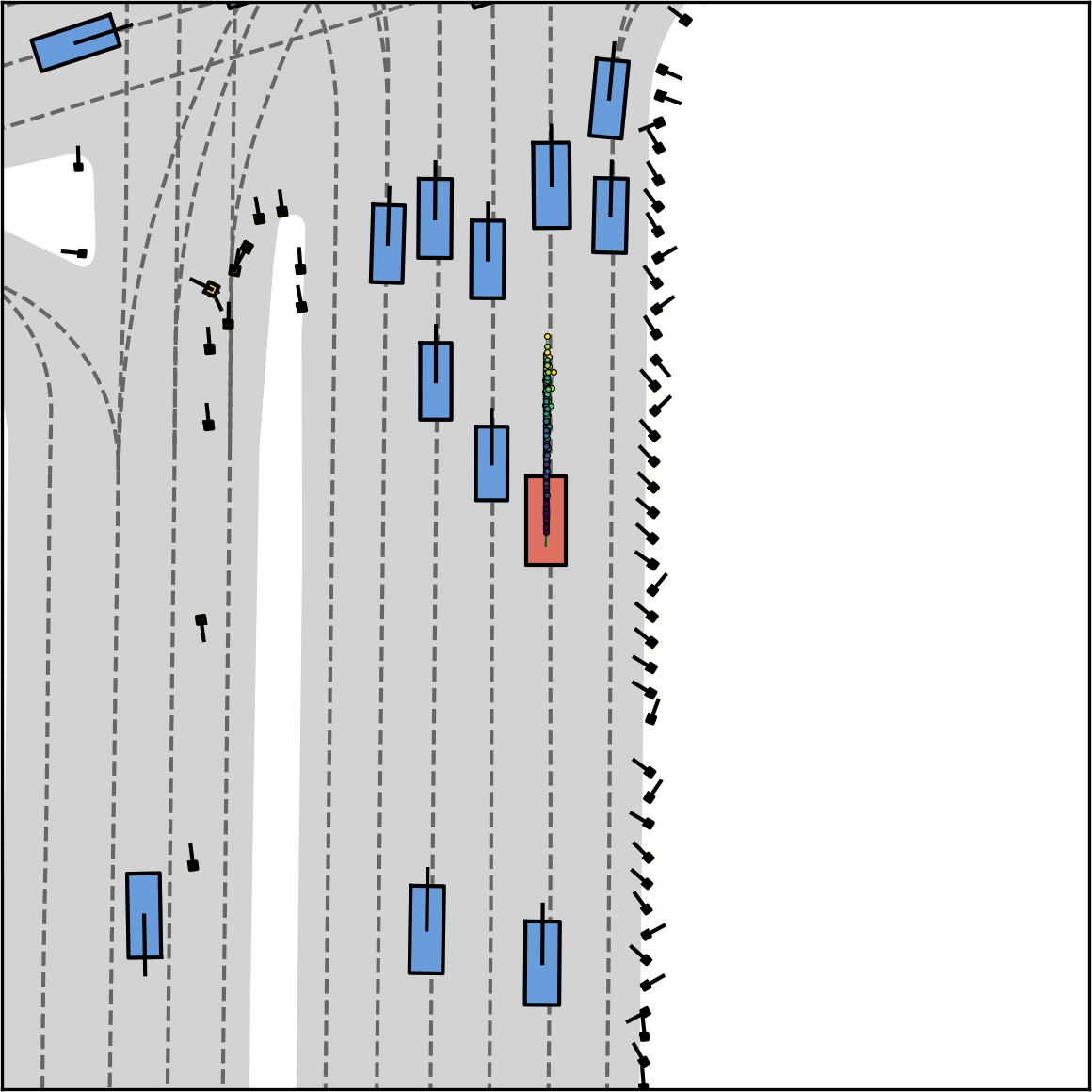} \\
 &
\vizimgall{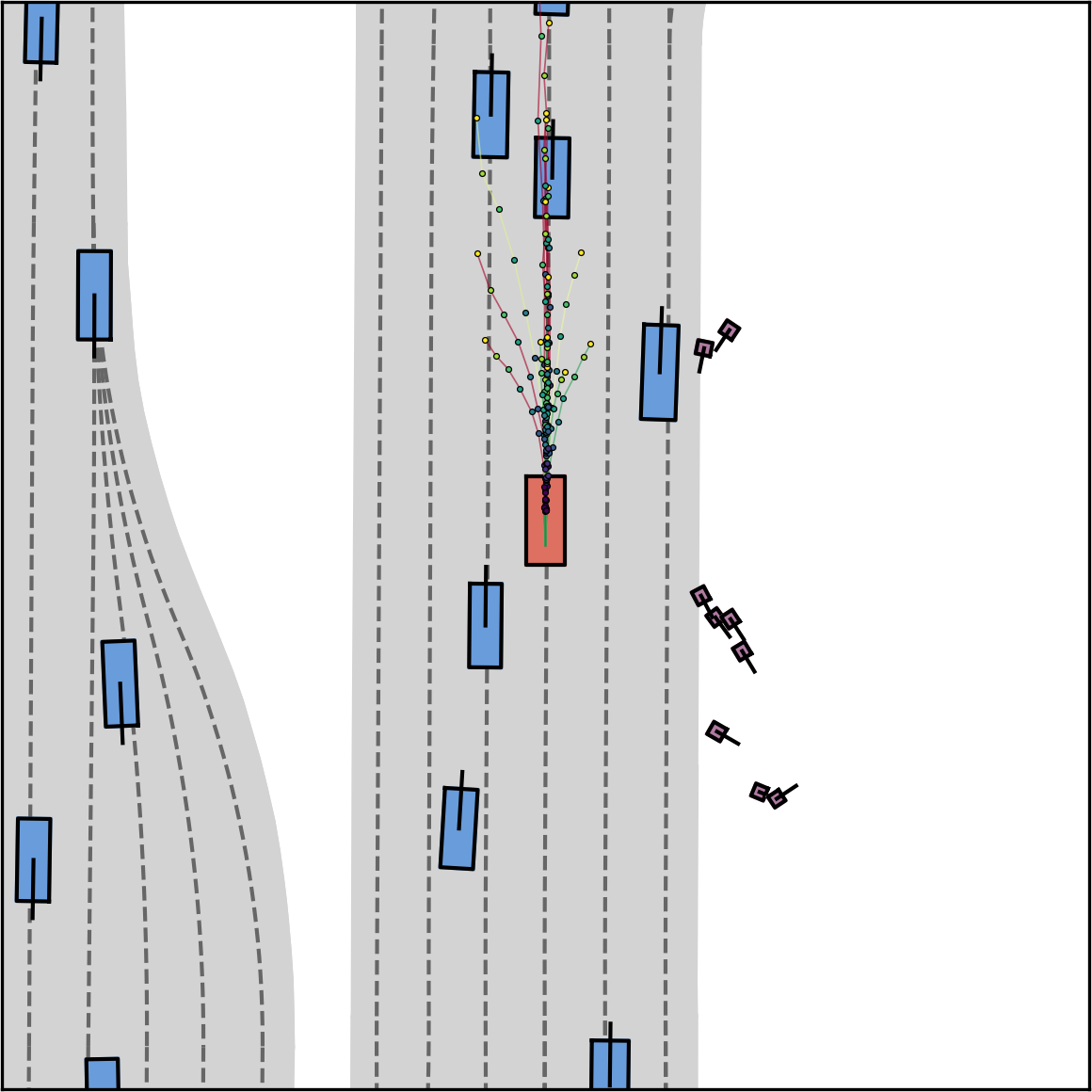} & \vizimgall{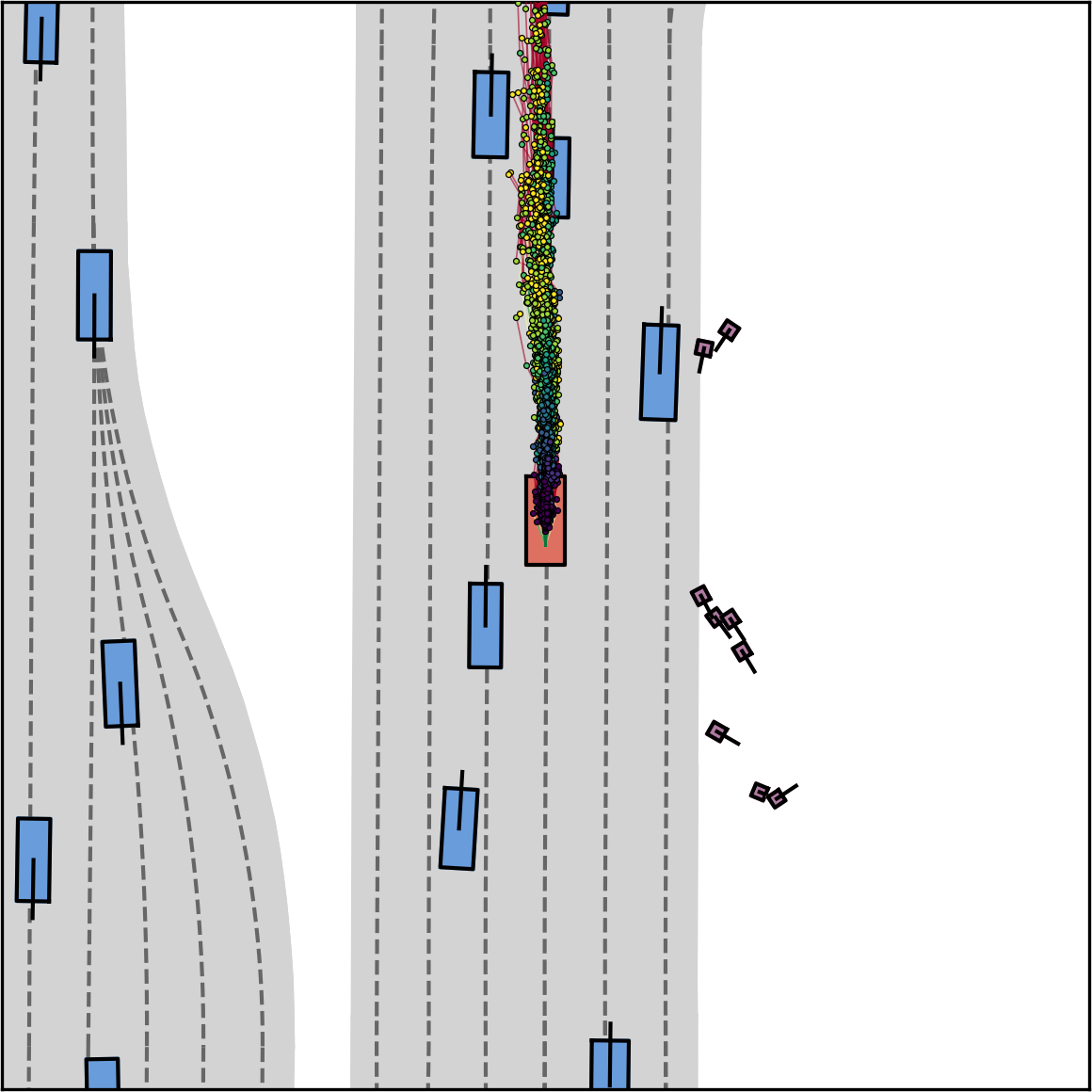} & \vizimgall{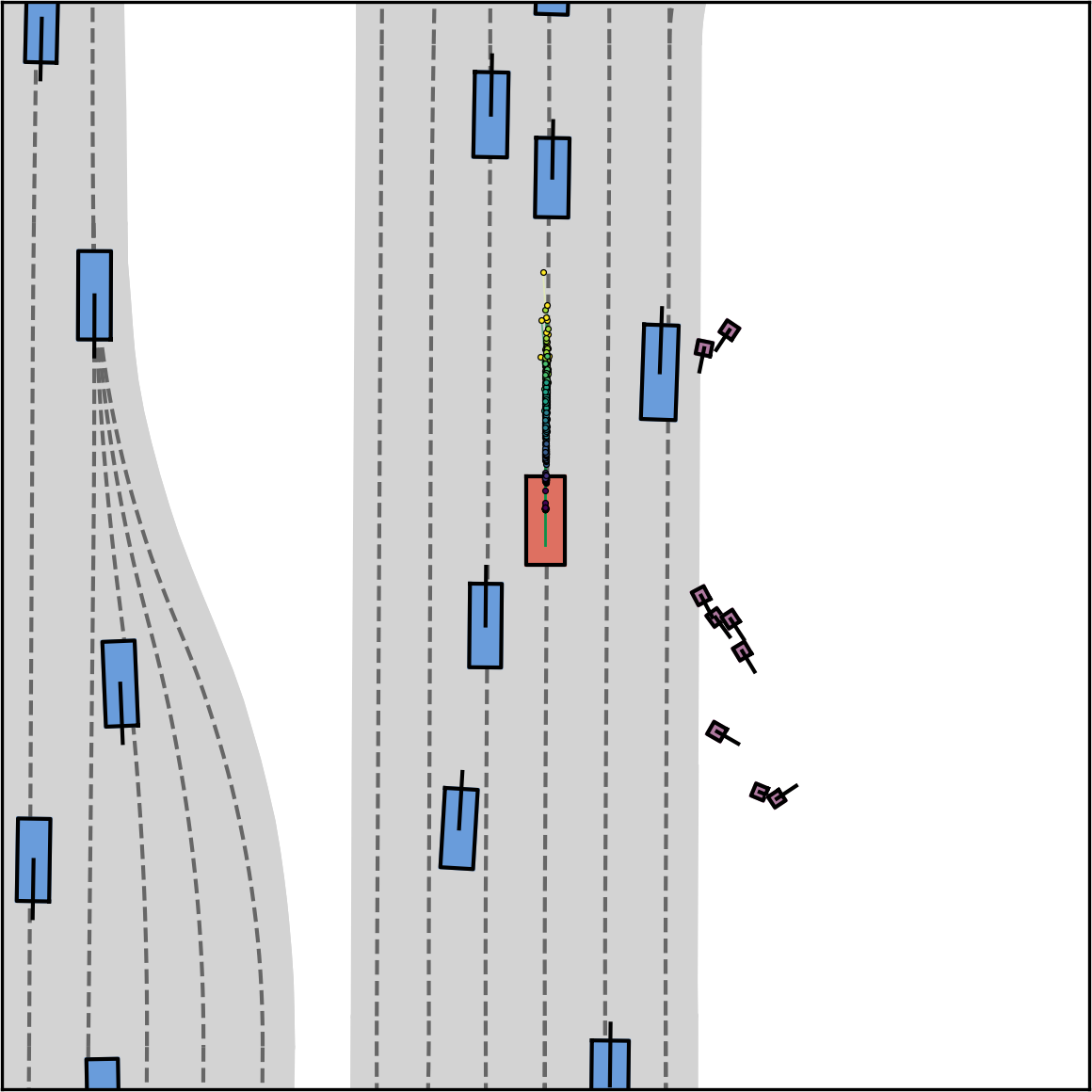} & \vizimgall{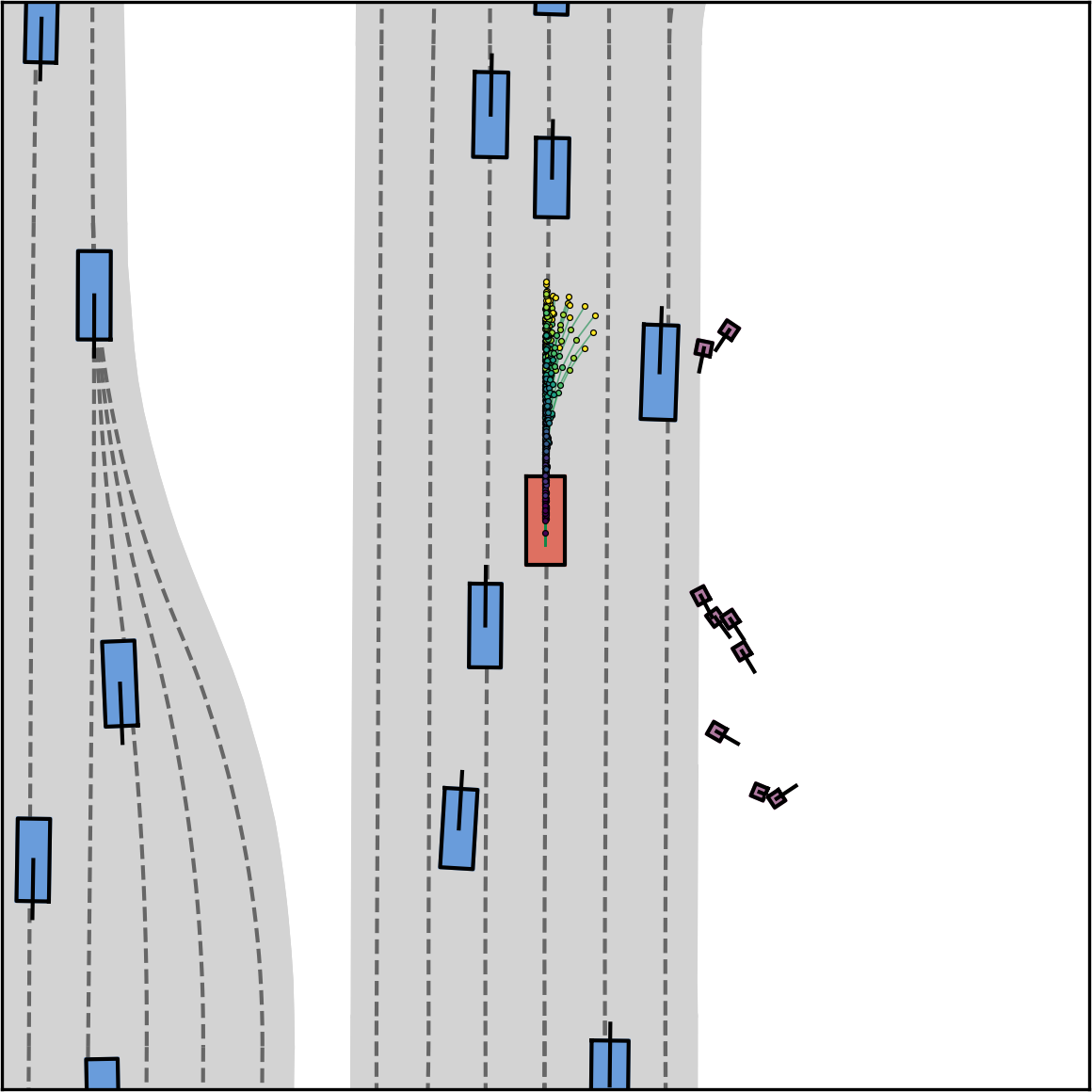} \\
\romarkall{Left} &
\vizimgall{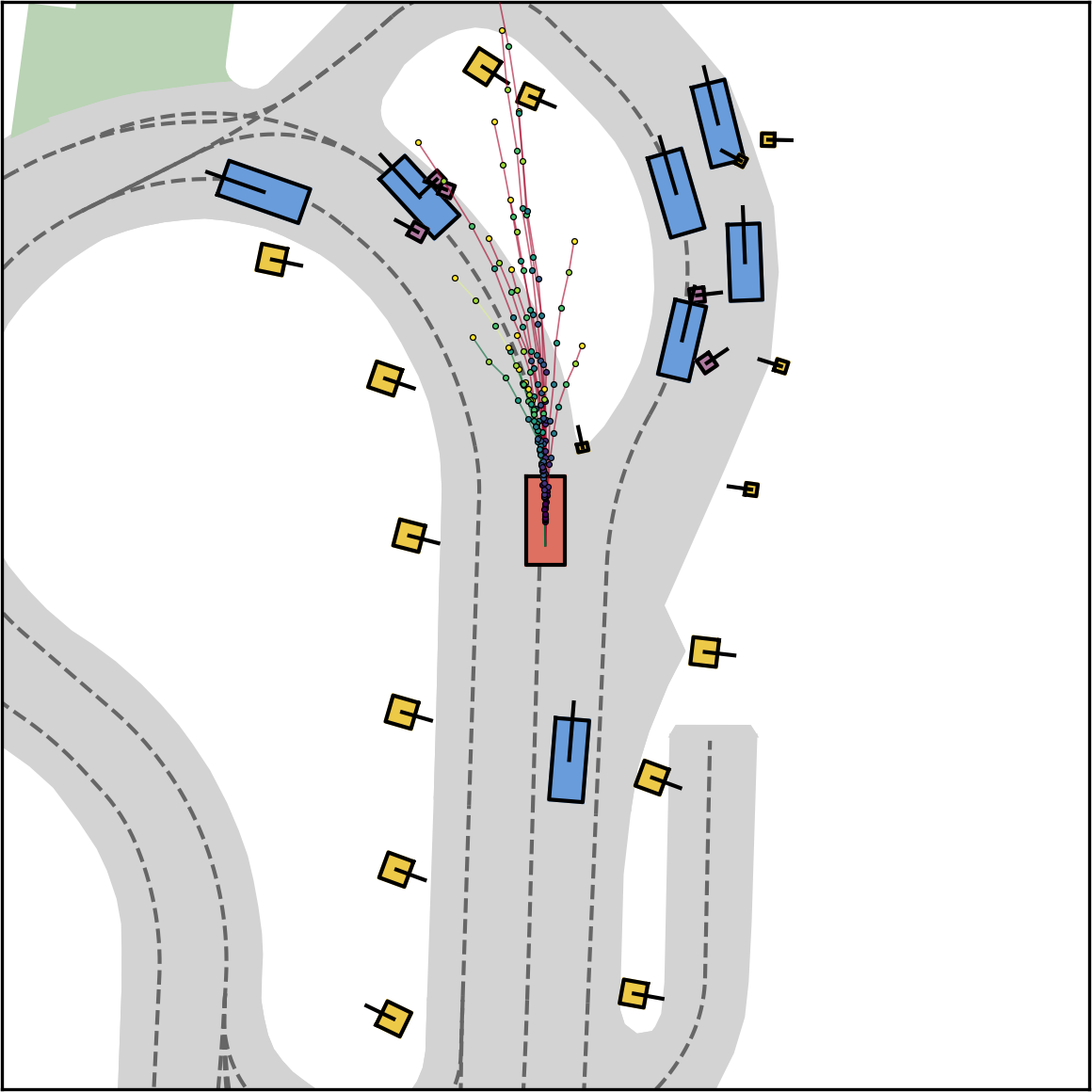} & \vizimgall{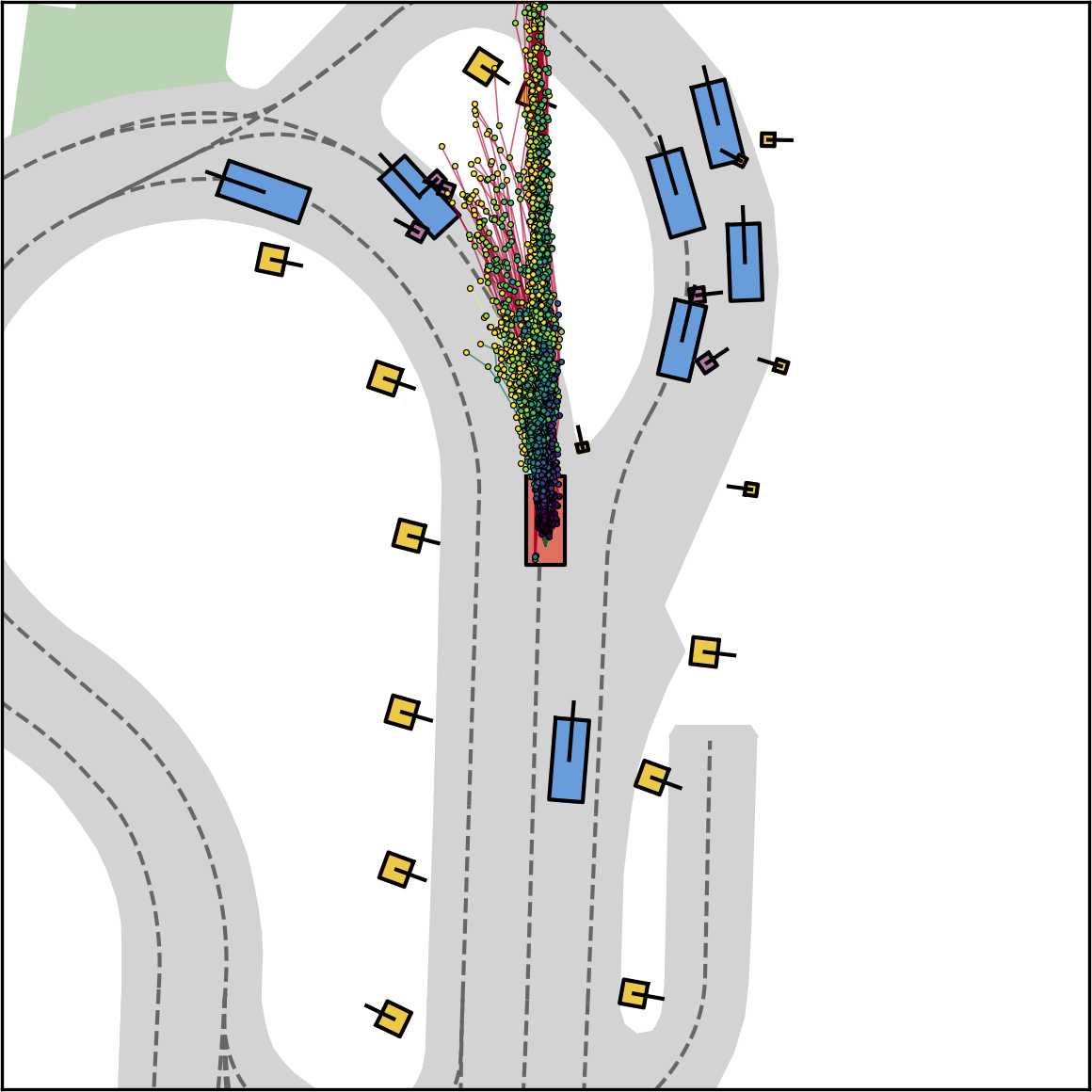} & \vizimgall{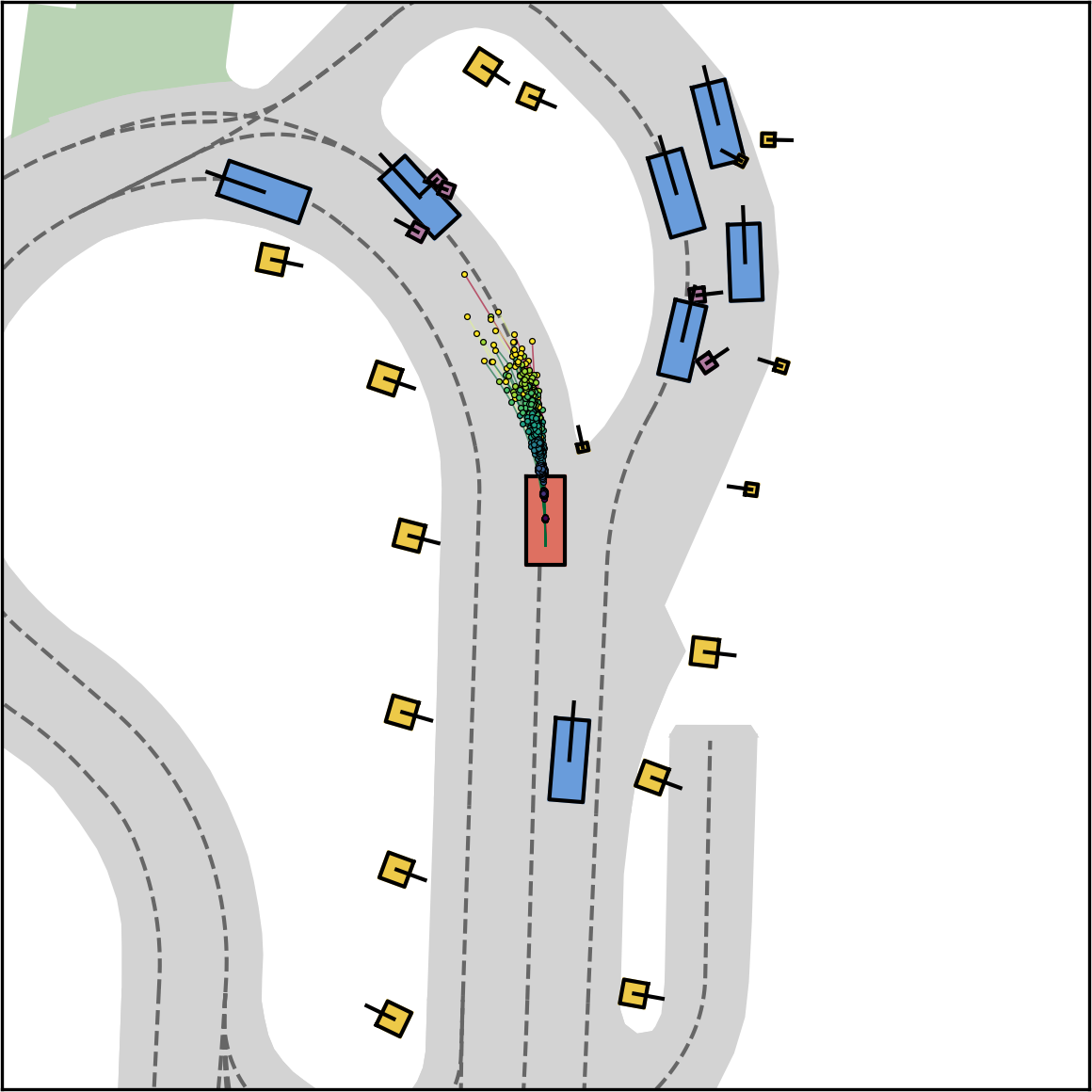} & \vizimgall{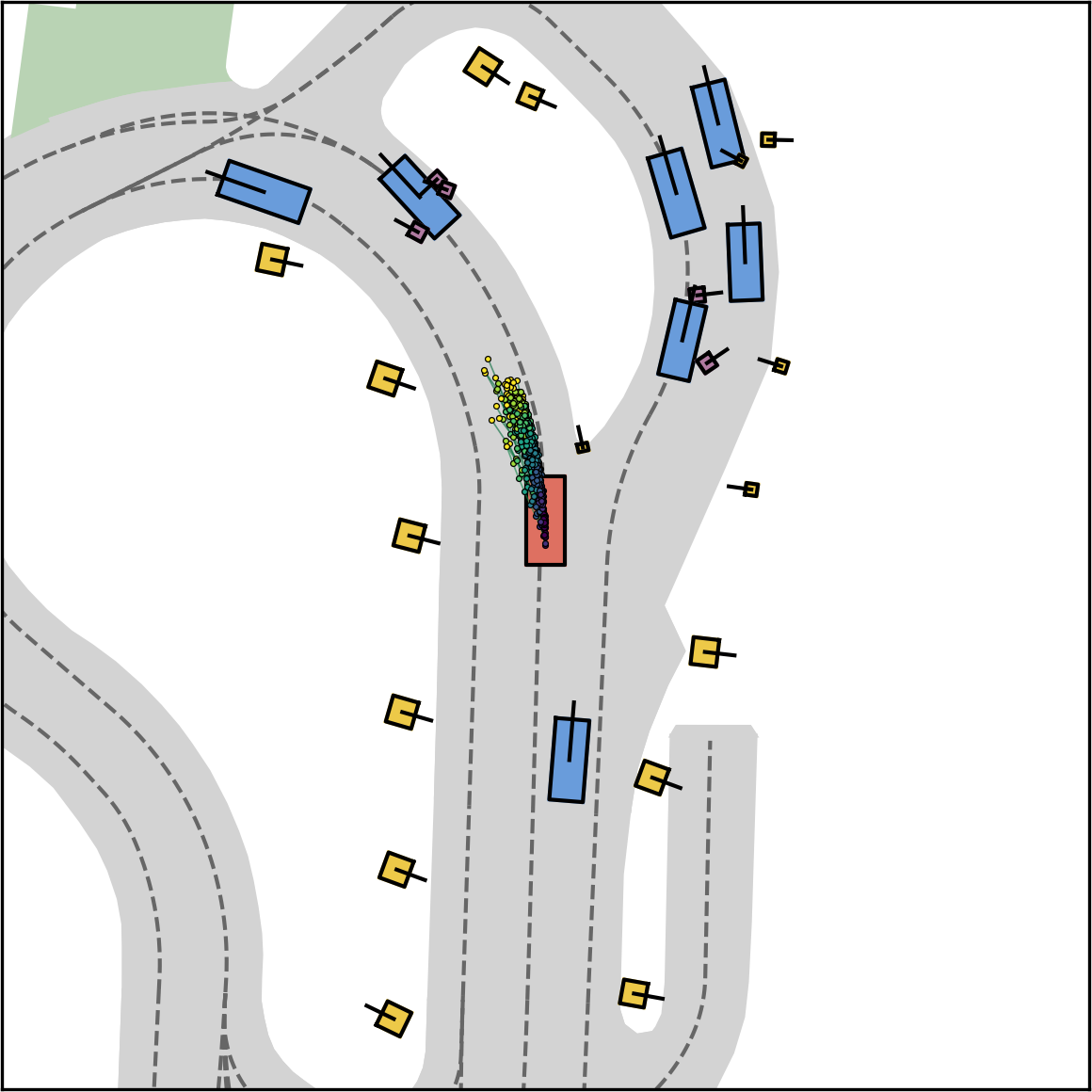} \\
 &
\vizimgall{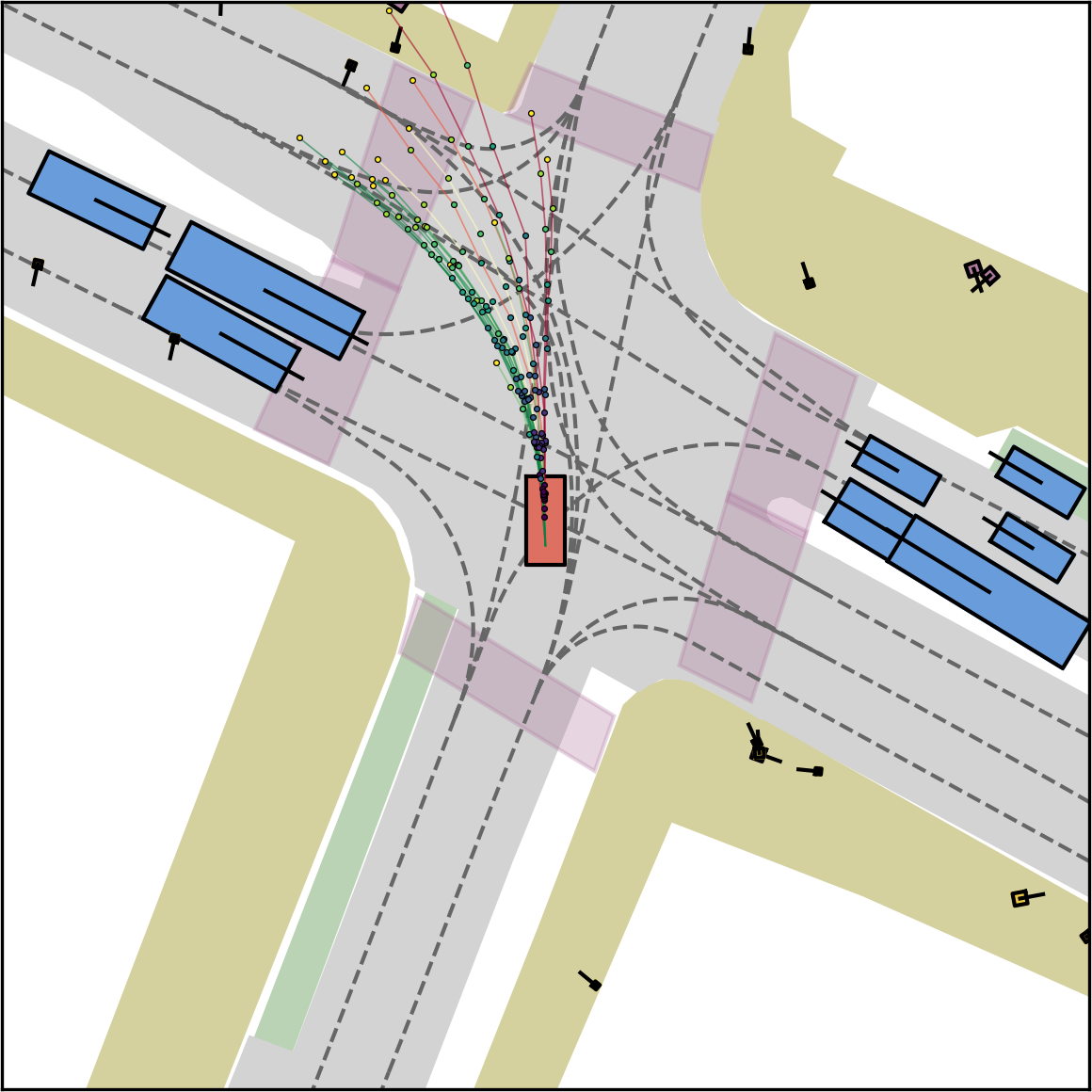} & \vizimgall{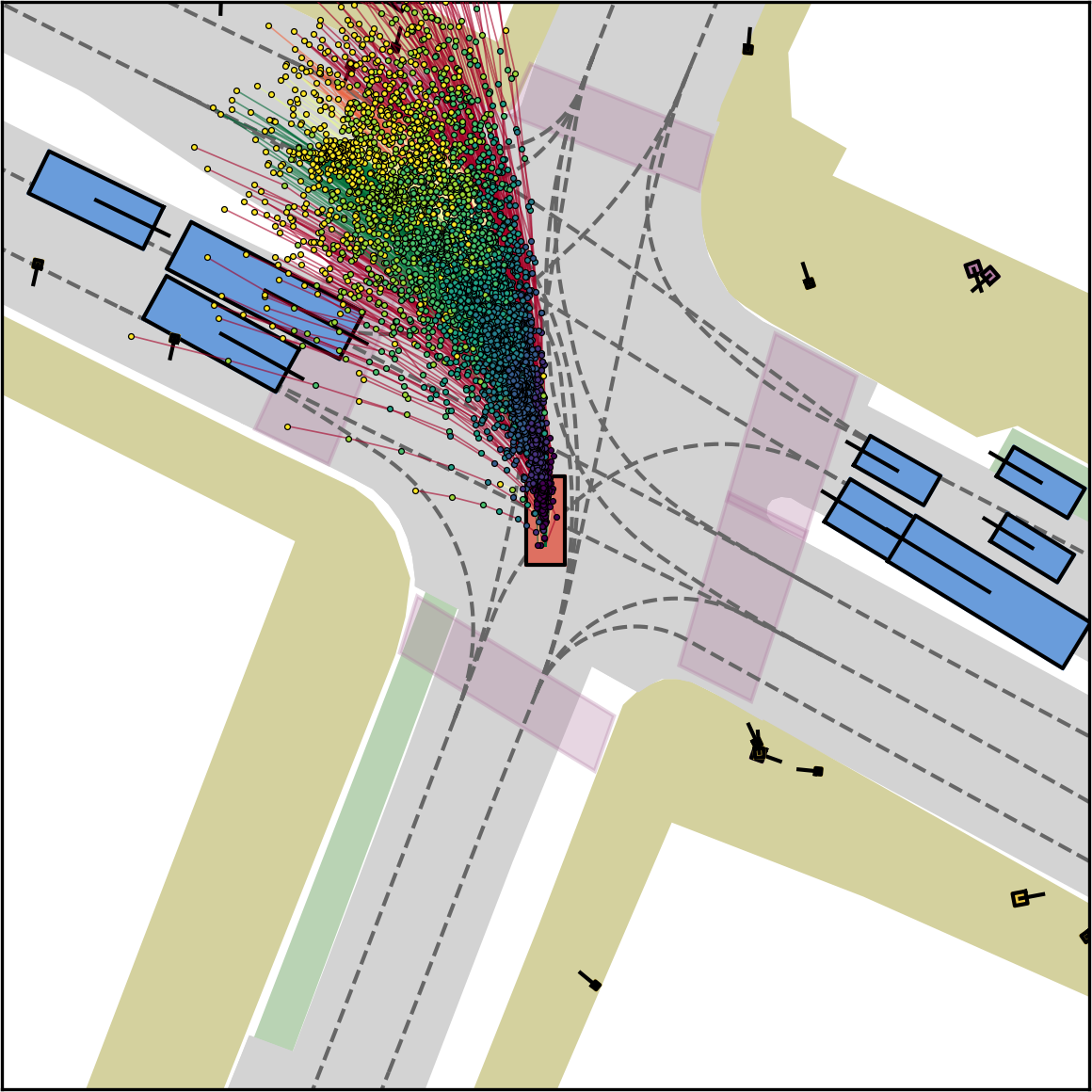} & \vizimgall{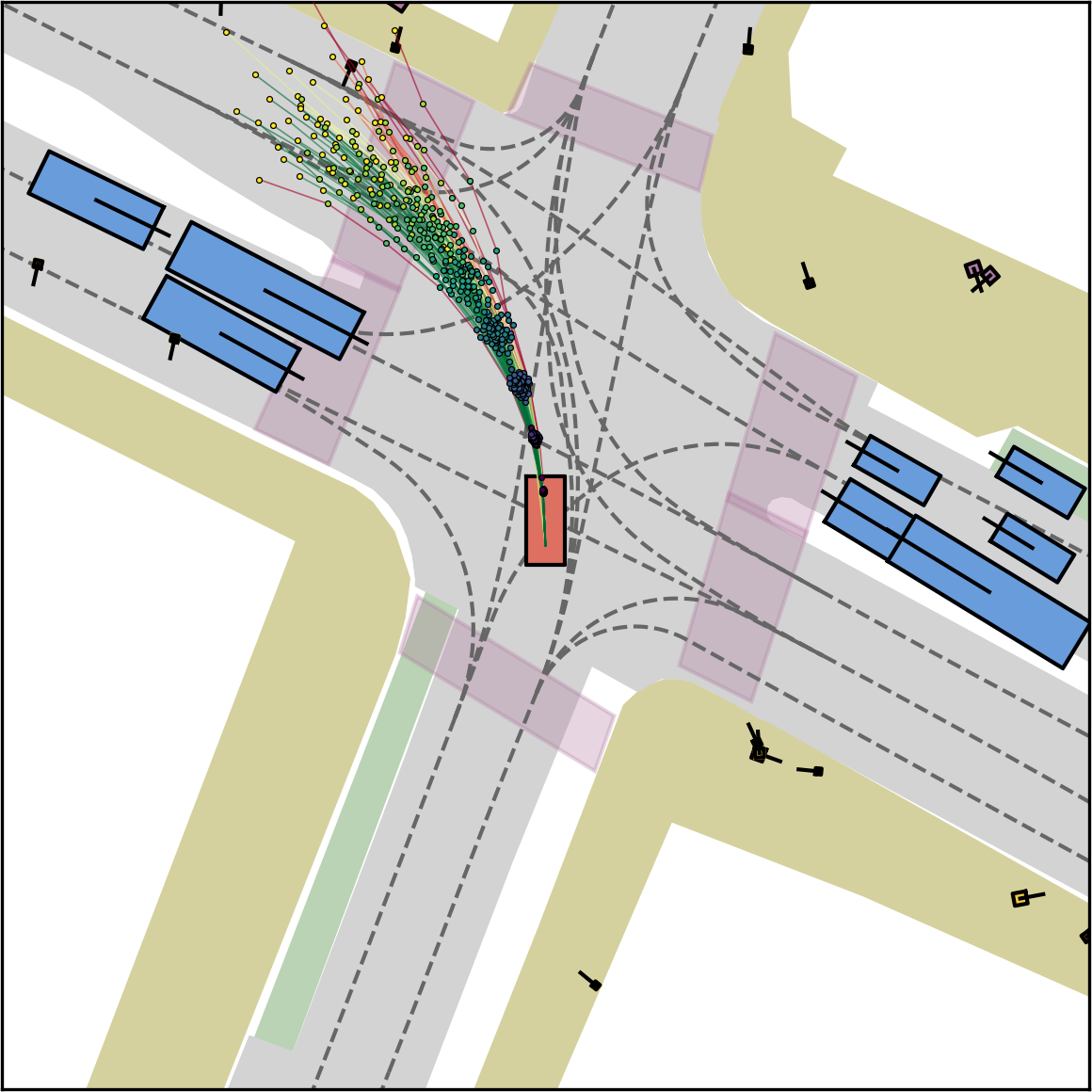} & \vizimgall{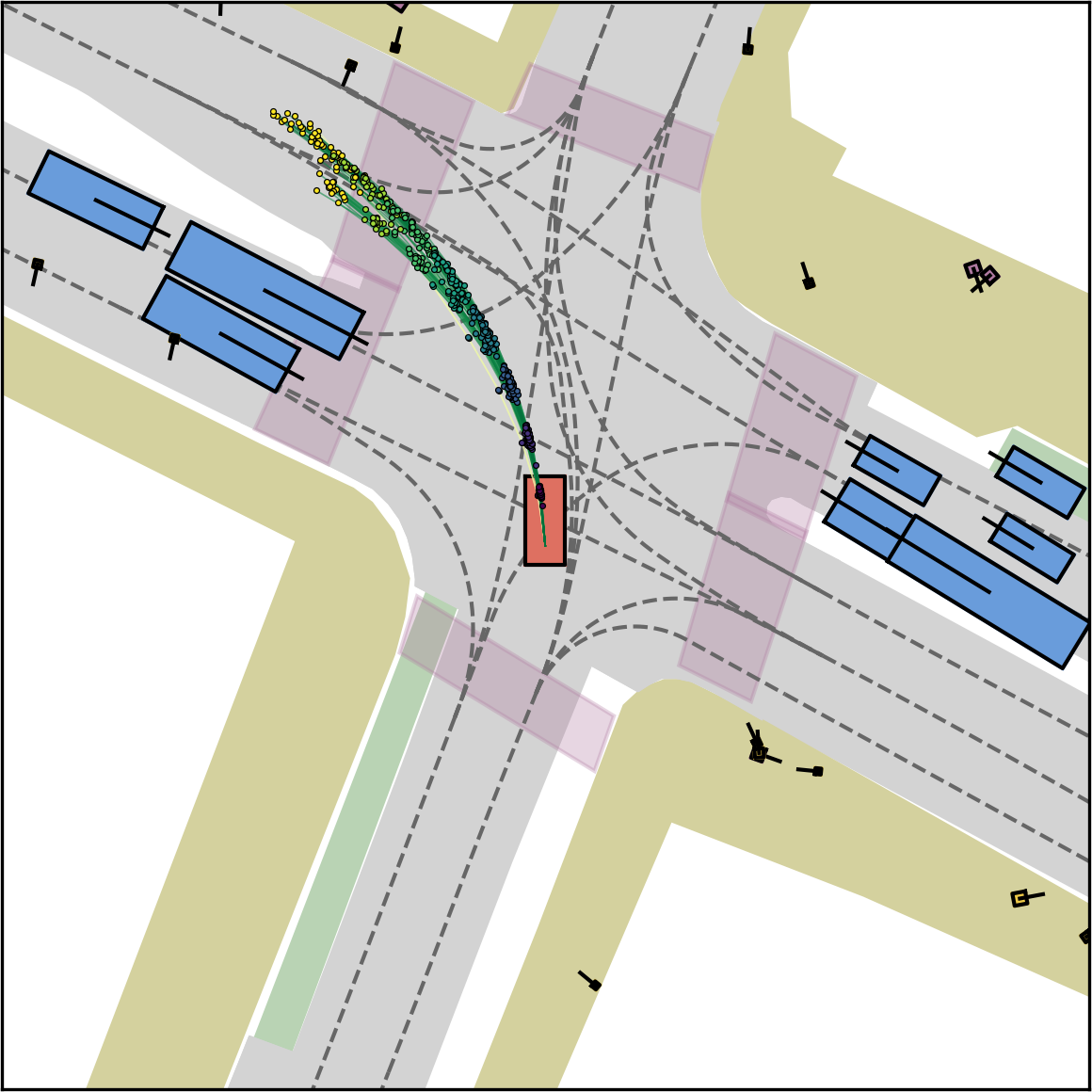} \\
\romarkall{Right} &
\vizimgall{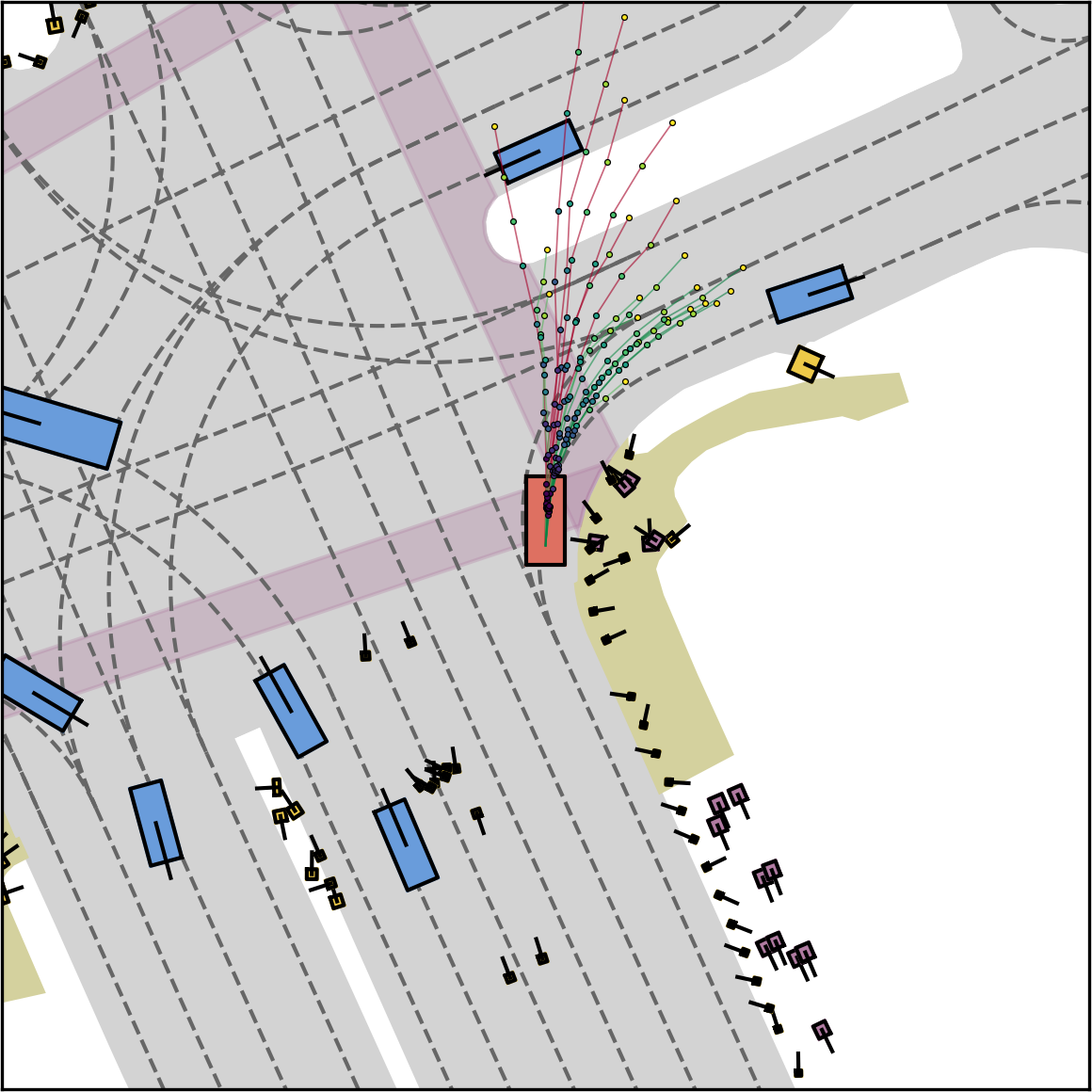} & \vizimgall{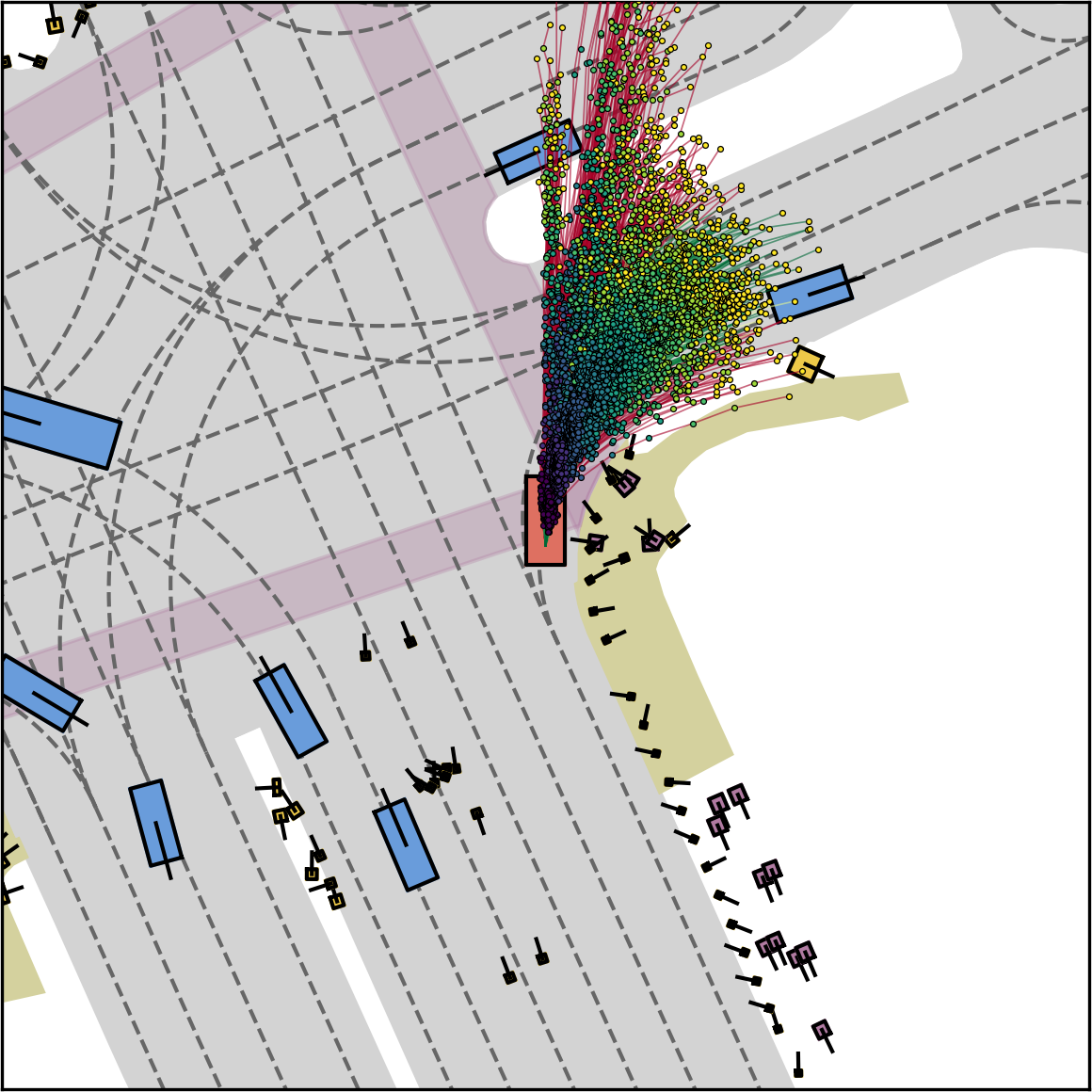} & \vizimgall{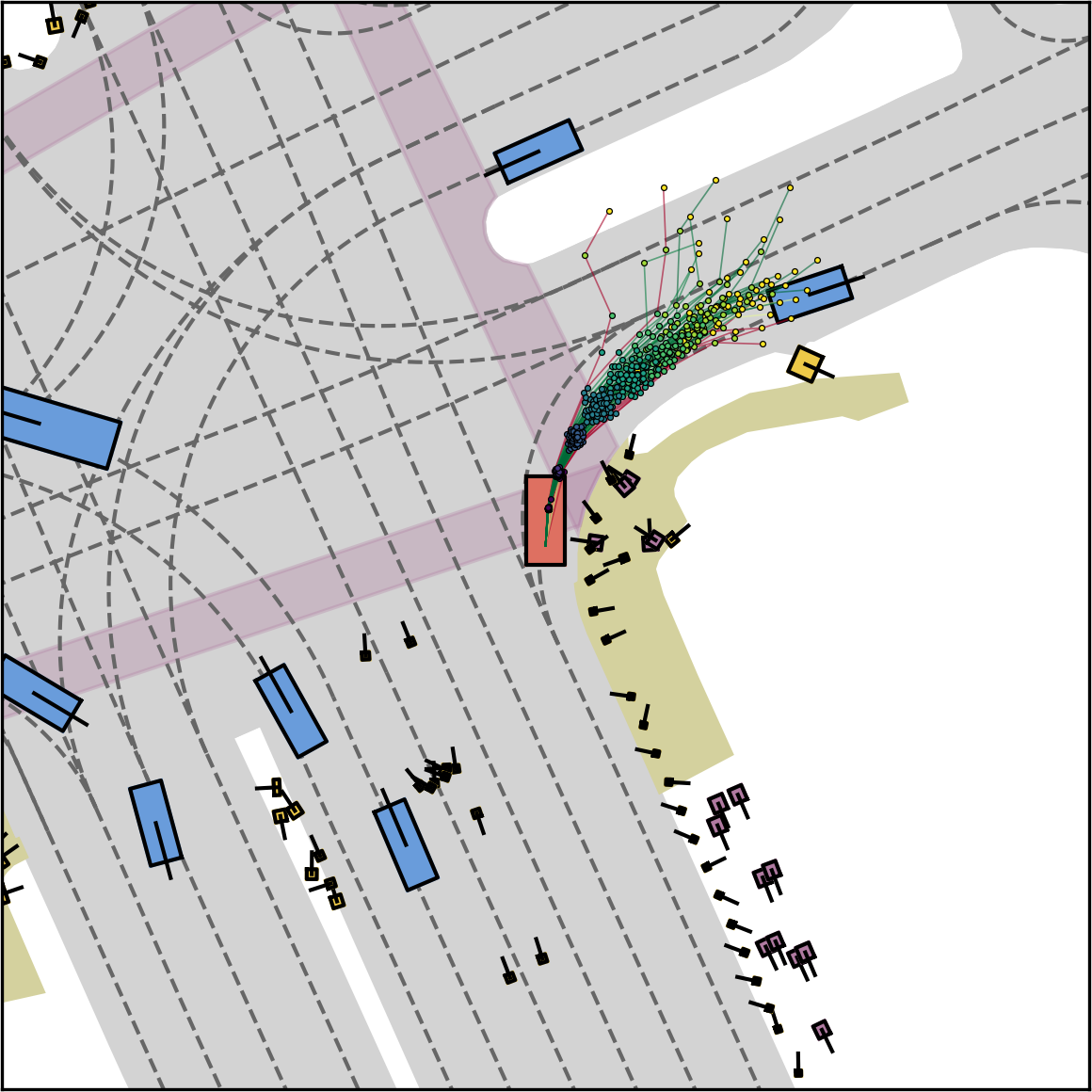} & \vizimgall{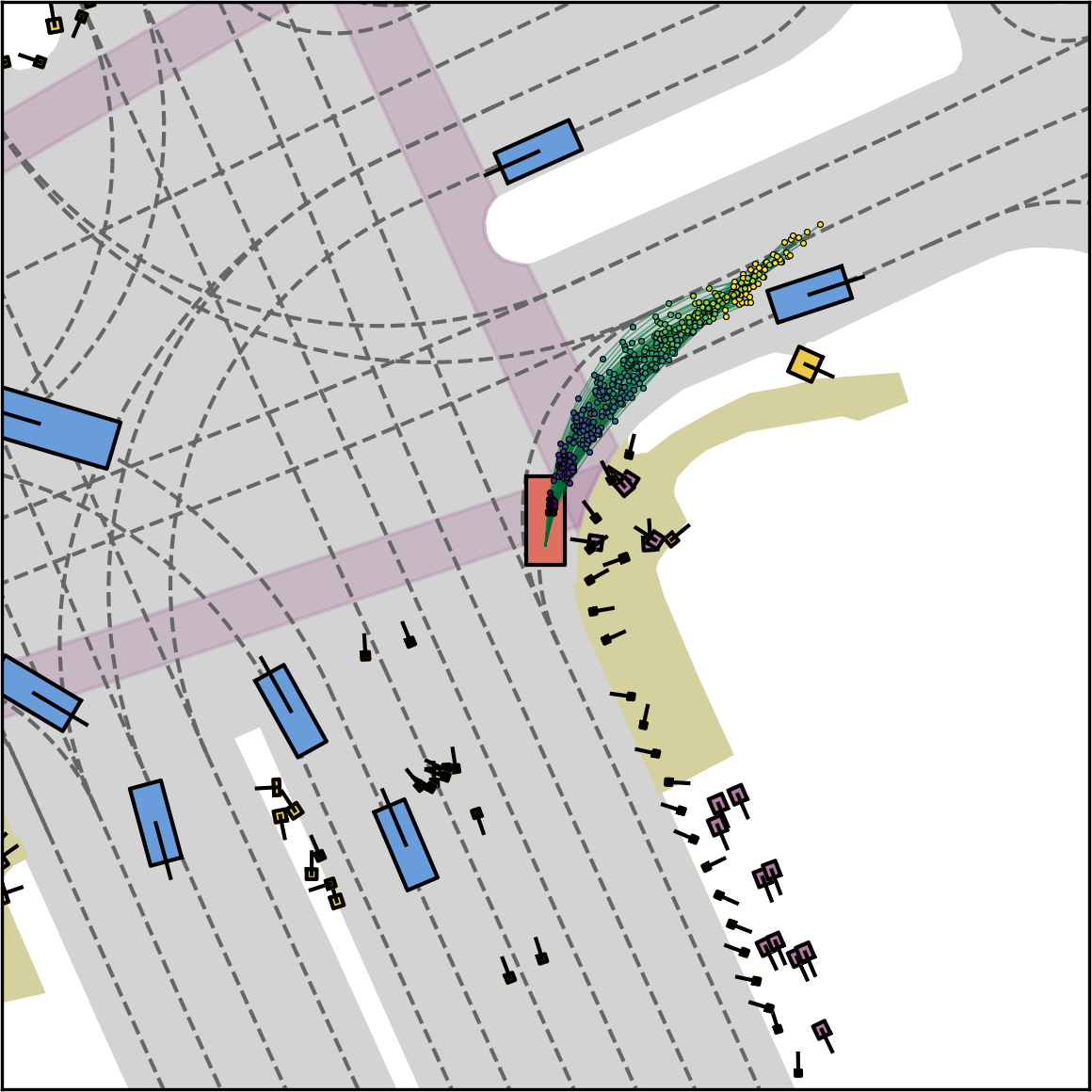} \\
 &
\vizimgall{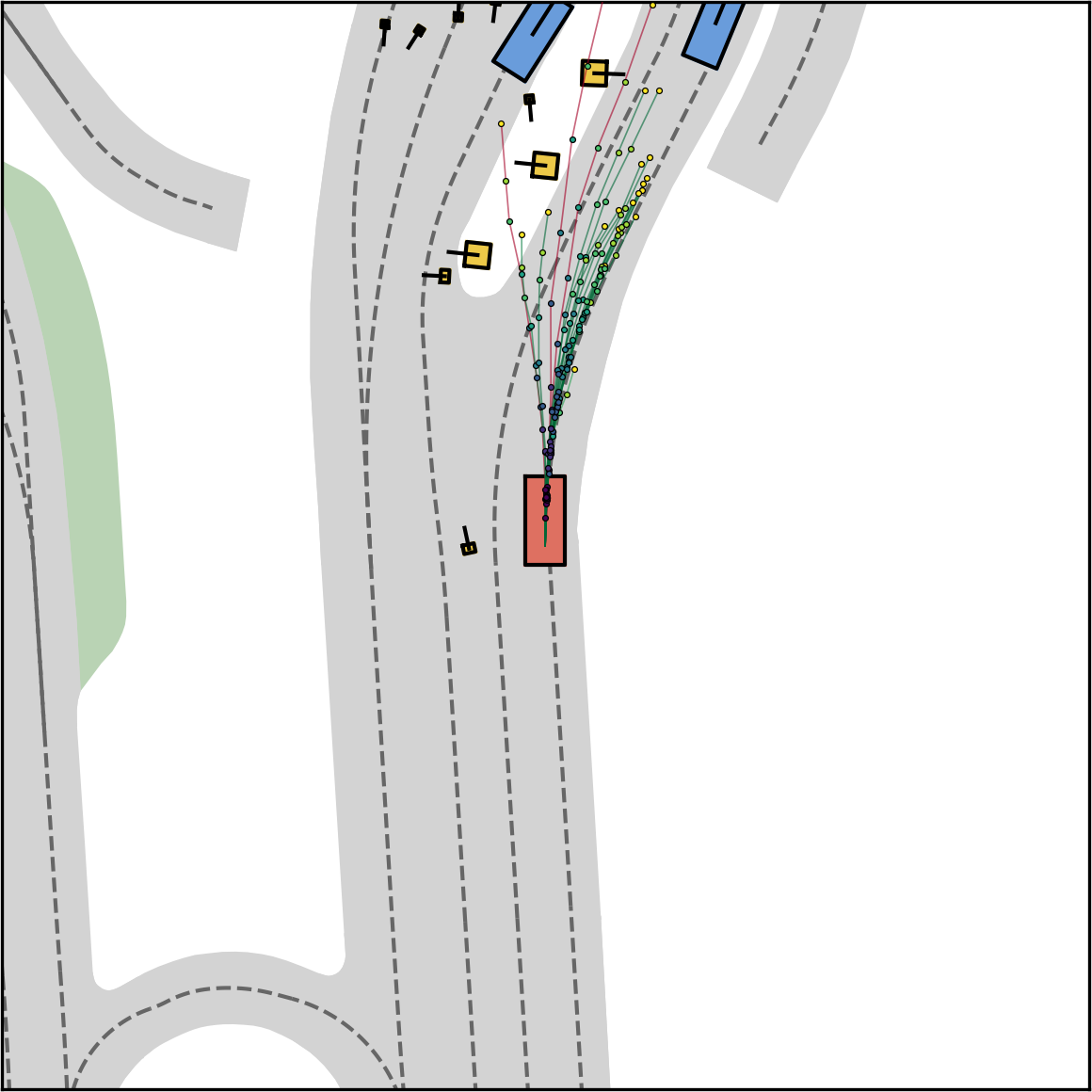} & \vizimgall{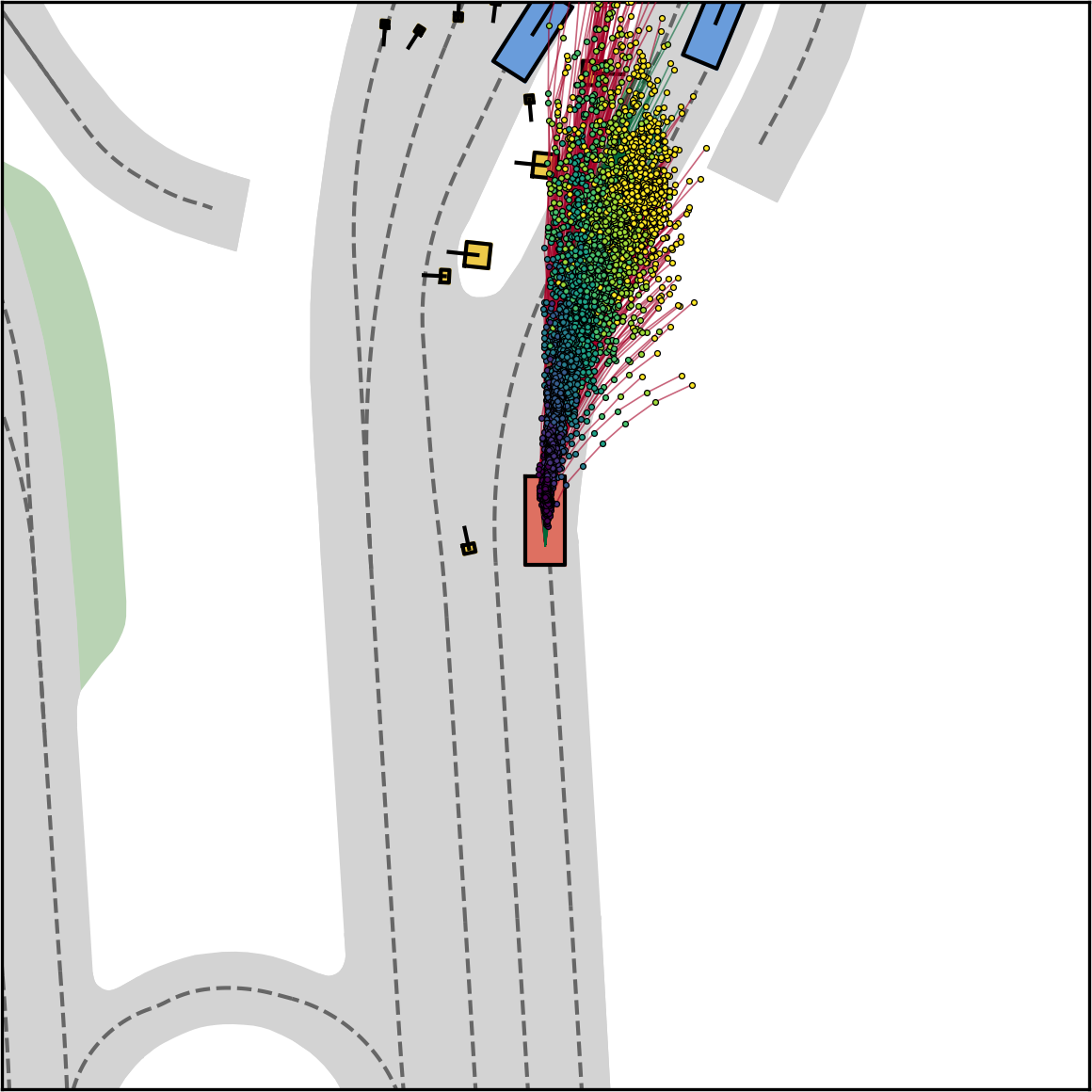} & \vizimgall{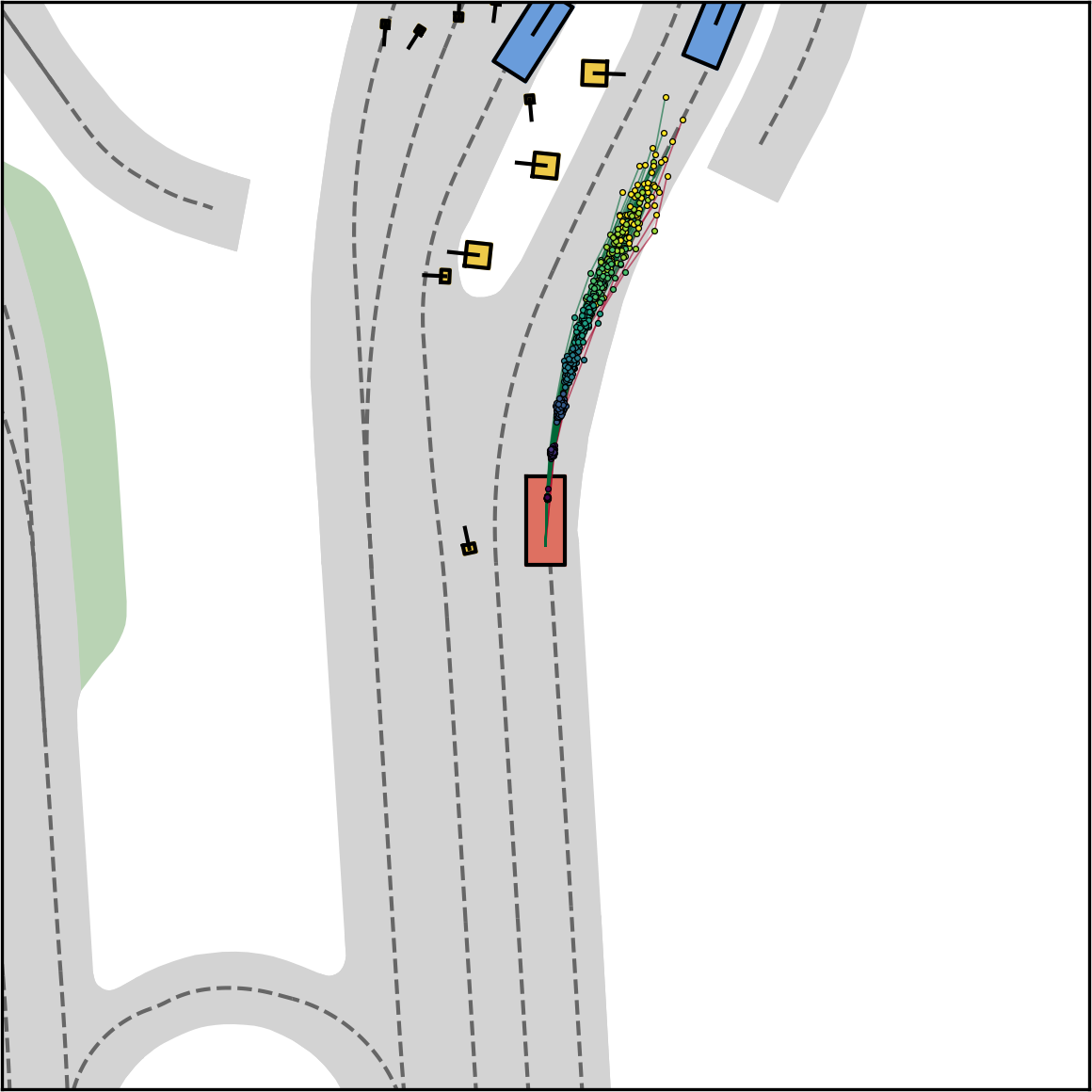} & \vizimgall{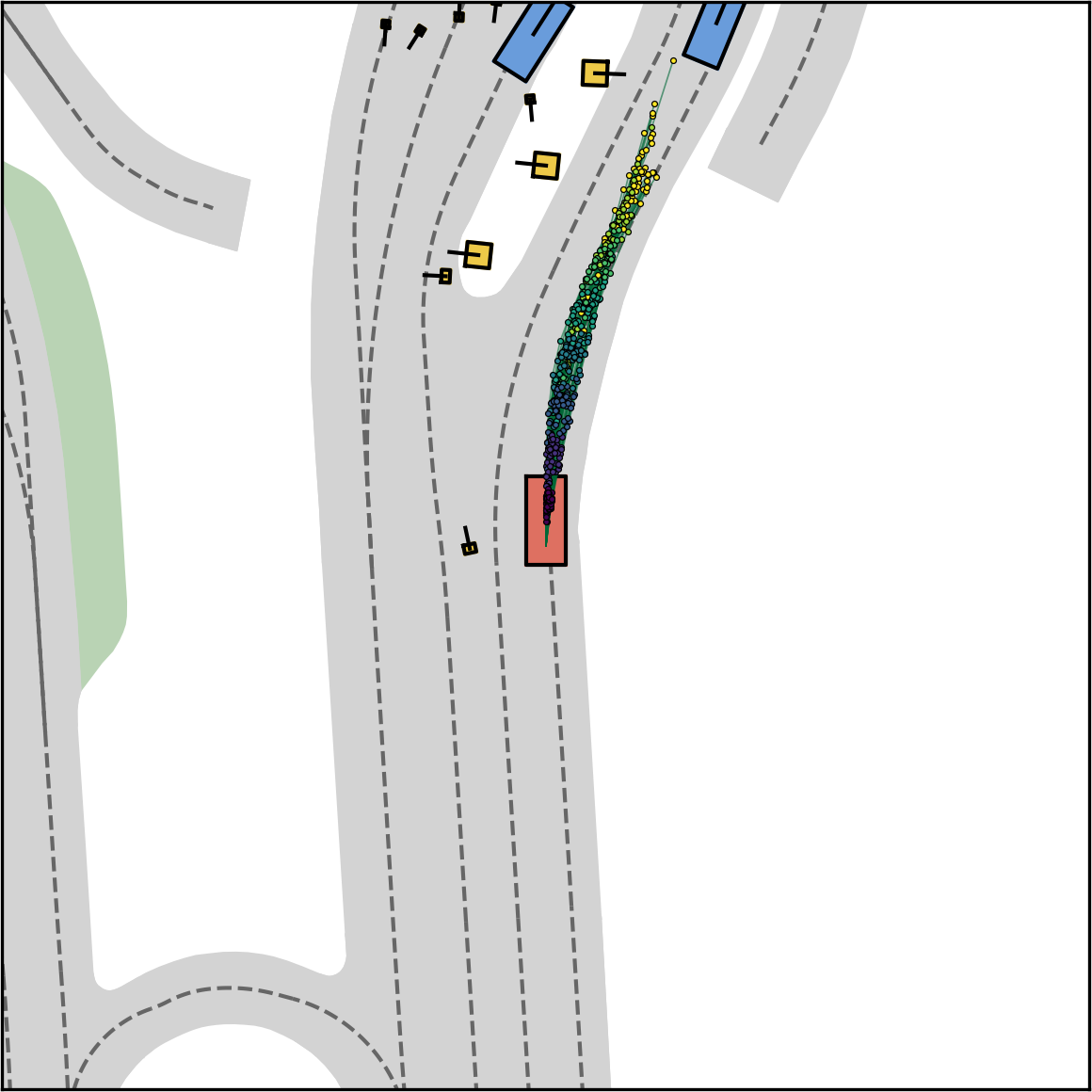} \\
\end{tabular}
\caption{Full qualitative comparison part 1 of 3. Trajectories are colored by PDMS from 0 ({\color[RGB]{200,30,30}red}) to 1 ({\color[RGB]{30,150,30}green}). Scenes are grouped by driving command labeled left.}
\label{fig:qual_all_1}
\end{figure*}

\begin{figure*}[t]
\centering
\setlength{\tabcolsep}{1pt}
\renewcommand{\arraystretch}{0.3}
\begin{tabular}{@{}c cccc@{}}
& \textbf{DiffusionDrive} & \textbf{DiffusionDriveV2} & \textbf{iPad} & \textbf{\ourmethod{} (Ours)} \\[2pt]
\romarkall{Forward} &
\vizimgall{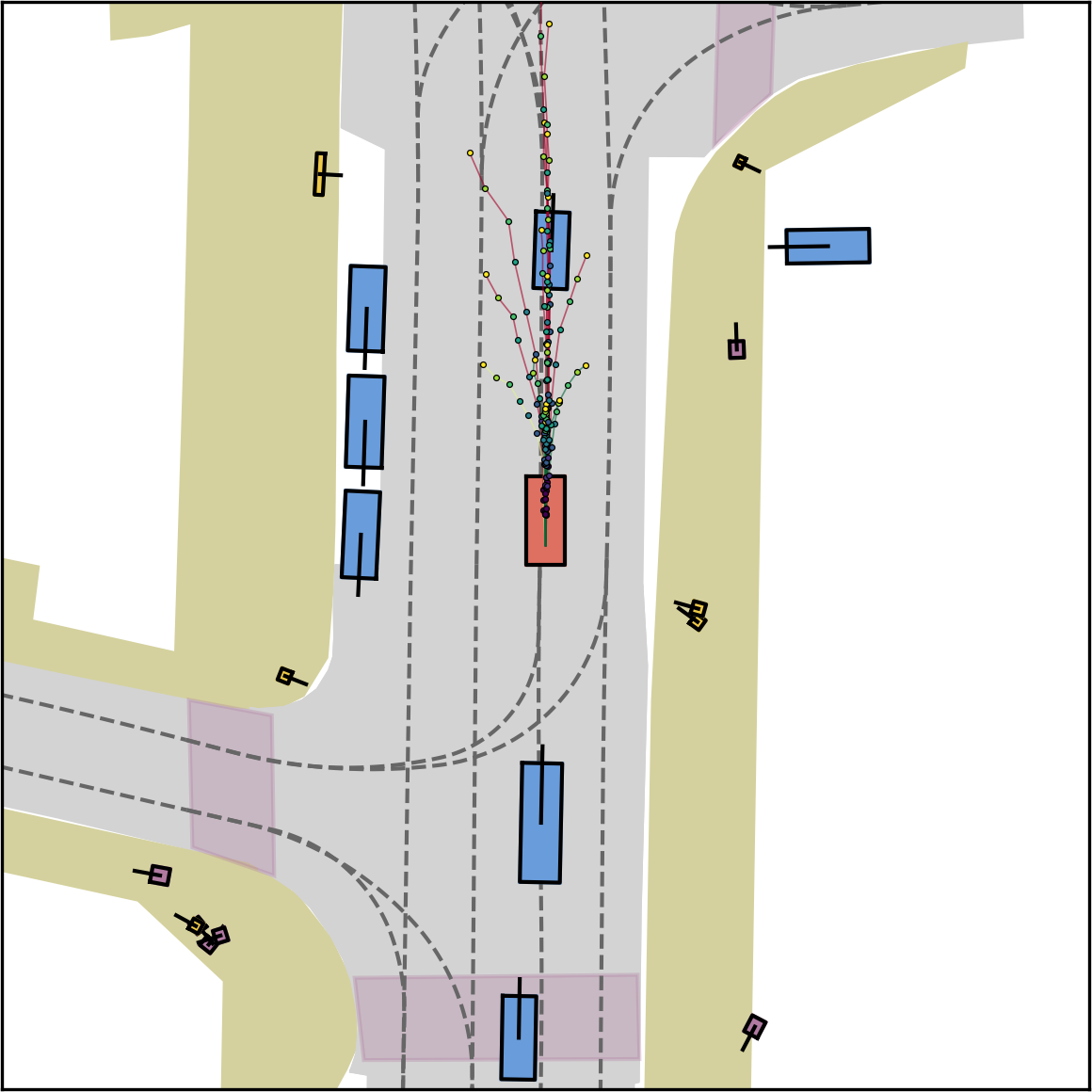} & \vizimgall{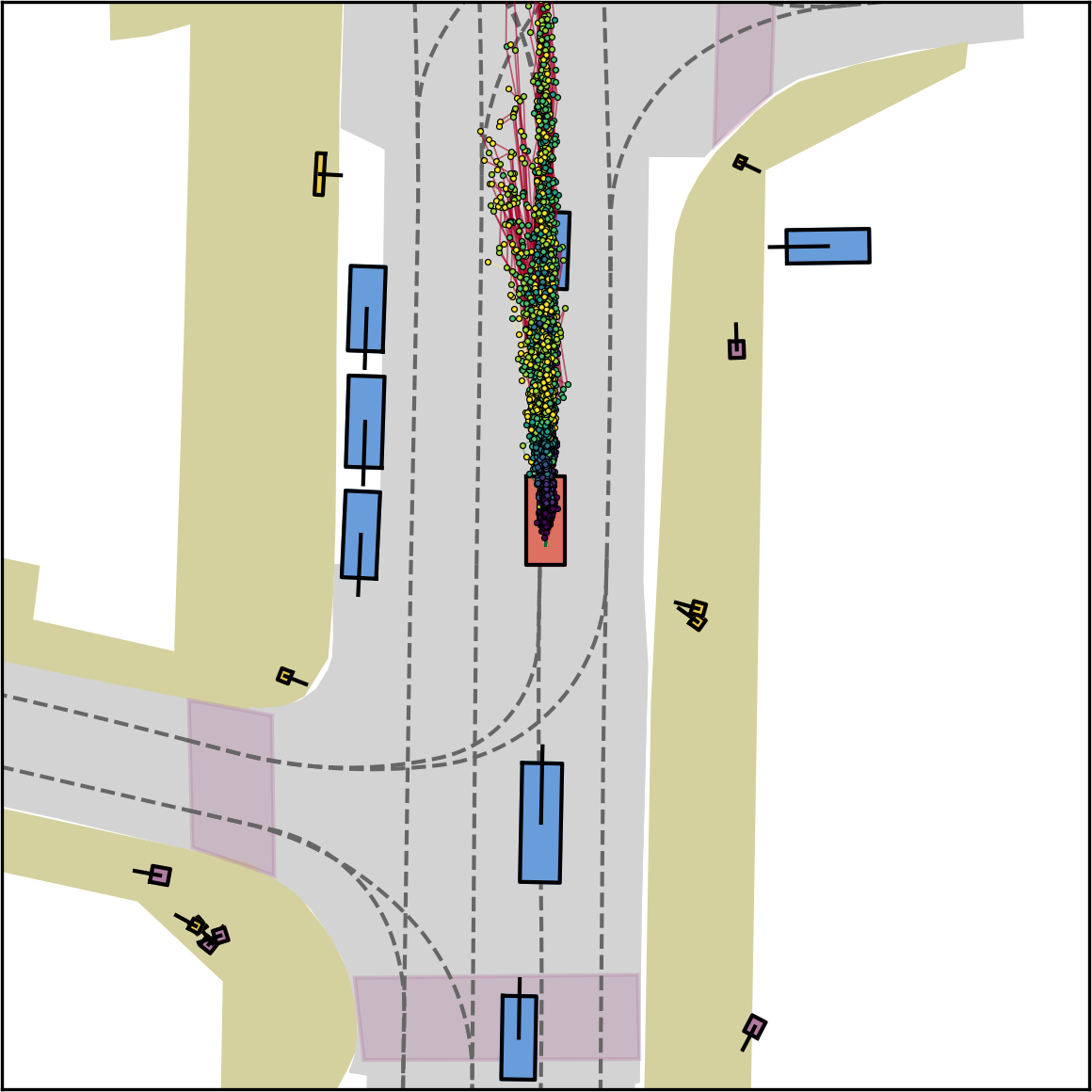} & \vizimgall{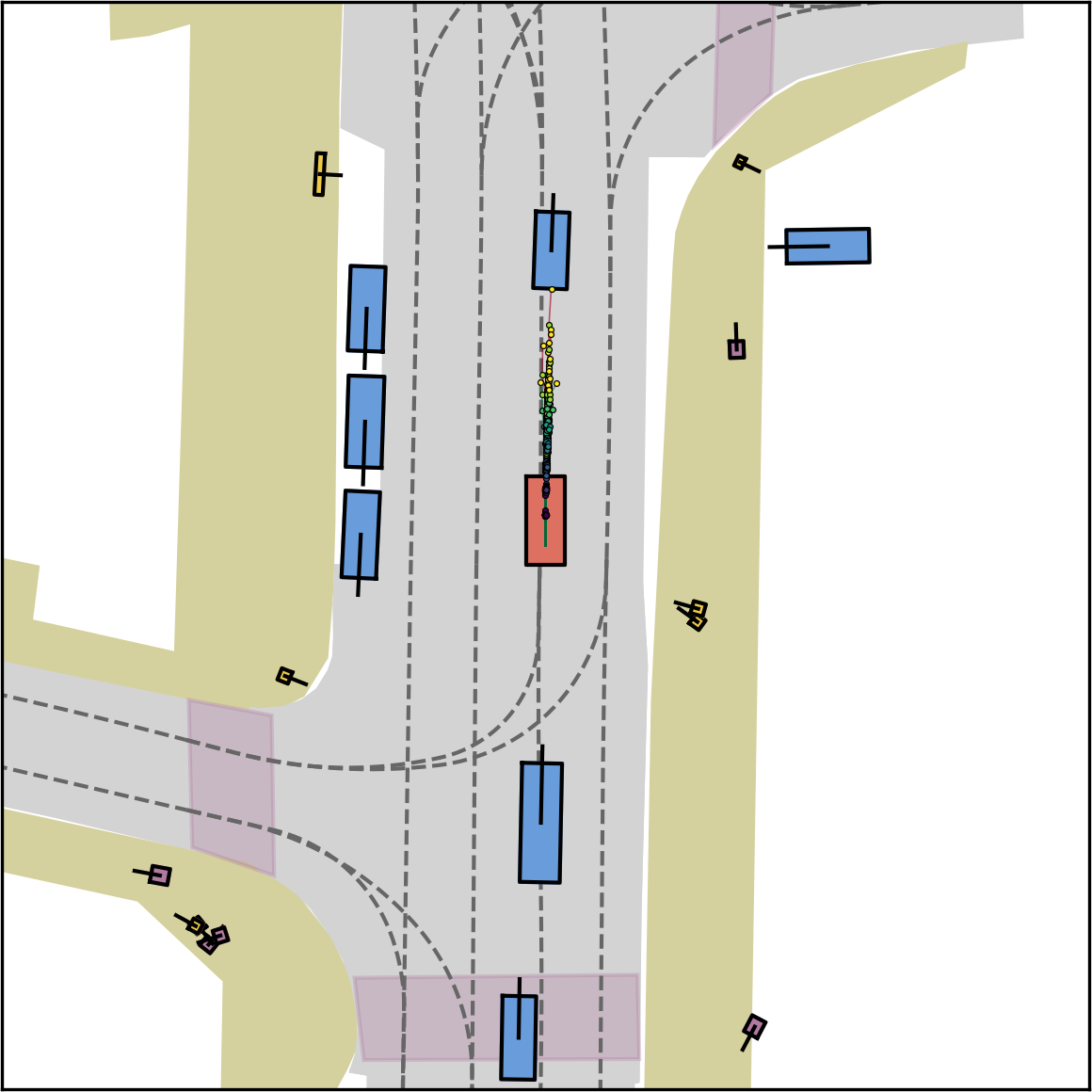} & \vizimgall{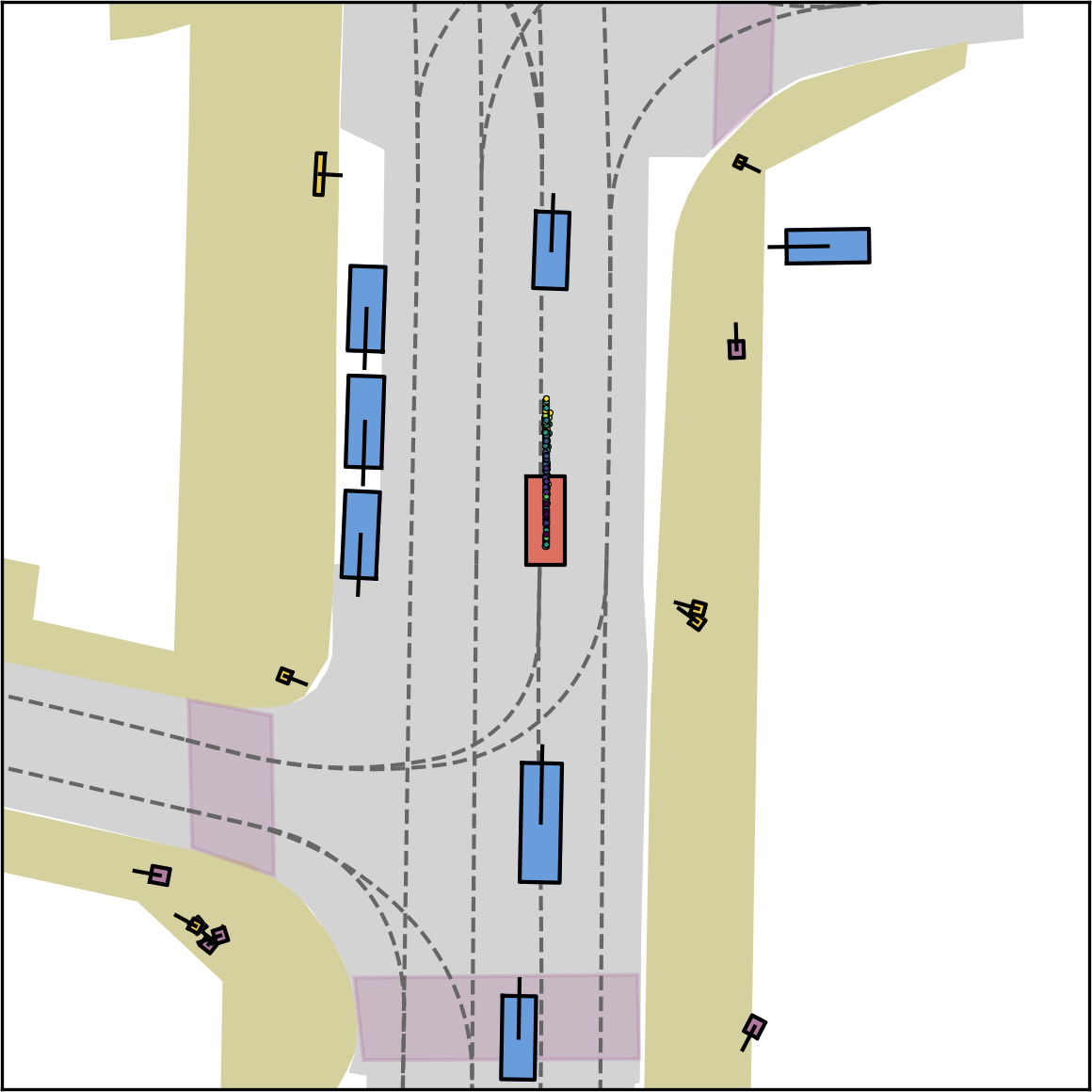} \\
 &
\vizimgall{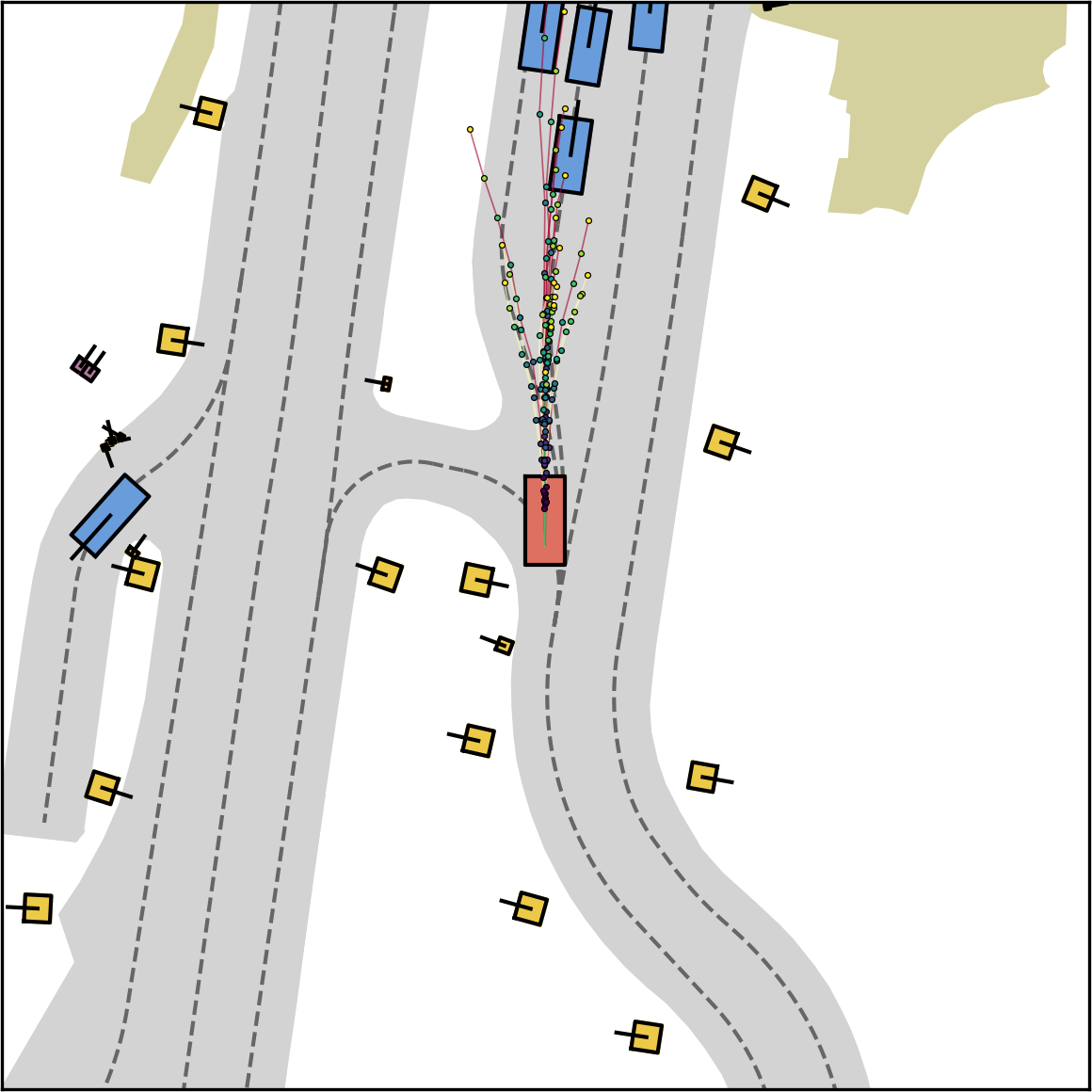} & \vizimgall{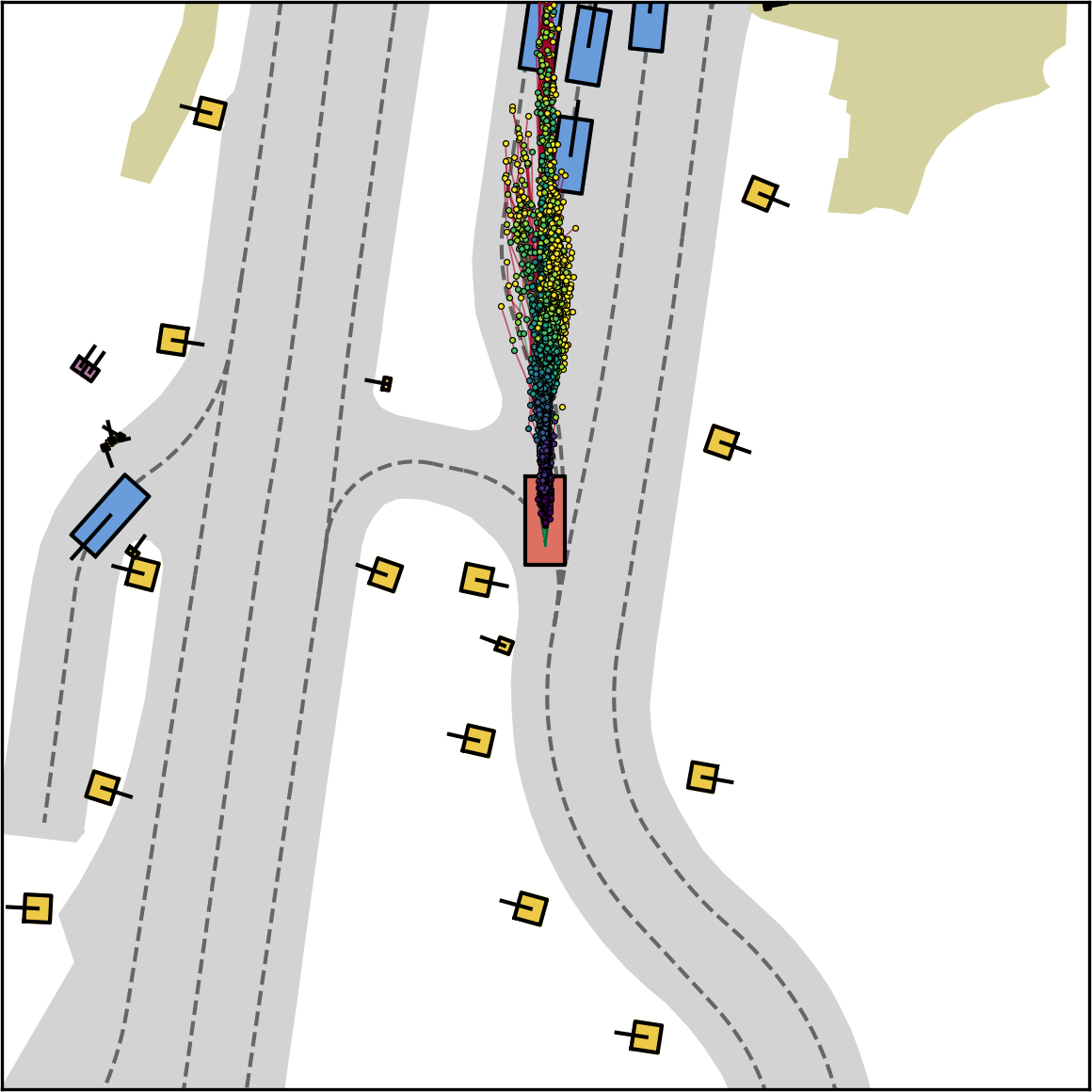} & \vizimgall{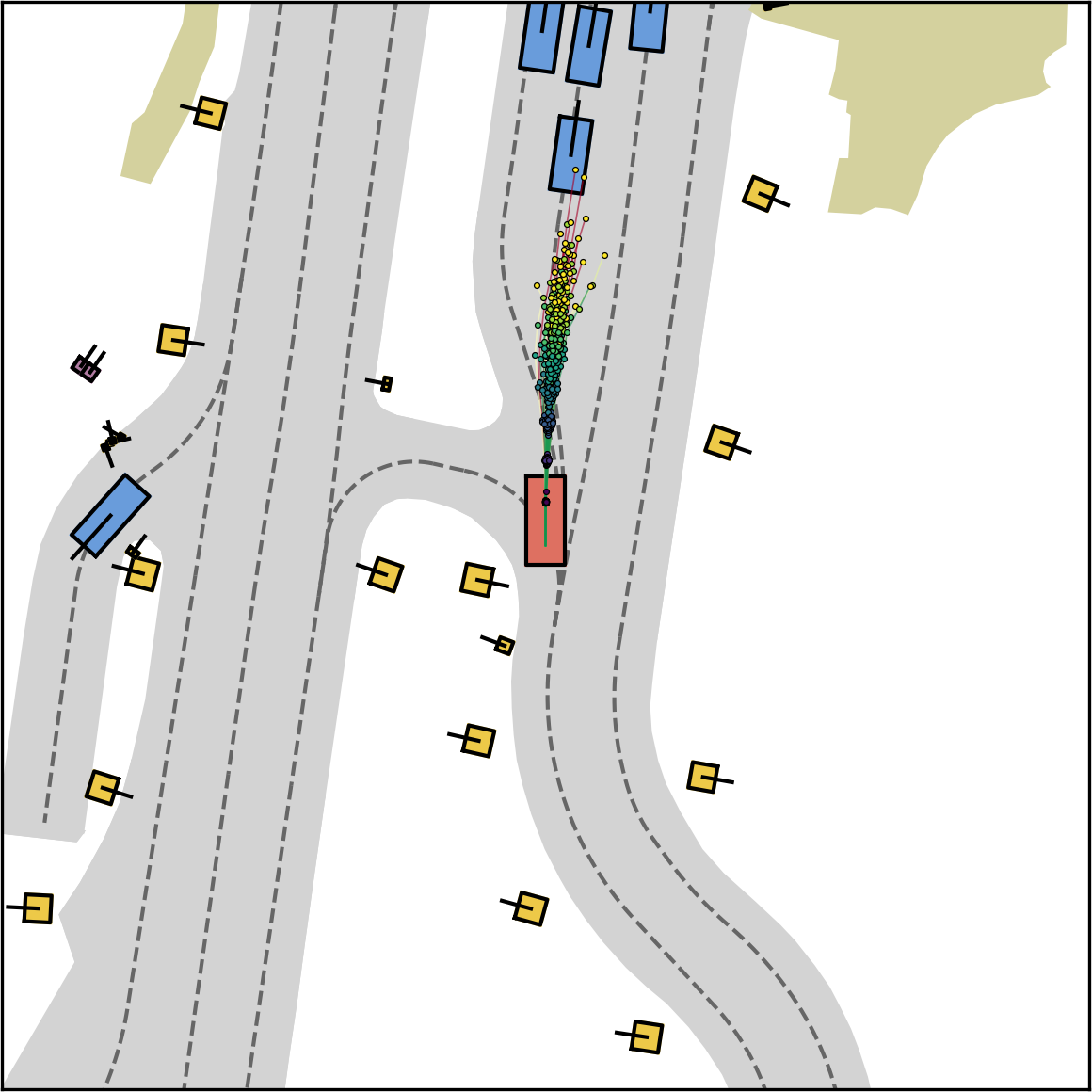} & \vizimgall{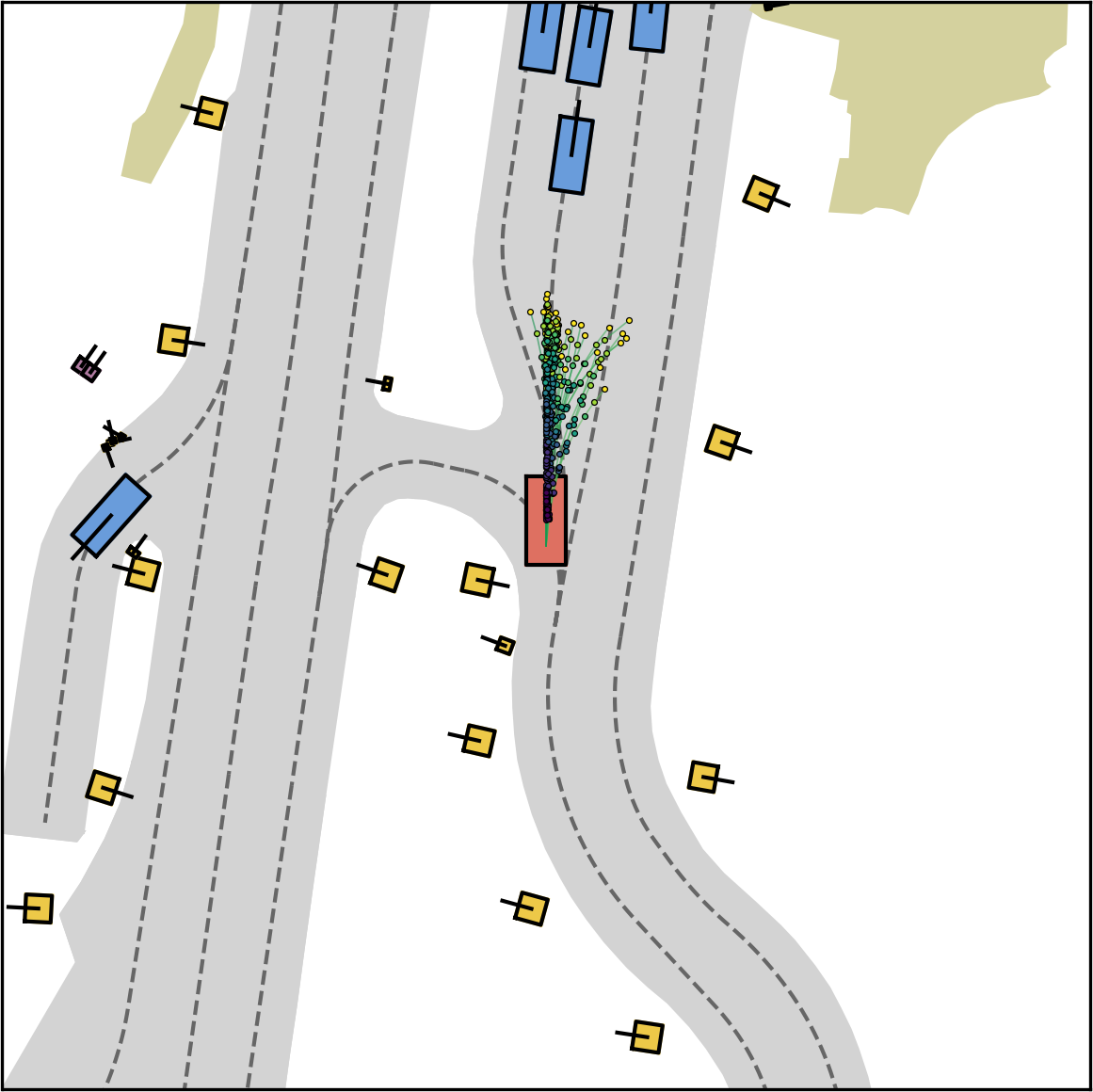} \\
 &
\vizimgall{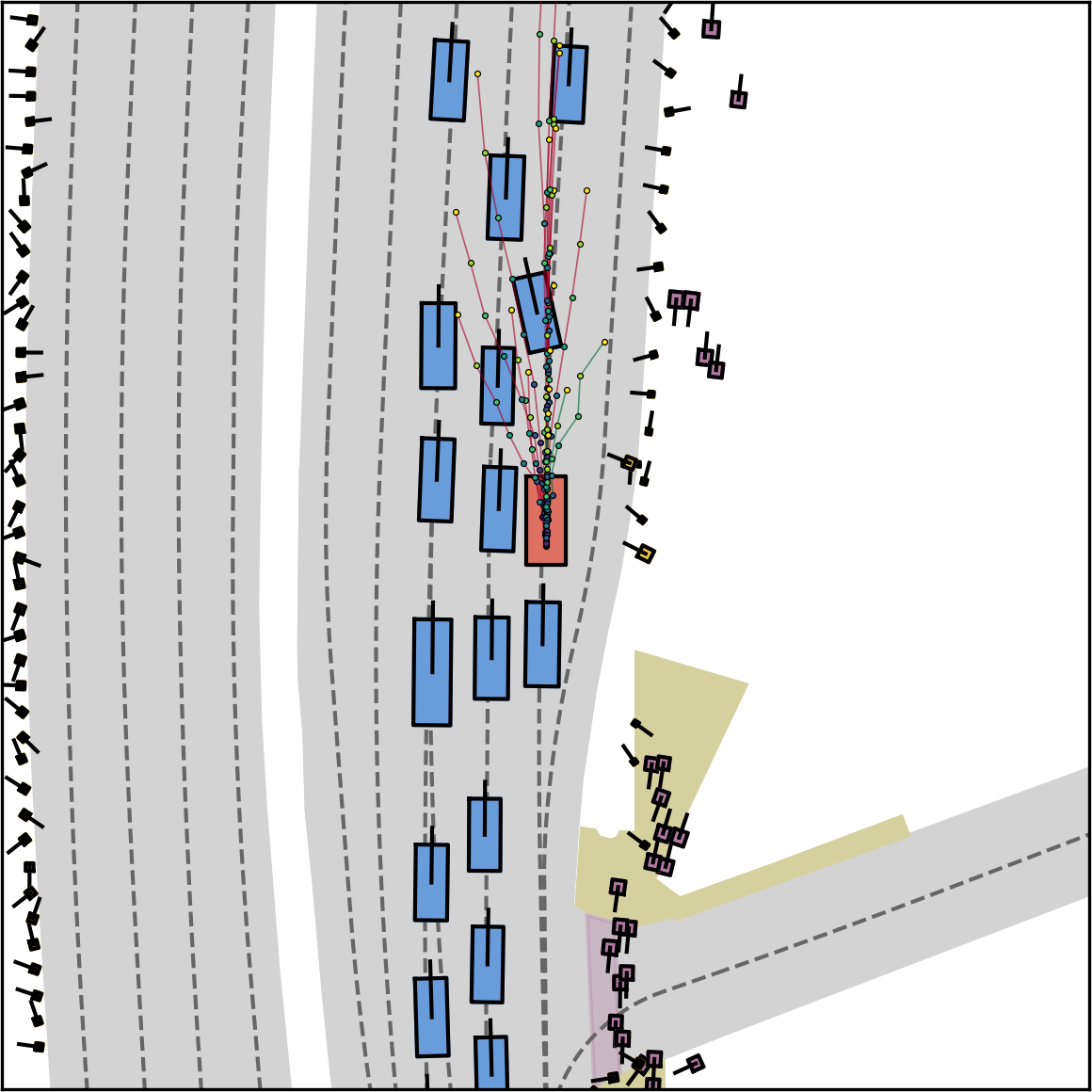} & \vizimgall{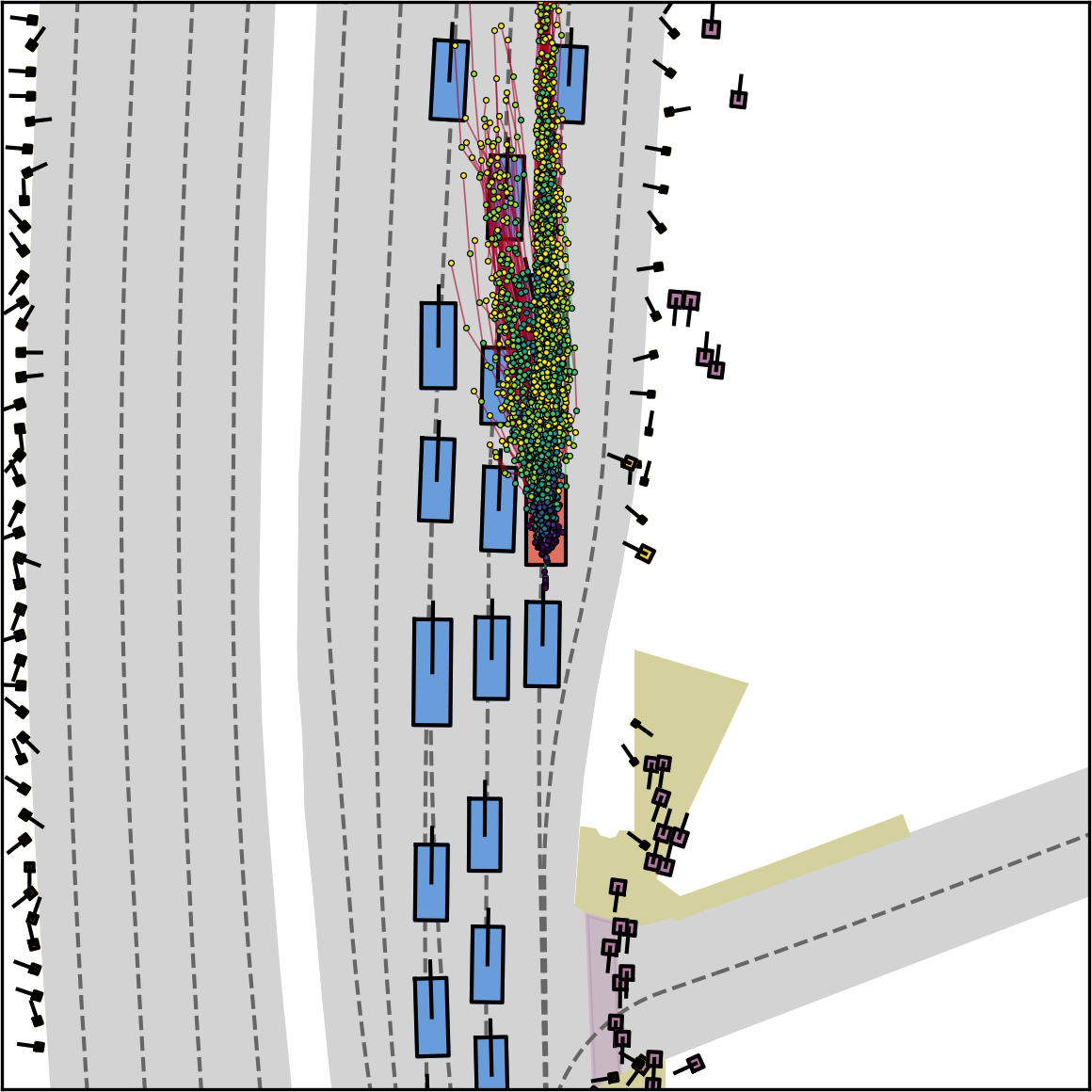} & \vizimgall{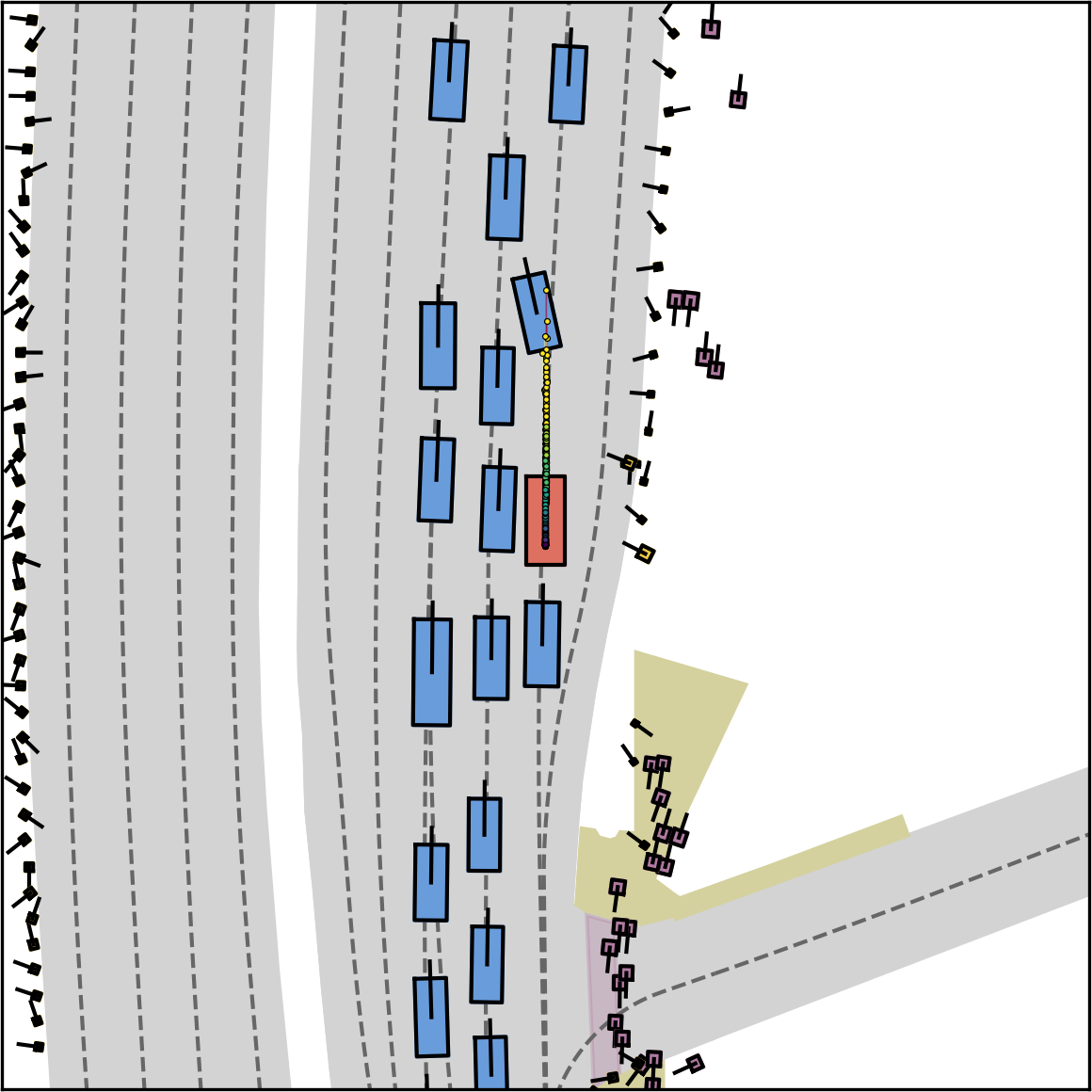} & \vizimgall{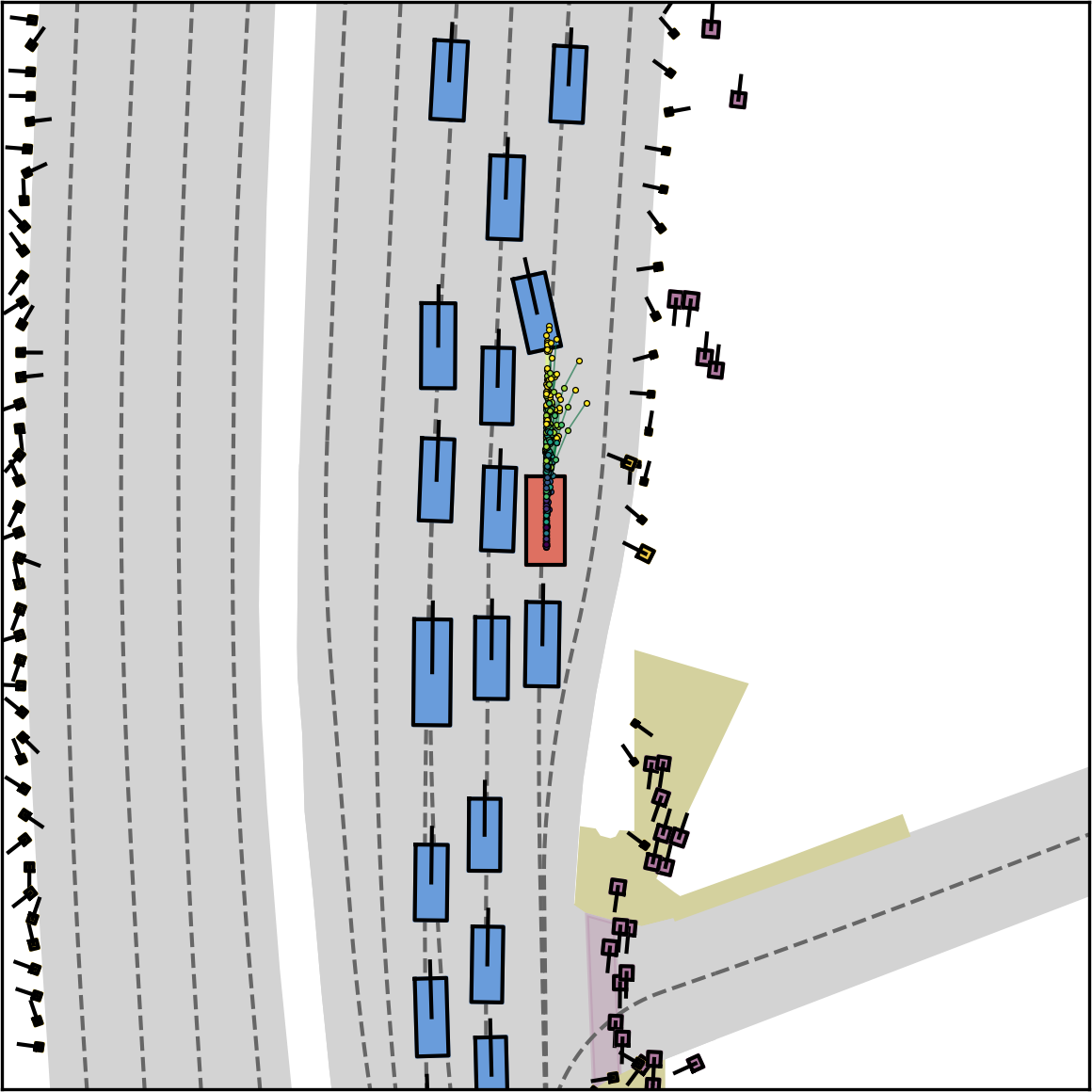} \\
 &
\vizimgall{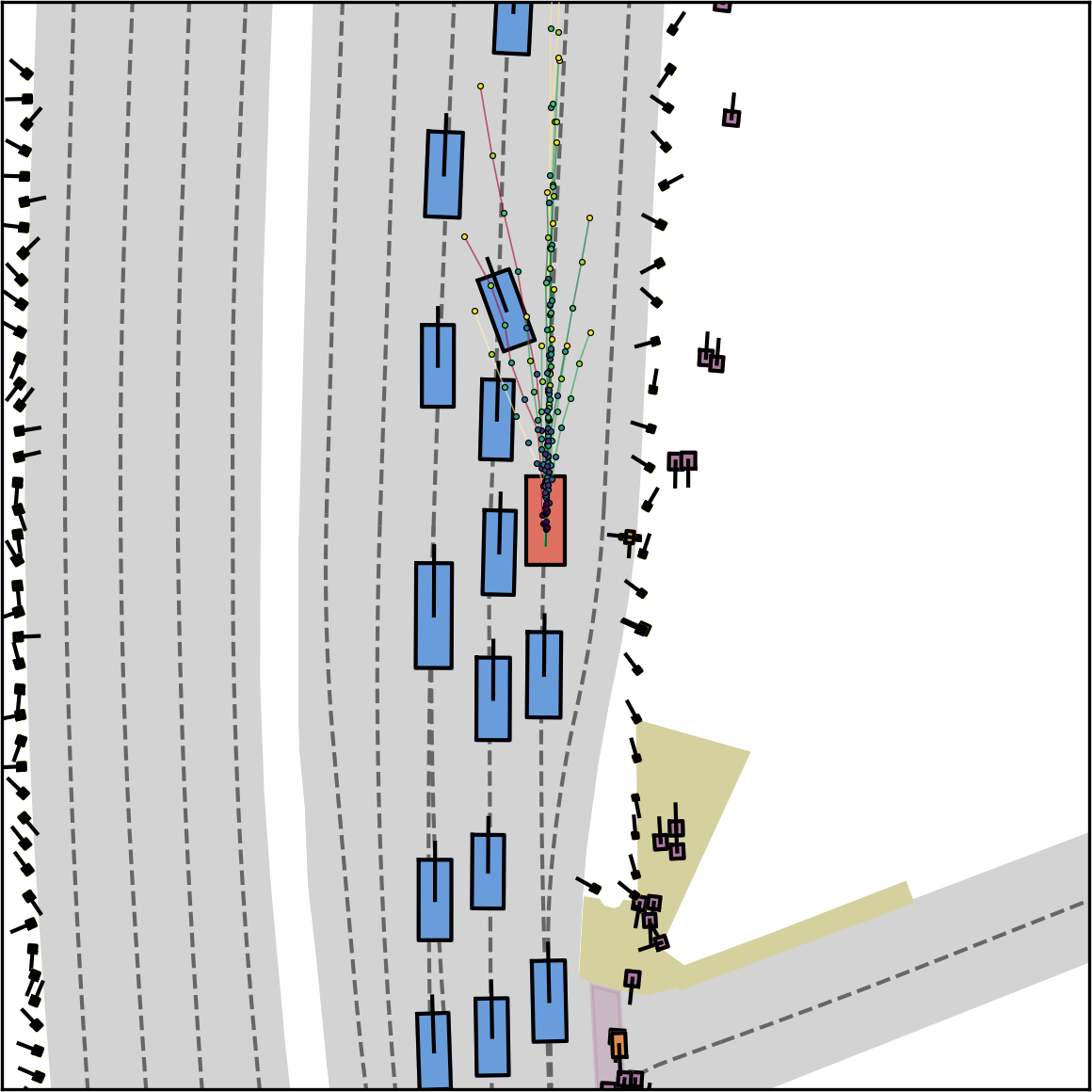} & \vizimgall{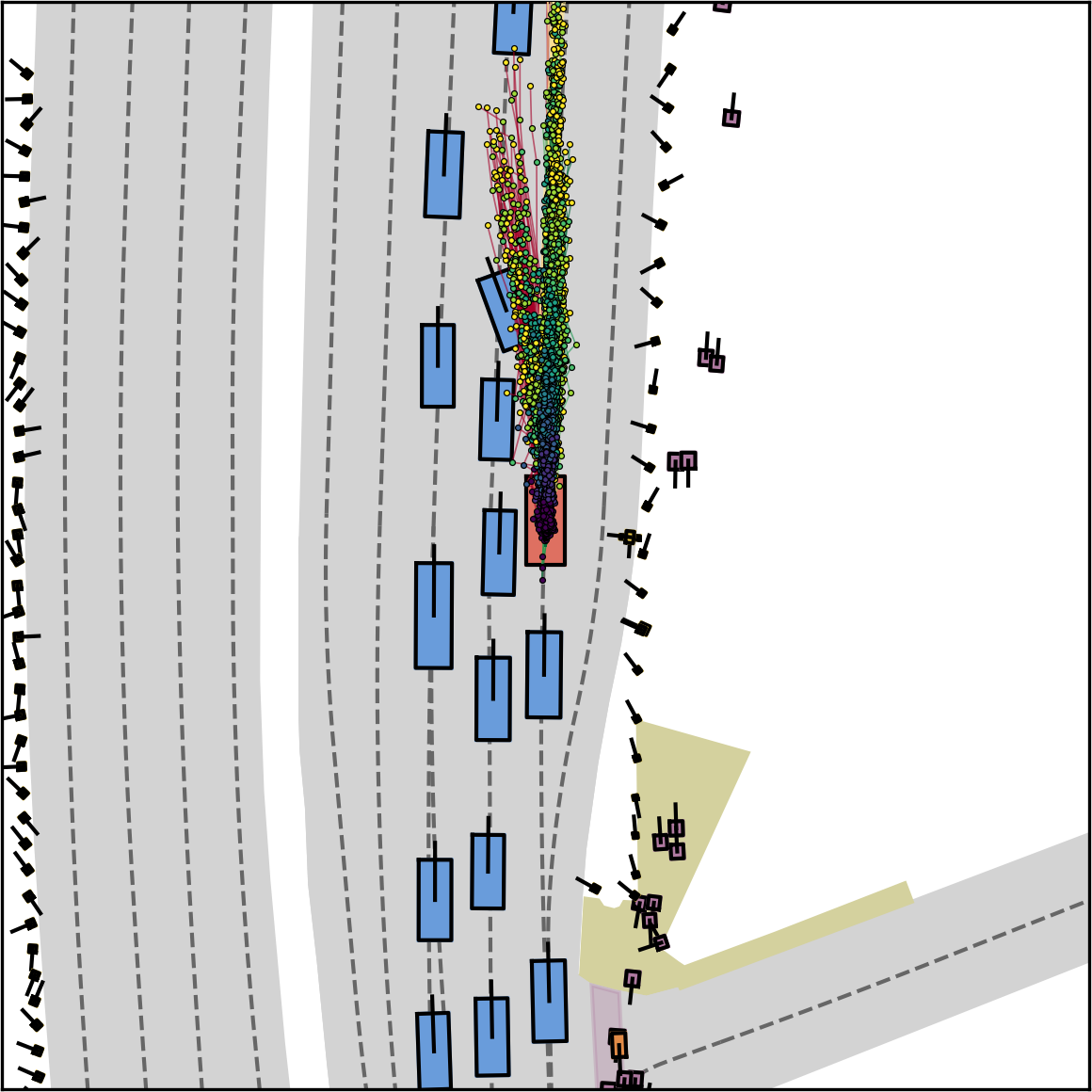} & \vizimgall{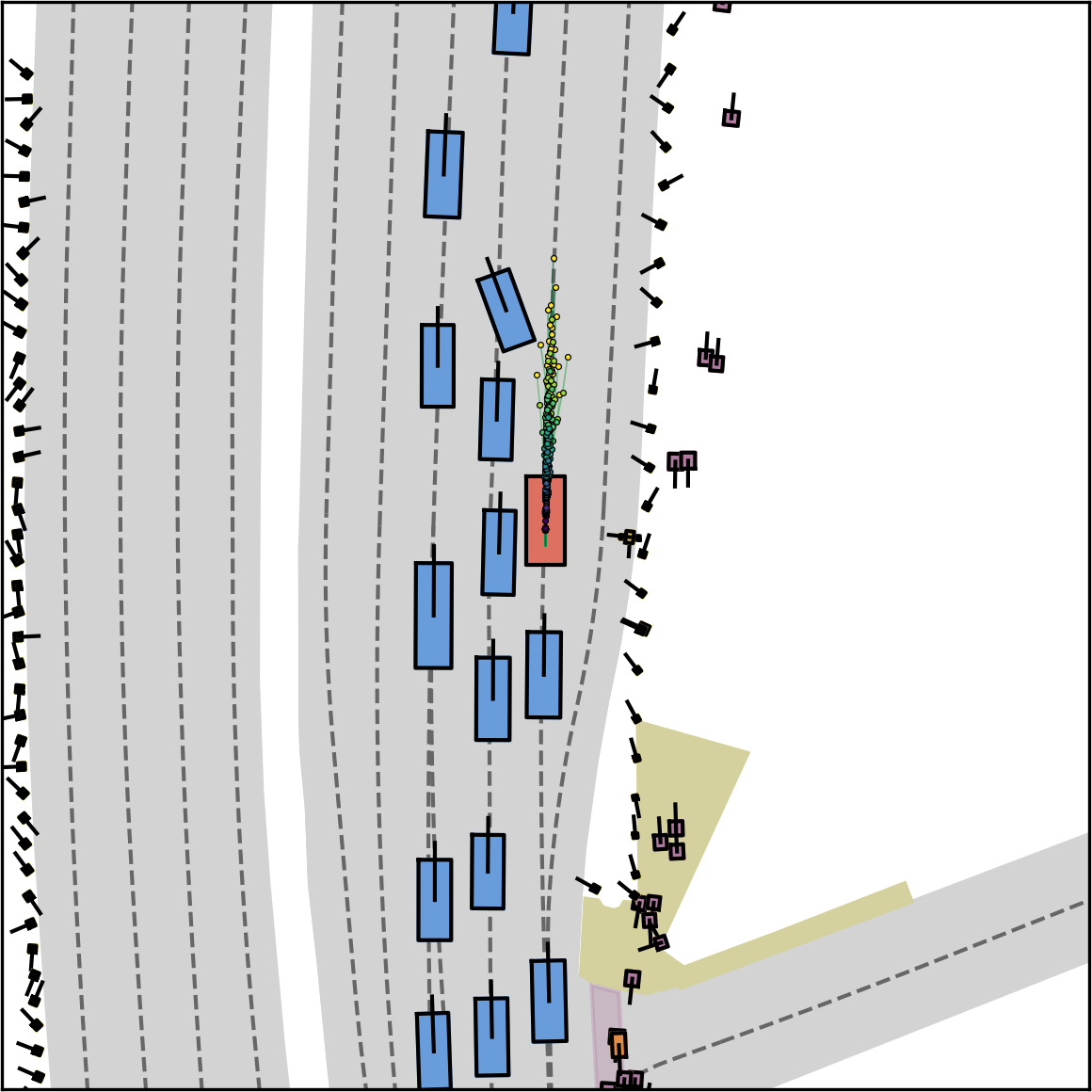} & \vizimgall{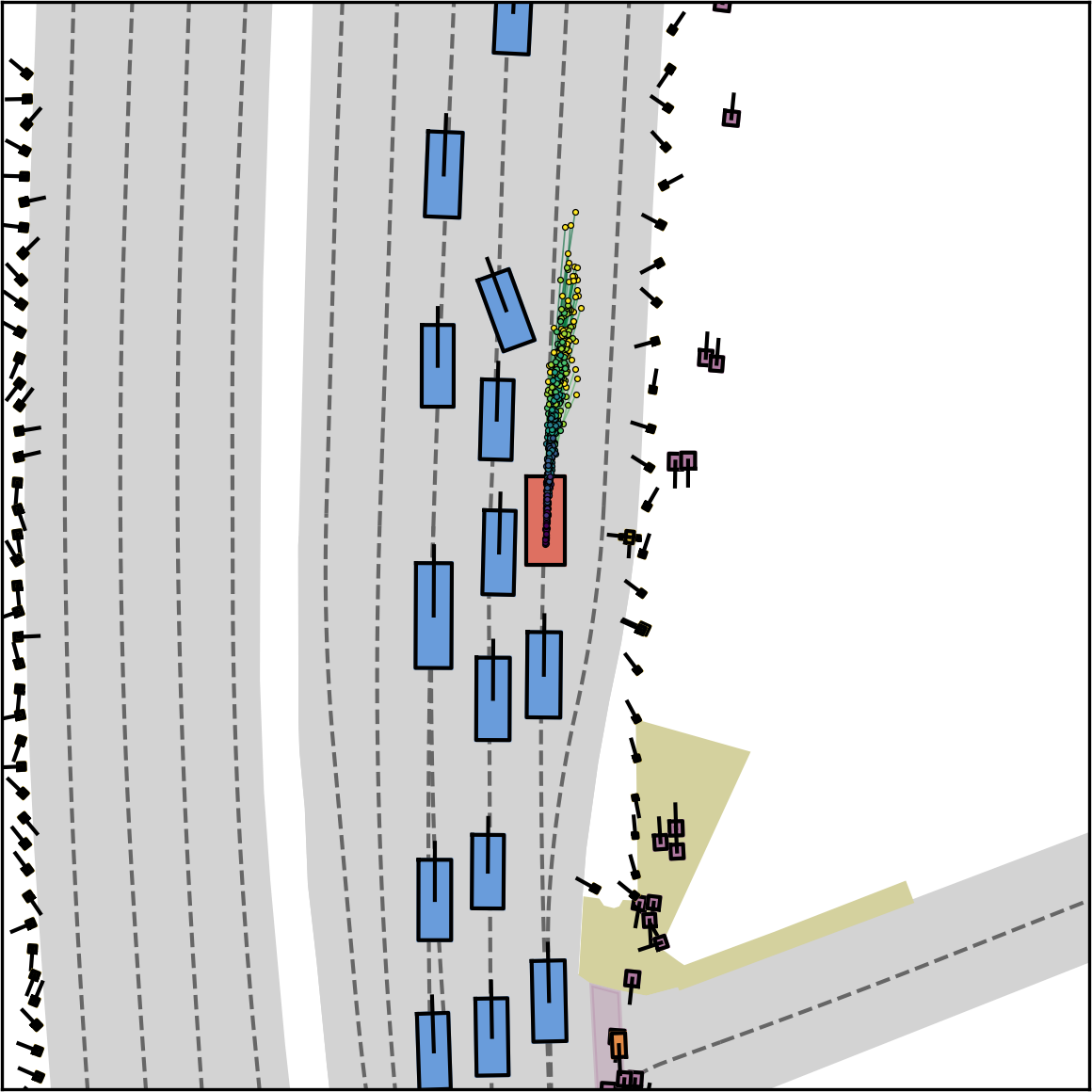} \\
\romarkall{Left} &
\vizimgall{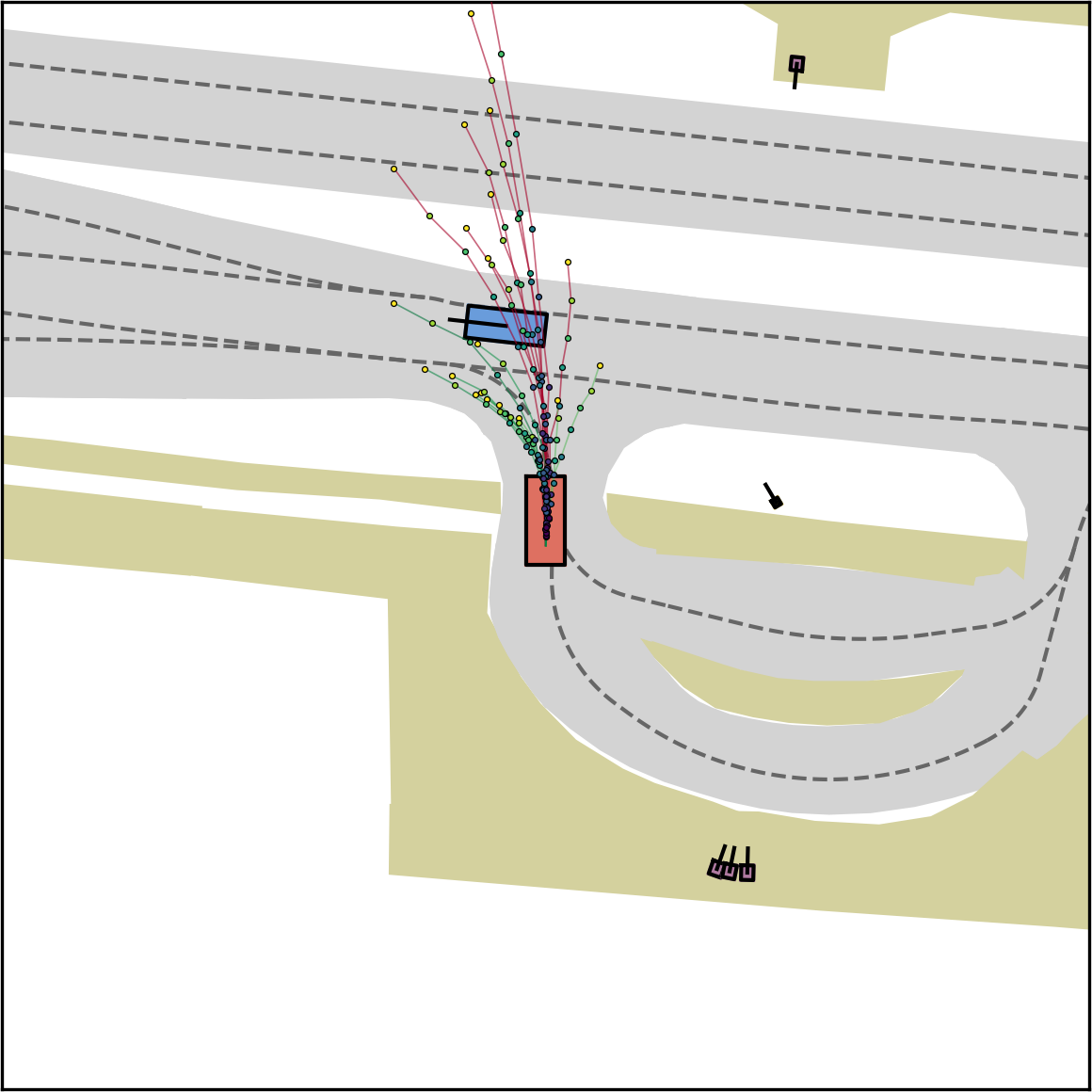} & \vizimgall{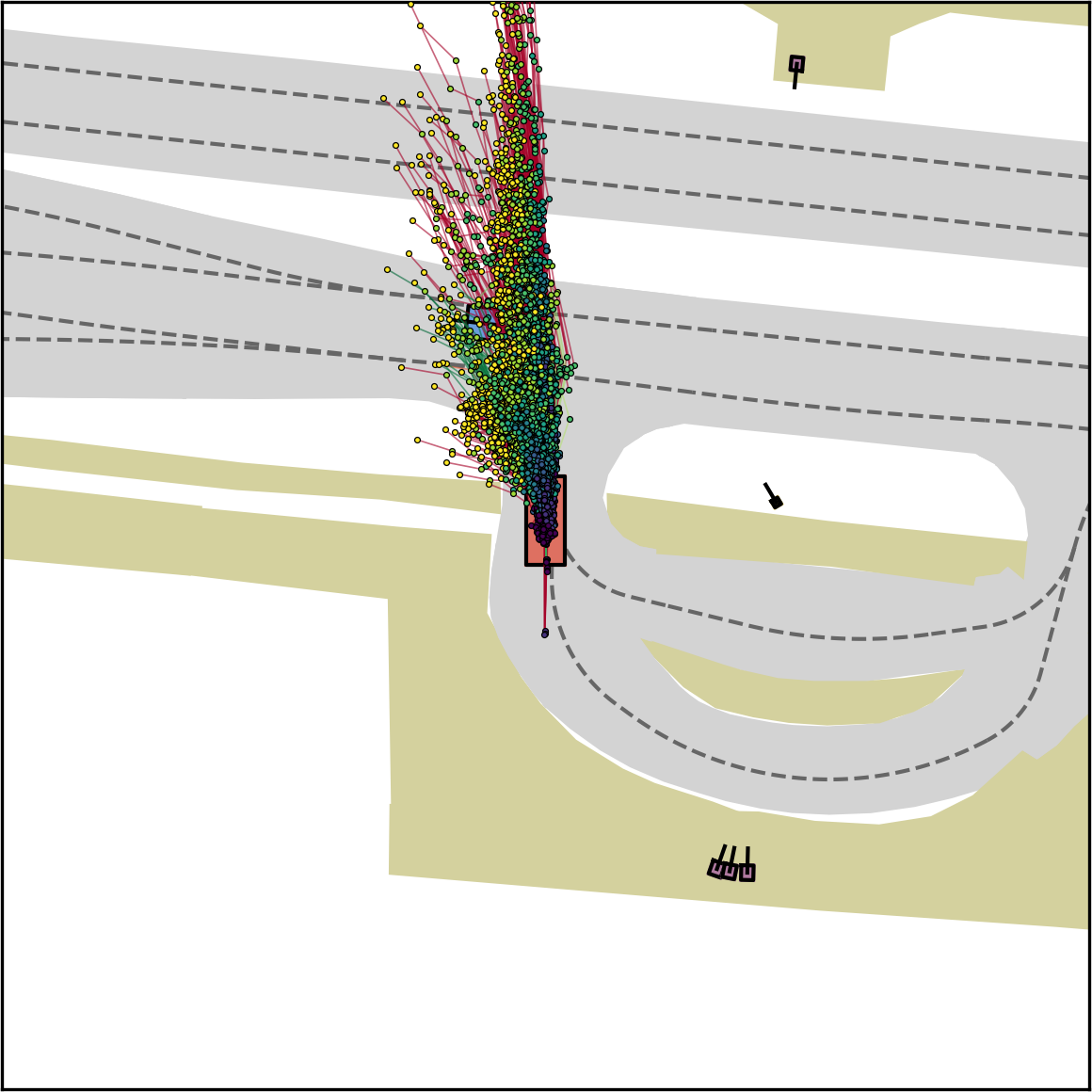} & \vizimgall{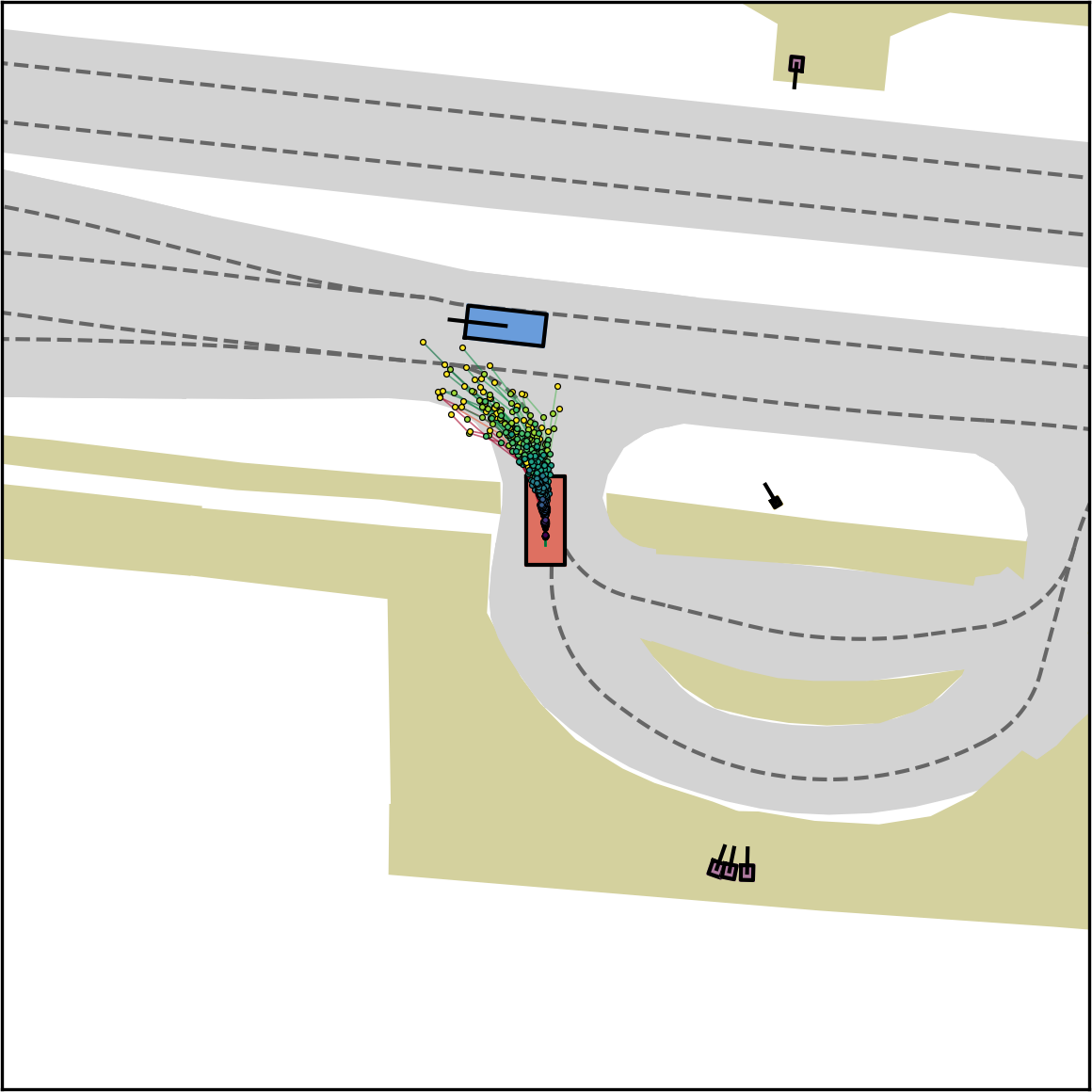} & \vizimgall{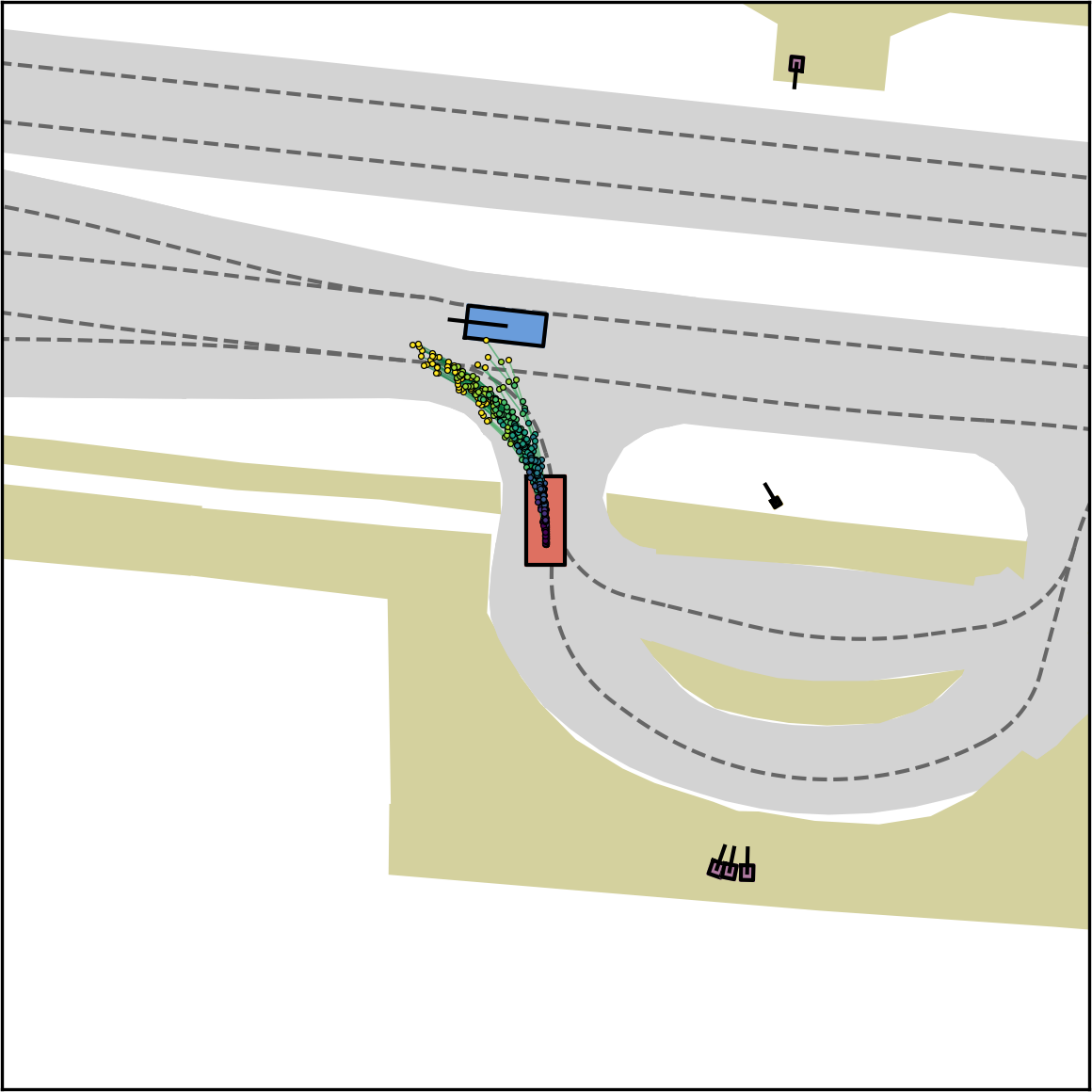} \\
 &
\vizimgall{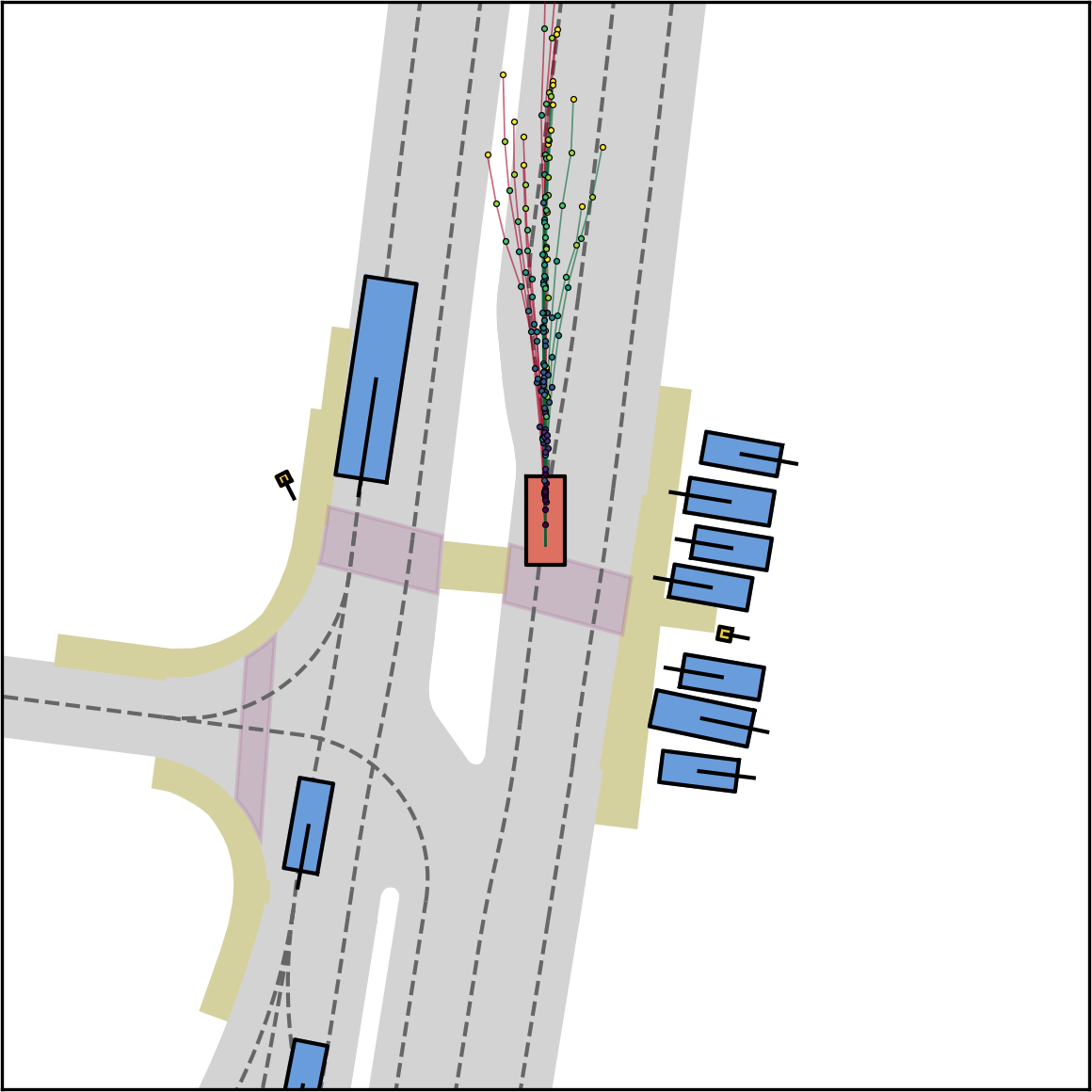} & \vizimgall{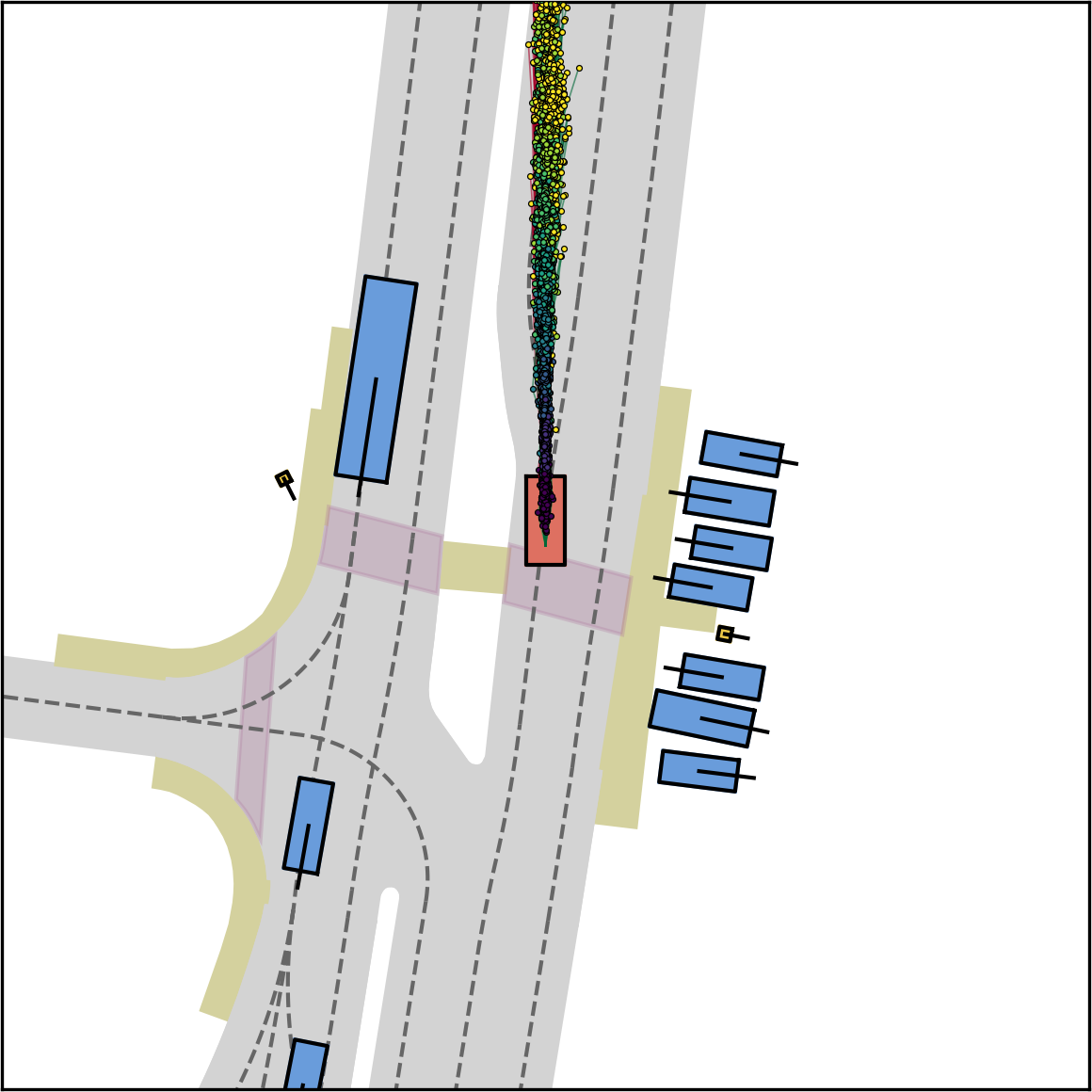} & \vizimgall{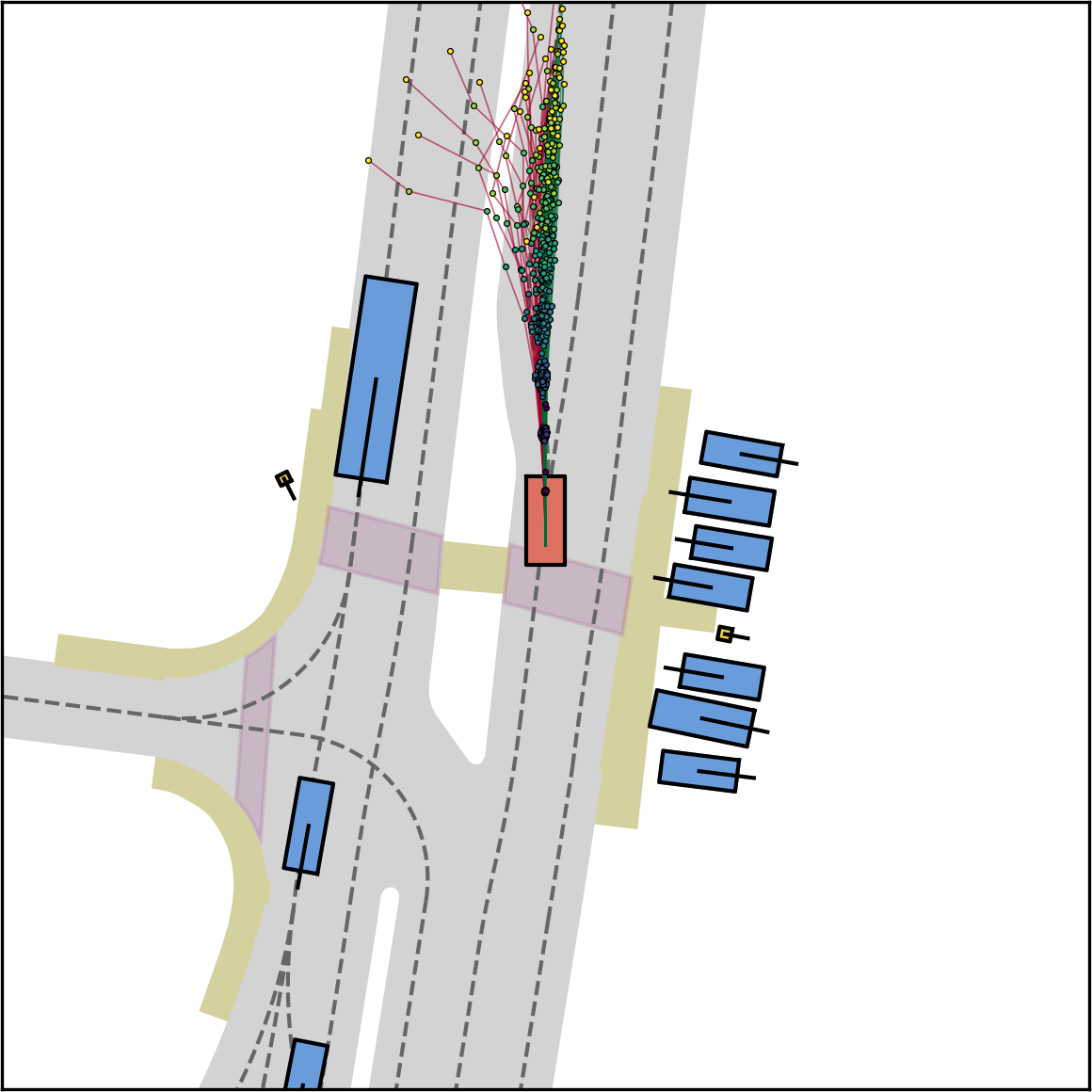} & \vizimgall{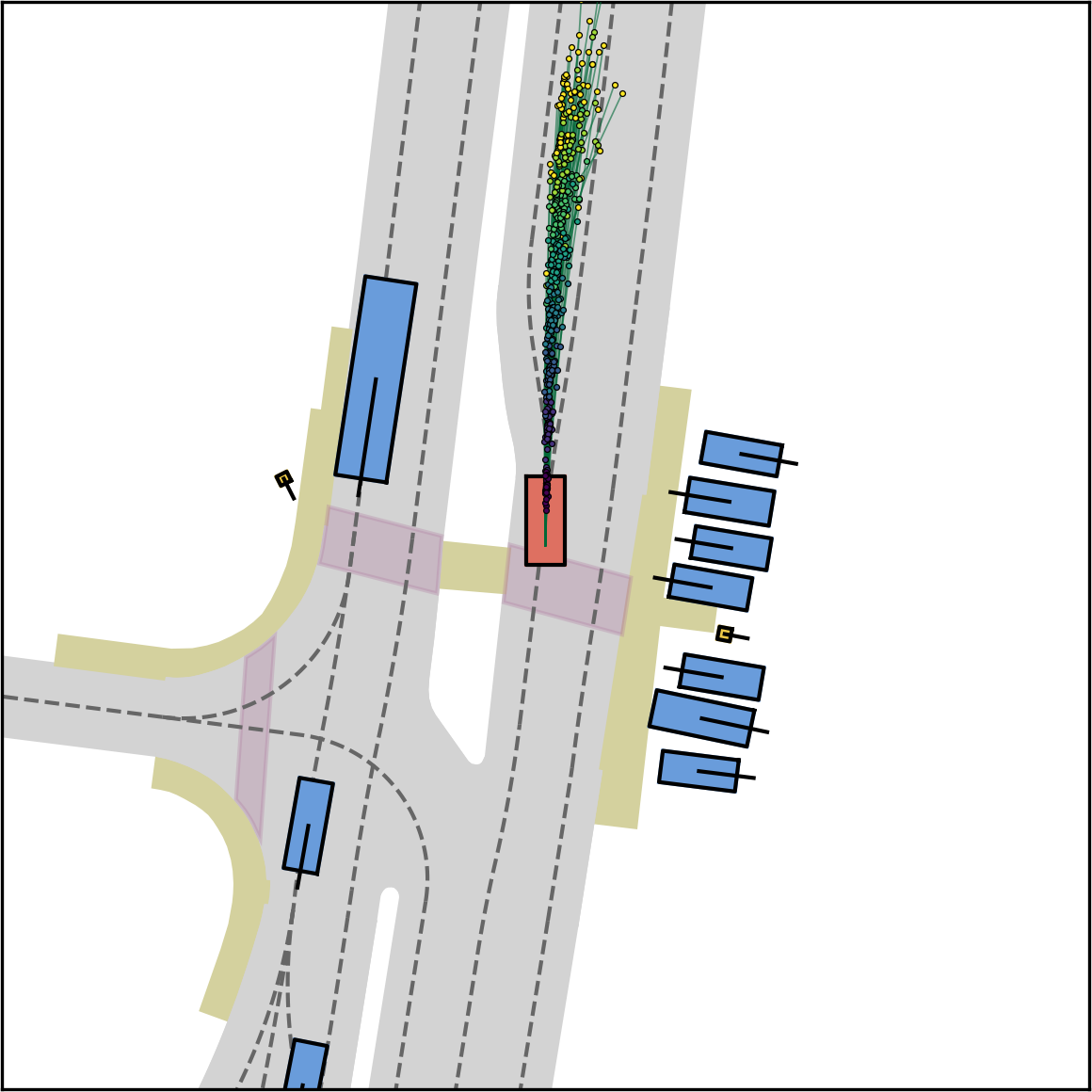} \\
\romarkall{Right} &
\vizimgall{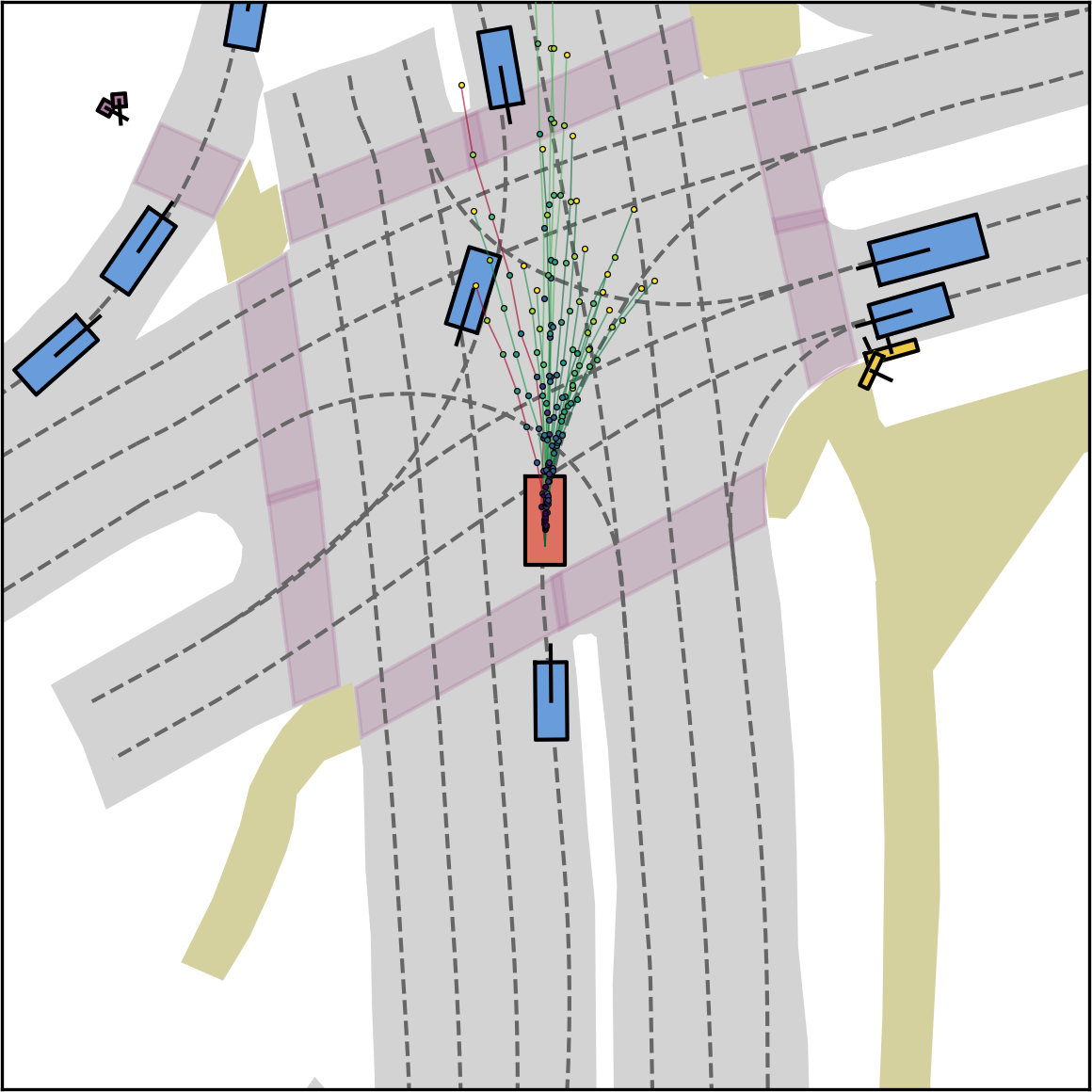} & \vizimgall{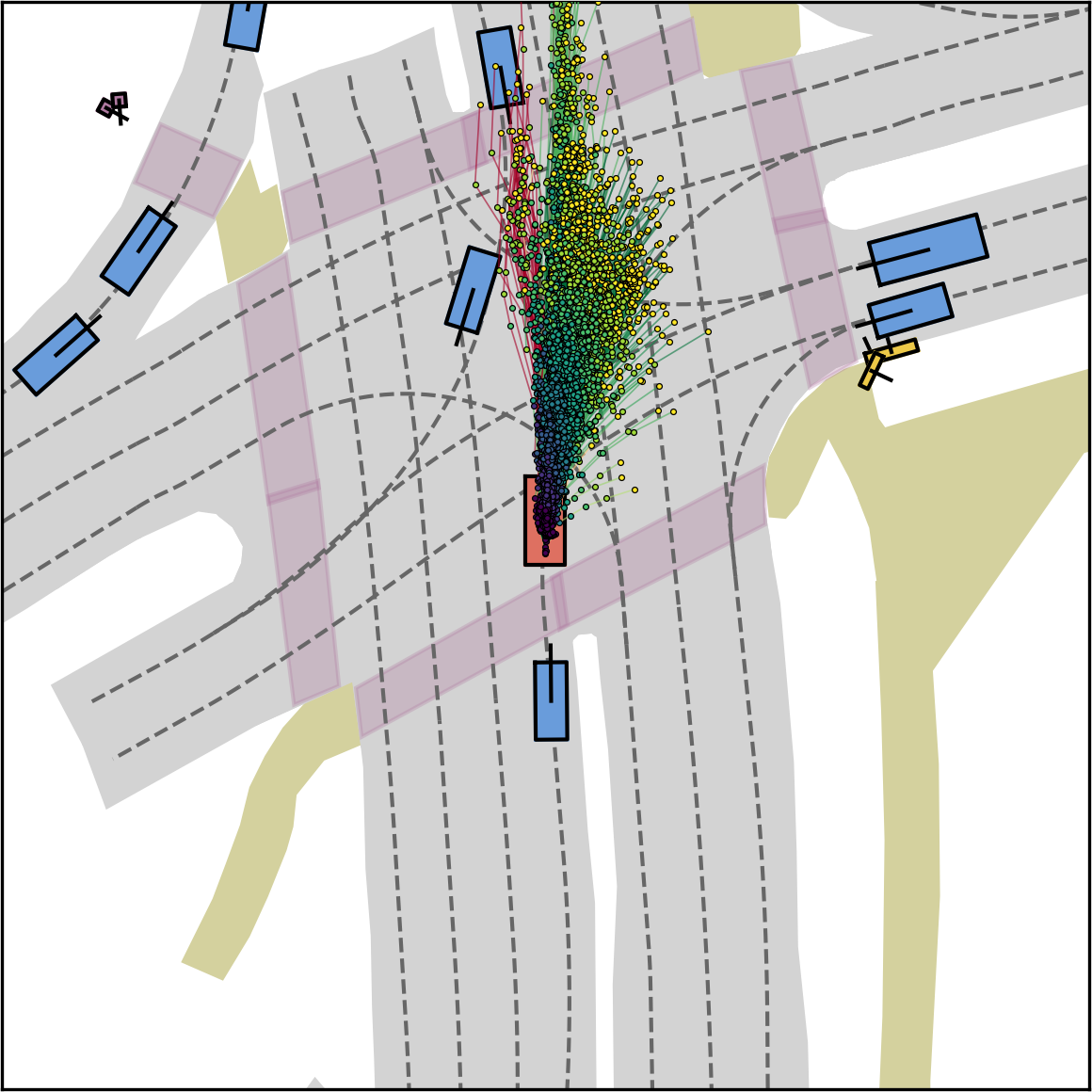} & \vizimgall{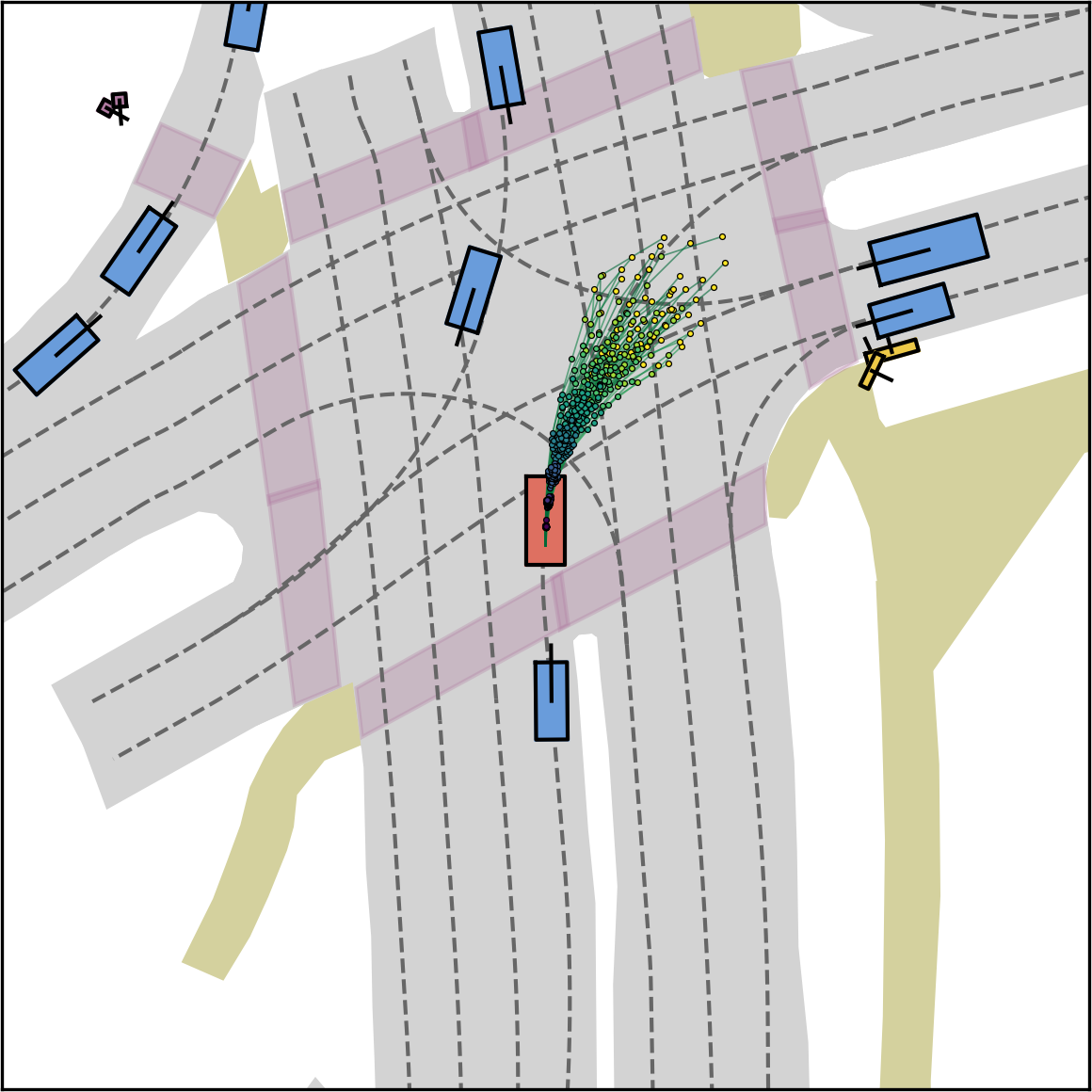} & \vizimgall{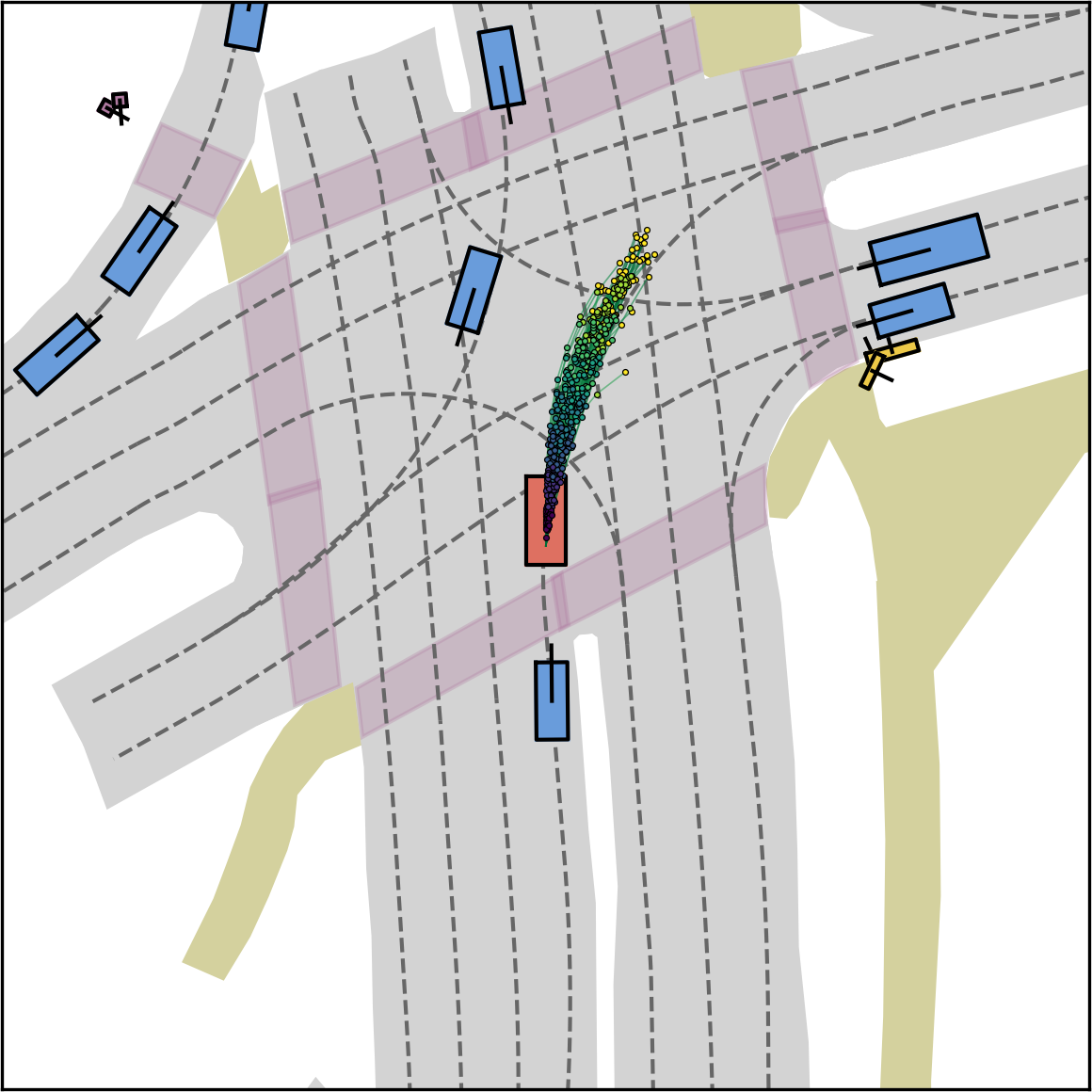} \\
 &
\vizimgall{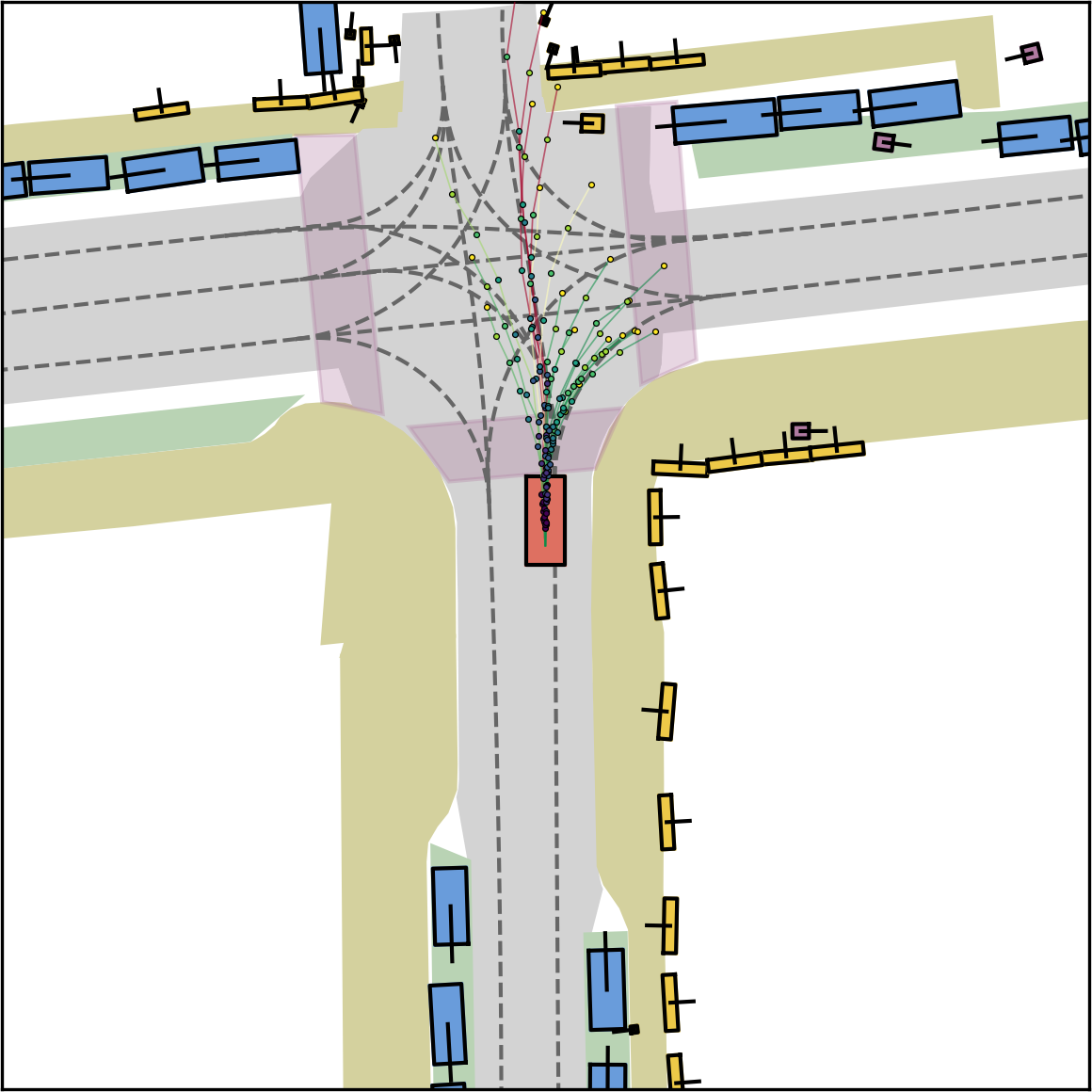} & \vizimgall{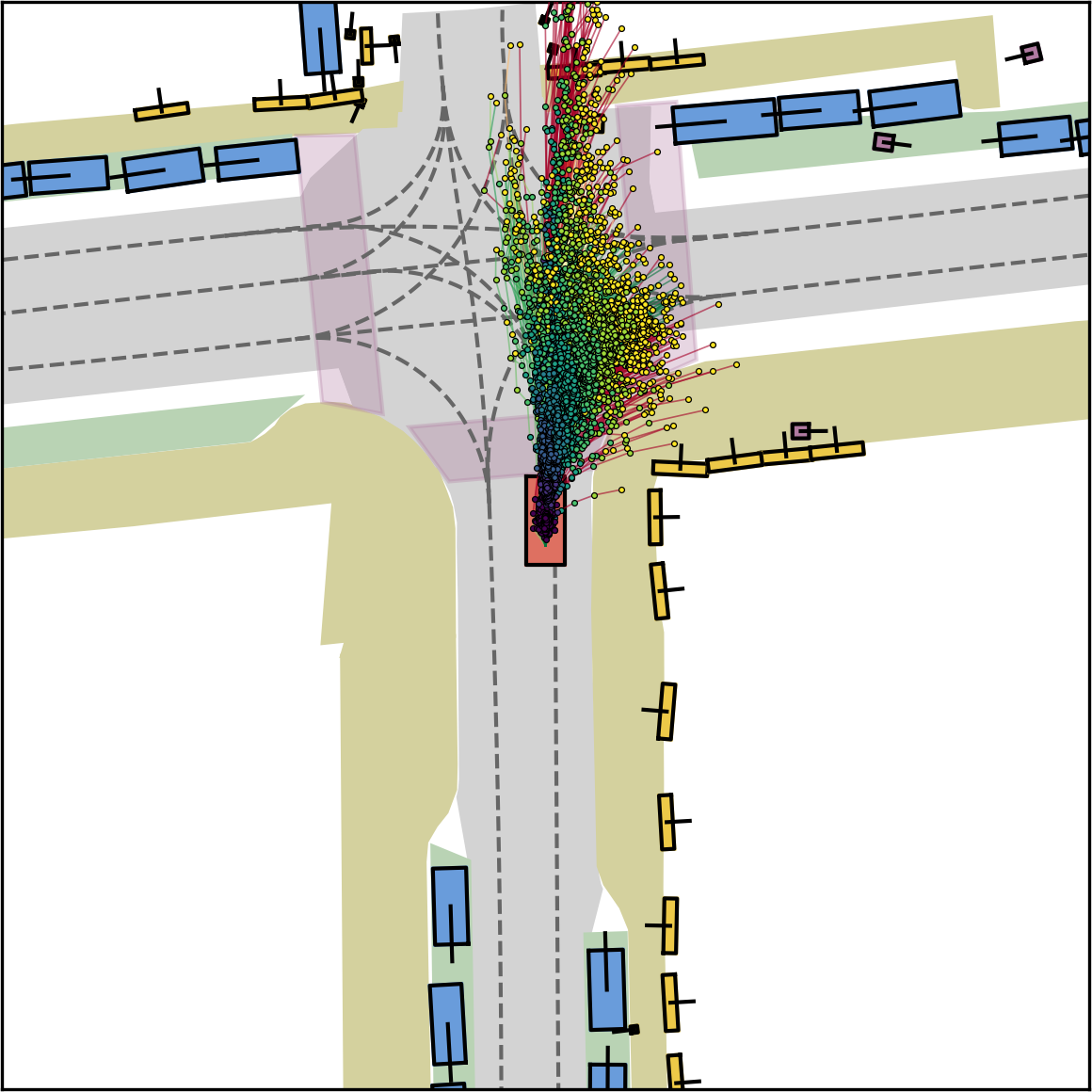} & \vizimgall{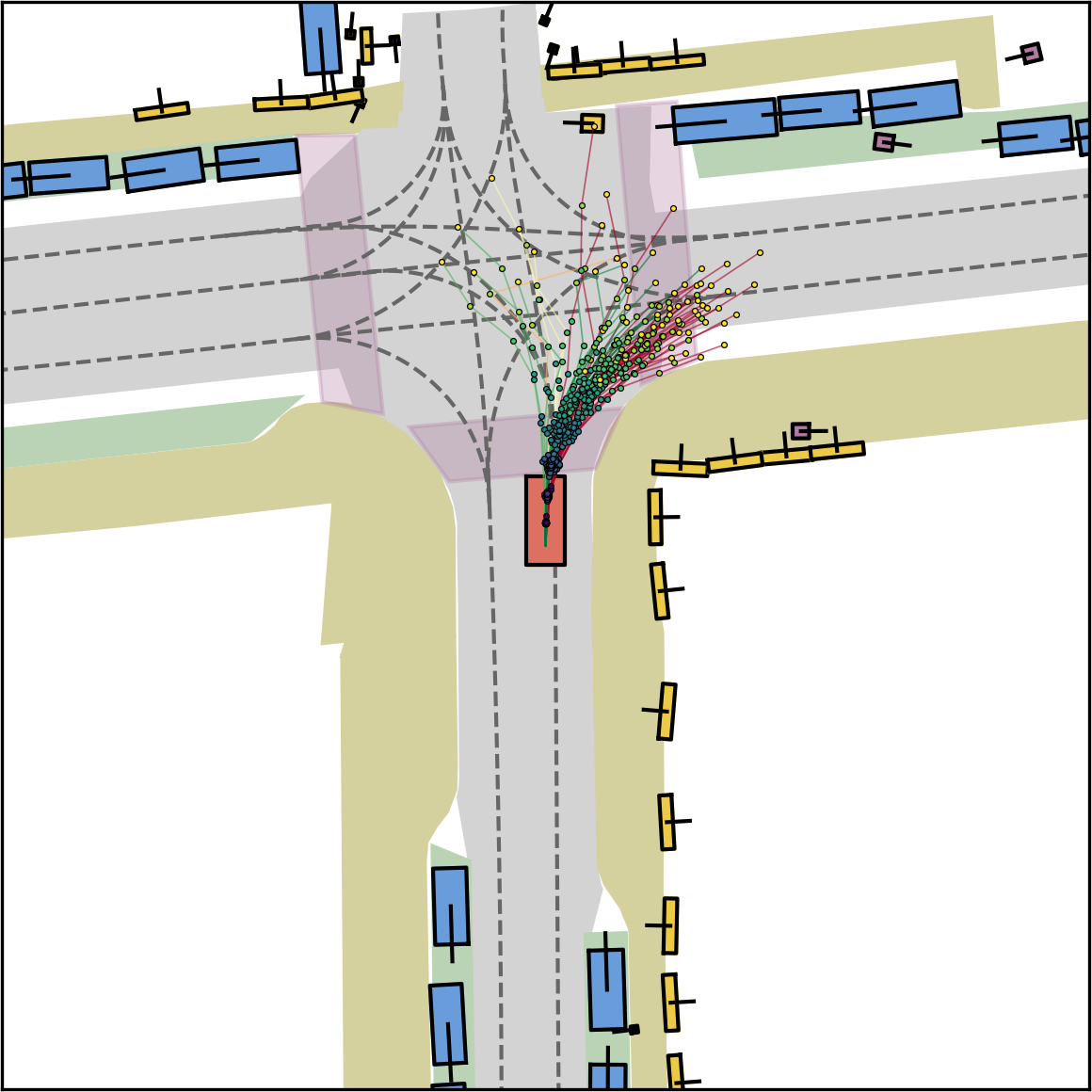} & \vizimgall{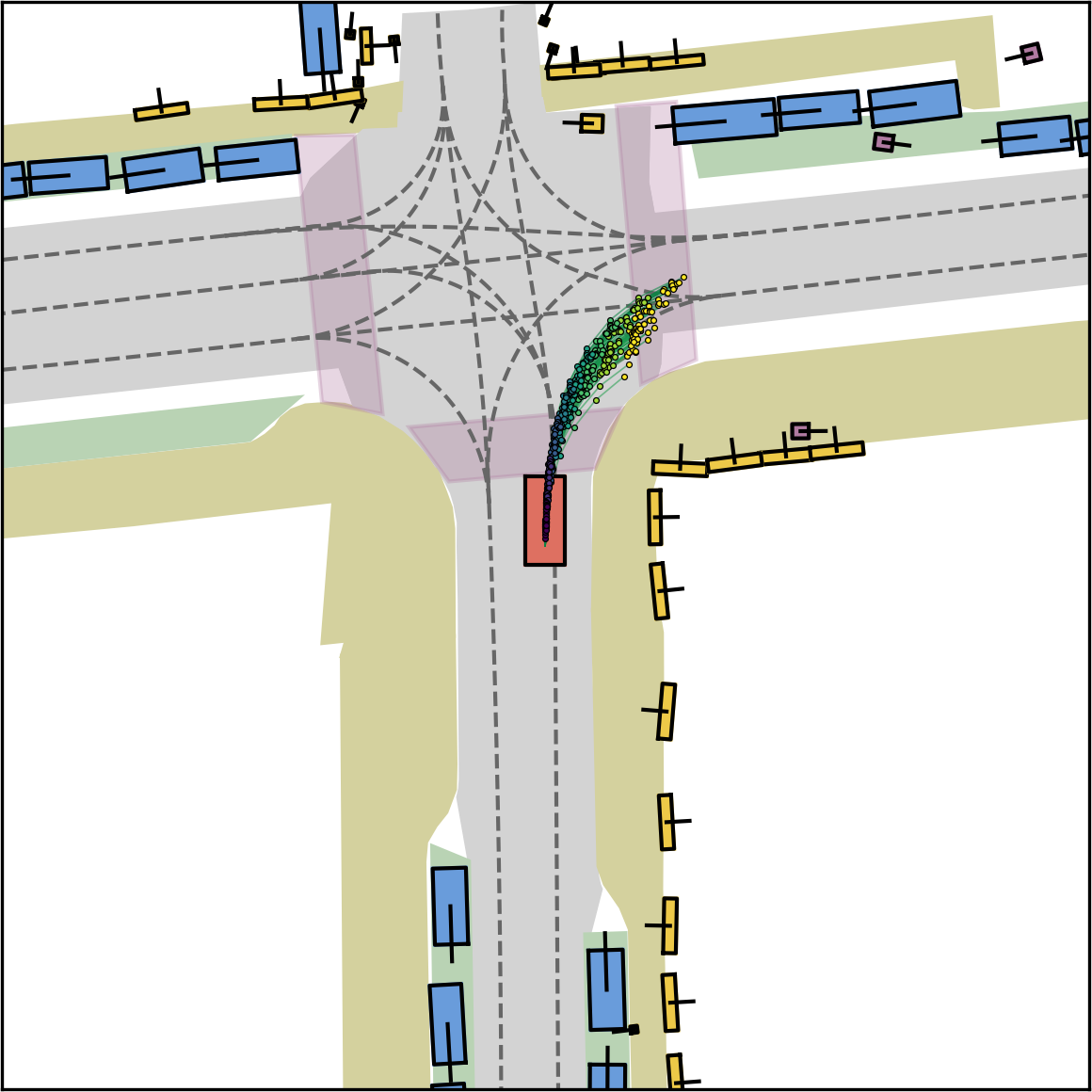} \\
\end{tabular}
\caption{Full qualitative comparison part 2 of 3. Trajectories are colored by PDMS from 0 ({\color[RGB]{200,30,30}red}) to 1 ({\color[RGB]{30,150,30}green}). Scenes are grouped by driving command labeled left.}
\label{fig:qual_all_2}
\end{figure*}

\begin{figure*}[t]
\centering
\setlength{\tabcolsep}{1pt}
\renewcommand{\arraystretch}{0.3}
\begin{tabular}{@{}c cccc@{}}
& \textbf{DiffusionDrive} & \textbf{DiffusionDriveV2} & \textbf{iPad} & \textbf{\ourmethod{} (Ours)} \\[2pt]
\romarkall{Forward} &
\vizimgall{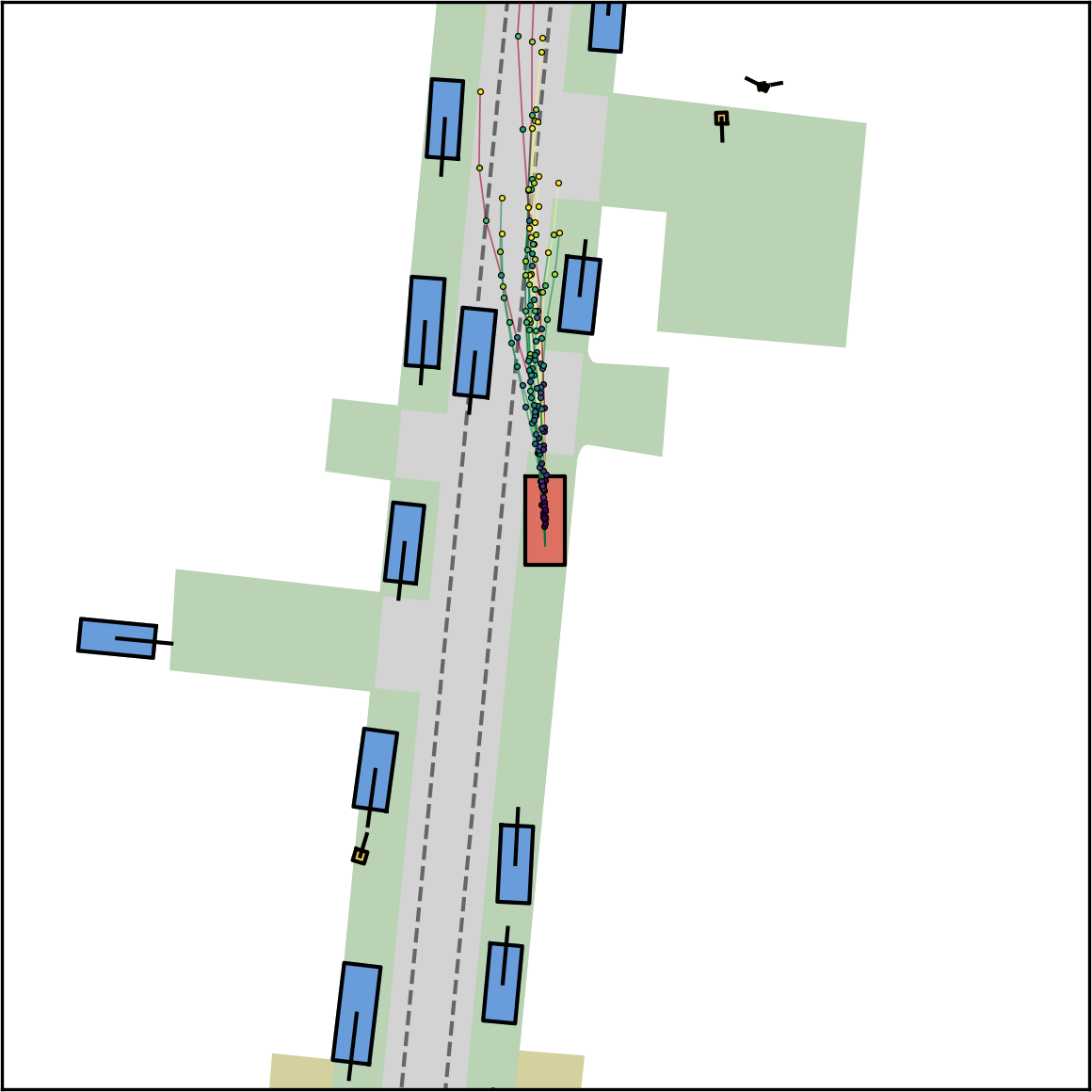} & \vizimgall{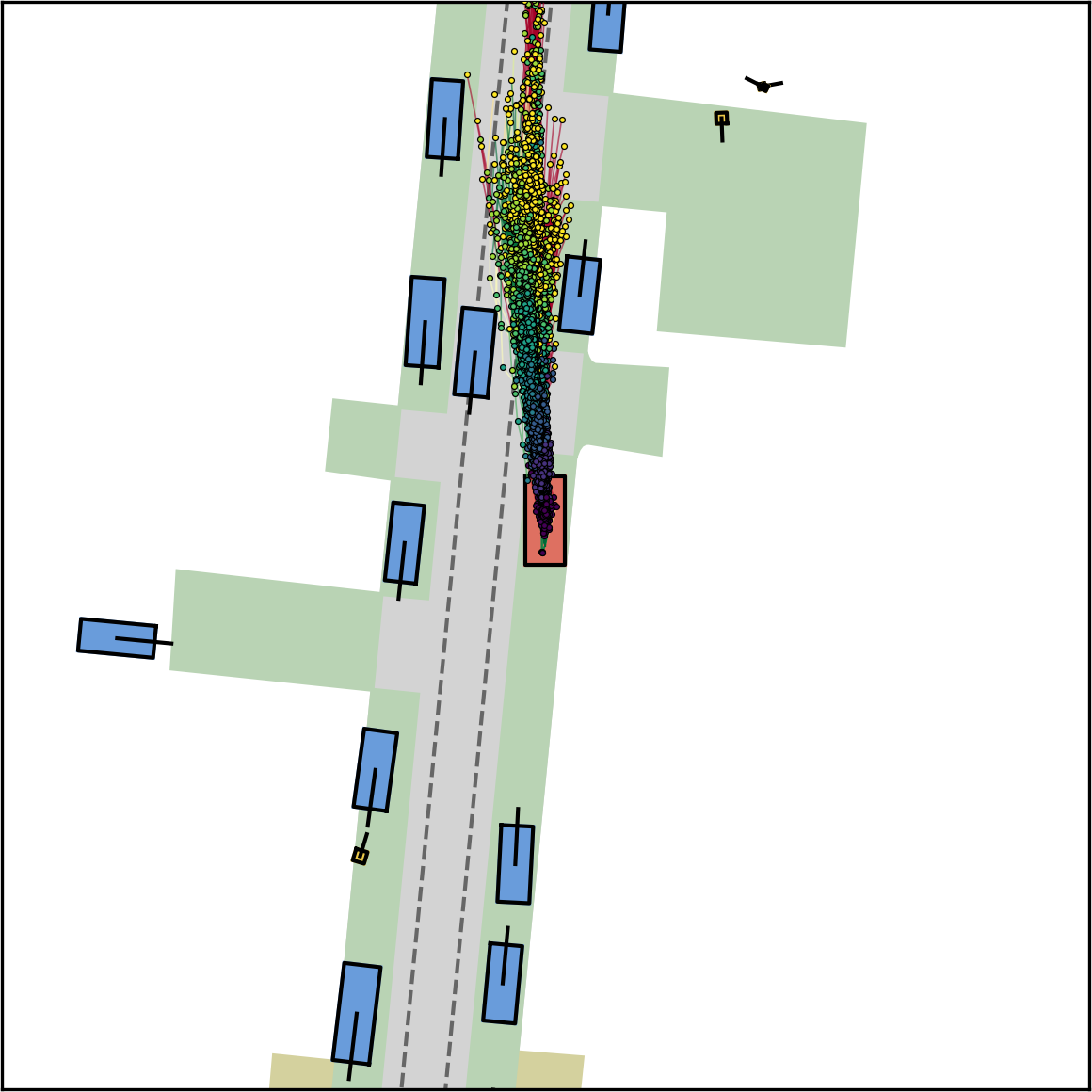} & \vizimgall{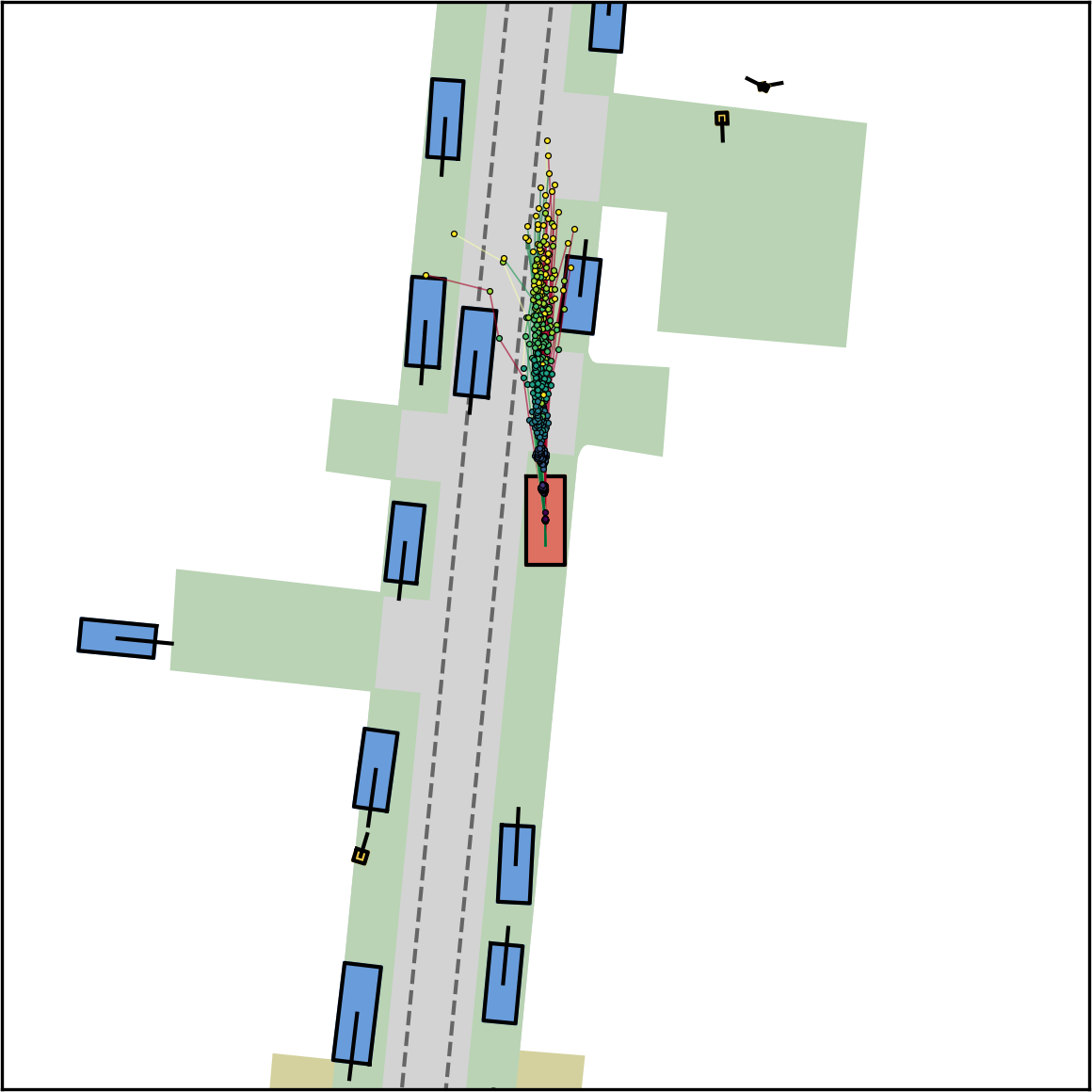} & \vizimgall{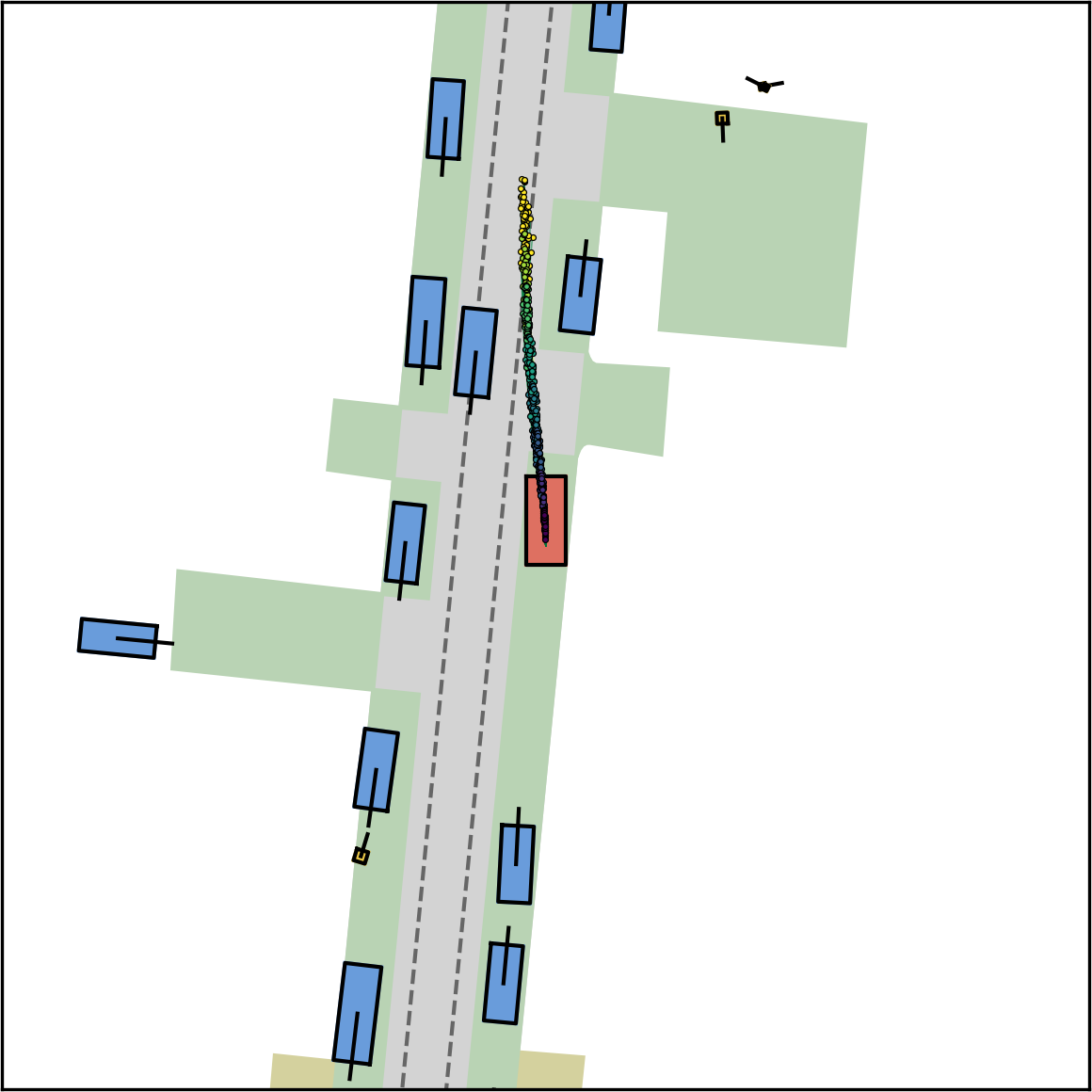} \\
 &
\vizimgall{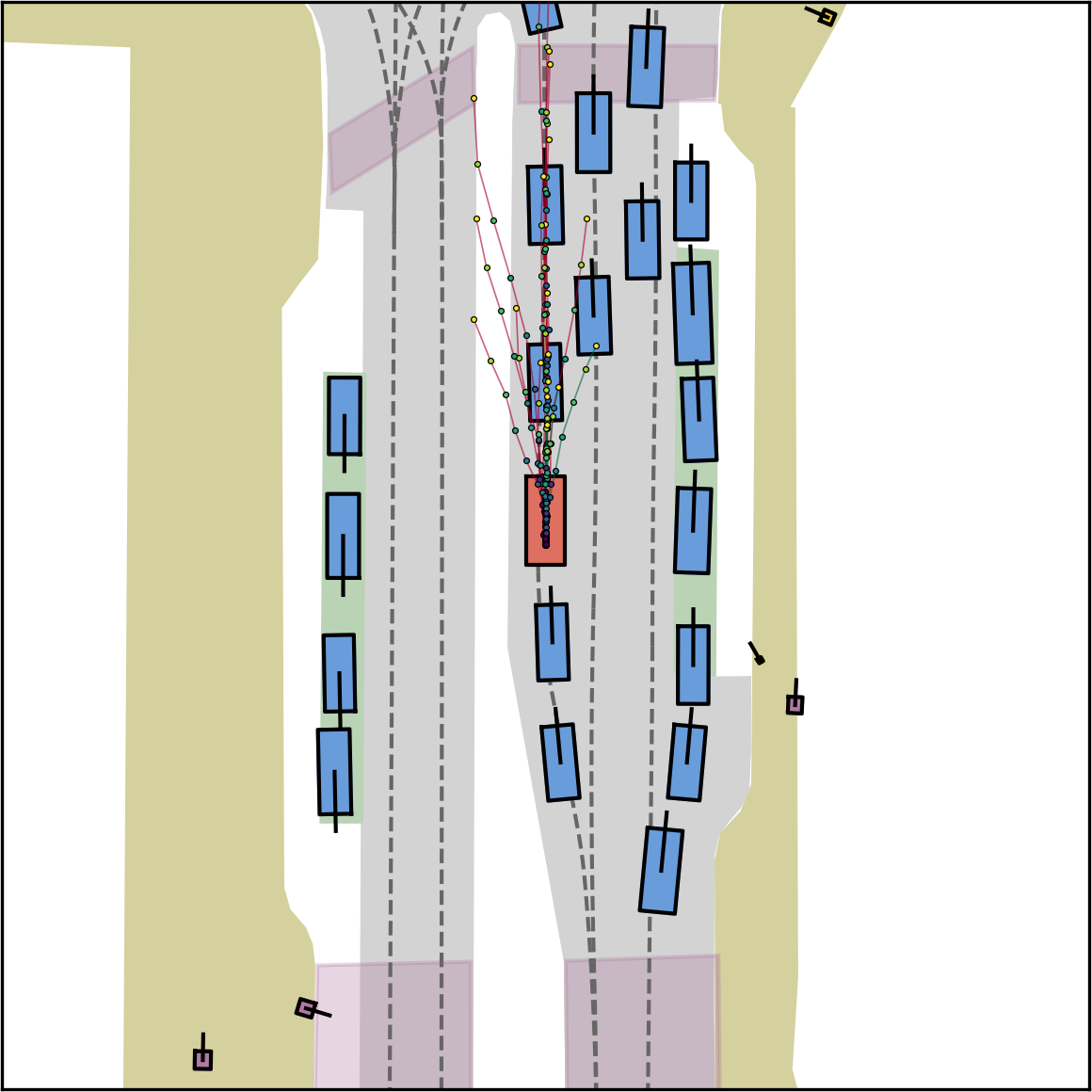} & \vizimgall{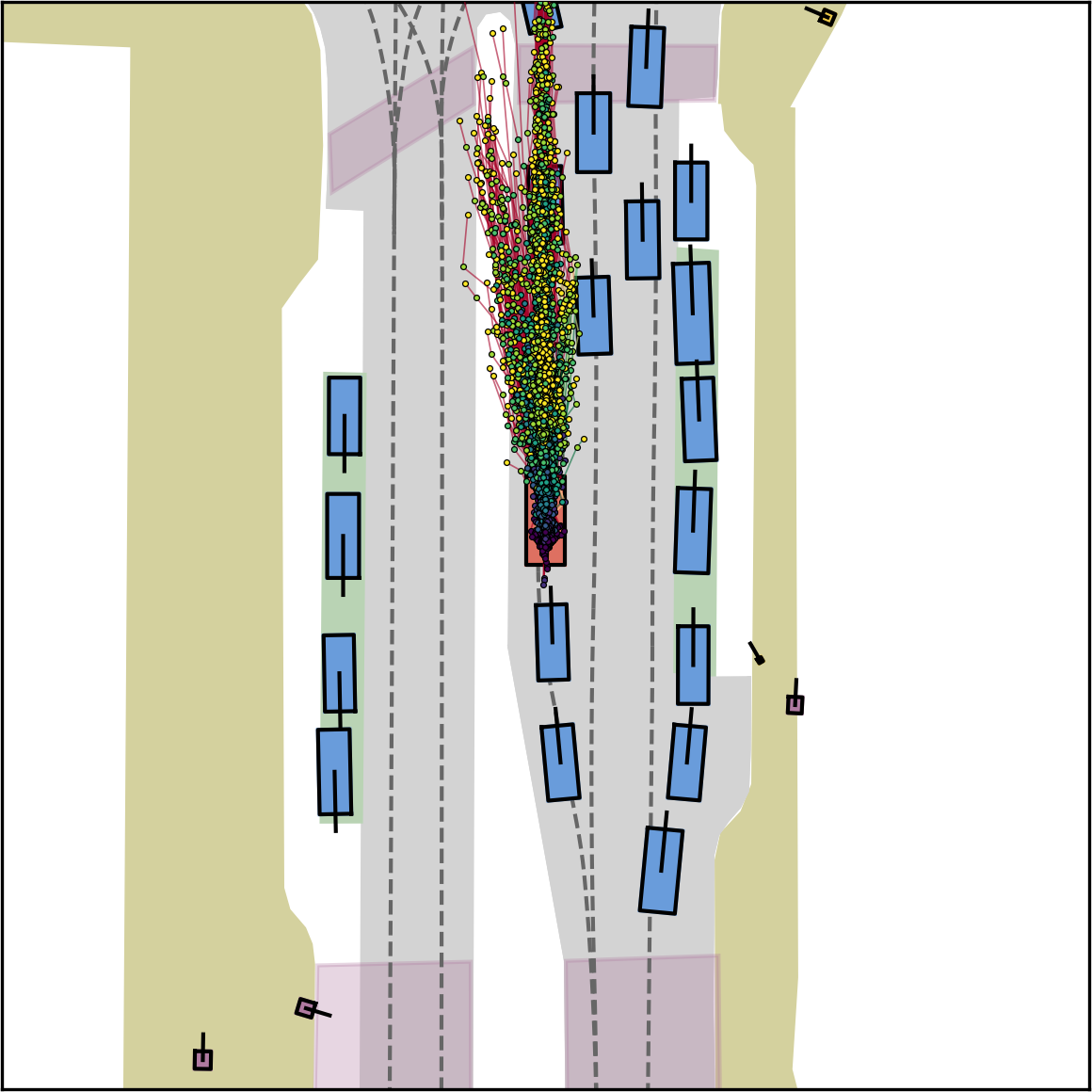} & \vizimgall{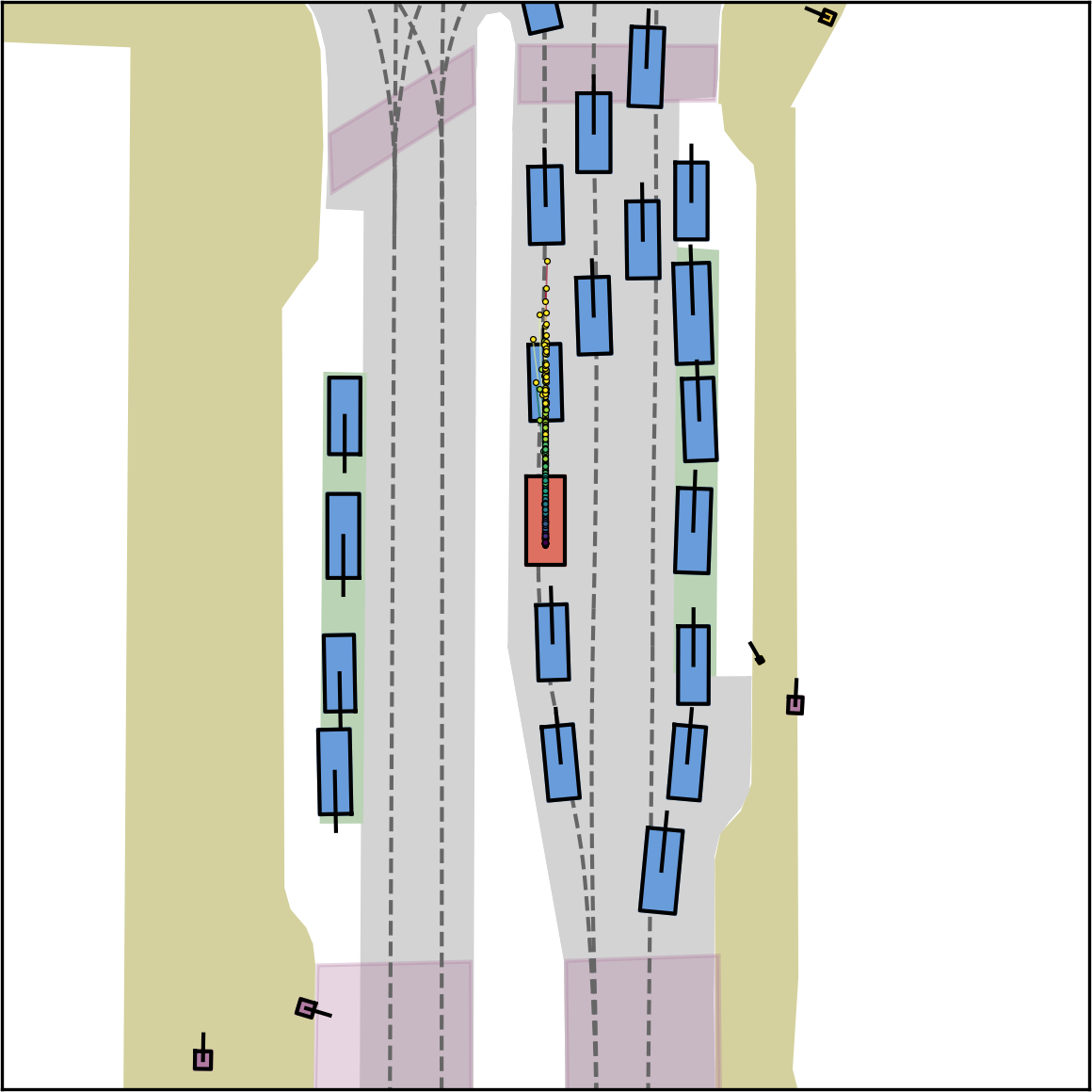} & \vizimgall{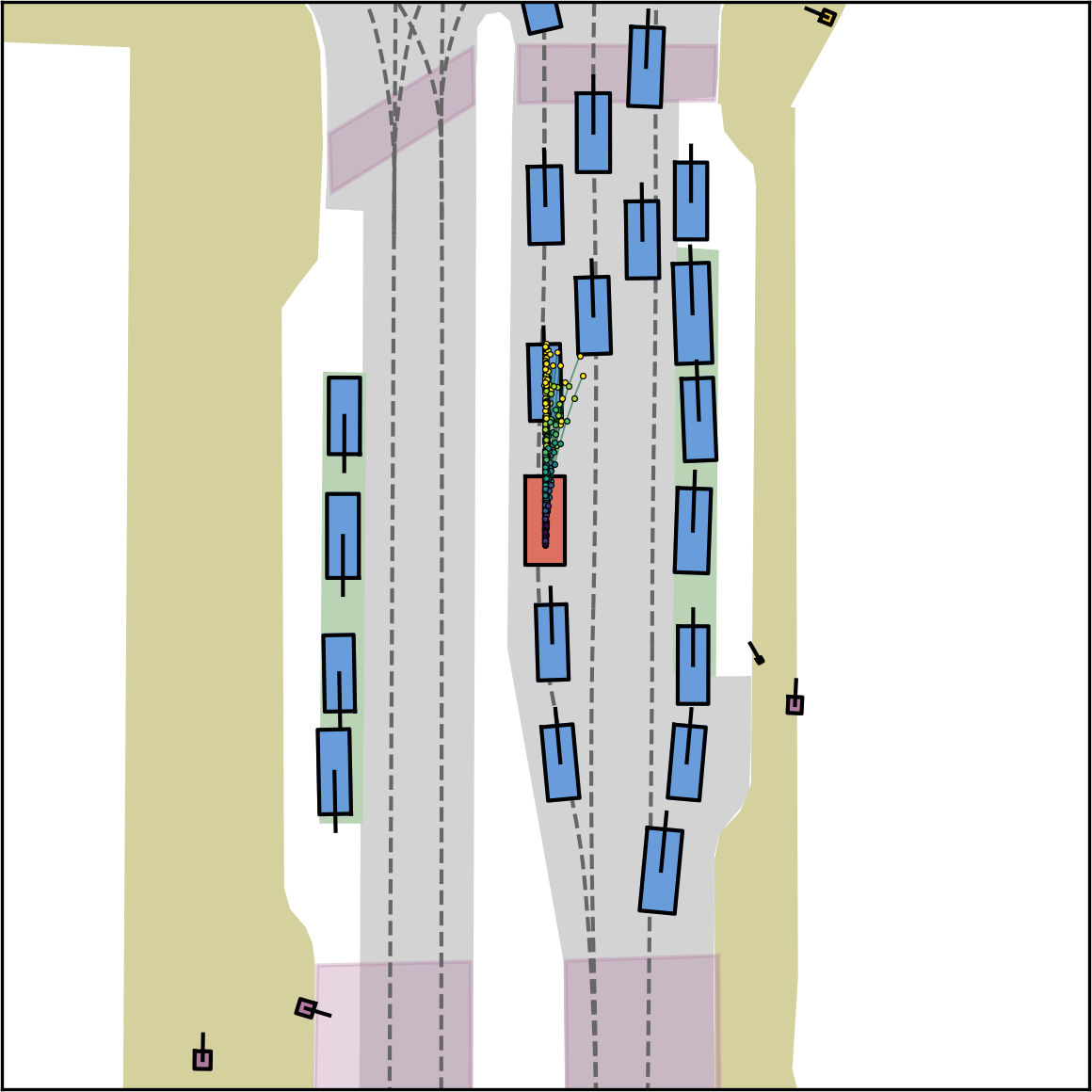} \\
 &
\vizimgall{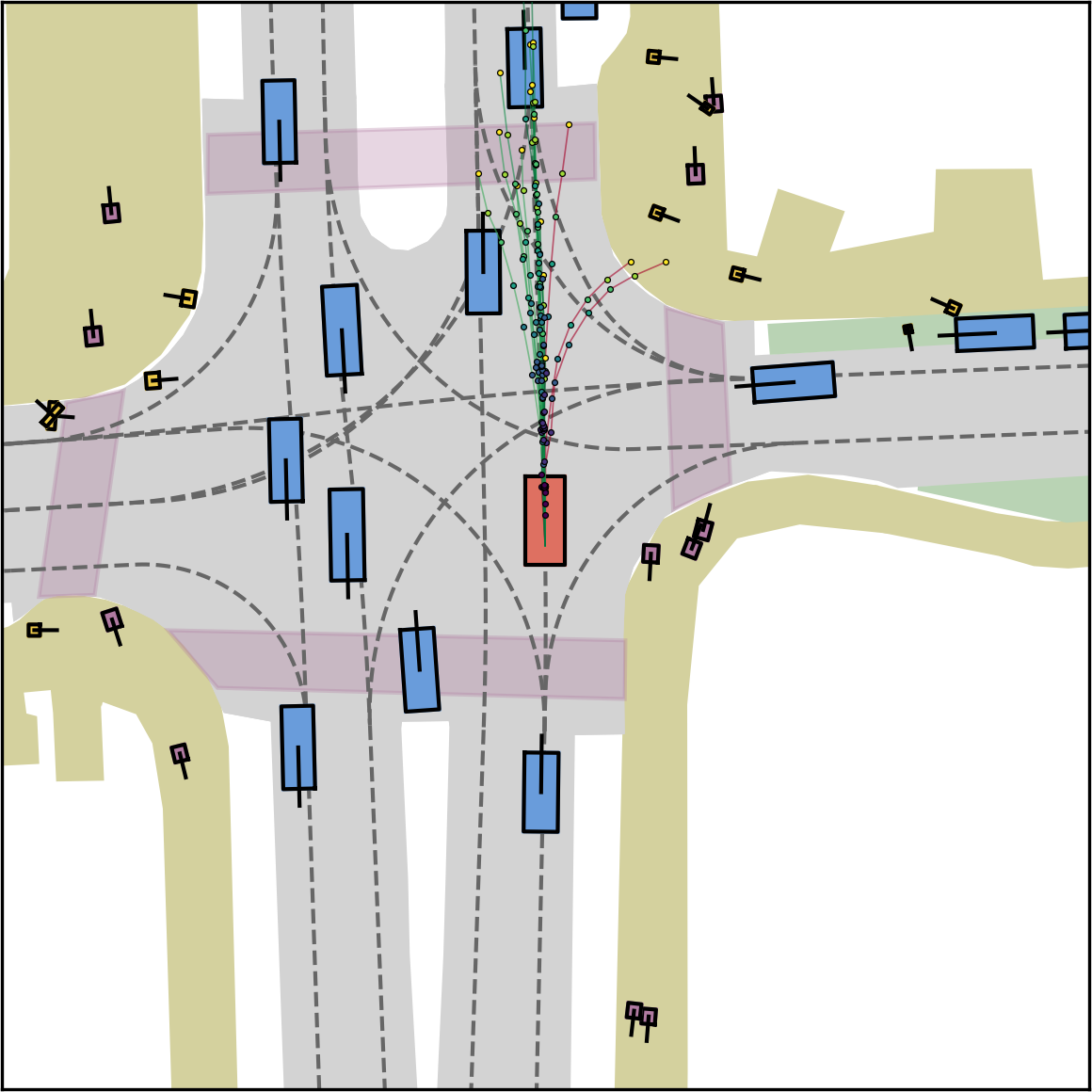} & \vizimgall{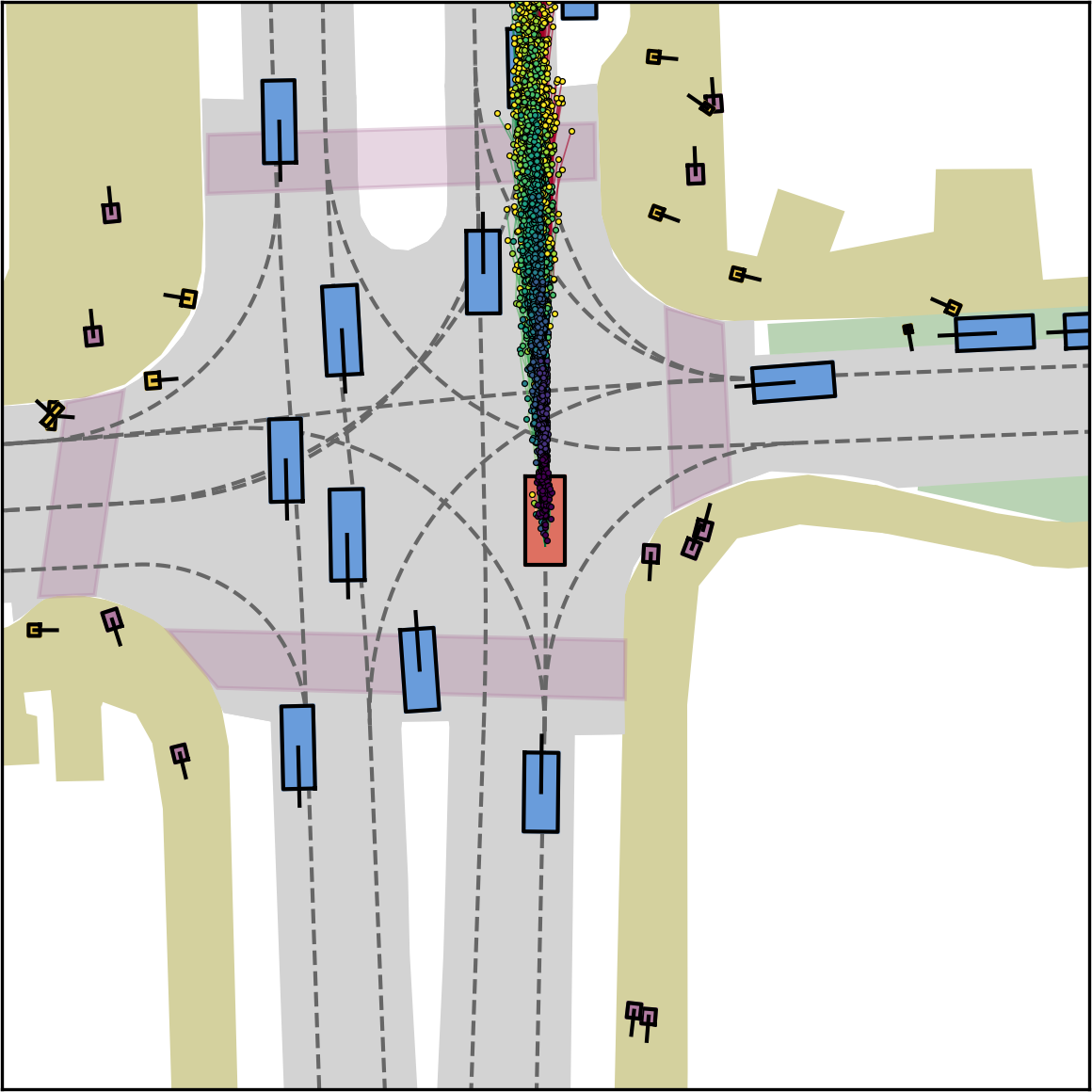} & \vizimgall{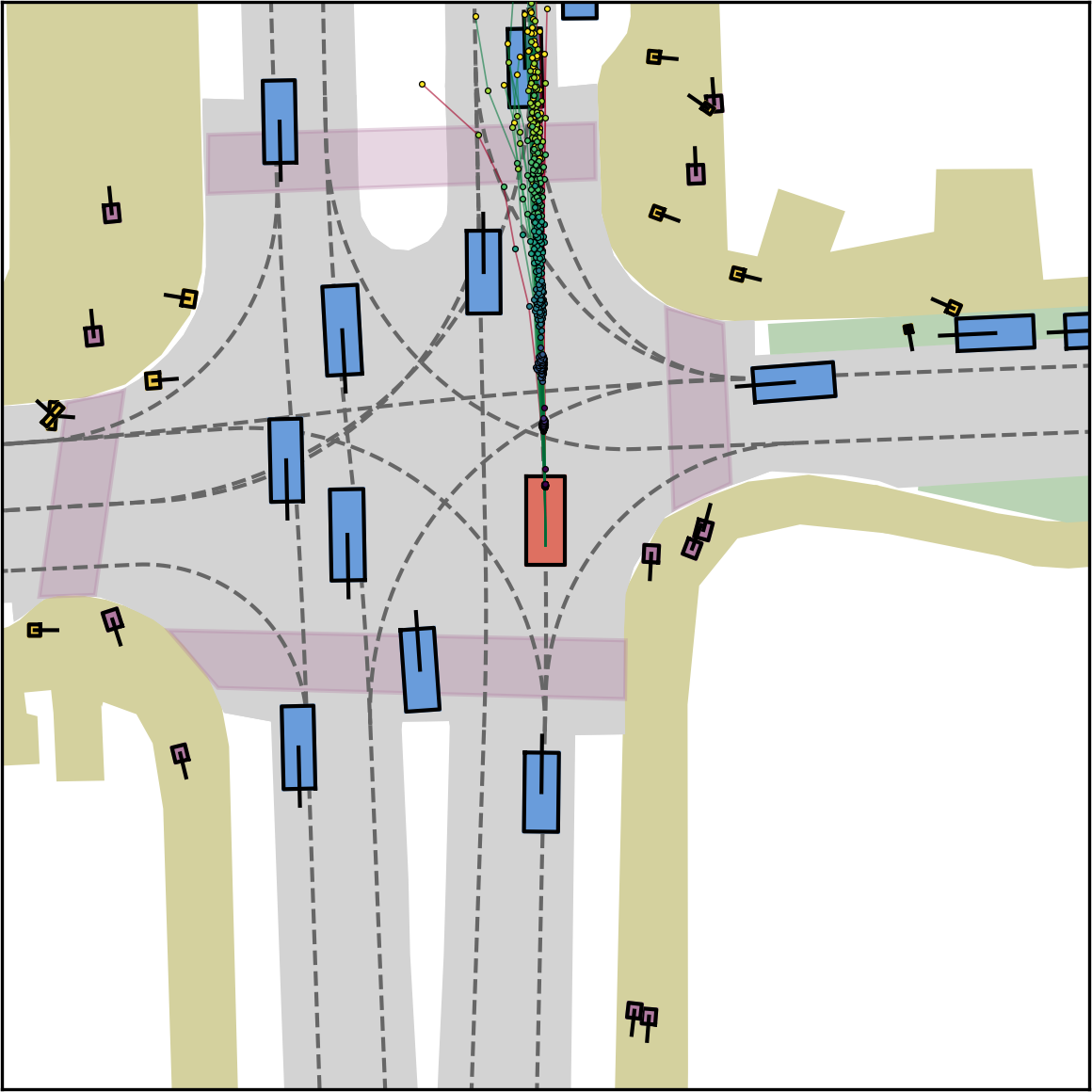} & \vizimgall{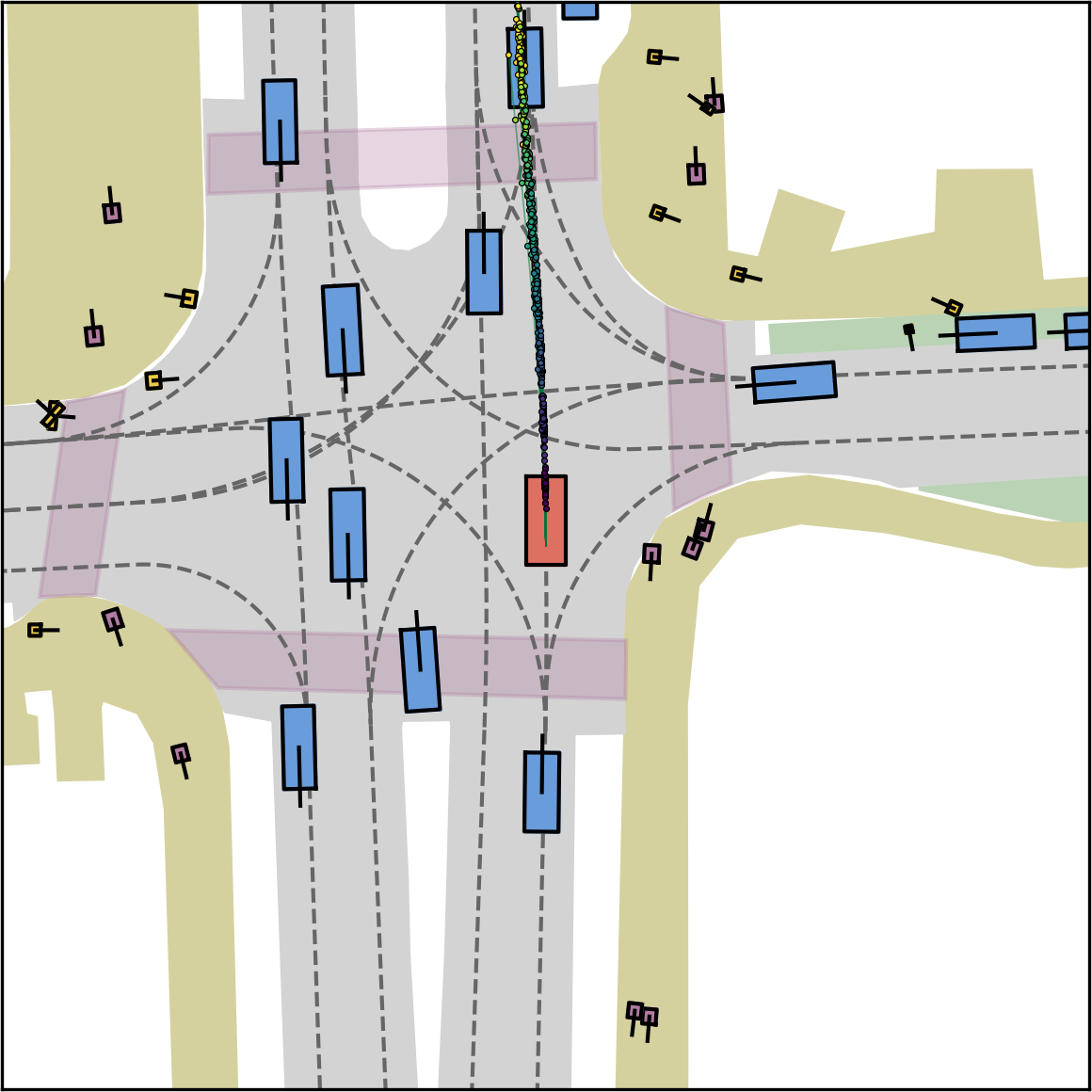} \\
 &
\vizimgall{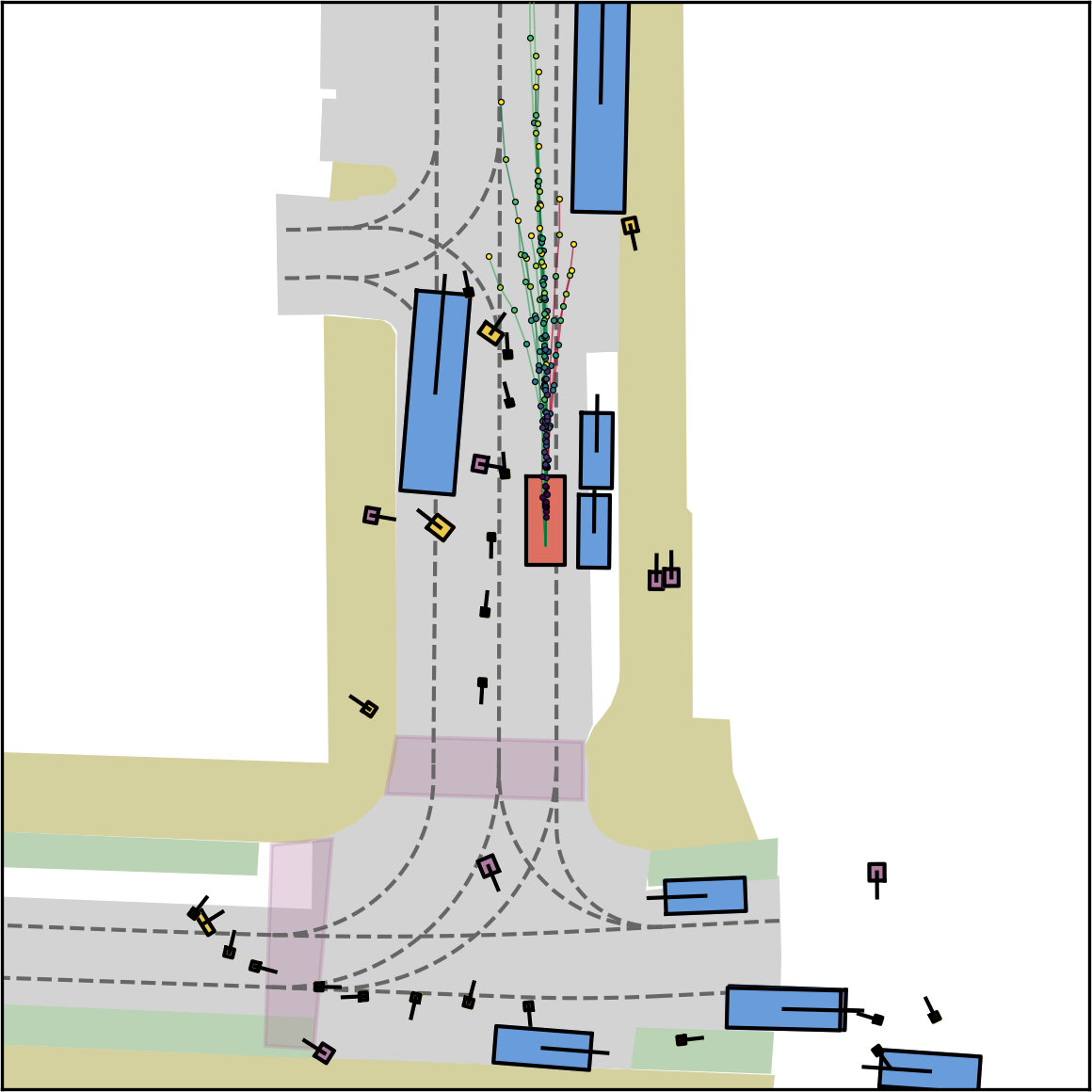} & \vizimgall{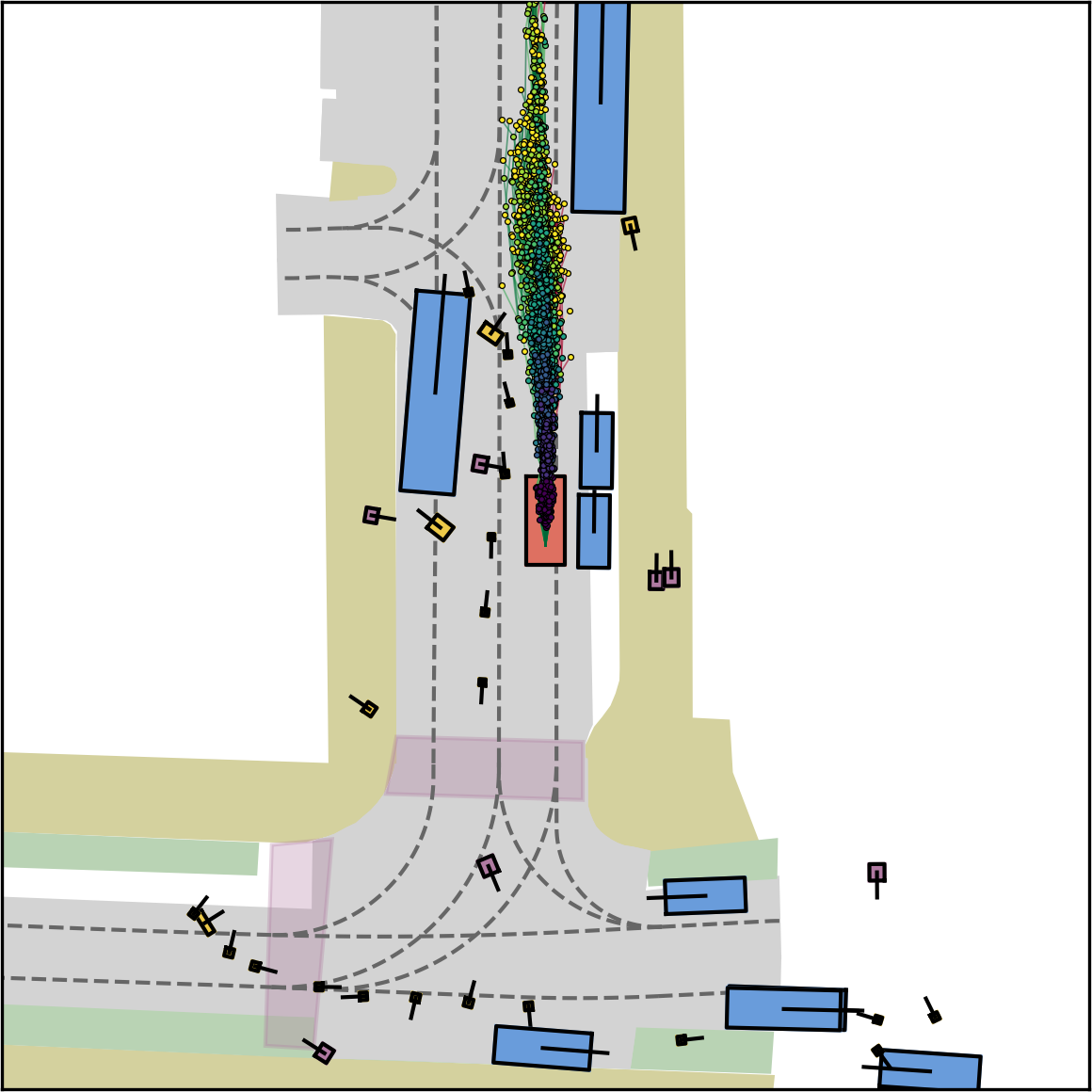} & \vizimgall{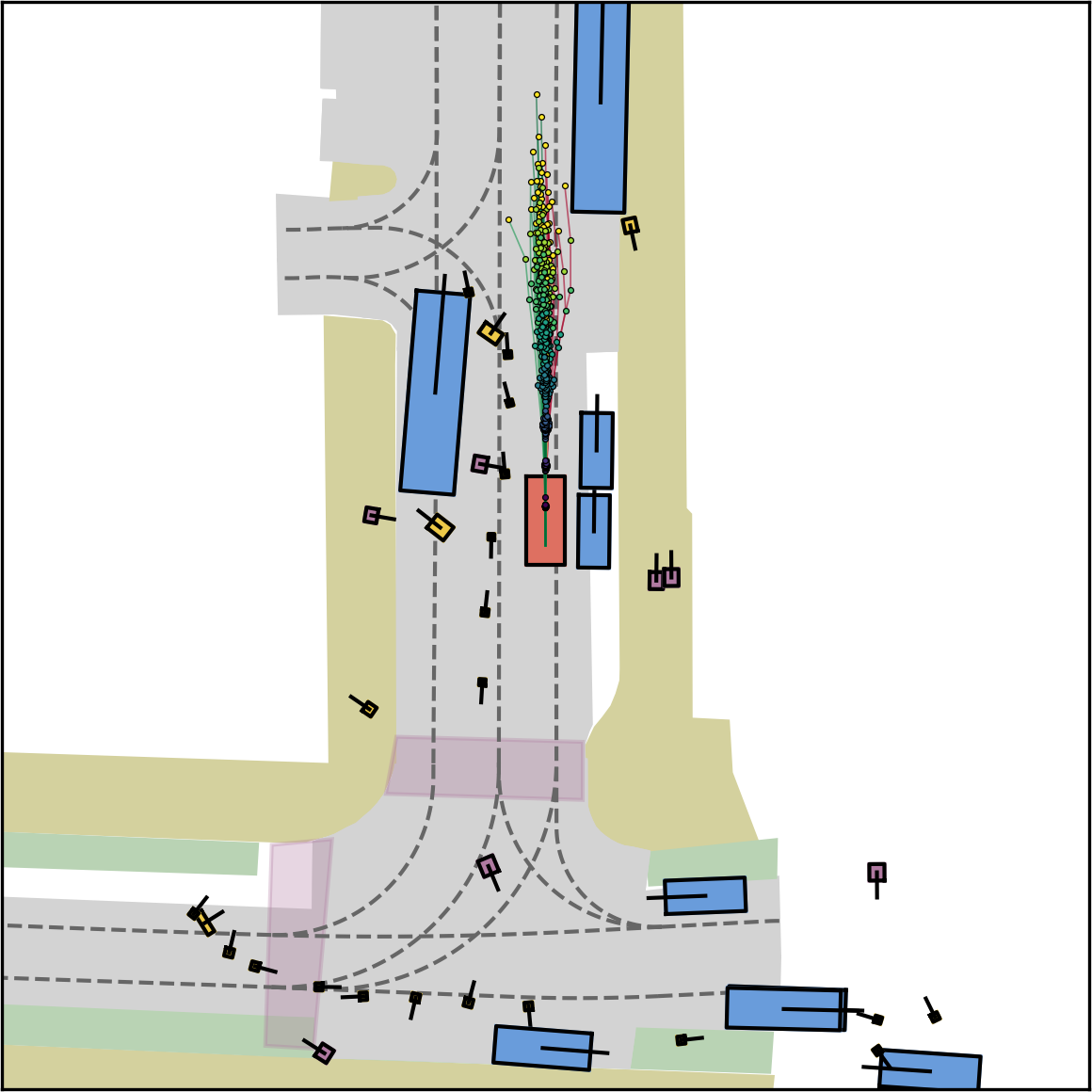} & \vizimgall{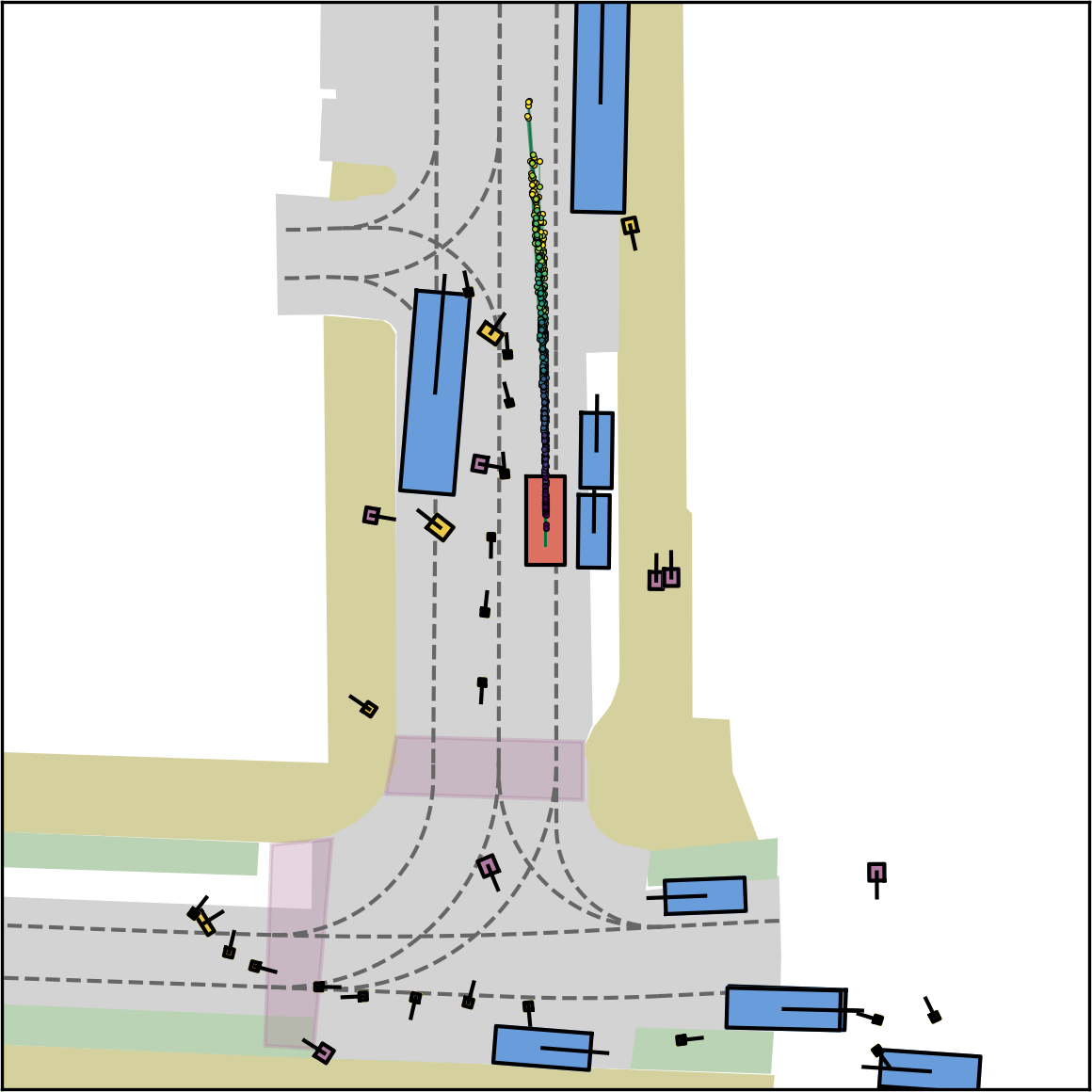} \\
\romarkall{Left} &
\vizimgall{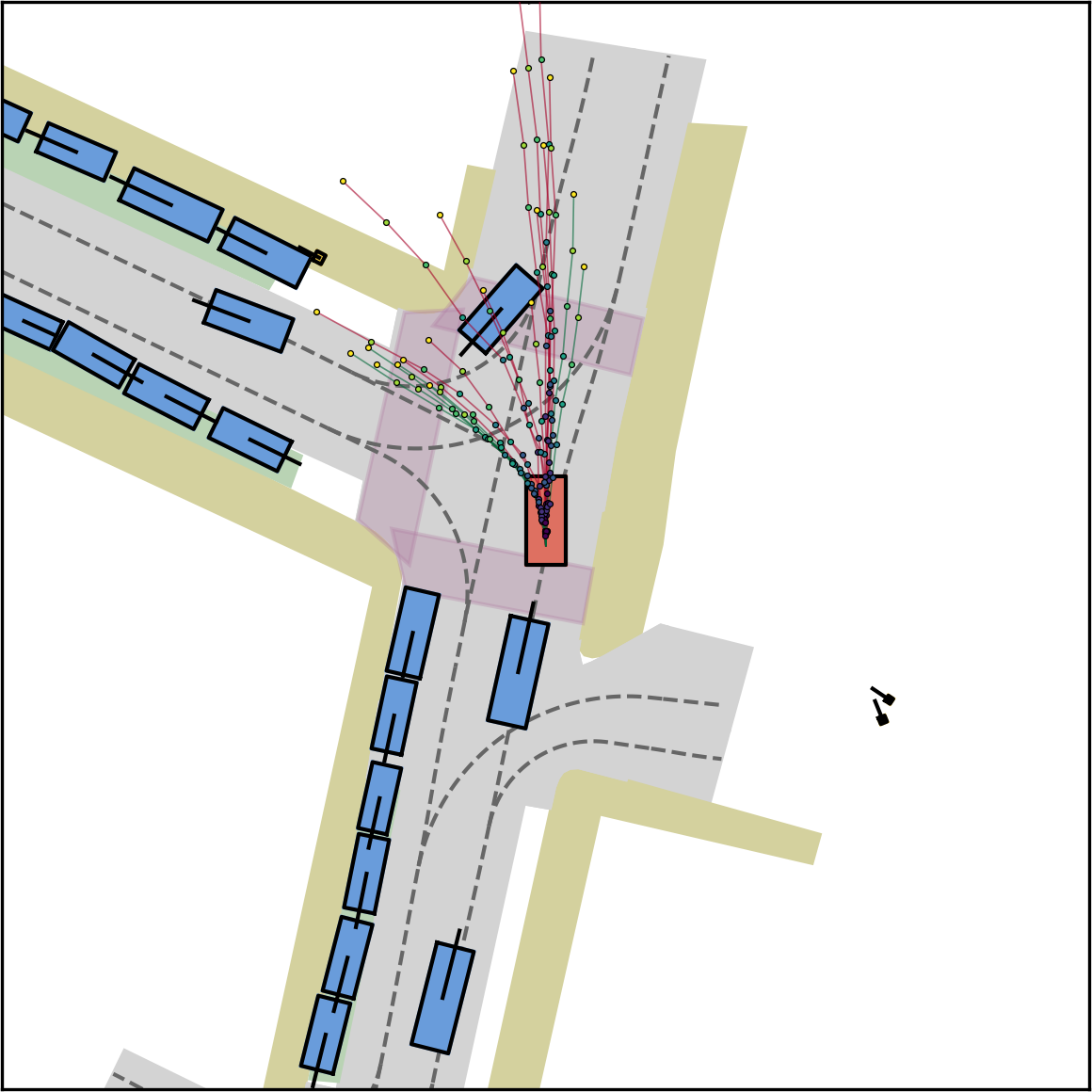} & \vizimgall{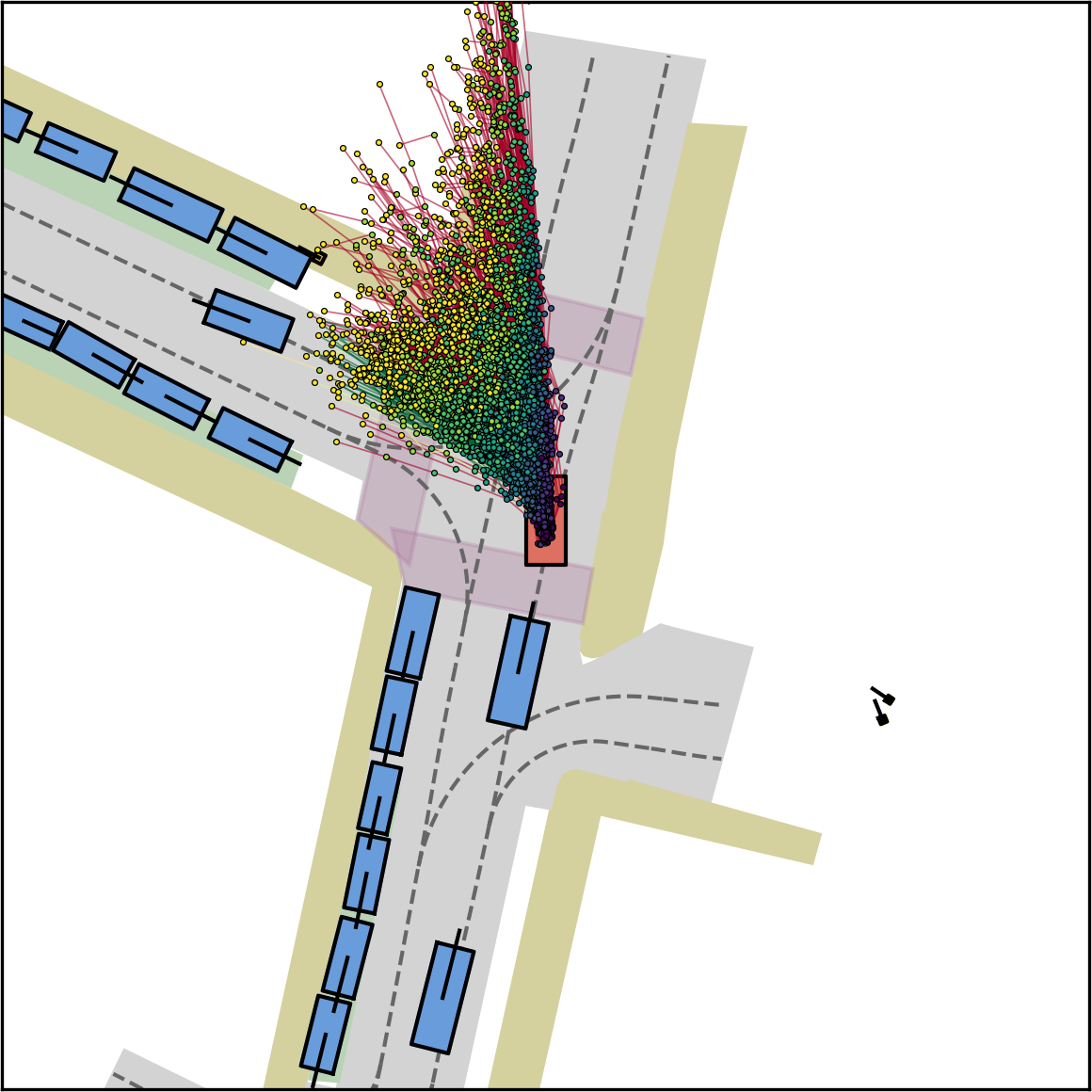} & \vizimgall{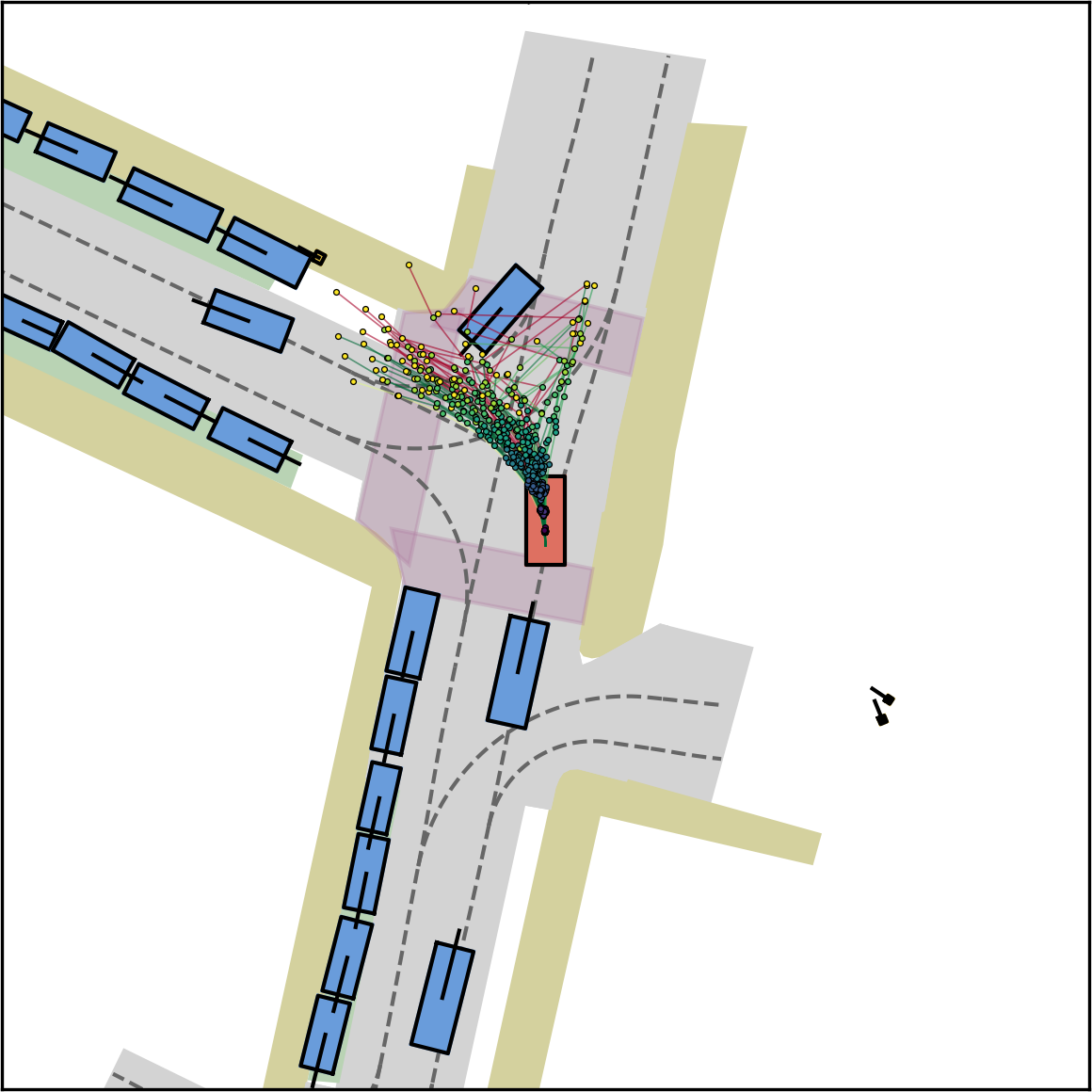} & \vizimgall{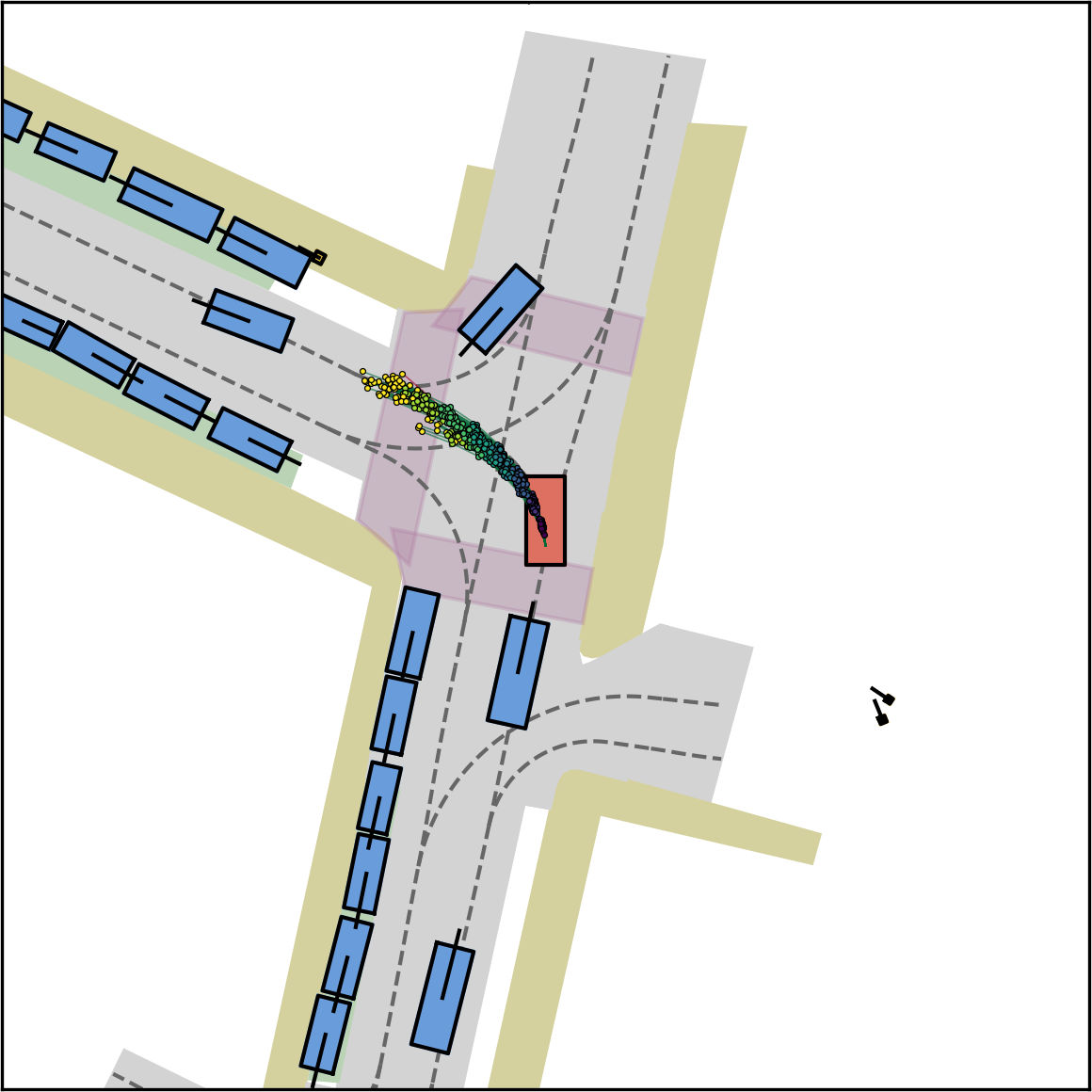} \\
 &
\vizimgall{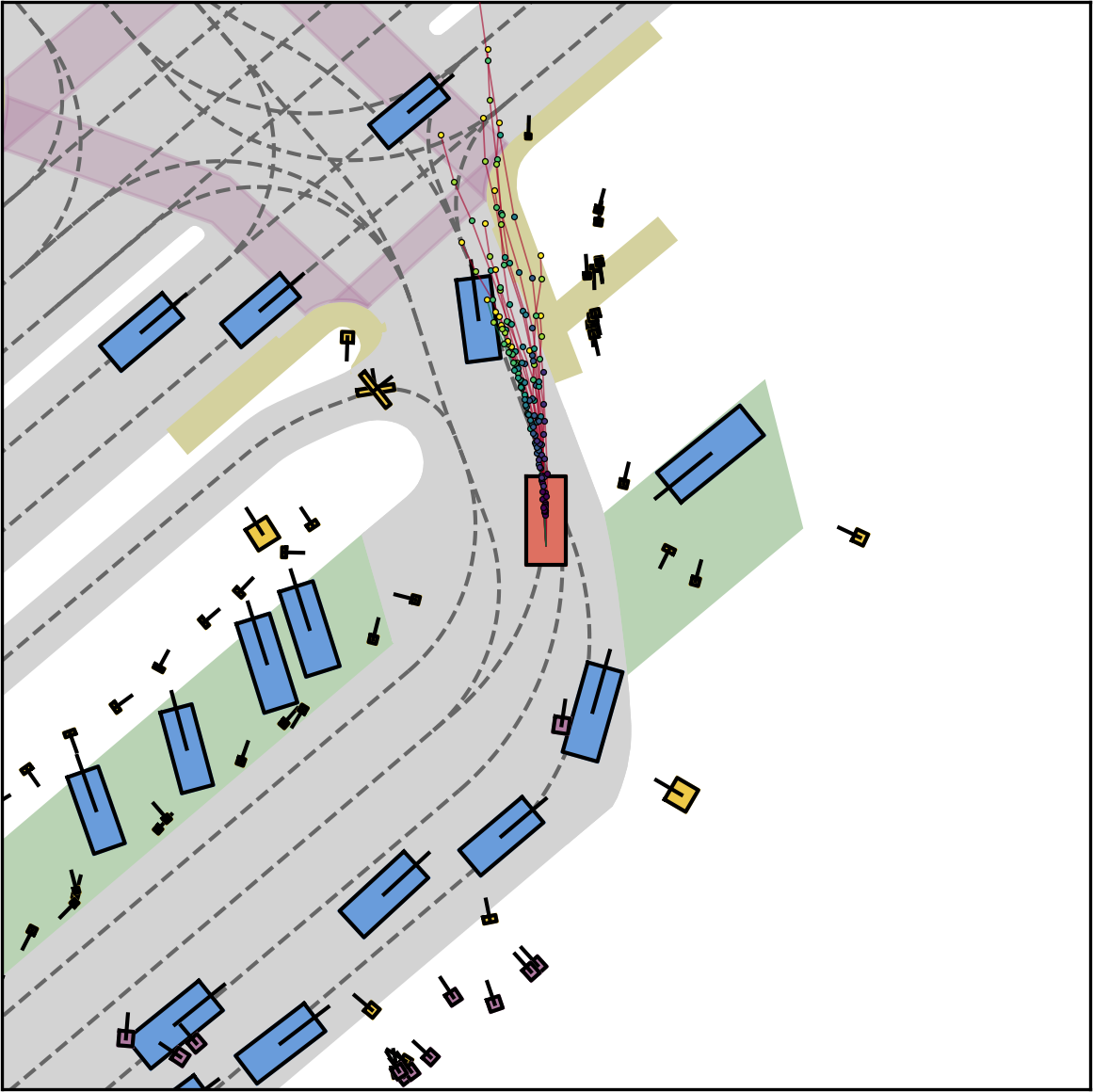} & \vizimgall{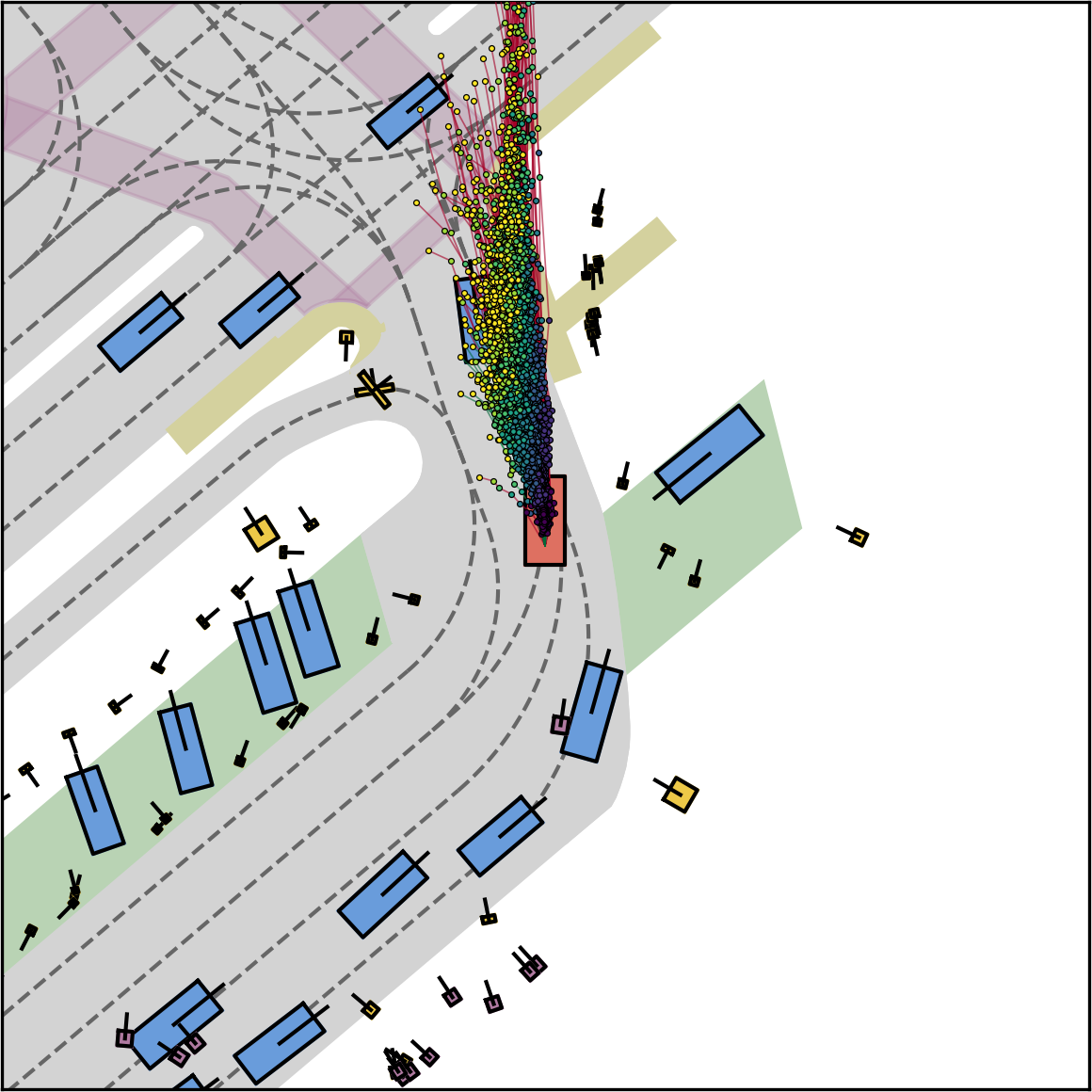} & \vizimgall{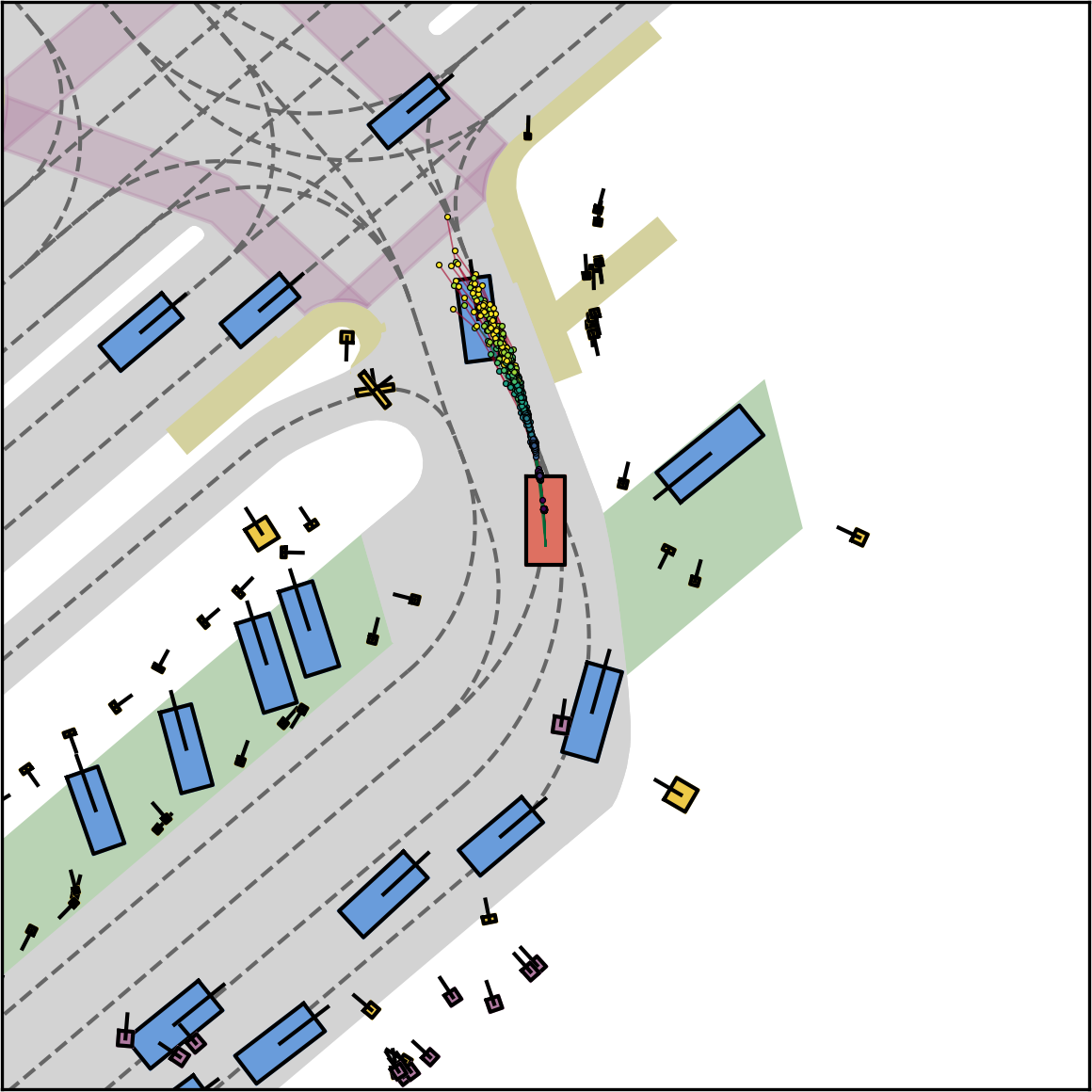} & \vizimgall{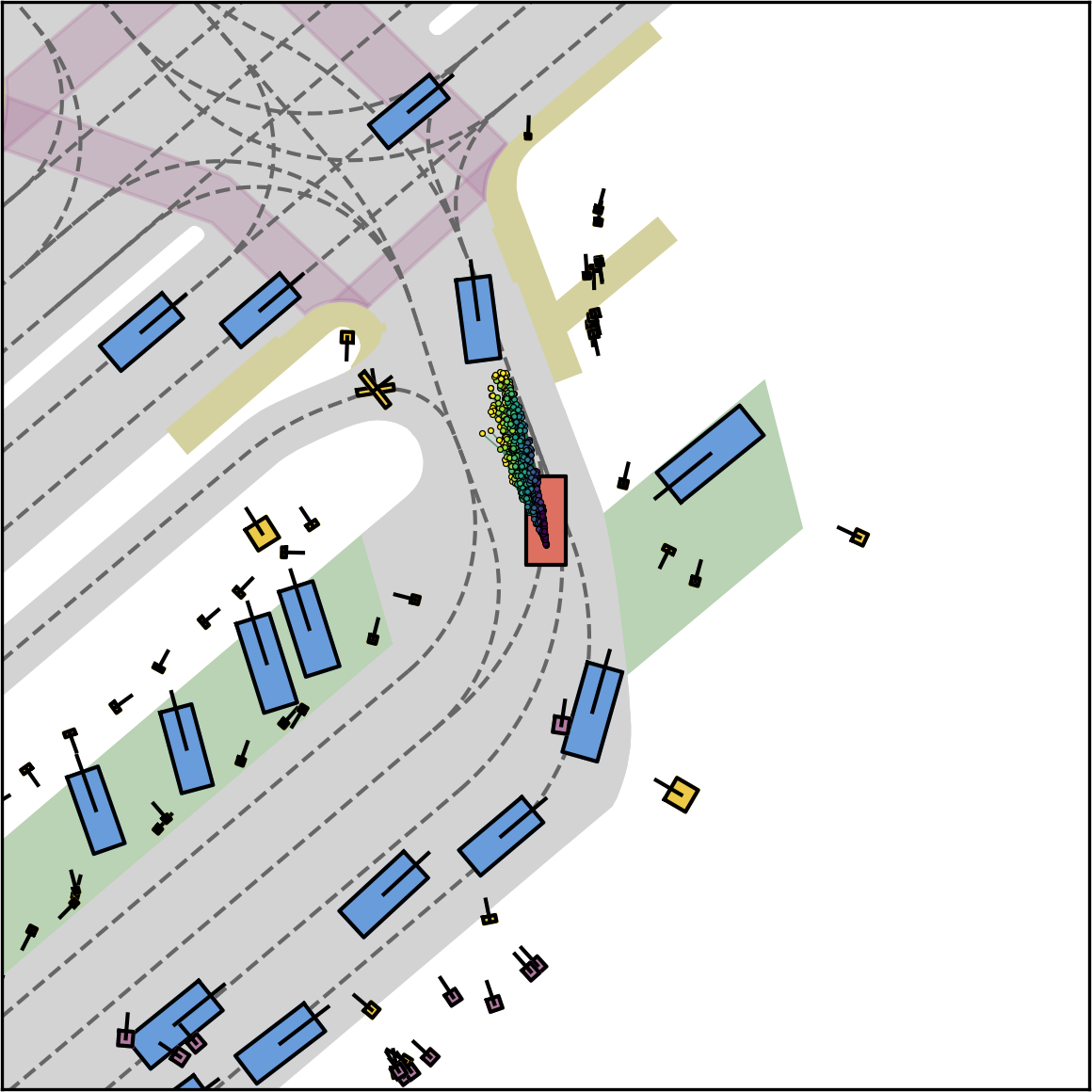} \\
\romarkall{Right} &
\vizimgall{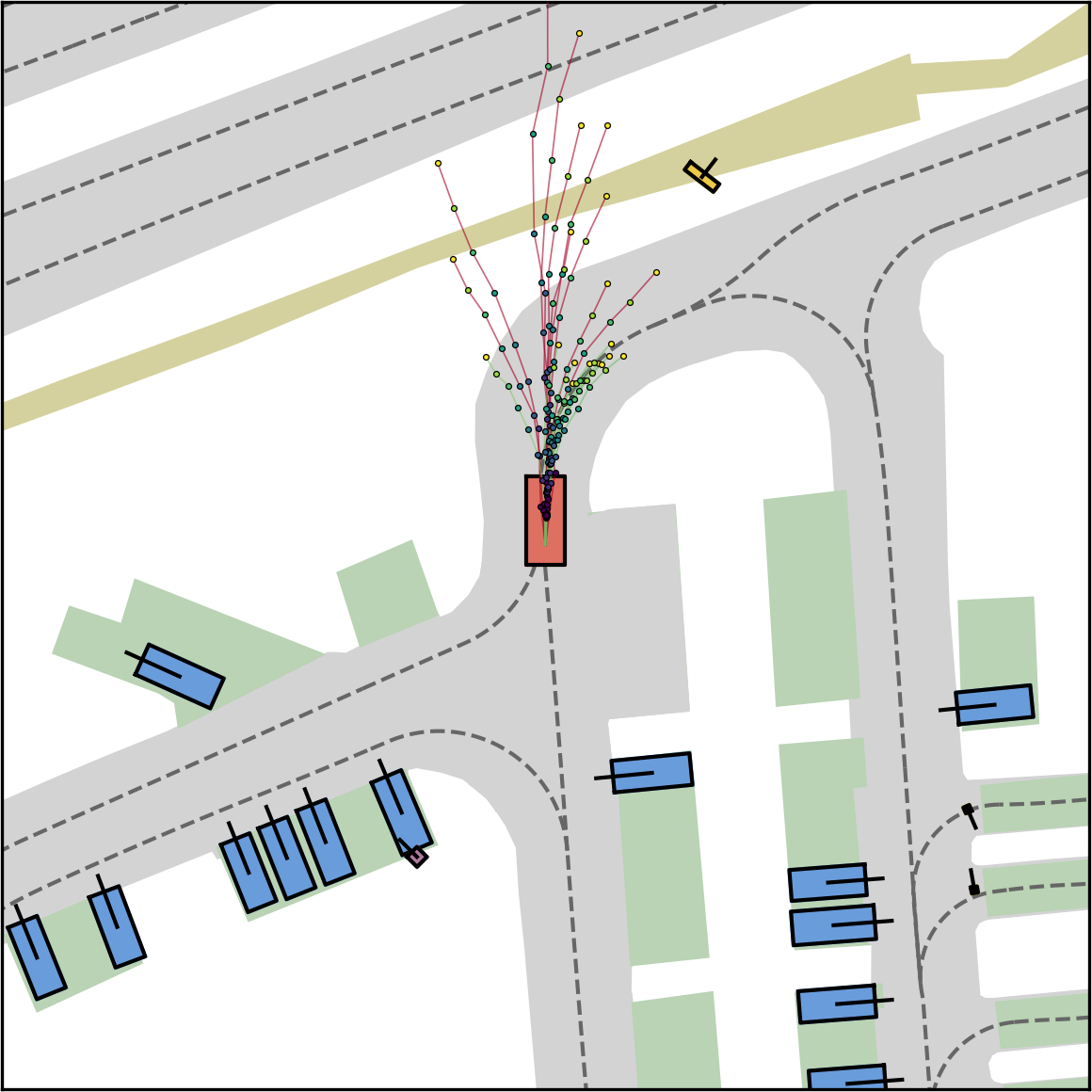} & \vizimgall{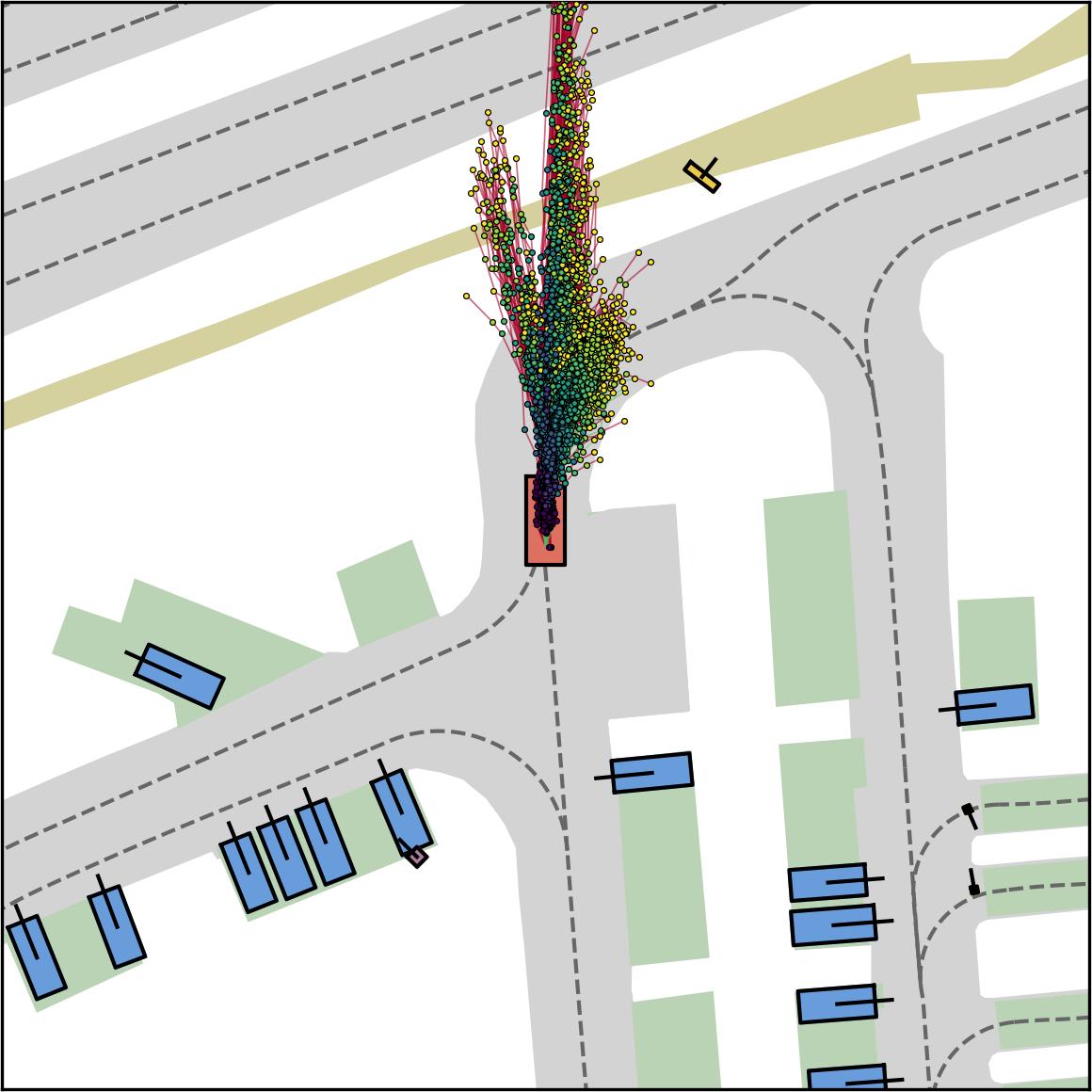} & \vizimgall{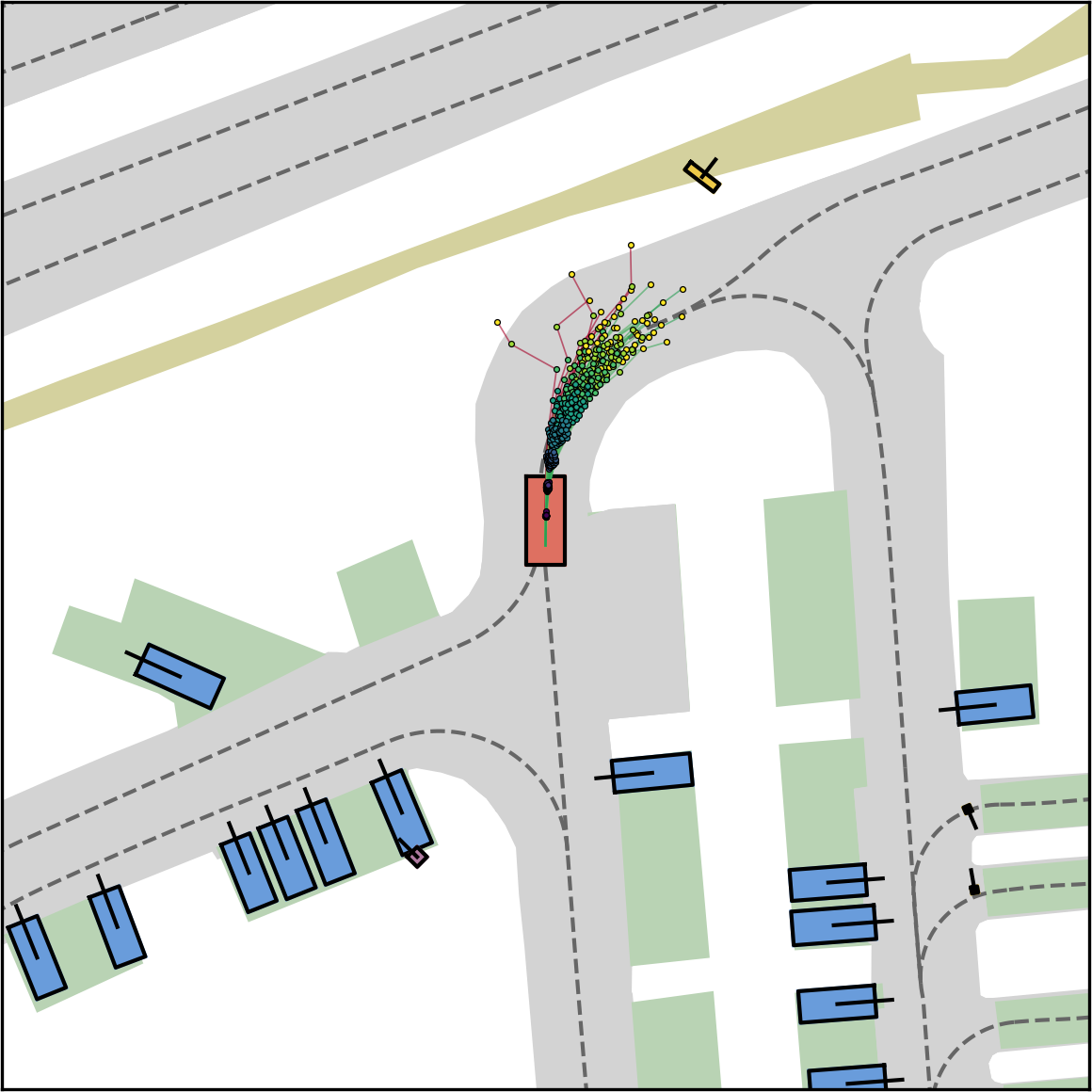} & \vizimgall{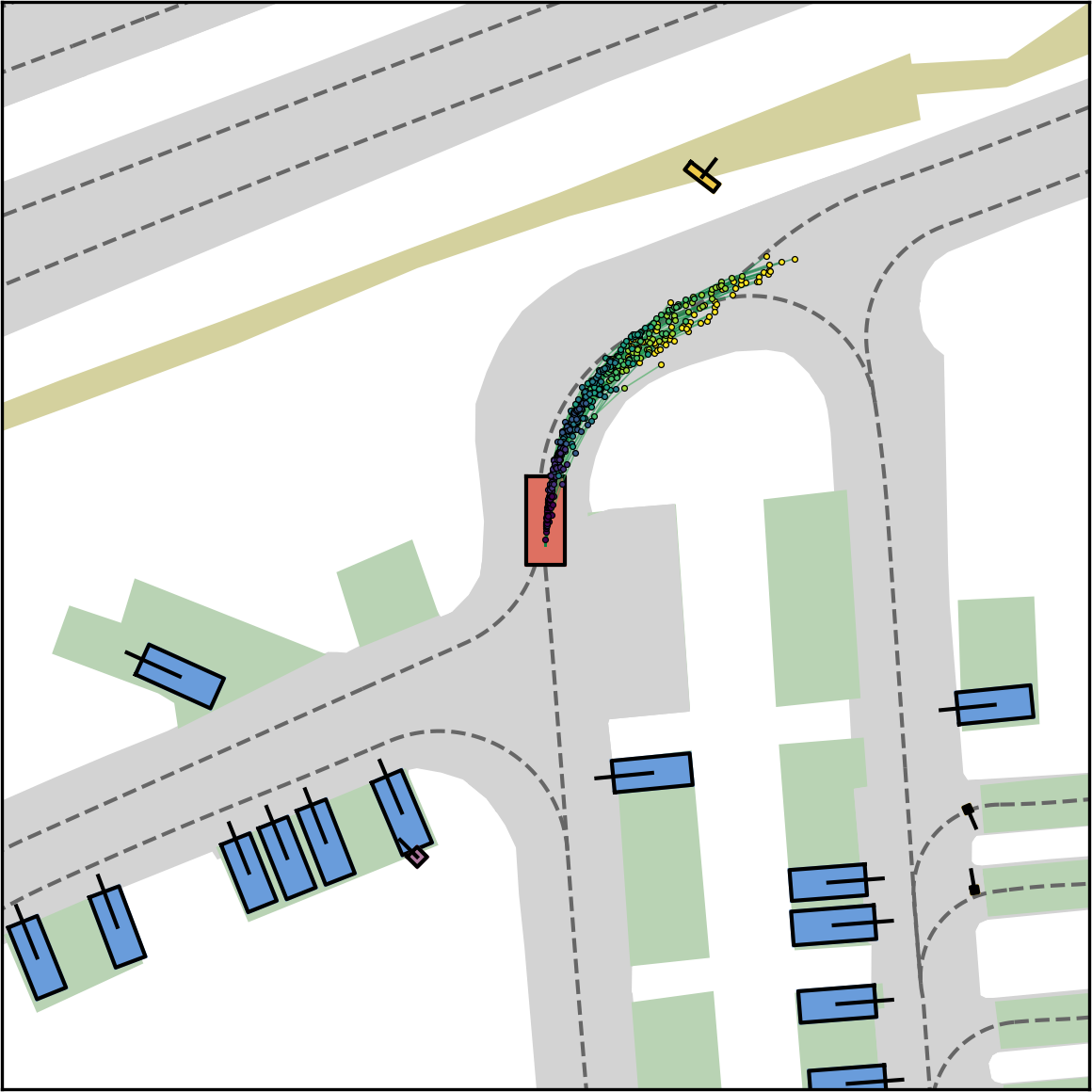} \\
 &
\vizimgall{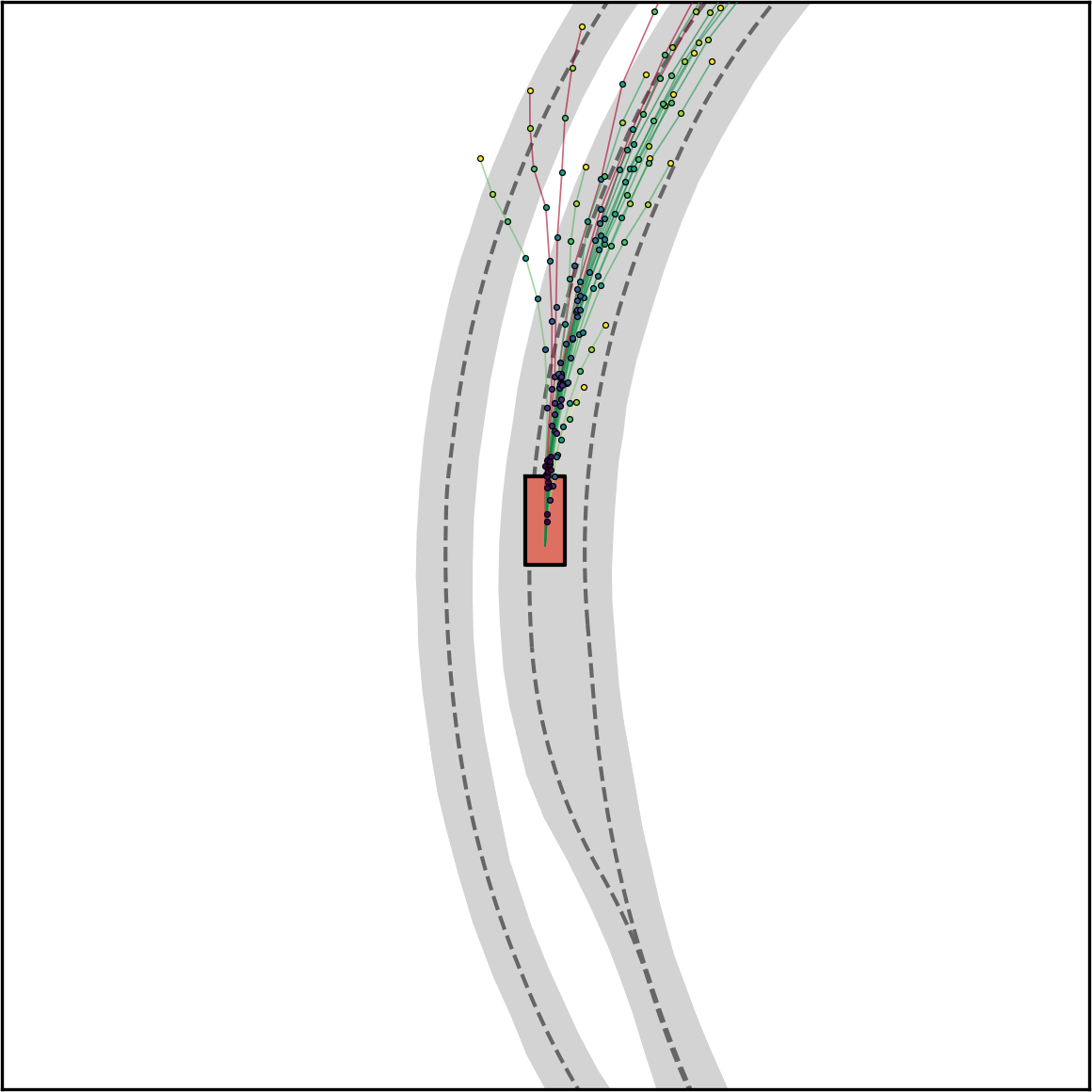} & \vizimgall{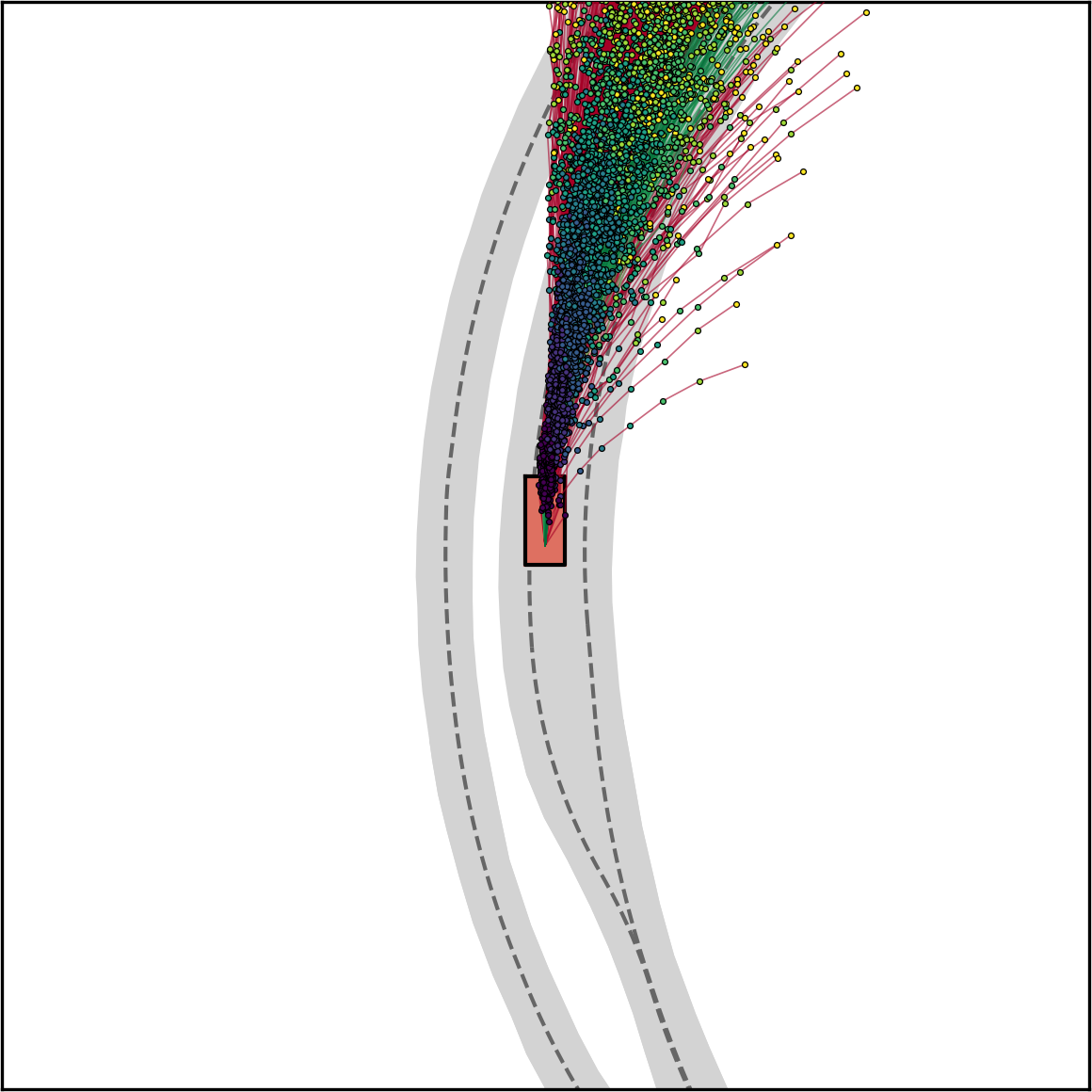} & \vizimgall{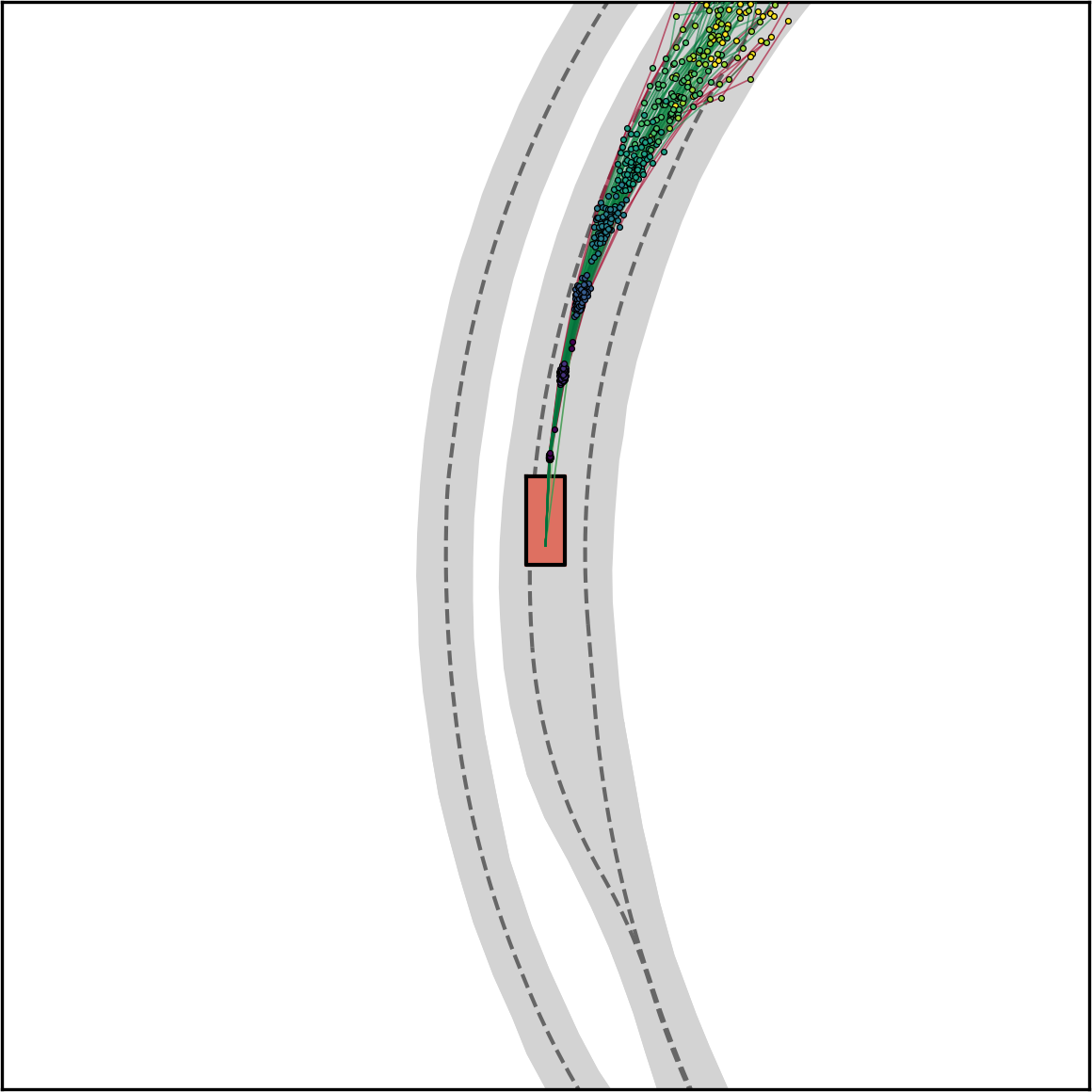} & \vizimgall{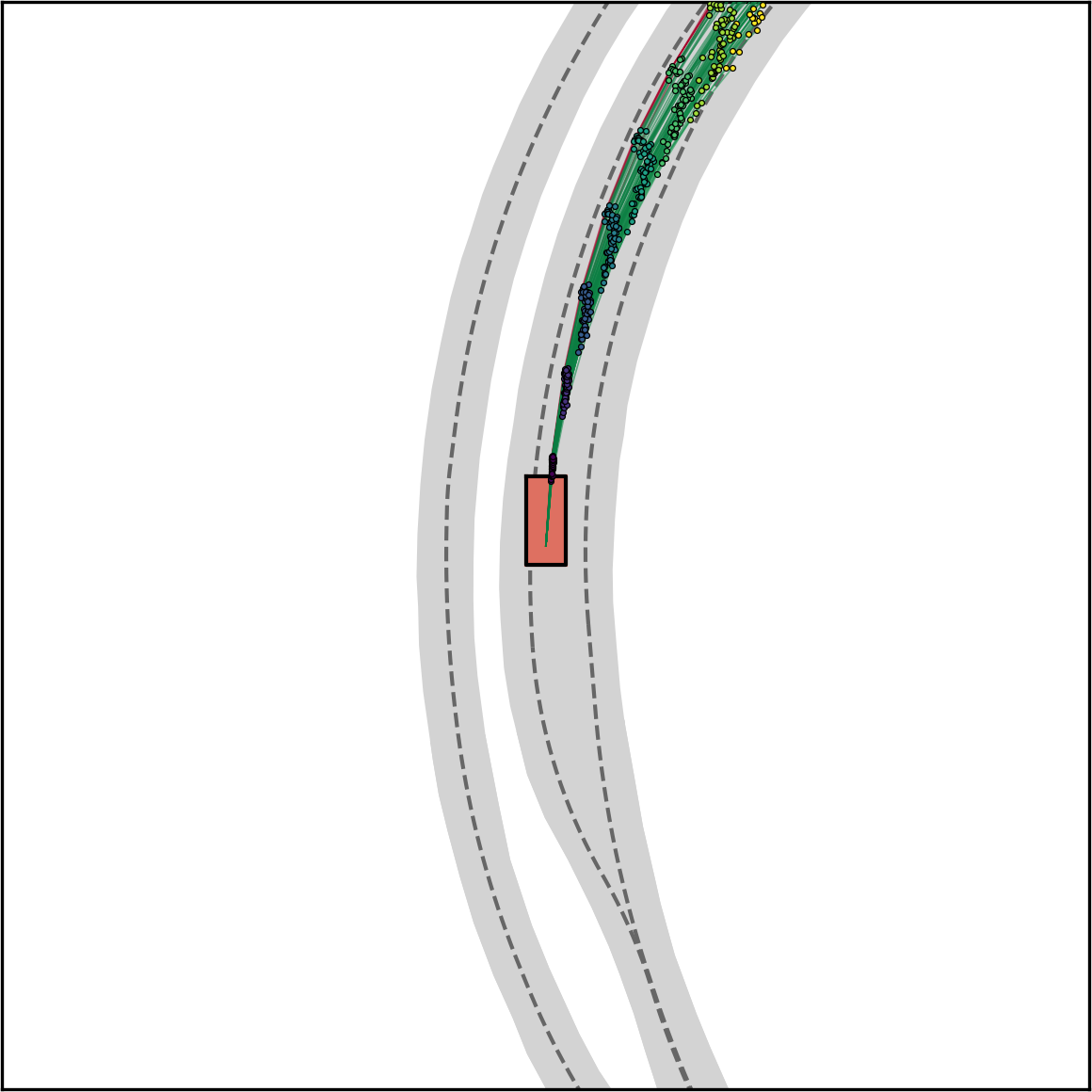} \\
\end{tabular}
\caption{Full qualitative comparison part 3 of 3. Trajectories are colored by PDMS from 0 ({\color[RGB]{200,30,30}red}) to 1 ({\color[RGB]{30,150,30}green}). Scenes are grouped by driving command labeled left.}
\label{fig:qual_all_3}
\end{figure*}


\end{document}